\documentclass[11pt,letter]{article}
\usepackage[papersize={8.5in,11in},margin=1in]{geometry}
\newcommand{\sepline}{\line(1,0){430}}
\newcommand{\seplineb}{\line(1,0){470}}
\usepackage[format=hang,font={footnotesize}]{caption}
\usepackage[numbers,sort,compress]{natbib} 
\usepackage{graphicx}

\usepackage[title]{appendix}
\usepackage{amsthm}
\usepackage{amsmath}
\usepackage{amssymb}
\usepackage{mathtools}
\usepackage{url}
\usepackage{subfigure}
\usepackage{wrapfig}
\usepackage{color}
\usepackage{enumitem}
\usepackage{grffile}
\usepackage{marvosym}
\usepackage{complexity}
\usepackage{setspace}
\usepackage{multicol}
\usepackage{multirow}
\usepackage{lettrine,Zallman}

\usepackage{calligra}

\usepackage{tcolorbox}
\usepackage{url}
\usepackage{color}
\usepackage{tikz}
\usetikzlibrary{positioning,shapes,arrows}
\usepackage{algorithm,algorithmicx} \usepackage[noend]{algpseudocode}

\usepackage{xspace}
\usepackage{dsfont} \usepackage{pifont} \usepackage{musicography}
\usepackage{bbm}
\usepackage{comment}
\usepackage{rotating}
\usepackage{stmaryrd}

\usepackage{tikz}
\usetikzlibrary{shapes,arrows.meta}
\usetikzlibrary{arrows}

\graphicspath{{canon-figs7-arxiv/}}

\newcommand{\dtlinkcolor}{{0.8 0.8 1}} \usepackage[hyperindex=true,pdfpagemode=UseOutlines,bookmarksnumbered=true,bookmarksopen=true,bookmarksopenlevel=2,pdfstartview=FitH,pdfborder={0 0 1},linkbordercolor=\dtlinkcolor,citebordercolor=\dtlinkcolor,urlbordercolor=\dtlinkcolor,pagebordercolor=\dtlinkcolor]{hyperref}
\hypersetup{pageanchor=false,pdfpagelabels}
\usepackage{hyperxmp} \hypersetup{keeppdfinfo,pdfmetalang={en}} \hypersetup{pdftitle={Physics of Language Models: Part 4.1, Architecture Design and the Magic of Canon Layers},pdfauthor={Zeyuan Allen-Zhu},pdfsubject={Language model architecture design},pdfkeywords={Physics of Language Models, Canon layers, architecture design, linear attention, state space models, Mamba, GLA, GatedDeltaNet, Transformer}}
\usepackage[align=center,shadow=true,shadowsize=5pt,nobreak=true,framemethod=tikz,style=0,skipabove=2pt,skipbelow=1pt,innertopmargin=-3pt,innerbottommargin=3pt,innerleftmargin=5pt,innerrightmargin=5pt,leftmargin=-2pt,rightmargin=-2pt]{mdframed}
\usetikzlibrary{shadows}

\allowdisplaybreaks[1]
\theoremstyle{plain} \setitemize{itemsep=0mm, topsep=1mm, leftmargin=8mm}
\setenumerate{itemsep=0mm, topsep=1mm, leftmargin=8mm}
\AtBeginDocument{%
 \abovedisplayskip=6pt minus 1pt
 \abovedisplayshortskip=4pt plus 1pt
 \belowdisplayskip=6pt minus 1pt
 \belowdisplayshortskip=4pt plus 1pt
}

\newcommand{\parhead}[1]{\smallskip \noindent {\bfseries\boldmath\ignorespaces #1.}\hskip 0.9em plus 0.3em minus 0.3em}
\newcommand{\parheadno}[1]{\smallskip \noindent {\bfseries\boldmath\ignorespaces #1}\hskip 0.9em plus 0.3em minus 0.3em}
\newcommand{\parheadsc}[1]{\smallskip \noindent {\boldmath\ignorespaces \textsc{\underline{#1}}.}\hskip 0.9em plus 0.3em minus 0.3em}

\newtheorem{innercustomres}{Result}
\newenvironment{sresult}[1]
  {\renewcommand\theinnercustomres{#1}\innercustomres}
  {\endinnercustomres}

\newtheorem*{theorem*}{Theorem}
\newtheorem{theorem}{Theorem}[section]

\newtheorem*{assumption*}{Assumption}
\newtheorem*{question*}{Question}

\newtheorem*{rep@theorem}{\rep@title}
\newcommand{\newreptheorem}[2]{%
\newenvironment{rep#1}[1]{%
 \def\rep@title{#2 \ref{##1}}%
 \begin{rep@theorem}}%
 {\end{rep@theorem}}}

\newreptheorem{theorem}{Theorem}
\newreptheorem{lemma}{Lemma}
\newreptheorem{proposition}{Proposition}
\newreptheorem{claim}{Claim}
\newreptheorem{definition}{Definition}
\newreptheorem{assumption}{Assumption}

\theoremstyle{definition}

\theoremstyle{remark}
\newtheorem{remark}[theorem]{Remark}
\newtheorem*{remark*}{Remark}

\numberwithin{equation}{section}

\newcommand{\namedref}[2]{\mbox{\hyperref[#2]{#1~\ref*{#2}}}}

\newcommand{\sectionref}[1]{\namedref{Section}{#1}}
\newcommand{\appendixref}[1]{\namedref{Appendix}{#1}}

\newcommand{\resultref}[1]{\namedref{Result}{#1}}

\newcommand{\figureref}[1]{\namedref{Figure}{#1}}
\newcommand{\figurerefb}[2]{\mbox{\hyperref[#1]{Figure~\ref*{#1}#2}}}

\newcommand{\footnoteref}[1]{\namedref{Footnote}{#1}}
\newcommand{\equationref}[1]{\mbox{\hyperref[#1]{(\ref*{#1})}}}
\renewcommand{\eqref}{\equationref}

\makeatletter
\newcommand\xLongLeftRightArrow[2][]%
  {\ext@arrow 0099{\LongLeftRightArrowfill@}{#1}{#2}}
\def\LongLeftRightArrowfill@
  {\arrowfill@\Leftarrow\Relbar\Rightarrow}
\newcommand\xLongRightArrow[2][]%
  {\ext@arrow 0099{\LongRightArrowfill@}{#1}{#2}}
\def\LongRightArrowfill@
  {\arrowfill@\Relbar\Relbar\Rightarrow}
\makeatother

\newcommand{\defem}[1]{\textsf{\emph{#1}}}

\renewcommand{\hbar}{\breve{h}}

\newcommand{\veps}{\varepsilon}

\usetikzlibrary{tikzmark}
\usetikzlibrary{calc}

\errorcontextlines\maxdimen

\newcommand{\ALGtikzmarkcolor}{black}\newcommand{\ALGtikzmarkextraindent}{4pt}\newcommand{\ALGtikzmarkverticaloffsetstart}{-.5ex}\newcommand{\ALGtikzmarkverticaloffsetend}{-.5ex}\makeatletter
\newcounter{ALG@tikzmark@tempcnta}

\newcommand\ALG@tikzmark@start{%
    \global\let\ALG@tikzmark@last\ALG@tikzmark@starttext%
    \expandafter\edef\csname ALG@tikzmark@\theALG@nested\endcsname{\theALG@tikzmark@tempcnta}%
    \tikzmark{ALG@tikzmark@start@\csname ALG@tikzmark@\theALG@nested\endcsname}%
    \addtocounter{ALG@tikzmark@tempcnta}{1}%
}

\def\ALG@tikzmark@starttext{start}
\newcommand\ALG@tikzmark@end{%
    \ifx\ALG@tikzmark@last\ALG@tikzmark@starttext
                    \else
        \tikzmark{ALG@tikzmark@end@\csname ALG@tikzmark@\theALG@nested\endcsname}%
        \tikz[overlay,remember picture] \draw[\ALGtikzmarkcolor] let \p{S}=($(pic cs:ALG@tikzmark@start@\csname ALG@tikzmark@\theALG@nested\endcsname)+(\ALGtikzmarkextraindent,\ALGtikzmarkverticaloffsetstart)$), \p{E}=($(pic cs:ALG@tikzmark@end@\csname ALG@tikzmark@\theALG@nested\endcsname)+(\ALGtikzmarkextraindent,\ALGtikzmarkverticaloffsetend)$) in (\x{S},\y{S})--(\x{S},\y{E});%
    \fi
    \gdef\ALG@tikzmark@last{end}%
}

\apptocmd{\ALG@beginblock}{\ALG@tikzmark@start}{}{\errmessage{failed to patch}}
\pretocmd{\ALG@endblock}{\ALG@tikzmark@end}{}{\errmessage{failed to patch}}
\makeatother

\definecolor{mygreen}{RGB}{80,180,0}

\newcommand{\bblue}[1]{{\color{blue}\textbf{#1}}}

\usepackage{lineno}

\usepackage{varwidth}
\usepackage{CJKutf8}

\newcommand{\bioS}{\textsf{bioS}}

\begin{document}

\newlength{\imgwidthBase}

\title{Physics of Language Models: Part 4.1, \\ Architecture Design and the Magic of Canon Layers}

\date{May 2, 2025\\
\medskip
\normalsize (version 3.0)%
\thanks{{\bf V1} appeared on \href{https://www.ssrn.com/abstract=5240330}{\texttt{SSRN}} this date; {\bf V1.1} (May 18, 2025) improves writing, adds the $\textrm{relu}^2$ experiments, and is accepted by NeurIPS 2025; {\bf V2} adds GDN experiments, tightens others for a stronger, fairer comparison, and re-organizes sections; {\bf V3} adds \resultref{res:2.1} and \sectionref{sec:trans-canon:why} on how Canon layers improve hierarchical feature learning, from our \href{https://youtu.be/8tUIaSOTA7g?t=1361}{\textcolor{blue}{\emph{Jan 2026 talk}}}.
\newline
\indent
\textbf{Future updates and code:} on \href{https://www.ssrn.com/abstract=5240330}{\texttt{SSRN}} and \href{https://physics.allen-zhu.com/part-4-architecture-design/part-4-1}{\texttt{physics.allen-zhu.com}}.
\textbf{YouTube tutorials:} Part 4.1a (\href{https://youtu.be/x3G8knjPDbM}{\texttt{youtu.be/x3G8knjPDbM}}) on our synthetic playground and Part 4.1b (\href{https://youtu.be/8tUIaSOTA7g}{\texttt{youtu.be/8tUIaSOTA7g}}) on our main results.
\newline
\indent
ZA sincerely thanks Vahab Mirrokni for the invitation to the Yale workshop (October 2023), where this research was sparked through enlightening discussions with him and Peilin Zhong.
Canon layers build on the uniform-attention idea explored in \cite{AL2023-cfg}.
ZA thanks Alberto Alfarano for introducing~\cite{press2021train,jelassi2024repeat,trockman2024mimetic,zhou2024transformers}.
At Meta, we sincerely thank Lin Xiao and Kristin Lauter for insightful discussions and unwavering support that made this research possible. Special thanks to Wangzhi Dai, Sam Doud, Dinesh Kannappan, Niki Kim, Junjie Qian, Ammar Rizvi, Travis Seevers, and Stephen Hartken at Meta, and Abraham Leal from $\textrm{W\&B}$, whose invaluable technical assistance made these experiments feasible. We are deeply grateful to Songlin Yang and Ali Behrouz for detailed instructions on replicating their academic-scale pretraining, and Fangcheng Sun for many helpful conversations on architecture design.
\newline
\textit{Contribution statement}.
ZA proposed all ideas, conducted all investigations, implemented all code, ran all experiments, wrote the entire manuscript, and managed all compliance reviews and social promotions; the term \emph{Canon Layers} was jointly conceived and designed with Xiaoli Xu.
}
}

\newcommand{\authorname}[1]{\makebox[6cm][c]{#1}}
\author{
\authorname{Zeyuan Allen-Zhu} \\
\texttt{\href{mailto:zeyuanallenzhu@meta.com}{\color{black}zeyuanallenzhu@meta.com}} \\
FAIR at Meta
}

\maketitle

\begin{abstract}

Understanding architectural differences in language models is challenging, especially at academic-scale pretraining (e.g., 1.3B parameters, 100B tokens), where results are often dominated by noise and randomness. To overcome this, we introduce controlled synthetic pretraining tasks that isolate and evaluate core model capabilities. Within this framework, we discover \emph{Canon layers}: lightweight architectural components—named after the musical term ``canon''—that promote horizontal information flow across neighboring tokens. Canon layers compute weighted sums of nearby token representations and integrate seamlessly into Transformers, linear attention, state-space models, or any sequence architecture.

We present 12 key results. These include how Canon layers enhance reasoning depth (e.g., by $2\times$), reasoning breadth, knowledge manipulation, etc. They lift weak architectures like NoPE to match RoPE, and linear attention to rival SOTA linear models like Mamba2/GDN—validated both through synthetic tasks and real-world academic-scale pretraining.
This synthetic playground offers an \emph{economical, principled path} to isolate core model capabilities often obscured at academic scales. Equipped with infinite high-quality data, it may even \emph{predict} how future architectures will behave as training pipelines improve—e.g., through better data curation or RL-based post-training—unlocking deeper reasoning and hierarchical inference.
\end{abstract}

\thispagestyle{empty}
\clearpage
\setcounter{page}{1}

\section{Introduction}
\lettrine[lines=3]{R}{}ecent
advances in large language models (LLMs) have sparked transformative progress across numerous tasks, including question answering, summarization, translation, code generation~\cite{brown2020language,chowdhery2023palm,touvron2023llama,openai2023gpt4}. Despite rapid progress, systematic understanding of effective neural architecture design has remained elusive, fundamentally hindered by some major challenges.

\parhead{Challenge 1: Pretraining loss as an unreliable proxy for intelligence}
Architectural comparisons often rely on perplexity or cross-entropy loss, but these metrics do not reliably reflect real-world capabilities—especially since natural data is \emph{skills-mixed}. For example, state-space architectures like Mamba~\cite{gu2023mamba,dao2024transformers} frequently achieve lower perplexity early in training due to rapid memorization, yet perform poorly on complex reasoning tasks.
Reliance on \emph{early stopping via perplexity} is thus problematic: it may lead to comparing models that have merely internalized surface-level linguistic patterns without developing deeper reasoning or factual understanding~\cite{ji2023survey}.

\begin{figure}[t!]
\centering
\vspace{-5mm}{\includegraphics[page=8,trim={0mm 90mm 38mm 0mm},clip,width=0.99\textwidth]{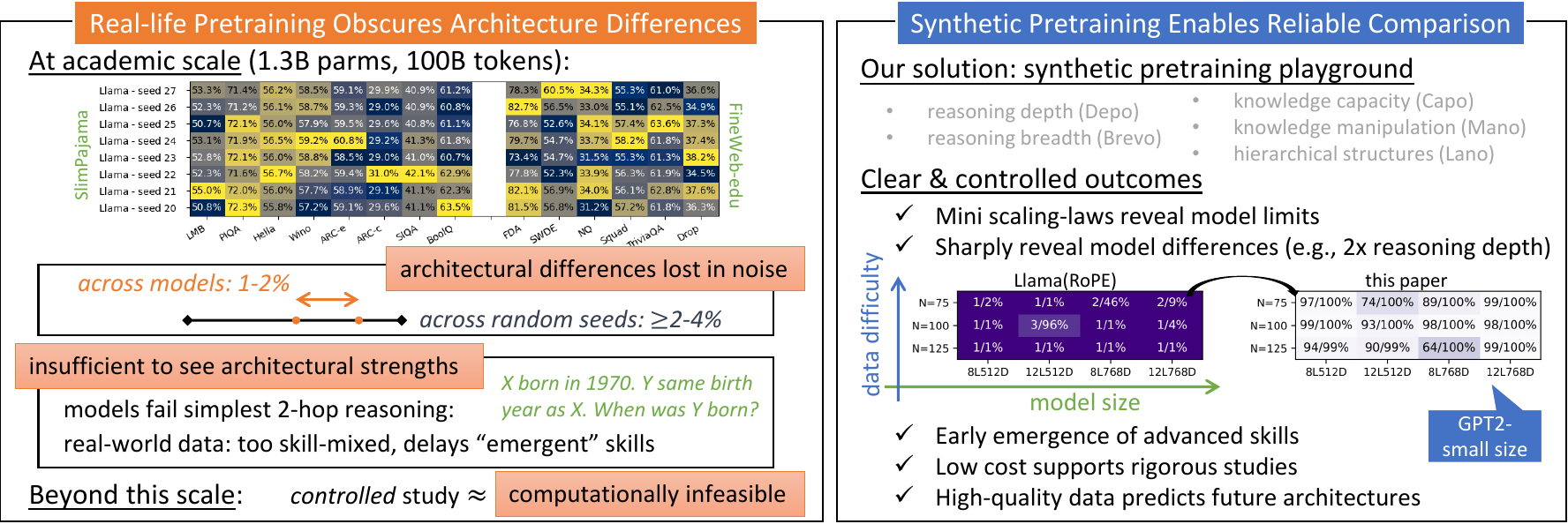}}
\caption{\label{fig:illustrate1}Architecture search in noisy real-life pretraining (good luck!) vs. our synthetic playground (scientific rigor).
\newline
\emph{See \figureref{fig:real-life:random} (Page~\pageref{fig:real-life:random}) for more benchmark variability, including fixed data and varied model random init.}}
\end{figure}

\parhead{Challenge 2: Noise below emergence thresholds}
Emergent abilities—complex skills that only arise in large-scale models (e.g., 7B parameters, 10T tokens~\cite{abdin2024phi})—complicate architectural comparisons at smaller, academic scales (e.g., 1.3B parameters, 100B tokens~\cite{golovneva2025multi,yang2024gated,behrouz2024titans}). At these scales, small benchmark gains (e.g., 2\%) often result from random initialization (and/or data shuffling)—variance that can cause 2–4\% swings in accuracy (see \figureref{fig:illustrate1}).
More fundamentally, \textbf{models fail even the simplest 2-hop reasoning} tasks, performing no better than random guessing.%
\footnote{In our simplest 2-hop reasoning tasks, birth years for 3 individuals are presented, followed by 3 ``[name2] was born in the same year as [name1]'' equivalences. The model is prompted to infer the second group’s birth years. Academic-scale pretrained models can only guess. See \resultref{res:12}.}
This basic reasoning floor \emph{masks architectural differences} in more advanced cognitive skills, making evaluation at this scale deeply unreliable.
While large-scale industry training might reveal these differences, its prohibitive cost blocks systematic ablations, impeding academic contributions to rigorous architecture science—and often reducing design choices to heuristics and guesswork.

\parhead{Challenge 3: Grokking, Data Quality, and Curriculum Learning}
Failures in complex reasoning tasks typically stem from deficiencies in training data, \emph{not} architectural limitations. Too few challenging samples and a lack of intermediate-complexity data often force models to rely on unstable grokking behavior—where generalization only emerges after unnecessarily long pretraining~\cite{power2022grokking}—and disrupt curriculum learning~\cite{bengio2009curriculum}. For instance, models lacking 2-hop reasoning data may unpredictably learn 3-hop tasks after extensive exposure to 1-hop and 3-hop examples. This makes training highly sensitive to randomness, further complicating architectural comparisons. Reinforcement learning (RL)-based post-training methods, such as GRPO~\cite{shao2024deepseekmath} and PPO~\cite{schulman2017proximal}, aim to address this by delivering tailored data at optimal difficulty levels. While effective, these methods introduce new experimental confounds—it becomes unclear whether performance gains stem from pretraining, RL fine-tuning, stochastic training dynamics, or architectural strength.

\parhead{Our approach: Atomic decomposition of intelligence}
To overcome the noise and cost of real-world pretraining—especially at academic scales where even 2-hop reasoning fails to emerge—we decompose intelligence into core (ideally atomic!) components, such as reasoning depth and breadth, and design synthetic, controllable \emph{pretrain} tasks to isolate and evaluate them independently.
This framework sharply characterizes architectural strengths and scalability under clean, idealized conditions (see \figureref{fig:illustrate1}), offering a principled and economical path for architecture design.

This directly addresses Challenge 1 by enabling \emph{single-skill evaluations}, minimizing the confounding factors prevalent in real-world pretraining data. For example, it allows rigorous comparisons of whether architecture A outperforms architecture B in reasoning depth, while ensuring modifications do not degrade other capabilities. By isolating intrinsic architectural biases, synthetic \emph{pretrain} tasks reveal properties often obscured by noise and mixed signals in typical real-life setups.

Challenge 2 is mitigated by \emph{lowering resource} needs for rigorous comparisons. Synthetic benchmarks yield infinite high-quality data, enabling meaningful pretraining even for smaller models (e.g., GPT2-small) where complex skills might otherwise not emerge. In these controlled environments, capabilities like deep multi-hop reasoning \emph{emerge clearly and reliably}, allowing rapid identification of architectural limitations, investigation of \emph{mini scaling-laws}, and the uncovering of trends that real-world pretrained models often fail to reveal due to noise or insufficient signal despite extensive training.

For Challenge 3, we manage data difficulty distributions to ensure adequate representation of intermediate-complexity samples, smoothing learning curves and enabling the \emph{early and consistent emergence} of advanced skills—unlike less predictable real-world data prone to grokking-driven instability. As training pipelines improve—via better data curation or RL-based continued pretraining—synthetic pretrain benchmarks may provide \emph{predictive insight} into which architectures best support scaling to more advanced tasks in the future.

\smallskip
We draw inspiration from physics, where idealized settings---such as frictionless planes or vacuum chambers---reveal first principles by removing confounding factors. Similarly, synthetic tasks eliminate the noise, randomness, and data contamination of real-world datasets, enabling clean, controlled, apples-to-apples architectural comparisons, much like Galileo's Pisa tower experiment.

This paper's key contributions are summarized below:

\parhead{Result 0: Building the Synthetic Playground (\sectionref{sec:data}+\ref{sec:init-compare})}
We introduce five synthetic pretraining tasks---\textsc{Depo} (reasoning depth), \textsc{Brevo} (reasoning breadth), \textsc{Capo} (knowledge capacity), \textsc{Mano} (knowledge manipulation), and \textsc{Lano} (hierarchical language structure). This controlled setup reveals clear, commonsense capability trends \emph{at small scale}: linear attention (e.g., GLA~\cite{yang2023gated}) underperforms consistently; state-space model Mamba2~\cite{dao2024transformers} excels at knowledge but struggles with reasoning; and GDN~\cite{yang2024gated} and Transformers dominate complex reasoning.

\parhead{Result 1: Canon Layers Add Horizontal Information Flow
(\sectionref{sec:canon-description})}
Transformers lack horizontal information flow within layers, leading to inefficiencies even on simple tasks like associative recall. Drawing on the musical canon (overlapping repetition), we introduce \emph{Canon layers}, horizontal ``residual links'' across neighboring tokens that can be flexibly inserted at multiple points --- before attention (Canon-A), inside attention (Canon-B), before MLP (Canon-C), inside MLP (Canon-D). While Canon layers can be implemented in many ways—even simple random averaging is highly effective—this paper focuses on trainable 1-d linear convolutions of kernel size 4. This is lightweight and integrates seamlessly into any sequence model with minimal code.

\parhead{Results 2–5: When Transformer Meets Canon
(\sectionref{sec:trans-canon})}
\begin{itemize}[leftmargin=8mm]
    \item \textsc{Boost performance.}
    In our playground, Canon layers improve reasoning depth (200–400\%), reasoning breadth (30\%), knowledge manipulation length (30\%), and more. These stem from enhanced hierarchical learning dynamics and come with minimal computational overhead.

    \item \textsc{Reviving NoPE.} Integrating Canon layers transforms NoPE models into strong performers, often matching or surpassing RoPE(+Canon). Canon layers outperform positional fixes like ALiBi~\cite{press2021train} or H-Alibi~\cite{jelassi2024repeat}, and reducing/removing RoPE usage improves length generalization.

\item \textsc{Ablation study.} Canon layers contribute cumulatively across sublayer positions (Canon-A/B/C/D), independently of attention or MLP components. \emph{Residual Canon} improves training efficiency; minimal parameter tuning is required without compromising stability.

    \item \textsc{MLP and MoE.} Canon layers can recover some knowledge capacity lost in gated MLP or mixture-of-expert (MoE) architectures, via improved training efficiency and stability.
\end{itemize}

\parhead{Results 6–9: When Linear Models Meet Canon
(\sectionref{sec:linear-models})}
\begin{itemize}[leftmargin=8mm]
    \item \textsc{Universal boost.}
    Across all linear architectures—\emph{GLA, Mamba2, and GDN}—Canon layers consistently enhance \emph{reasoning}: in-context (\textsc{Depo}/\textsc{Brevo}), knowledge (\textsc{Mano}), and structural (\textsc{Lano}), though by varying degrees.
    \begin{itemize}[topsep=0pt]
        \item For linear attention (GLA), Canon lifts reasoning depth from 1 to 4-hop, doubles reasoning breadth and knowledge length, and even surpasses Mamba2.
        \item Mamba2’s built-in \texttt{conv1d} (partial Canon-B) drives most of its gains; removing it drops performance to GLA, while replacing it with full Canon yields further improvements.
        \item GDN benefits least, as its gating and delta updates capture part of Canon-like behavior.
    \end{itemize}

    \item \textsc{Ablation findings.}
    Canon’s residual design ensures stability and never hurts performance.
    Canon-ACD alone often matches \texttt{conv1d}/Canon-B, showing horizontal context flow is universal—not limited to linear-attention or SSM sub-layers.

    \item \textsc{Architectural insight.}
    Most linear-model performance (for Mamba2/GDN) is achievable with the simple \textbf{GLA+Canon} design, suggesting that many modern refinements \emph{might largely replicate} Canon-like mixing rather than introduce new computation.
\end{itemize}

\parhead{Results 10–11: Comparing Transformers and Linear Models
(\sectionref{sec:final-compare})
}
\begin{itemize}[leftmargin=8mm]
    \item \textsc{Controlled comparisons.}
    Equipping all architectures with full Canon layers enables a fair, apple-to-apples evaluation.
    Linear models show $\sim$40\% higher knowledge capacity, but Transformers reach 2–4× greater reasoning depth and stronger structural reasoning.

    \item \textsc{Root cause of shallow reasoning.}
    Linear models fall short \emph{not from insufficient memory}—each layer’s recurrent state is vastly over-provisioned—but from cumulative compression and retrieval errors, pinpointing \defem{memory dynamics} as the main bottleneck.

    \item \textsc{Path forward.}
    Canon-equipped Transformer–linear hybrids can mitigate these limits, enabling deep reasoning with linear efficiency.
\end{itemize}

\parhead{Result 12: Academic-Scale Real-World Pretraining
(\sectionref{sec:real-life})
}
Pretraining 1.3B-parameter models on 100B tokens (context length 4096) shows high noise and limited resolution, making many architectural comparisons statistically unreliable.
\textbf{Still, several consistent patterns emerge.}
Canon layers markedly improve NoPE and GLA—raising them to match RoPE and Mamba2/GDN, respectively—while removing \texttt{conv1d} reduces Mamba2 to GLA level.
Linear models lag behind full Transformers on retrieval-heavy tasks even with Canon, and all models fail 2-hop reasoning, even in short (100-token) contexts, underscoring the limits of academic-scale pretraining.
Reducing or removing RoPE improves long-context generalization when Canon layers are present.
\textbf{These trends mirror our synthetic results} (Results~\ref{res:3},~\ref{res:6.1-gla},~\ref{res:7.1-mamba},~\ref{res:gdn.1},~\ref{res:9-linear},~\ref{res:10},~\ref{res:11}).

\smallskip
In summary, Canon layers fundamentally improve horizontal information flow across diverse architectures, enabling deeper reasoning and efficient scalability. Combined with synthetic benchmarks, they provide systematic insights into future opportunities in model design.

\parhead{Future research}
We plan to extend our study of Canon layers beyond the academic scale.
Preliminary results from larger pretrains (1–8B models on 1–2T tokens) closely align with the findings reported here.
Notably, several synthetic trends—such as Transformer+Canon strongly outperforming Transformer, GLA+Canon matching GDN and outperforming Mamba2—become \emph{clearly observable at these larger scales}.
Code is available on GitHub~\cite{PhysicsLM42}, some models on HuggingFace, and all resources are linked at \href{https://physics.allen-zhu.com/part-4-architecture-design/part-4-1}{\texttt{physics.allen-zhu.com}}.

\begin{figure}[t!]
\centering
\vspace{-5mm}{\includegraphics[page=3,trim={0mm 105mm 8mm 0mm},clip,width=0.99\textwidth]{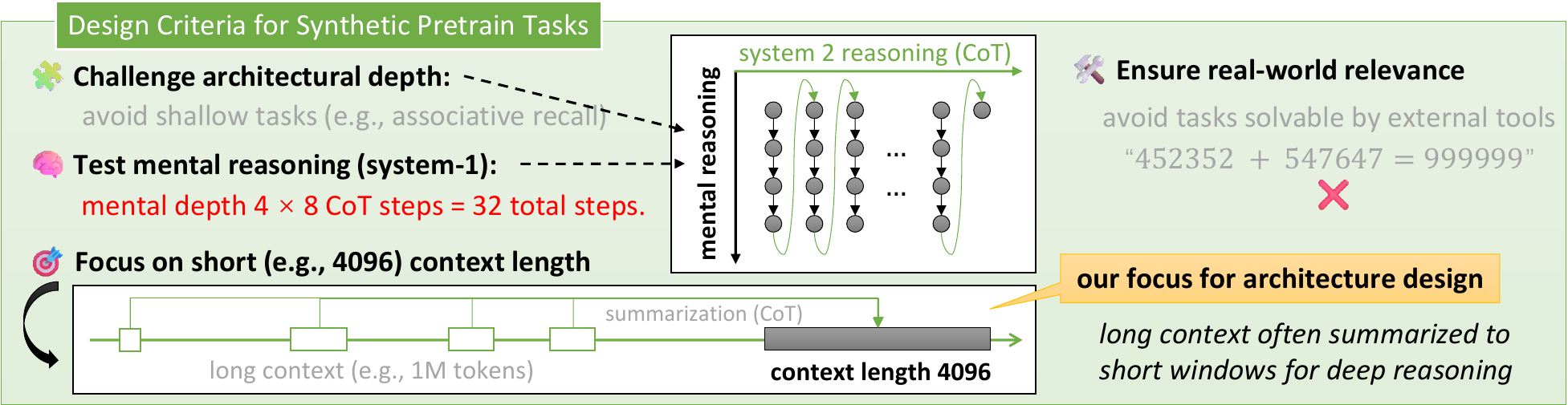}}
\caption{\label{fig:illustrate2}Our design criteria for synthetic pretrain tasks.}
\end{figure}

\section{Synthetic Tasks for Decomposing Intelligence}
\label{sec:data}

We design synthetic tasks to systematically evaluate specific capabilities of language model architectures under controlled conditions, minimizing confounds and enabling clean comparisons. Task selection is guided by four criteria:

\parhead{Criterion 1: Tasks must not be shallow}
Shallow tasks—like associative recall or copying—are easily solvable by small and shallow models, and do not meaningfully test architectural strength. Deep learning relies on stacked layers to progressively learn abstract features~\cite{allen2020backward}, so tasks involving hierarchical reasoning better evaluate architectural scalability and efficiency.

\parhead{Criterion 2: Emphasis on mental thinking}
Tasks should assess a model’s ability to reason internally without Chain-of-Thought (CoT). While CoT helps decompose problems, it does not reflect intrinsic ``system 1'' reasoning~\cite{yu2024distilling}. For example, a model reasoning 4 steps internally and 8 via CoT achieves 32 steps, but \emph{only internal ones reflect architectural strength}. Current models like o3/R1 produce verbose reasoning traces even for trivial prompts (e.g., ``Hello'')—revealing inefficiencies in system 1. To guide architectural progress, tasks must target mental reasoning.

\parhead{Criterion 3: Avoid emphasis on length generalization}
Length generalization is often unstable—sensitive to random seeds and training order~\cite{zhou2024transformers}—and thus unreliable for comparing architectures. While length generalization is important, models over-optimized for long contexts (e.g., 100k tokens) may exhibit reduced performance on standard lengths like 4096 tokens.%
\footnote{This is observed in methods like ALiBi~\cite{press2021train}, Halibi~\cite{jelassi2024repeat}, and Mimetic initialization~\cite{trockman2024mimetic}, whose performance degrades on shorter contexts, as we show in this paper.}
In practice, long inputs are typically summarized into shorter windows before reasoning, so we prioritize evaluating architectures on dense, 4096-token contexts, where critical reasoning unfolds.

\parhead{Criterion 4: Relevance to real-world skills}
Tasks should prioritize broadly applicable skills while avoiding capabilities better suited to external tools. For example, large-number arithmetic (e.g., adding 10-digit numbers) is theoretically interesting but can be delegated to Python interpreters; failures in this area typically reflect limited data exposure rather than architectural weaknesses (e.g., Llama3-70B miscalculates 452352 + 547647). Synthetic tasks should focus on universally relevant skills, aligned with real-world applications, to ensure meaningful assessments.

\begin{figure}[t!]
\centering
\vspace{-5mm}{\includegraphics[page=9,trim={0mm 110mm 48mm 0mm},clip,width=0.99\textwidth]{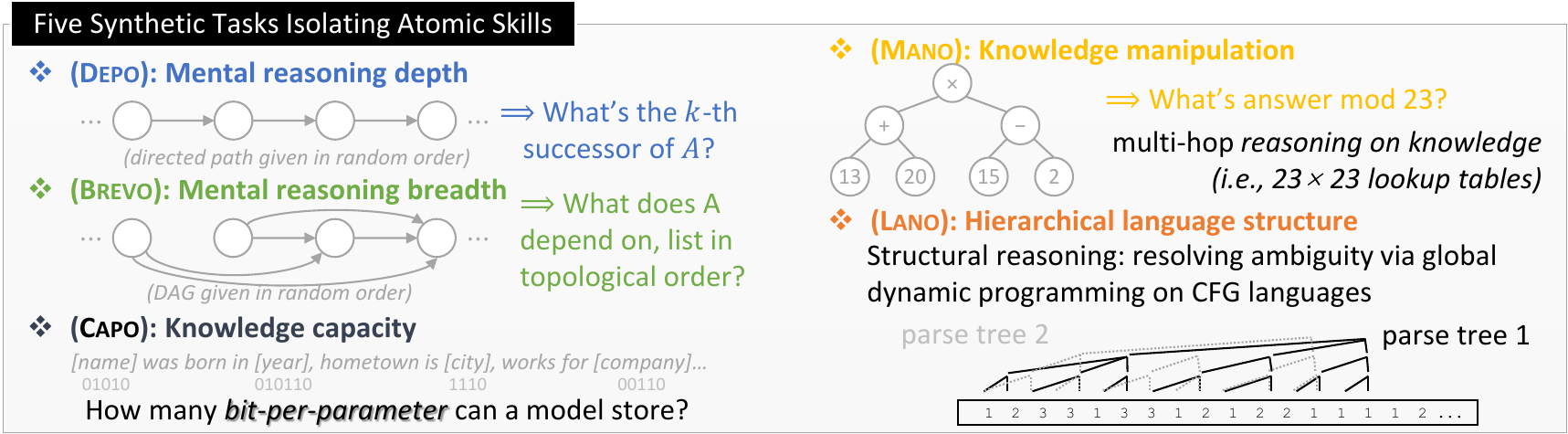}}
\caption{\label{fig:illustrate3}Overview of our five synthetic tasks, each isolating an atomic skill for  rigorous  architectural comparison.
}
\end{figure}

\subsection{Our First Set of Five Synthetic Pretrain Tasks}
\label{sec:data-choice}

To operationalize the criteria above, we design five synthetic tasks—each targeting a distinct dimension of language model capability. We name them \textsc{Depo}, \textsc{Brevo}, \textsc{Capo}, \textsc{Mano}, and \textsc{Lano}.

\parhead{Task \textsc{Depo}: Mental reasoning depth}
Reasoning depth represents a fundamental capability for LLMs, requiring models to retrieve information through multi-step computation. Task \textsc{Depo} evaluates reasoning depth as $k$-hop traversal over directed permutations, where models compute the $k$-th successor for each query $q$ entirely internally, without intermediate steps like Chain-of-Thought (CoT).\footnote{Using CoT would reduce the $k$-hop task to simpler 1-hop associative recall.} Each instance is formatted as:
\vspace{-2mm}
\begin{center}
\verb|<bos> x1 y1 x2 y2 ... xn yn <query_k1> q1 a1 <query_k2> q2 a2 ... <eos>|
\end{center}
\vspace{-2mm}
Here, $2n$ tokens encode $n$ directed edges $x_i \rightarrow y_i$, forming a random permutation of $n$ nodes.

The dataset is controlled by two parameters: $N$, the maximum permutation size, and $K$, the maximum reasoning depth. During training, $n$ is sampled from $[3, N]$, while $k \in [1, K]$. Context lengths are fixed to 2048 tokens.
We employ two variants of \textsc{Depo}:
\begin{itemize}
    \item \textsc{Depo1}: Each node spans 1--2 tokens from vocab size 50, with $N = 225$, $300$, $375$ and $K = 8$.
    \item \textsc{Depo2}: Each node spans 5--7 tokens from vocab size 4, with $N = 75$, $100$, $125$ and $K = 16$.
\end{itemize}
Evaluation focuses on both the hardest cases ($n = N$, $k = K$) and intermediate difficulty ($k = K/2$). For weaker models, we utilize \emph{reduced} training setups with $K = 4$, denoted \textsc{Depo1}$(K=4)$ and \textsc{Depo2}$(K=4)$. The full methodological details are provided in \appendixref{app:depo}.

\parhead{Task \textsc{Brevo}: Mental reasoning breadth}
This evaluates a model's ability to process multiple dependencies simultaneously, as required in tasks involving tree-like traversal or dependency graphs. For example, solving queries like ``Who are Alice's nephews?'' or GSM-like examples requires parallel reasoning across branches of a graph to process relationships bottom-up~\cite{YXLZ2024-gsm1}. Task \textsc{Brevo} isolates this capability using recursive traversal of directed acyclic graphs (DAGs), abstracting away natural language or arithmetic complexities.
Each task instance is formatted as:
\begin{center}
\verb|<bos> x1 y1 x2 y2 ... xm ym <query> q <ans> a1 a2 ... ap <eos>|
\end{center}
Here, $2m$ tokens define $m$ edges $x_i \rightarrow y_i$, representing dependencies where $y_i$ depends on $x_i$. Upon receiving a query vertex $q$, the model outputs all vertices recursively reachable from $q$, sorted in topological order starting from the leaves (e.g., $u \rightarrow v \rightarrow q$ yields output $u$ followed by $v$).

The dataset is parameterized by $N$, the maximum graph size, with DAGs created using $n\leq N$ nodes, each of degree at most 4. Pretraining data is sampled by varying graph sizes, while testing focuses on the hardest graphs ($n = N$). We employ two variants of \textsc{Brevo}:
\begin{itemize}[leftmargin=8mm]
    \item \textsc{Brevo1}: Each vertex name spans a single token, with $N = 70/90/110$, fit within 1024 tokens.
    \item \textsc{Brevo2}: Name spans 2--4 tokens of vocab size 4, with $N = 30/40/50$, fit within 1536 tokens.
\end{itemize}

A key discovery from \cite{YXLZ2024-gsm1}
revealed that, due to the non-uniqueness of valid outputs, language models must preprocess the entire topological order of the DAG \emph{mentally} before generating the first token $a_1$. This insight confirms that our synthetic data rigorously evaluates reasoning breadth by requiring models to globally process the underlying graph structure before producing outputs.

\parhead{Task \textsc{Capo}: Knowledge capacity}
Task \textsc{Capo} evaluates a model's efficiency in encoding factual knowledge directly within its parameters, quantified as \emph{bits per parameter}, which measures reliable storage capacity.
Following the framework in \cite{AL2024-knowledgeScaling}, synthetic datasets of (fake) biographies are constructed to test knowledge retention. Each biography includes several attributes (e.g., birthdate, university, employer, etc.) and is presented in diverse paraphrased formats to reduce surface-level memorization~\cite{AL2023-knowledge,AL2023-knowledgeUB}. Capacity is measured using the next-token prediction distribution, accounting for both exact correctness and partial accuracy.

To highlight architectural differences, we adopt an undertrained regime where each biography is exposed only 100 times during pretraining.\footnote{Exposing each biography 1000 times during pretraining diminishes architectural differences, as even transformers without MLP layers can achieve similar storage efficiency~\cite{AL2024-knowledgeScaling}. Uniform exposure ensures clean systematic comparisons while avoiding confounding effects tied to rare outliers and junk data~\cite{AL2024-knowledgeScaling}.} The dataset includes $N = 50\text{K}$ to $2\text{M}$ biographies, encoding $2 \times 10^6$ to $10^8$ total bits of information. Models of varying sizes are tested, and results are visualized via ``bit vs. model size'' plots. Additional details are provided in \appendixref{app:capo}.

\parhead{Task \textsc{Mano}: Knowledge manipulation}
Task \textsc{Mano} evaluates a distinct form of reasoning: the ability to manipulate stored knowledge internally, contrasting with in-context reasoning tasks like \textsc{Depo} or \textsc{Brevo}. While those tasks focus on reasoning over external tokens, \textsc{Mano} requires models to retrieve factual knowledge embedded in their parameters and perform hierarchical computation entirely mentally. This combination of retrieval and reasoning makes knowledge manipulation uniquely challenging and a skill that must be learned during pretraining.\footnote{For instance, questions like ``Was [name] born in an even or odd month?'' or derived 2-hop queries such as ``What is [name]’s sister's birthdate?'' demand reasoning layers over stored knowledge. These skills cannot reliably emerge through supervised fine-tuning alone~\cite{AL2023-knowledgeUB} and require development during pretraining or continued pretraining.}

To test this capability, \textsc{Mano} employs synthetic modular arithmetic expressions inspired by human mental computation, particularly small-number arithmetic like the $9 \times 9$ multiplication table. Models solve multi-step arithmetic problems without intermediate steps like Chain-of-Thought. For example, given:
\verb|<bos> + * a b - c d <ans>|
the task requires evaluating $((a \times b) + (c - d)) \bmod 23$ for $\ell=3$, where operands $a, b, c, d$ are sampled uniformly from $[0, 22]$. Modular arithmetic provides the foundational factual knowledge ($23 \times 23$ operation tables), while the task challenges hierarchical reasoning by recursively composing operations.
Additional details are provided in \appendixref{app:mano}.

The dataset is parameterized by a maximum expression length $L$, with $\ell$ sampled uniformly from $[1, L]$. We prepare three \textsc{Mano} datasets across difficulty levels: $L = 10$, $13$, and $16$.

\parhead{Task \textsc{Lano}: Hierarchical language structure}
Task \textsc{Lano} evaluates structural reasoning over hierarchical relationships and long-range dependencies. Unlike \textsc{Depo}, \textsc{Brevo}, and \textsc{Mano}, which rely on explicit key-value pairs (in-context or knowledge), \textsc{Lano} challenges models to infer implicit recursive structures across sequences and resolve global ambiguities within them.

To test this, \textsc{Lano} leverages synthetic datasets built from context-free grammars (CFGs). Training sequences consist of CFG-valid sentences separated by \verb|<bos>| tokens. For example:
\begin{center}
\verb|<bos> 3 3 2 2 1 ... 3 3 1 2 <bos> 1 2 3 3 1 ... 1 2 2 1 <bos> ...|
\end{center}
CFGs are designed with token-level ambiguity, where local tokens (e.g., 1, 2, 3) provide insufficient information to directly infer their mapping to CFG rules. Resolving this requires dynamic programming to globally map the entire sequence to a valid recursive application of CFG rules, which must also be learned during training. This reasoning grows in worst-case complexity ($O(n^3)$) as sequence lengths increase. Details are in \appendixref{app:lano}.

Building upon \textsf{cfg3f}~\cite{AL2023-cfg}, which includes sequences of lengths 100--500, we introduce extended datasets \textsf{cfg3j} and \textsf{cfg3k}, with sequences ranging up to 200--1000 tokens to increase recursive depth and test models on more nested rules and longer dependencies. Training uses context lengths of 1536 for \textsf{cfg3j} and \textsf{cfg3k}, compared to 512 for \textsf{cfg3f}. Evaluation prompts models with \verb|<bos>| to generate CFG-valid sentences, validated via a dynamic programming parser. KL divergence is also used to compare token distributions against ground truth.

\parhead{In summary}
This set of five synthetic tasks covers non-overlapping skills and distinct aspects of accuracy—token-level (\textsc{Depo}, \textsc{Mano}), generative (\textsc{Brevo}, \textsc{Lano}), and distributional (\textsc{Capo}, \textsc{Lano}). While this pool can be further enriched, it serves as a strong starting point for deriving meaningful architectural insights, as demonstrated in the following sections.

\begin{figure}[t!]
\centering
\setlength{\imgwidthBase}{0.242\textwidth}
\vspace{-5mm}%
\includegraphics[page=1,trim={2.5mm 1.5mm 2.5mm 1.5mm},clip,width=\imgwidthBase]{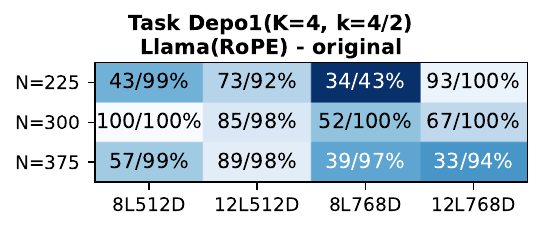}
\includegraphics[page=1,trim={2.5mm 1.5mm 2.5mm 1.5mm},clip,width=\imgwidthBase]{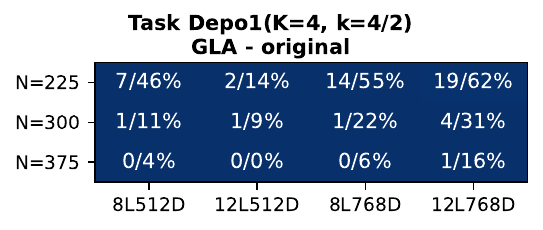}
\includegraphics[page=1,trim={2.5mm 1.5mm 2.5mm 1.5mm},clip,width=\imgwidthBase]{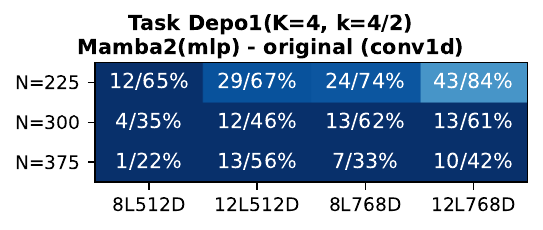}
\includegraphics[page=1,trim={2.5mm 1.5mm 2.5mm 1.5mm},clip,width=\imgwidthBase]{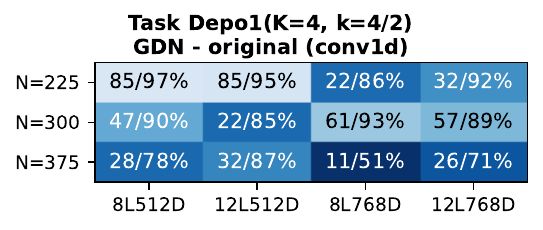}
\\
\includegraphics[page=1,trim={2.5mm 1.5mm 2.5mm 1.5mm},clip,width=\imgwidthBase]{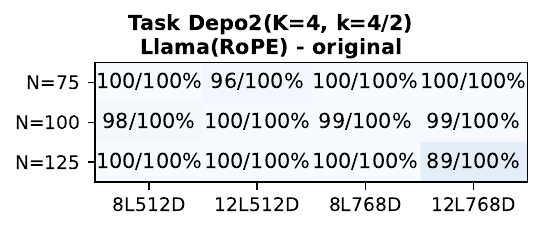}
\includegraphics[page=1,trim={2.5mm 1.5mm 2.5mm 1.5mm},clip,width=\imgwidthBase]{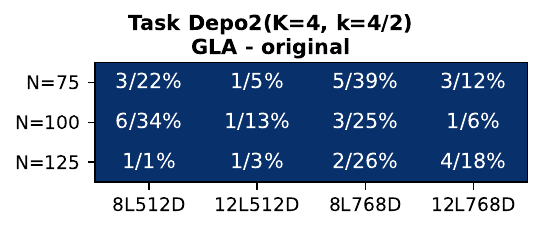}
\includegraphics[page=1,trim={2.5mm 1.5mm 2.5mm 1.5mm},clip,width=\imgwidthBase]{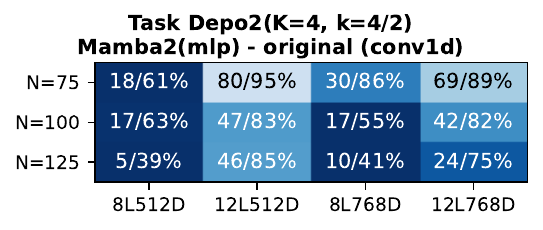}
\includegraphics[page=1,trim={2.5mm 1.5mm 2.5mm 1.5mm},clip,width=\imgwidthBase]{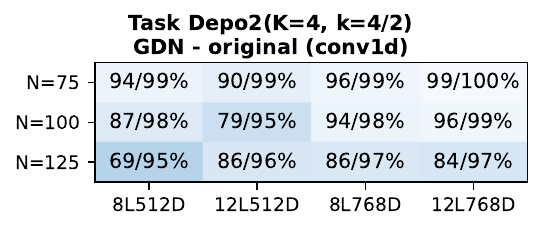}
\\
\includegraphics[page=1,trim={2.5mm 1.5mm 2.5mm 1.5mm},clip,width=\imgwidthBase]{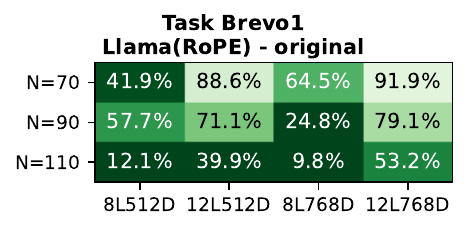}
\includegraphics[page=1,trim={2.5mm 1.5mm 2.5mm 1.5mm},clip,width=\imgwidthBase]{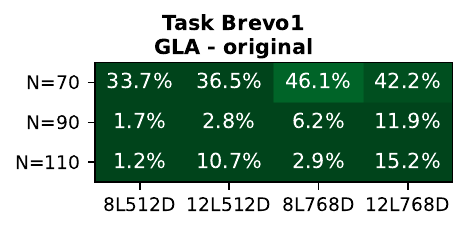}
\includegraphics[page=1,trim={2.5mm 1.5mm 2.5mm 1.5mm},clip,width=\imgwidthBase]{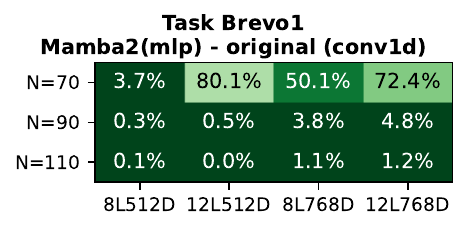}
\includegraphics[page=1,trim={2.5mm 1.5mm 2.5mm 1.5mm},clip,width=\imgwidthBase]{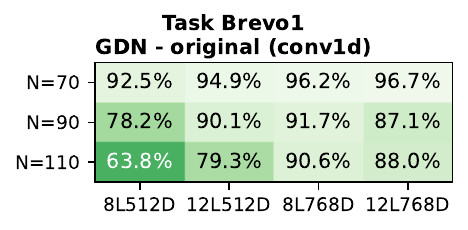}
\\
\includegraphics[page=1,trim={2.5mm 1.5mm 2.5mm 1.5mm},clip,width=\imgwidthBase]{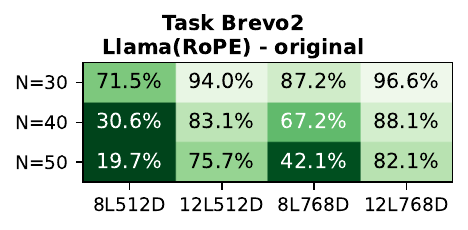}
\includegraphics[page=1,trim={2.5mm 1.5mm 2.5mm 1.5mm},clip,width=\imgwidthBase]{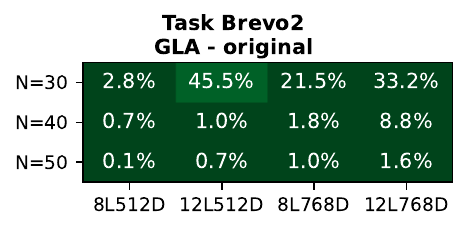}
\includegraphics[page=1,trim={2.5mm 1.5mm 2.5mm 1.5mm},clip,width=\imgwidthBase]{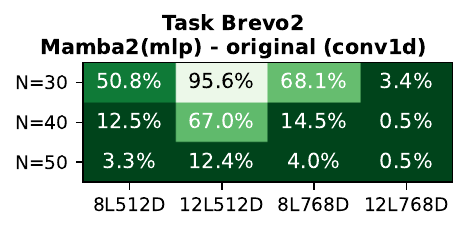}
\includegraphics[page=1,trim={2.5mm 1.5mm 2.5mm 1.5mm},clip,width=\imgwidthBase]{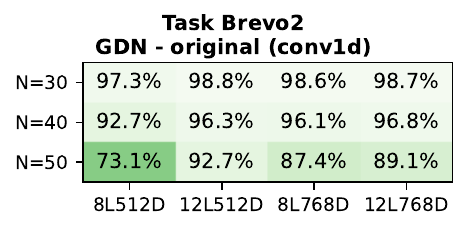}
\\
\includegraphics[page=1,trim={2.5mm 1.5mm 2.5mm 1.5mm},clip,width=\imgwidthBase]{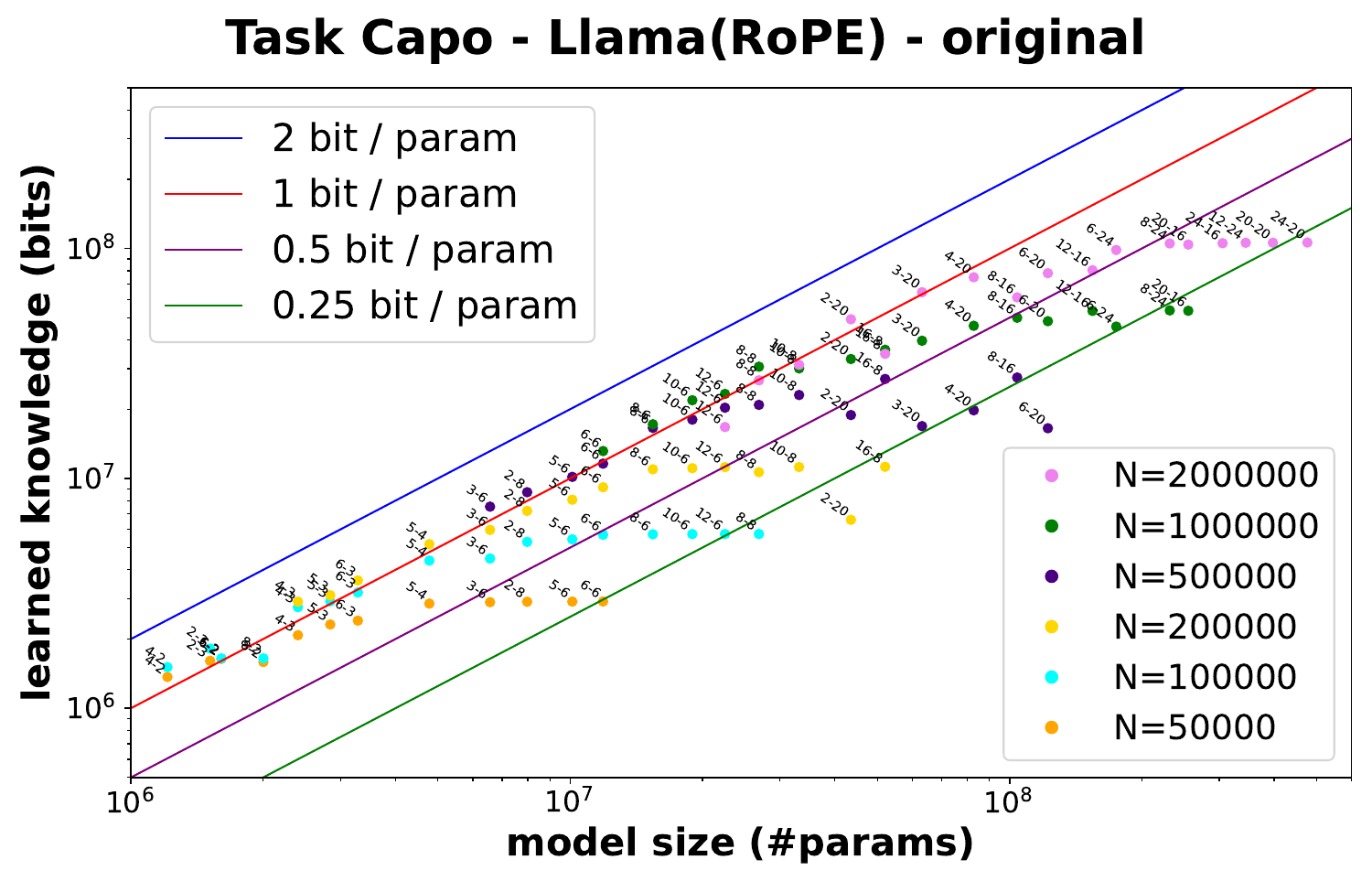}
\includegraphics[page=1,trim={2.5mm 1.5mm 2.5mm 1.5mm},clip,width=\imgwidthBase]{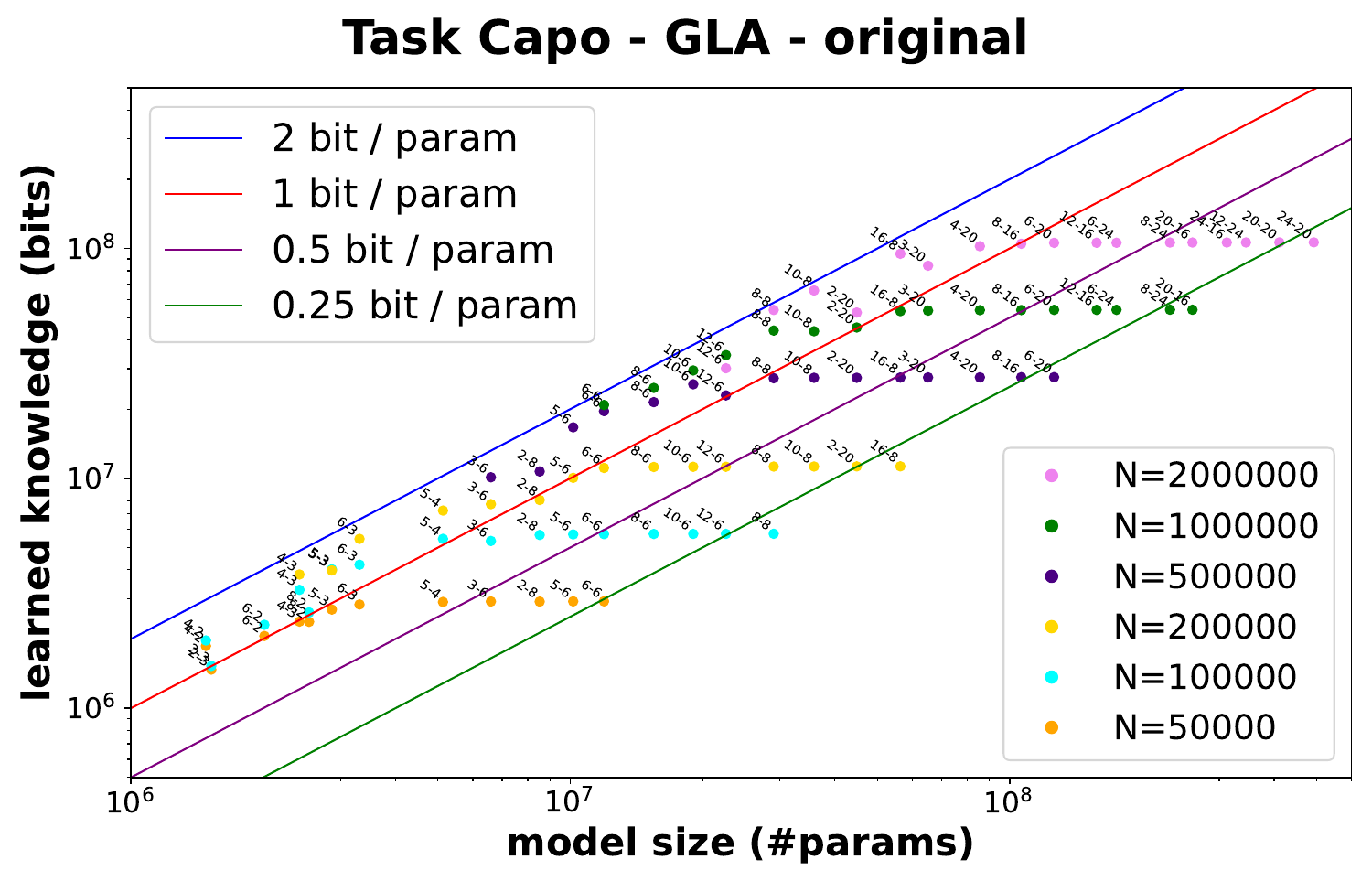}
\includegraphics[page=1,trim={2.5mm 1.5mm 2.5mm 1.5mm},clip,width=\imgwidthBase]{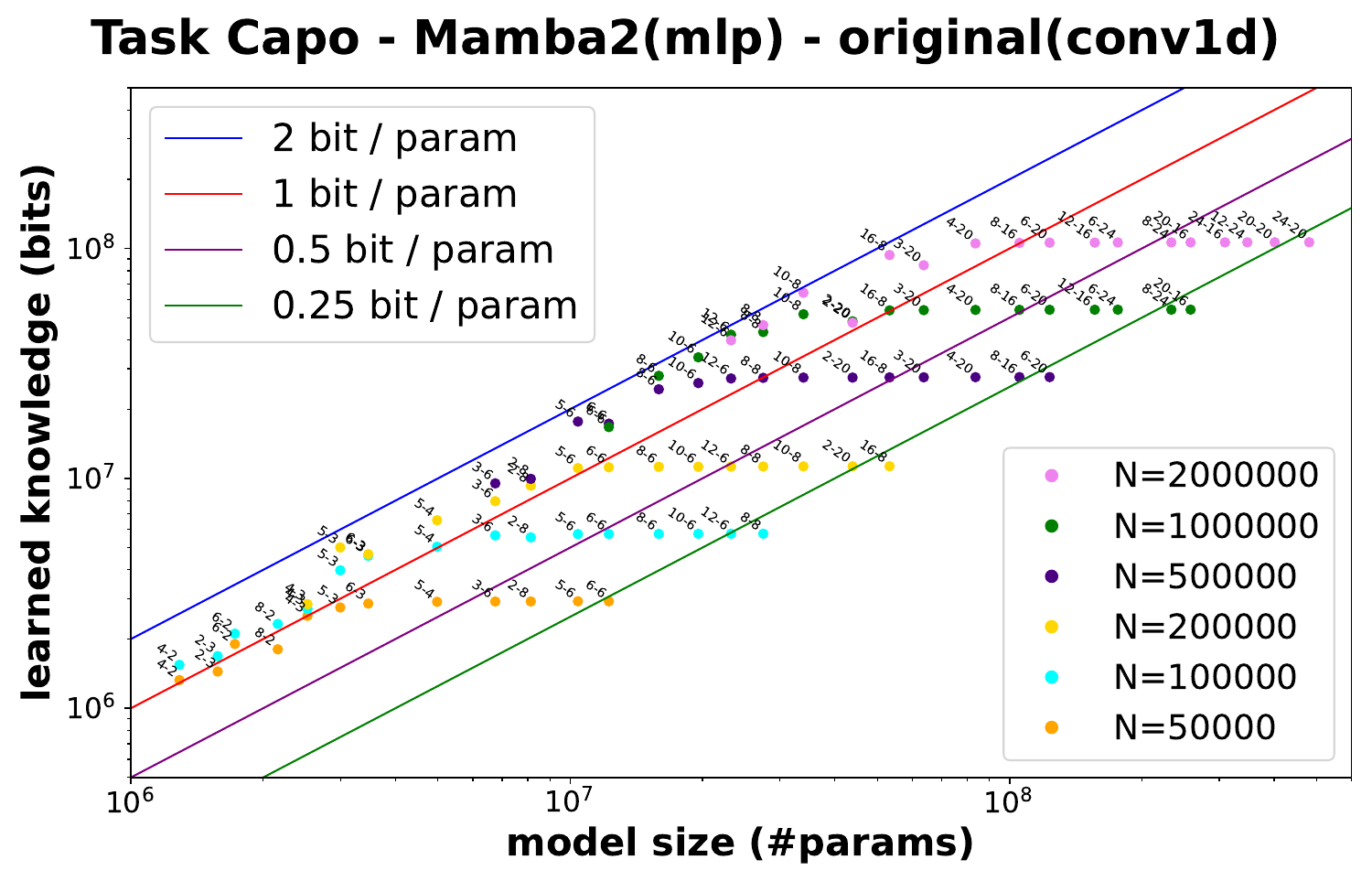}
\includegraphics[page=1,trim={2.5mm 1.5mm 2.5mm 1.5mm},clip,width=\imgwidthBase]{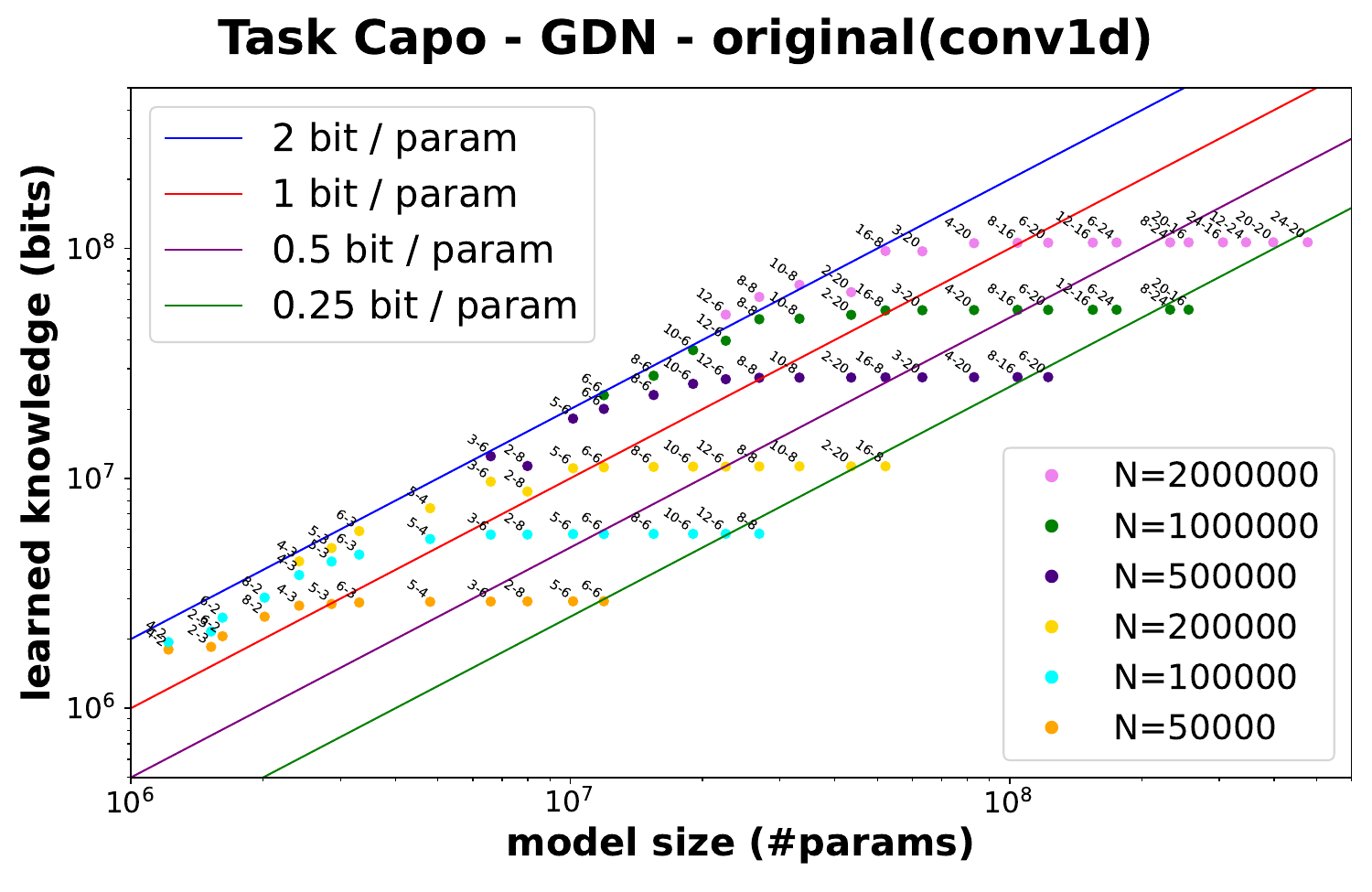}
\\
\includegraphics[page=1,trim={2.5mm 1.5mm 2.5mm 1.5mm},clip,width=\imgwidthBase]{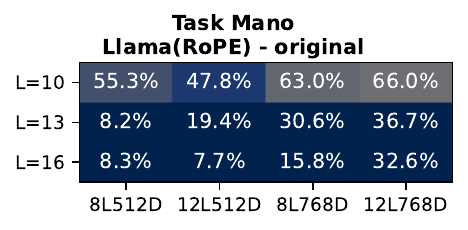}
\includegraphics[page=1,trim={2.5mm 1.5mm 2.5mm 1.5mm},clip,width=\imgwidthBase]{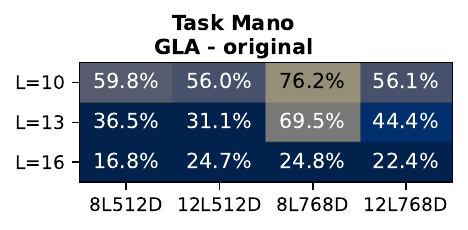}
\includegraphics[page=1,trim={2.5mm 1.5mm 2.5mm 1.5mm},clip,width=\imgwidthBase]{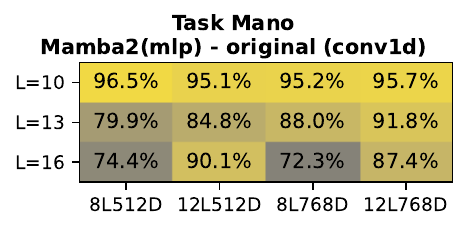}
\includegraphics[page=1,trim={2.5mm 1.5mm 2.5mm 1.5mm},clip,width=\imgwidthBase]{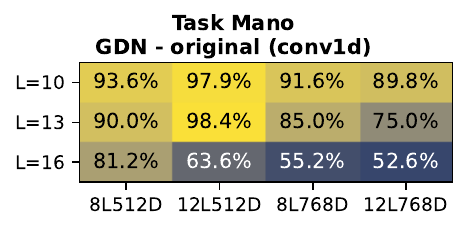}
\\
\includegraphics[page=1,trim={2.5mm 1.5mm 2.5mm 1.5mm},clip,width=\imgwidthBase]{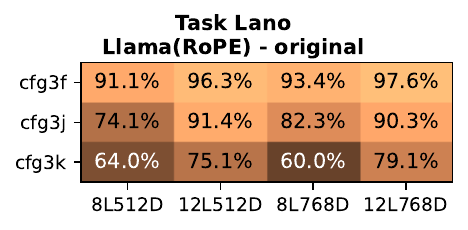}
\includegraphics[page=1,trim={2.5mm 1.5mm 2.5mm 1.5mm},clip,width=\imgwidthBase]{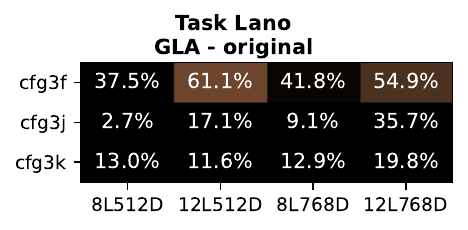}
\includegraphics[page=1,trim={2.5mm 1.5mm 2.5mm 1.5mm},clip,width=\imgwidthBase]{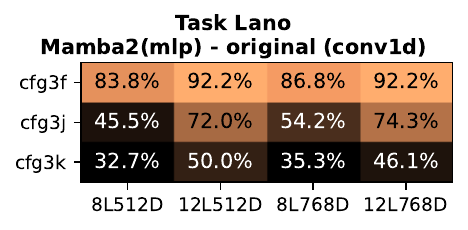}
\includegraphics[page=1,trim={2.5mm 1.5mm 2.5mm 1.5mm},clip,width=\imgwidthBase]{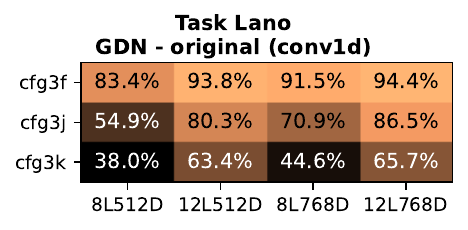}
\caption{\label{fig:initial}\textbf{Initial comparison of base models on five synthetic tasks.} GLA performs weakest; Mamba2(mlp) excels in knowledge (\textsc{Capo}, \textsc{Mano}); GDN strengthens reasoning and surpasses Llama(RoPE) on \textsc{Brevo} (reasoning breadth), while RoPE remains best on \textsc{Depo}+\textsc{Lano} (depth and structural reasoning). These results confirm our synthetic playground \textbf{as effective} for architectural comparison, but adding Canon layers (see later) will build a ``Pisa tower''—enabling controlled, fair comparisons where the landscape \textbf{shifts drastically} and reasoning depth improves 2–4×.}
\end{figure}

\section{Initial Comparison on Well-Known Base Architectures}
\label{sec:init-compare}

Language model architectures have evolved significantly since Transformers~\cite{vaswani2017attention}, giving rise to three major families distinguished by their computational mechanisms.

\emph{Quadratic-time attention} models include BERT~\cite{kenton2019bert} and GPT-2~\citep{radford2019language}. Refinements such as Rotary Position Embeddings (RoPE)~\cite{su2021roformer,gpt-neox-20b} and gated MLPs~\cite{shazeer2020glu} define their modern variants. We use the HuggingFace implementation of Llama, denoted \bblue{Llama(RoPE)}, which includes both refinements, and \bblue{Llama(NoPE)}, which omits positional embeddings. When clear, we refer to them as RoPE and NoPE.
Relative positional embeddings (e.g., \cite{he2020deberta}) are omitted due to limited empirical benefit but added computational cost~\cite{AL2023-cfg}.

RoPE models often generalize poorly beyond training context lengths, whereas NoPE generalizes better but achieves lower overall performance. Recent attention-score variants such as \bblue{ALiBi}~\cite{press2021train} and \bblue{Hard-ALiBi}~\cite{jelassi2024repeat} partially mitigate this, and we shall investigate them closely in this paper.

\emph{Linear-time attention} reduces computation by compressing sequences into fixed-length representations. Notable architectures include Linformer~\cite{wang2020linformer}, Performer~\cite{choromanski2020rethinking}, and Linear Transformer~\cite{katharopoulos2020transformers}. We focus on more recent Gated Linear Attention (\bblue{GLA})~\cite{yang2023gated} for its efficiency and scalability.

\emph{Recurrent and state-space models (SSM)} process long sequences via evolving hidden states rather than full attention. Mamba~\cite{gu2023mamba,dao2024transformers} exemplifies this family; we study its 2nd generation (\bblue{Mamba2}). Another key model is Gated DeltaNet (\bblue{GDN})~\cite{yang2024gated}, which we also analyze. Other notable variants include S4~\cite{gu2021efficiently}, S5~\cite{smith2022simplified}, RetNet~\cite{sun2023retentive}, RWKV~\cite{peng2023rwkv}, HGRN~\cite{qin2023hierarchically}, GSA~\cite{zhang2024gated}, and DeltaNet~\cite{yang2024parallelizing}.

\parhead{Exclusion of hybrid architectures}
We omit hybrid models integrating attention with linear or state-space mechanisms—e.g., Griffin~\cite{de2024griffin}, Samba~\cite{ren2024samba}, GDN-H1/H2~\cite{yang2024gated}, or sliding-window attention—to preserve clarity. Although such hybrids may excel on long contexts (up to 1M tokens), our focus is precision within standard windows (e.g., 4096 tokens). In practice, long contexts are often compressed (e.g., via CoTs) for detailed reasoning, making local precision the key concern.

Hybrids can \emph{obscure architectural trade-offs}, as aggregated results blur the contributions of individual modules. For instance, Mamba2 performs well on memory tasks but underperforms on structured reasoning; hybridization may conceal such contrasts. To ensure transparency, we analyze isolated \emph{base} architectures to reveal their intrinsic strengths and weaknesses.

Notably, Falcon-H1~\cite{tiifalconh1} (May 2025, 32B) combines Mamba2 with full attention, while Qwen3-Next~\cite{Qwen3next} (Sep 2025, 80B) combines GDN with full attention. These recent hybrids validate our choice of Mamba2 and GDN as representative base linear models.

\parhead{Architecture size standardization}
To ensure fair comparison, we standardize model sizes and evaluate Llama, GLA, Mamba2, and GDN as representatives of their respective families.

For all tasks except \textsc{Capo}, we test four sizes: Llama models with 12 or 8 layers and hidden dimensions of 768 or 512 (12 or 8 heads), denoted \texttt{12L768D}, \texttt{12L512D}, \texttt{8L768D}, and \texttt{8L512D}. (\texttt{12L768D} matches GPT-2-small.) These configurations are \emph{translated} to GLA, Mamba2(mlp), and GDN for comparable parameter counts.\footnote{See \appendixref{app:arch} for details. Briefly, with hidden size $d$, GLA follows the $4d^2 + 8d^2$ design (linear attention $4d^2$, MLP $8d^2$), while Mamba2(mlp) and GDN use $6d^2 + 6d^2$. We also test Mamba2 without MLP, reported separately in the appendix and referred to as Mamba2.}

For \textsc{Capo} (bit-per-parameter knowledge capacity), we vary model and data scales more broadly. Following~\cite{AL2024-knowledgeScaling}, model size is denoted $\ell$-$h$: for Llama, $\ell$ layers, hidden size $64h$, and $h$ heads. This notation extends consistently to GLA, Mamba2, and GDN (see \appendixref{app:arch}).

\parhead{Training}
All architectures share identical training settings (batch size, steps, learning rate, etc.) to ensure fairness. Full details appear in \appendixref{app:all-tasks}. Random seeds are fixed so that all models pre-train on identical data sequences.

\subsection{Initial Comparison Results}
\label{sec:init-compare:compare}

From \figureref{fig:initial}, linear-attention GLA performs weakest overall. Mamba2 excels on knowledge tasks (\textsc{Capo}, \textsc{Mano}) but lags in reasoning. GDN improves Mamba2’s reasoning and occasionally surpasses Llama(RoPE) on certain reasoning tasks (e.g., \textsc{Brevo}), though not others. These patterns align with real-world observations on natural data, supporting the validity of our synthetic playground. We defer deeper interpretation, as both Llama and GLA later prove to lack a critical architectural component—\bblue{making this comparison incomplete and partially unfair}.

For now, we highlight several \emph{key remarks}.

\parhead{3×4 mini scaling laws}
Randomness can affect outcomes, especially on hard tasks where \emph{grokking emerges}. In \textsc{Mano}, even with two seeds and four learning rates, smaller models sometimes outperform larger ones. This reflects staged reasoning: a model must learn $k$-hop reasoning (e.g., \texttt{Mano}, \texttt{Depo}) before advancing to $k\!+\!1$, and the transition often depends on random training dynamics. To reduce such variance, we test all tasks across \emph{three} data scales and \emph{four} model sizes (more for \textsc{Capo}). These ``3×4'' mini scaling laws yield more stable and interpretable comparisons.

\parhead{Benefits of synthetic tasks}
Synthetic tasks clarify architectural differences starkly (e.g., 90\% vs 5\%), clearly exposing strengths and weaknesses. By contrast, real-world experiments often produce modest differences (e.g., 2\%) buried in noise. Thus, synthetic pretraining environments allow clean evaluations of architectures' scalability and true capabilities.

\parhead{Interpreting task failures}
If a specific architecture (of a given size) fails at a certain difficulty level (e.g., large $N$ or $k$), it does not imply the model cannot learn the skill given infinite training. Our comparison uses a fixed, limited training budget: all architectures train for the same number of steps with identical data and shuffling, reporting best accuracy across multiple learning rates. Thus, results should be seen as differences in the \emph{speed of skill acquisition}, not absolute capability.\footnote{Faster learning is practically important—for example, a model ideally learns reasoning skills quicker than pure memorization. Similar observations arise in knowledge capacity tasks~\cite{AL2024-knowledgeScaling}, where architectural differences vanish with ample training but become pronounced when training budgets are limited.}

\parhead{Predicting future pipelines}
Synthetic tasks simulate idealized, high-quality pretraining conditions targeting core skills like multi-hop reasoning (\textsc{Depo}). Unlike datasets such as FineWeb-Edu or SlimPajama, which contain sparse reasoning examples obscured by simpler content, synthetic tasks highlight core capabilities. Currently, 100B-token pretraining fails even the simplest 2-hop reasoning (\resultref{res:12}). As training pipelines evolve---via improved data curation or RL-based post-training---synthetic tasks like \textsc{Depo} may better predict models' potential and guide architectural choices.

\section{Canon Layers: Enhancing Horizontal Information Flow}
\label{sec:canon-description}

Attention-based Transformers are widely recognized for their ability to perform associative recall—e.g., predicting \texttt{?} in the sequence \texttt{[A] [B] ... [A] [?]} where \texttt{?} = \texttt{[B]}. One might expect the second \texttt{[A]} could simply attend to the first to retrieve \texttt{[B]}, but causal masking makes this impossible: the first occurrence of \texttt{[A]} sees no future tokens. Accurate recall thus ``requires'' two attention layers—the first copies the first \texttt{[A]} into its neighbor \texttt{[B]}; the second uses this enriched representation, querying by \texttt{[A]} to retrieve value = \texttt{[B]} (via key = \texttt{[A]}). Using global attention just to pass information between adjacent tokens is, in effect, \emph{shooting a bird with a cannon}.%

\begin{figure}[t!]
\centering
\vspace{-3mm}
\begin{minipage}[t]{0.4\textwidth}
  \vspace{0pt}    \boxed{\includegraphics[page=5,trim={2mm 132mm 192mm 2mm},clip,width=0.98\textwidth]{plots}}
\end{minipage}
\hspace{5mm}
\begin{minipage}[t]{0.5\textwidth}
  \vspace{0pt}    \includegraphics[page=1,trim={2.5mm 1.5mm 2.5mm 1.5mm},clip,width=\textwidth]{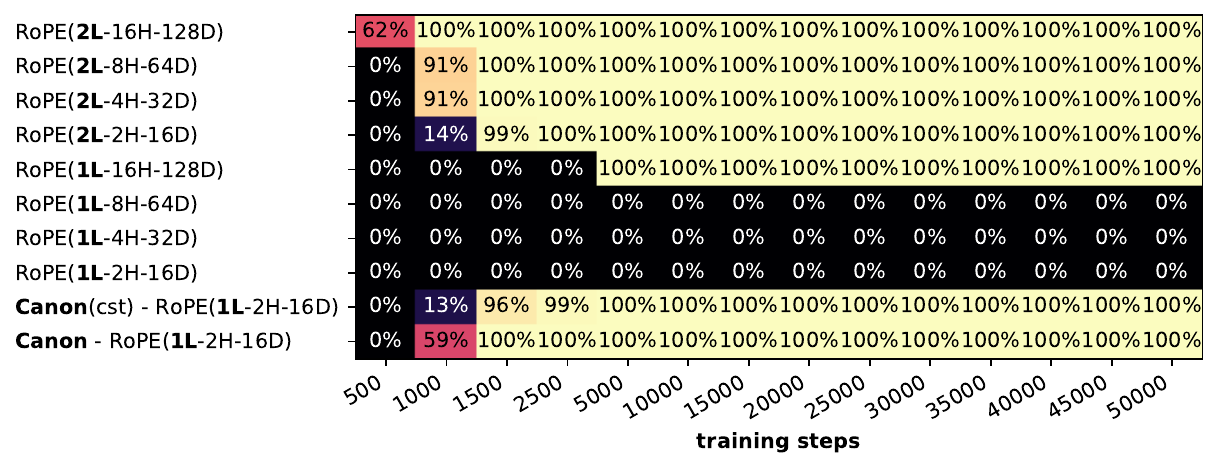}
\end{minipage}
\caption{\label{fig:copy}\textbf{A trivial token-copying experiment for 500 tokens}, added for completeness. 1-layer RoPE requires $d \geq 128$, while 2-layer RoPE or 1-layer RoPE + Canon achieves 100\% with $d=16$.}
\end{figure}

\begin{remark}
This is not a strict lower bound. A 1-layer Transformer is Turing-complete and can perform recall by blindly aggregating most (or all) context into one position, allowing the MLP to do local query/key/value computations. But this is inefficient: \figureref{fig:copy} shows that a 1-layer Transformer needs hidden size 128 to recall length-500 sequences, while 2 layers succeed with size 16.
\end{remark}

\parhead{The importance of local context}
Even simple tasks like token recall require careful mixing of local context—\emph{not to mention} more complex ones or when words span multiple tokens. Since MLP layers don’t mix tokens, attention must handle all communication.
Rotary and relative positional encodings help by biasing attention toward nearby tokens, but they remain tied to attention and still ``shoot birds with cannons.'' Similar issues arise in GLA~\cite{yang2023gated} and Mamba2, where recent-token information must be retrieved via compression mechanisms not optimized for local detail.

\parhead{Canon layers: general form}
Inspired by (vertical) residual connections, we introduce \emph{Canon layers} to enhance horizontal information flow across neighboring tokens. Canon layers aggregate nearby hidden states into the current position, enabling lightweight local mixing within a fixed window (e.g., size 4), unlike attention-based global aggregation or recurrent compression.

Formally, for any hidden states $h_t \in \mathbb{R}^m$ at token position $t$, a Canon layer computes:
$$
h'_t = w_0 \odot h_t + w_1 \odot h_{t-1} + w_2 \odot h_{t-2} + w_3 \odot h_{t-3},
$$
where $\odot$ denotes element-wise multiplication, $w_i \in \mathbb{R}^m$ ($i=0,1,2,3$) are weights, and padding zeros are used for boundary conditions. We call this \textit{Canon}, borrowing from the musical term, as it resembles melodies played sequentially at fixed temporal delays.\footnote{In Pachelbel's Canon in D, violins sequentially play the same melody with delays, creating overlapping horizontal repetition patterns analogous to Canon layers.}

\parhead{Flexible Integration}
Canon layers integrate at multiple points within each Transformer block:
\begin{itemize}
    \item \emph{Canon-A}: Before the attention block ($m=d$ if hidden size is $d$), after RMSnorm.
    \item \emph{Canon-B}: Inside the attention block, applied after Q/K/V projections ($m=3d$).
    \item \emph{Canon-C}: Before the MLP block ($m=d$), after RMSnorm.
    \item \emph{Canon-D}: Within MLP ($m=4d$ for standard, $m=\tfrac{16}{3}d$ for gated MLP), before activation.
\end{itemize}
Combining all four points gives \textit{Canon-ABCD} (full-score Canon); partial combinations (Canon-A/B/ABC) can also be explored.
Canon layers integrate flexibly across diverse architectures, including linear-attention and state-space models. For Mamba2 (without standard MLP layers), Canon layers appear at Canon-A and Canon-B positions (yielding Canon-AB); for Mamba2(mlp), the complete Canon-ABCD applies. Canon-B in Mamba2 scales as $m=4d+o(d)$.\footnote{For example, Mamba2 settings with \texttt{ssm\_state\_size=64}, \texttt{num\_heads=16} result in $m=4d+144$ dimensions.}

\begin{figure}[t!]
\centering
{\includegraphics[page=6,trim={3mm 93mm 2mm 0mm},clip,width=0.99\textwidth]{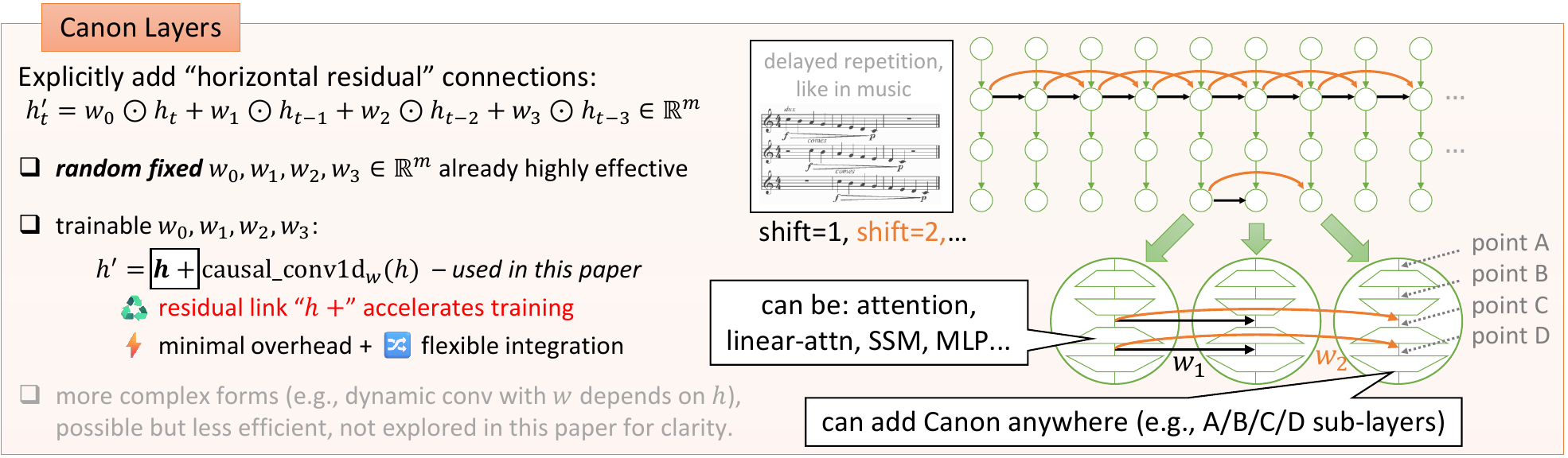}}
\caption{\label{fig:illustrate4}\textbf{Illustration of Canon layers}.}
\end{figure}

\parhead{Canon layers: Implementation variants}
Canon layers can be implemented in many ways. Even a simple version with \textbf{fixed, random weights}—aggregating the past three tokens as \emph{horizontal residual links}—\textbf{already notably enhances performance} (\figureref{fig:gpt} on Page~\pageref{fig:gpt}).\footnote{Unlike vertical residual links ($h' = h + \sigma(\mathbf{W}h)$), Canon layers aggregate multiple token vectors from different relative positions ($t-1$, $t-2$, $t-3$). Assigning fixed orthogonal directions effectively provides each position a unique ``ID'' for aggregation. Simple scalar weighting (e.g., $h'_t = h_t + 0.4 h_{t-1} + 0.2 h_{t-2} + 0.1 h_{t-3}$) can degrade performance.}
More complex variants—e.g., dynamic convolutions with input-dependent weighting—are possible but not studied here, as it remains unclear whether such additional cost is justified.

In this paper, for simplicity and efficiency, we implement Canon layers as a 1-d causal convolution with kernel size 4, available through efficient CUDA kernels implemented by the open-source H3 library (pip package \texttt{causal\_conv1d})~\cite{fu2022hungry}. We also incorporate explicit residual connections:
\begin{equation}\label{eqn:canon-res}
h'_t = h_t + \mathrm{conv1d}\bigl([h_t, h_{t-1}, h_{t-2}, h_{t-3}]\bigr)\enspace,
\end{equation}
denoted as Canon(res). Without residual connections, we denote it Canon(no-res). Minimal code changes (just a few lines) are needed for integration.
Even fully enabled (Canon-ABCD), Canon layers increase the parameter count minimally.\footnote{Fewer than 0.45\% parameters for GPT2-small. For a 1.3B-parameter Llama with Canon-ABCD enabled, parameters increase by 0.0063\%, runtime overhead on an H100 GPU with naive implementation (PyTorch bf16, flash attention, causal conv1d kernels) is 12.4\%, 14.1\%, and 20.8\% for forward, backward, and generation respectively. For Canon-AC, overheads reduce to 5.8\%, 5.8\%, and 7.0\%. Further runtime efficiencies are possible (e.g., consolidating multiple Canon operations across layers), though these optimizations remain beyond this paper's scope.} Our emphasis is on clearly demonstrating Canon layers' substantial performance benefits; detailed runtime optimizations remain future work.

\parhead{Related Work}
A precursor to Canon layers appears in~\cite{AL2023-cfg}, which studied uniform attention—i.e., averaging the past $k$ tokens—for $k \in \{1,2,4,8,\dots\}$ on CFG tasks. Surprisingly, this simple mixing outperformed GPT2 with absolute positional embeddings and closely approached GPT2(RoPE).%
\footnote{One ICML reviewer rejected the paper, commenting that the results were ``too surprising to be true.'' We invite curious readers to try it themselves—it really works.}
Canon layers generalize this idea: we apply learned, position-specific mixing over a short window (typically 4 tokens), removing value and projection matrices for better efficiency and modularity.

Our use of \texttt{causal\_conv1d} is inspired by Mamba~\cite{gu2023mamba,dao2024transformers} and GLA~\cite{yang2023gated}, which trace back to H3~\cite{fu2022hungry}, where the component was introduced as ``shift-SSM.'' After the initial release of our paper, we also became aware of Primer~\cite{so2109primer}, which proposes ``multi-dconv-head'' attention. These models apply \texttt{conv1d} (often with SiLU activation) within SSM or attention modules, without residual connections. In our terminology, these roughly correspond to Canon-B(no-res).

Our work generalizes and isolates this design as the Canon layer, and systematically evaluates its effect across all types of sequential models and sublayers (A/B/C/D). By studying Canon under \emph{controlled} synthetic pretraining, we can clearly attribute performance gains to the conv1d-based mixing mechanism, rather than to other architectural components such as attention or state-space recurrence. Moreover, we show that Canon layers are intrinsically \emph{not tied to attention or SSMs}—and in fact, may not benefit from being tightly coupled to them.

Convolutions have been used in Transformers for different goals. Conformer~\cite{gulati2020conformer} and CvT~\cite{wu2021cvt} integrate heavier convolutional modules for feature extraction in speech and vision. In contrast, Canon layers are lightweight and designed to enhance horizontal information flow—like horizontal ``residual links.'' Notably, even random-weight Canon layers yield substantial improvements.

Concurrent work on Multi-Token Attention (MTA)~\cite{golovneva2025multi} explores more complex 2D convolutional layers within attention heads. While MTA improves associative recall, it is heavier and more attention-specific. Investigating whether such designs offer further gains when combined with Canon, or whether Canon alone suffices for most settings, is an interesting direction for future work.

\begin{figure}[t!]
\centering
\setlength{\imgwidthBase}{0.193\textwidth}
\includegraphics[page=1,trim={2.5mm 1.5mm 2.5mm 1.5mm},clip,width=\imgwidthBase]{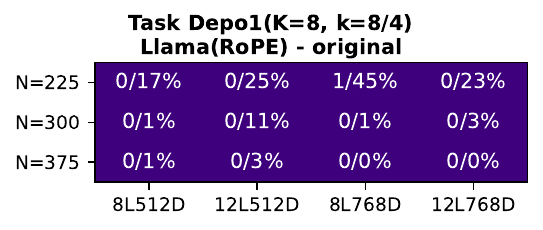}
\includegraphics[page=1,trim={2.5mm 1.5mm 2.5mm 1.5mm},clip,width=\imgwidthBase]{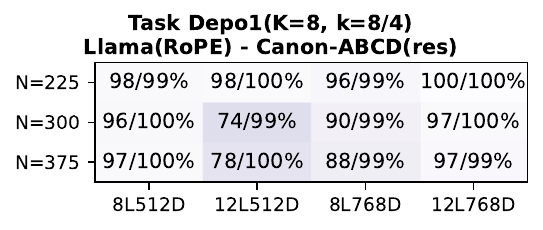}
\includegraphics[page=1,trim={2.5mm 1.5mm 2.5mm 1.5mm},clip,width=\imgwidthBase]{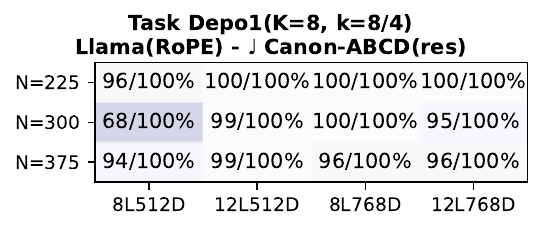}
\includegraphics[page=1,trim={2.5mm 1.5mm 2.5mm 1.5mm},clip,width=\imgwidthBase]{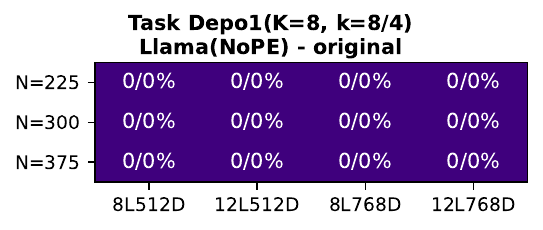}
\includegraphics[page=1,trim={2.5mm 1.5mm 2.5mm 1.5mm},clip,width=\imgwidthBase]{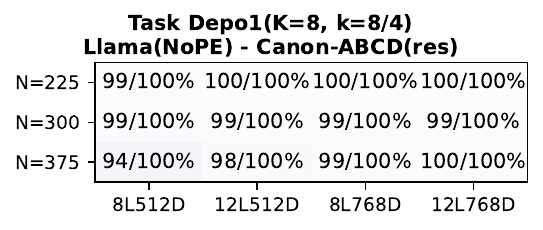}
\\
\includegraphics[page=1,trim={2.5mm 1.5mm 2.5mm 1.5mm},clip,width=\imgwidthBase]{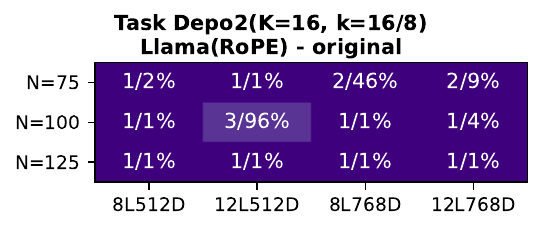}
\includegraphics[page=1,trim={2.5mm 1.5mm 2.5mm 1.5mm},clip,width=\imgwidthBase]{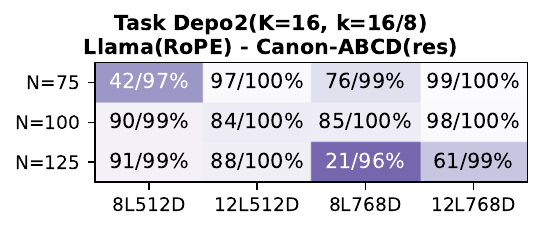}
\includegraphics[page=1,trim={2.5mm 1.5mm 2.5mm 1.5mm},clip,width=\imgwidthBase]{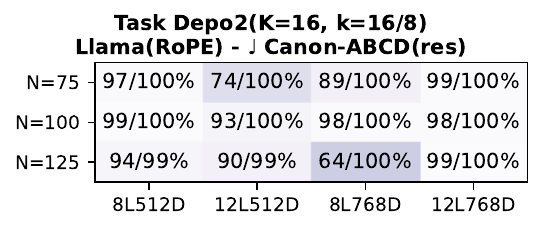}
\includegraphics[page=1,trim={2.5mm 1.5mm 2.5mm 1.5mm},clip,width=\imgwidthBase]{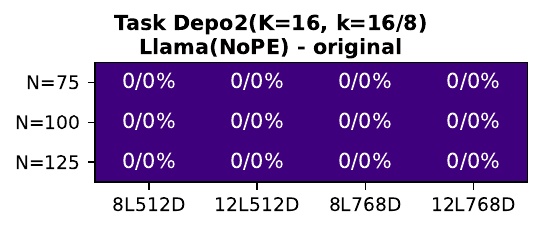}
\includegraphics[page=1,trim={2.5mm 1.5mm 2.5mm 1.5mm},clip,width=\imgwidthBase]{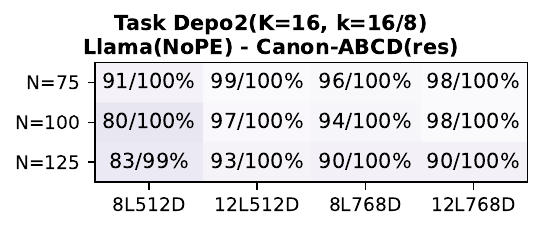}
\\
\includegraphics[page=1,trim={2.5mm 1.5mm 2.5mm 1.5mm},clip,width=\imgwidthBase]{top_sort/Llama_RoPE_-original}
\includegraphics[page=1,trim={2.5mm 1.5mm 2.5mm 1.5mm},clip,width=\imgwidthBase]{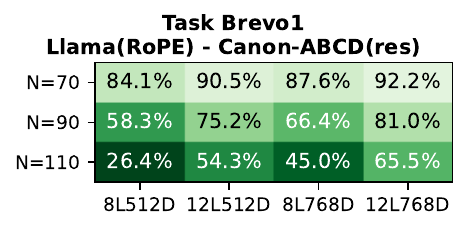}
\includegraphics[page=1,trim={2.5mm 1.5mm 2.5mm 1.5mm},clip,width=\imgwidthBase]{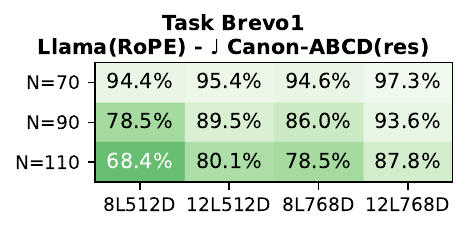}
\includegraphics[page=1,trim={2.5mm 1.5mm 2.5mm 1.5mm},clip,width=\imgwidthBase]{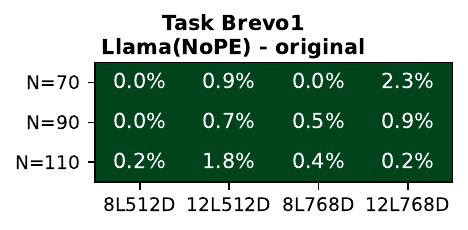}
\includegraphics[page=1,trim={2.5mm 1.5mm 2.5mm 1.5mm},clip,width=\imgwidthBase]{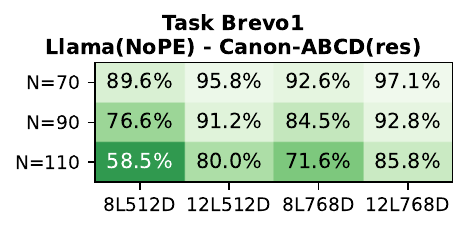}
\\
\includegraphics[page=1,trim={2.5mm 1.5mm 2.5mm 1.5mm},clip,width=\imgwidthBase]{top_sort_multi/Llama_RoPE_-original}
\includegraphics[page=1,trim={2.5mm 1.5mm 2.5mm 1.5mm},clip,width=\imgwidthBase]{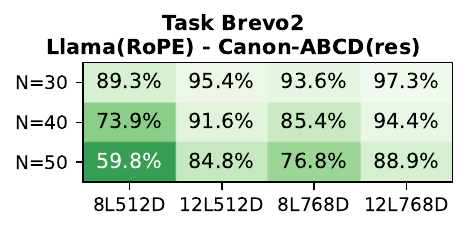}
\includegraphics[page=1,trim={2.5mm 1.5mm 2.5mm 1.5mm},clip,width=\imgwidthBase]{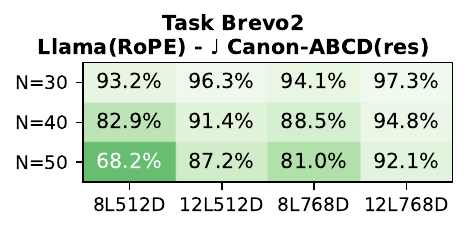}
\includegraphics[page=1,trim={2.5mm 1.5mm 2.5mm 1.5mm},clip,width=\imgwidthBase]{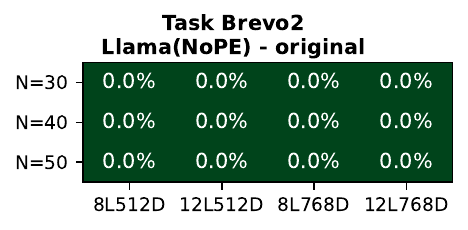}
\includegraphics[page=1,trim={2.5mm 1.5mm 2.5mm 1.5mm},clip,width=\imgwidthBase]{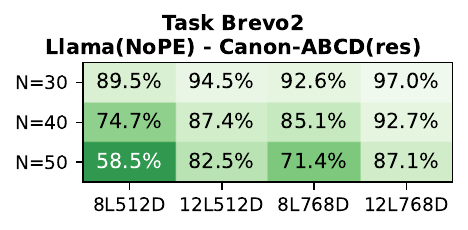}
\\
\includegraphics[page=1,trim={2.5mm 1.5mm 2.5mm 1.5mm},clip,width=\imgwidthBase]{biocap/Llama-original}
\includegraphics[page=1,trim={2.5mm 1.5mm 2.5mm 1.5mm},clip,width=\imgwidthBase]{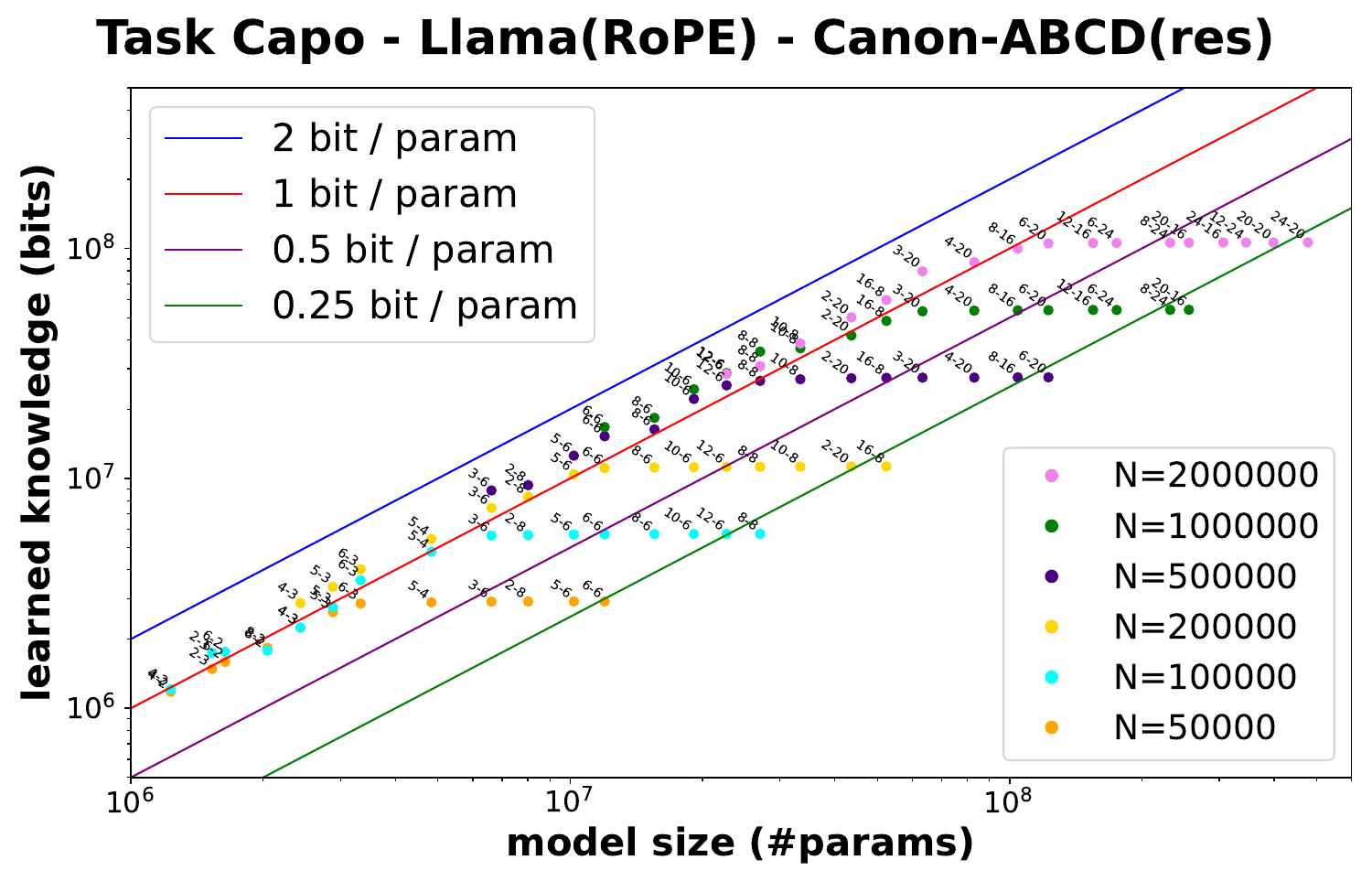}
\includegraphics[page=1,trim={2.5mm 1.5mm 2.5mm 1.5mm},clip,width=\imgwidthBase]{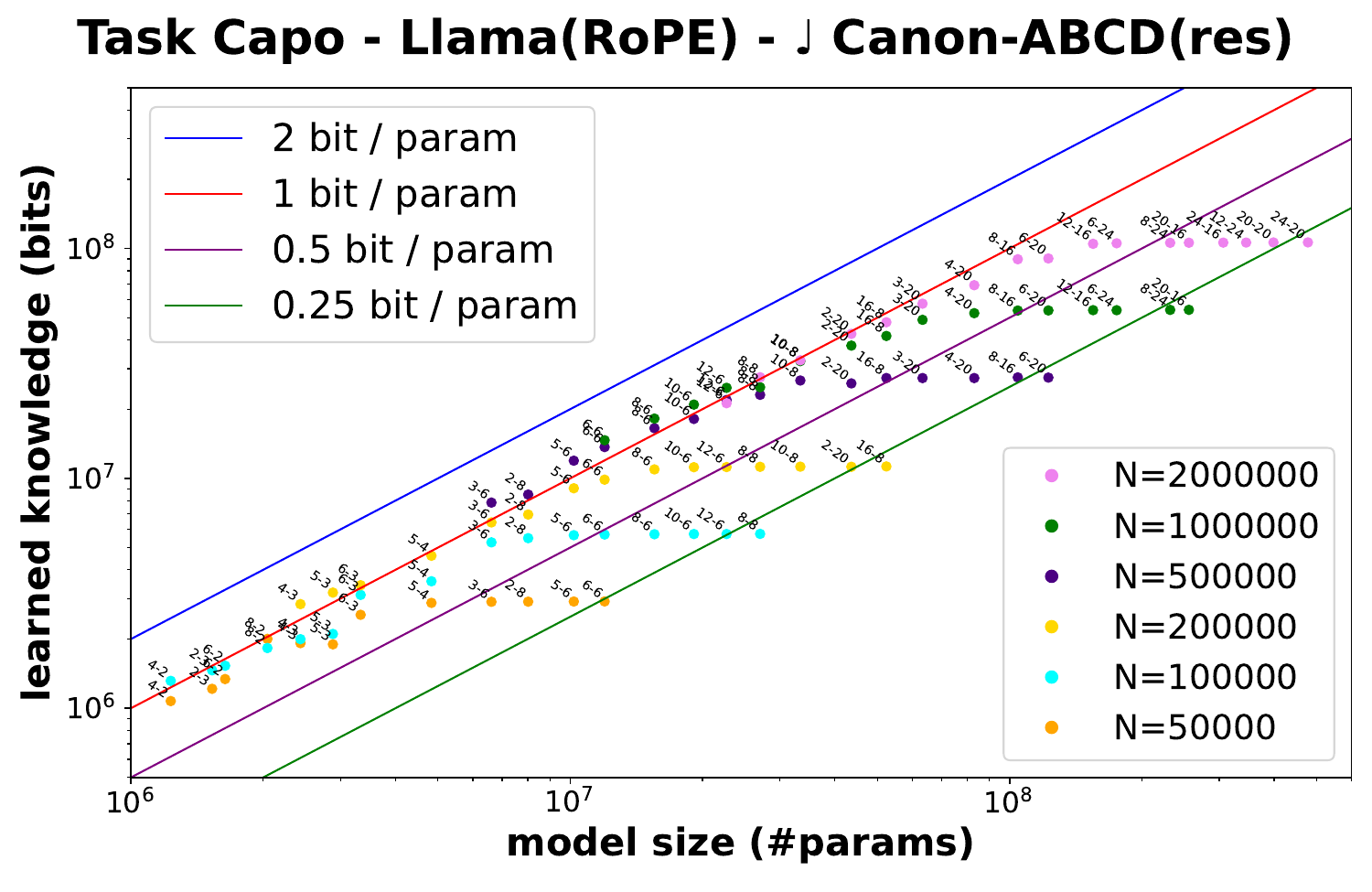}
\includegraphics[page=1,trim={2.5mm 1.5mm 2.5mm 1.5mm},clip,width=\imgwidthBase]{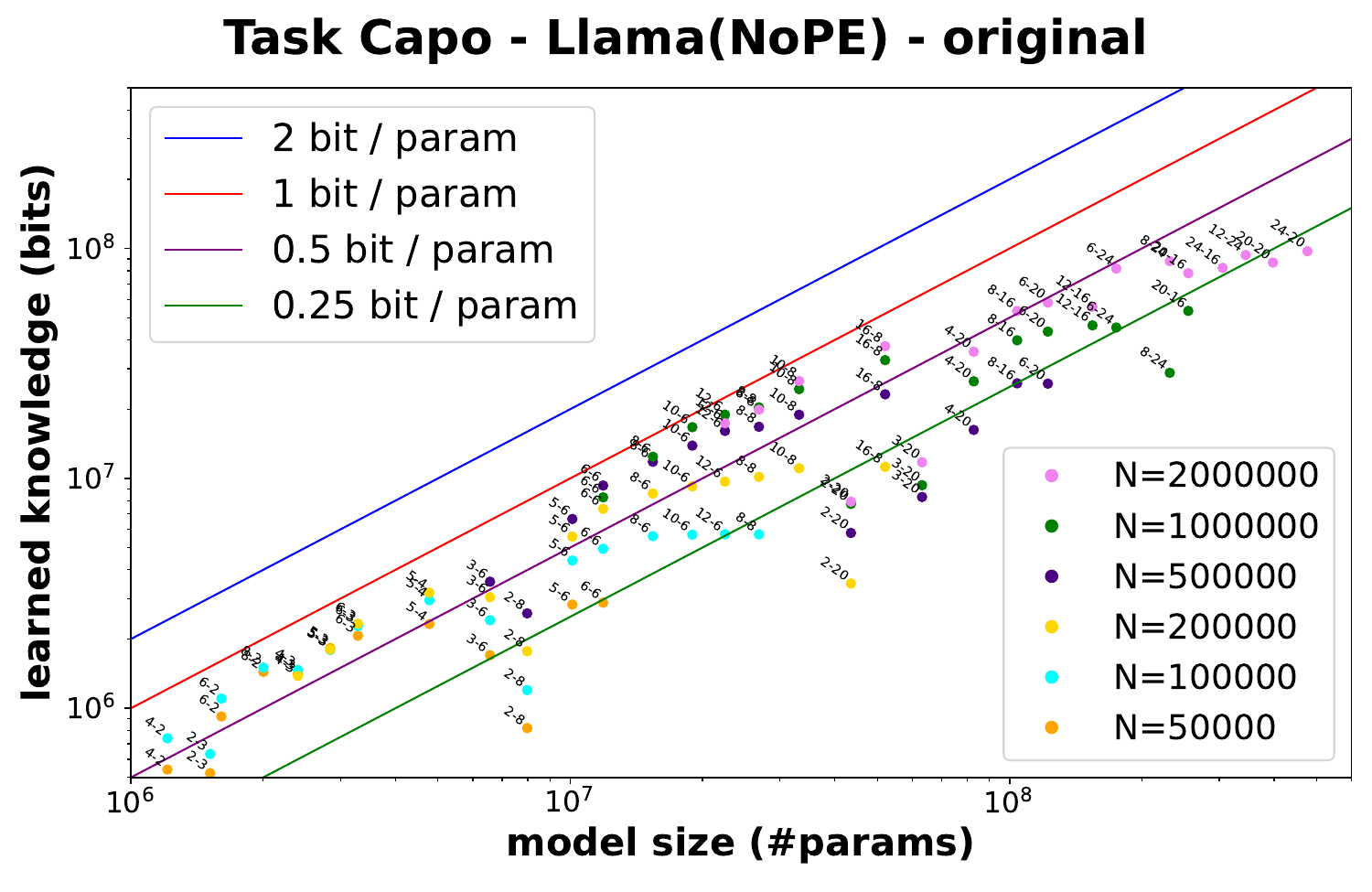}
\includegraphics[page=1,trim={2.5mm 1.5mm 2.5mm 1.5mm},clip,width=\imgwidthBase]{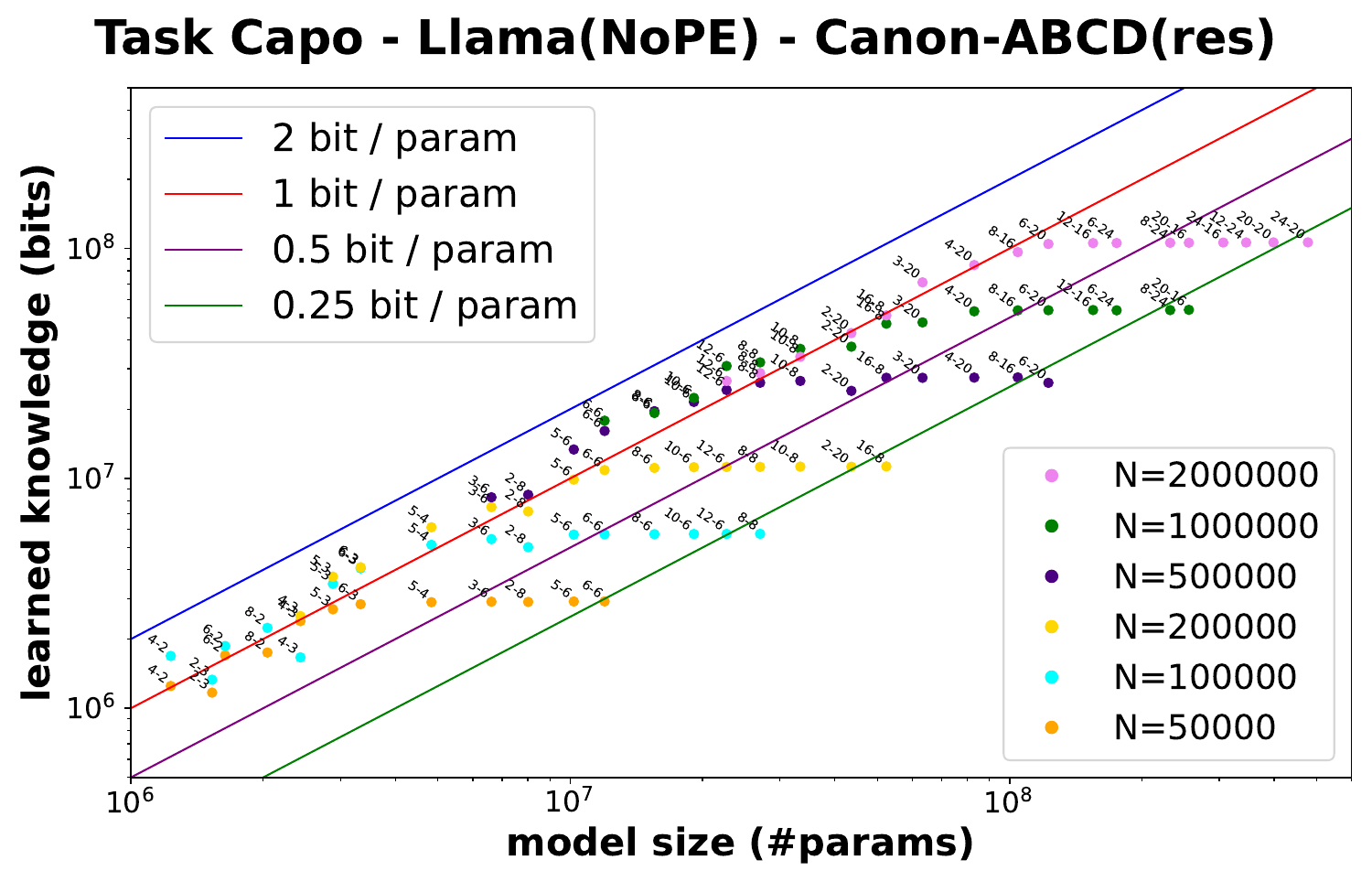}
\\
\includegraphics[page=1,trim={2.5mm 1.5mm 2.5mm 1.5mm},clip,width=\imgwidthBase]{arith/Llama_RoPE_-original}
\includegraphics[page=1,trim={2.5mm 1.5mm 2.5mm 1.5mm},clip,width=\imgwidthBase]{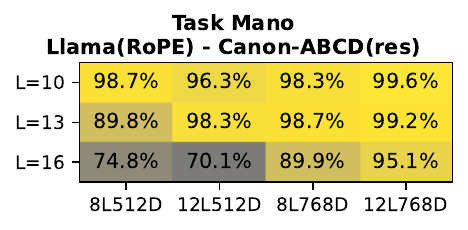}
\includegraphics[page=1,trim={2.5mm 1.5mm 2.5mm 1.5mm},clip,width=\imgwidthBase]{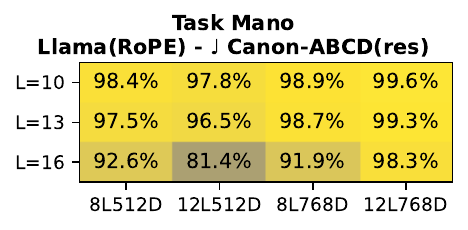}
\includegraphics[page=1,trim={2.5mm 1.5mm 2.5mm 1.5mm},clip,width=\imgwidthBase]{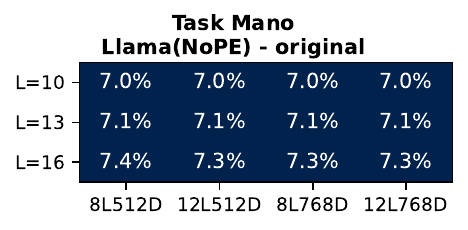}
\includegraphics[page=1,trim={2.5mm 1.5mm 2.5mm 1.5mm},clip,width=\imgwidthBase]{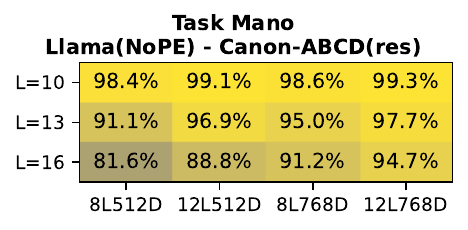}
\\
\includegraphics[page=1,trim={2.5mm 1.5mm 2.5mm 1.5mm},clip,width=\imgwidthBase]{cfg/Llama_RoPE_-original}
\includegraphics[page=1,trim={2.5mm 1.5mm 2.5mm 1.5mm},clip,width=\imgwidthBase]{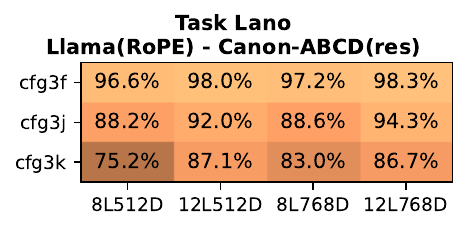}
\includegraphics[page=1,trim={2.5mm 1.5mm 2.5mm 1.5mm},clip,width=\imgwidthBase]{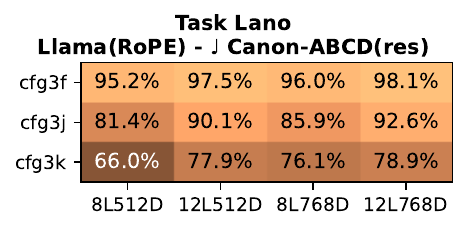}
\includegraphics[page=1,trim={2.5mm 1.5mm 2.5mm 1.5mm},clip,width=\imgwidthBase]{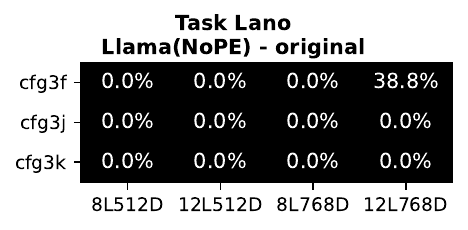}
\includegraphics[page=1,trim={2.5mm 1.5mm 2.5mm 1.5mm},clip,width=\imgwidthBase]{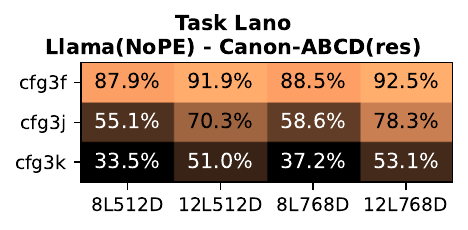}
\caption{\label{fig:trans-canon}\textbf{Column 1→2}: Canon layers dramatically enhance RoPE, improving reasoning depth by 2–4×.
\newline
\textbf{Column 4→5}:  Canon transforms NoPE into a strong performer on par with RoPE-based models.
\newline
\textbf{Column 2+5→3}: With Canon, RoPE usage can be reduced — RoPE + \musQuarter{}Canon (RoPE enabled for 1/4 dimensions) outperforms both RoPE/NoPE + Canon, \bblue{great news for length generalization!}
\newline
\sepline
\newline
\textit{Remark.} This figure uses \textsc{Depo1}(K=8) and \textsc{Depo2}(K=16). Earlier results in \figureref{fig:initial} were based on \textsc{Depo1}(K=4) and \textsc{Depo2}(K=4), because model performances were weaker.
}
\end{figure}

\section{When Transformer Meets Canon}
\label{sec:trans-canon}

\figureref{fig:initial}+\ref{fig:trans-canon} show that a 12-layer, 768-dimension Llama(RoPE) model trained on our ideal data can only handle 4-hop retrieval in contexts of length 2048. Can this be any better?

\subsection{RoPE with Canon Layers}
\label{sec:trans-canon:rope}

\begin{mdframed}
\begin{sresult}{2}[\figureref{fig:trans-canon} --- 1st vs. 2nd column]\label{res:2}
In our controlled playground, Canon layers (ABCD) introduce substantial improvements: with a 0.5\% increase in trainable parameters, reasoning depth of RoPE increases by 2-4×, reasoning breadth by 30\%, knowledge capacity by 10--15\%, knowledge manipulation length by 30\%, measurable gains in hierarchical language structure reasoning.
\end{sresult}
\end{mdframed}

\parhead{Task \textsc{Depo}}
In reasoning depth, RoPE pretrained on \textsc{Depo1}($K=8$)—covering $(k\le8)$--hop instances—achieves near-zero accuracy even at $k=4$, whereas RoPE+Canon-ABCD exceeds 50\% at $k=8$. On \textsc{Depo2}($K=16$)—a more challenging setup where each directed edge spans 10--14 tokens, far beyond a 4-token Canon window—RoPE completely fails, while RoPE+Canon-ABCD attains high accuracy at $k=16$. This demonstrates that Canon layers are not merely for single-token recall: by enriching local representations of multi-token segments, they empower the global attention to more effectively chain information across hops.\footnote{\textsc{Depo2} is designed so a 4-token window cannot resolve key–value pairs spanning 10–14 tokens, posing a substantial challenge even for Canon.}

These gains may seem surprising. For $k=1$ associative recall, one Canon\,+\,attention layer already does the job of two attention layers (recall \figureref{fig:copy}), so Canon saves at most one attention layer. Yet, \textbf{one extra layer \bblue{\emph{cannot explain}} a 2--4$\times$ gain} in reasoning depth. The answer lies in learning dynamics rather than representational power, as we explain in \sectionref{sec:trans-canon:why}.

\begin{figure}[t!]
\centering
\setlength{\imgwidthBase}{0.32\textwidth}
\vspace{-3mm}
\includegraphics[page=1,trim={2.5mm 1.5mm 2.5mm 1.5mm},clip,width=\imgwidthBase]{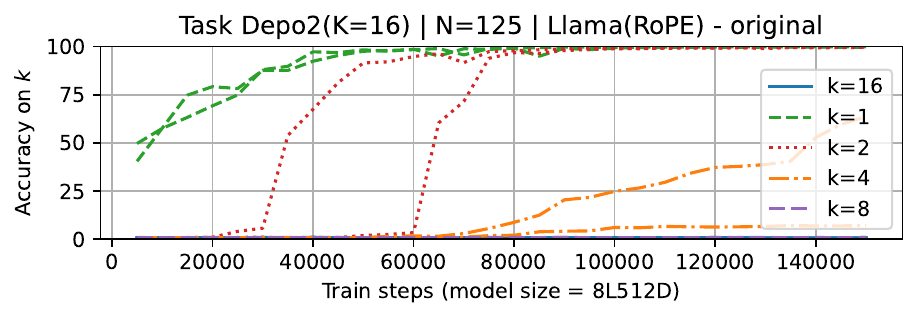}
\includegraphics[page=1,trim={2.5mm 1.5mm 2.5mm 1.5mm},clip,width=\imgwidthBase]{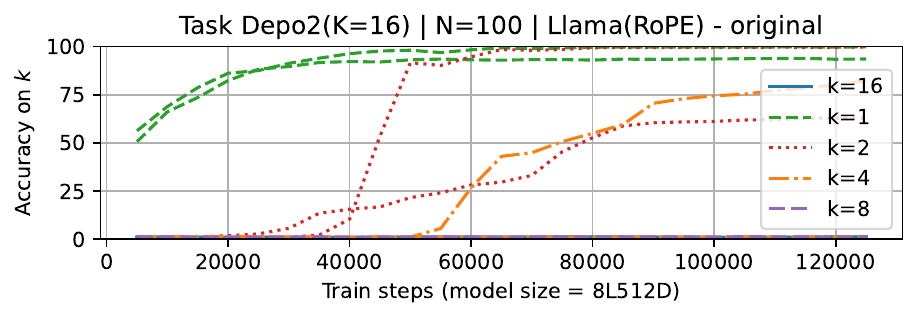}
\includegraphics[page=1,trim={2.5mm 1.5mm 2.5mm 1.5mm},clip,width=\imgwidthBase]{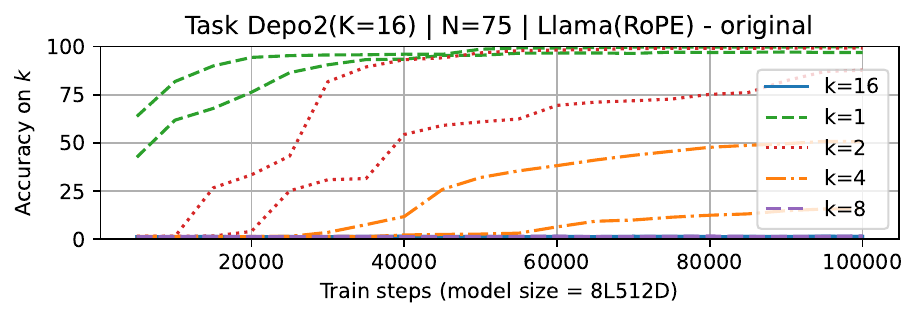}
\\
\includegraphics[page=1,trim={2.5mm 1.5mm 2.5mm 1.5mm},clip,width=\imgwidthBase]{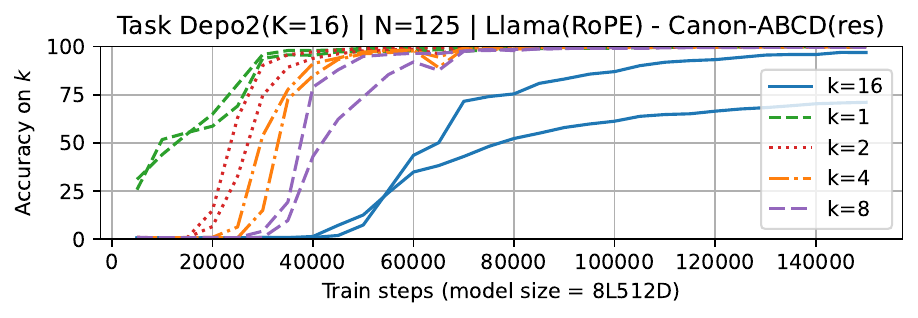}
\includegraphics[page=1,trim={2.5mm 1.5mm 2.5mm 1.5mm},clip,width=\imgwidthBase]{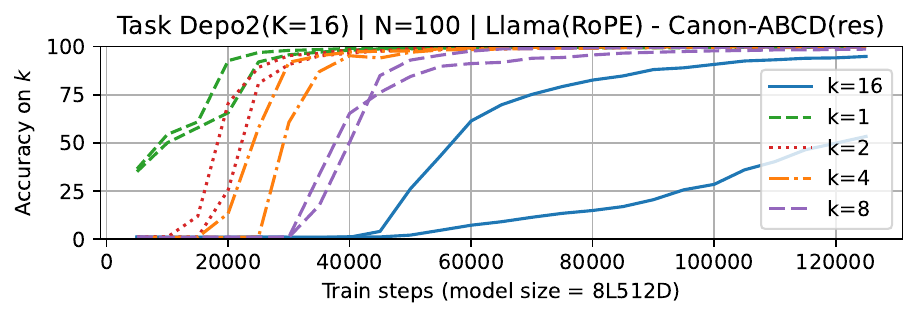}
\includegraphics[page=1,trim={2.5mm 1.5mm 2.5mm 1.5mm},clip,width=\imgwidthBase]{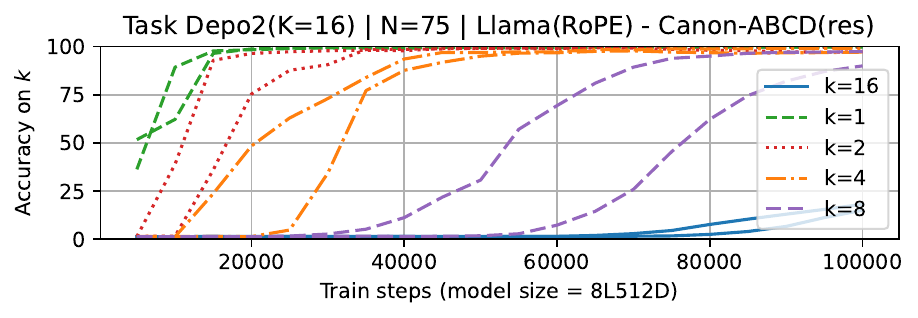}
\caption{\label{fig:depo2-curve}\textbf{Training curves for RoPE models w/+w/o Canon}, on \textsc{Depo2}($K=16$), evaluated at $k=1, 2, 4, 8, 16$ and maximum size $n=N$, shown in two best LRs. More model sizes/data are in \figureref{fig:depo2-curve:ext} on Page~\pageref{fig:depo2-curve:ext}.}
\end{figure}

\begin{figure}[t!]
\centering
\setlength{\imgwidthBase}{0.99\textwidth}
\includegraphics[page=1,trim={2.5mm 1.5mm 2.5mm 1.5mm},clip,width=\imgwidthBase]{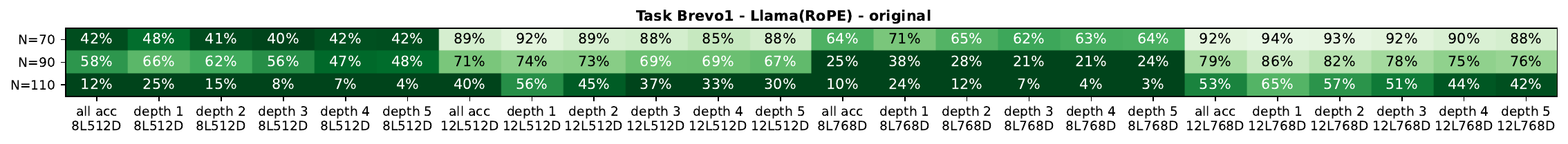}
\\
\includegraphics[page=1,trim={2.5mm 1.5mm 2.5mm 1.5mm},clip,width=\imgwidthBase]{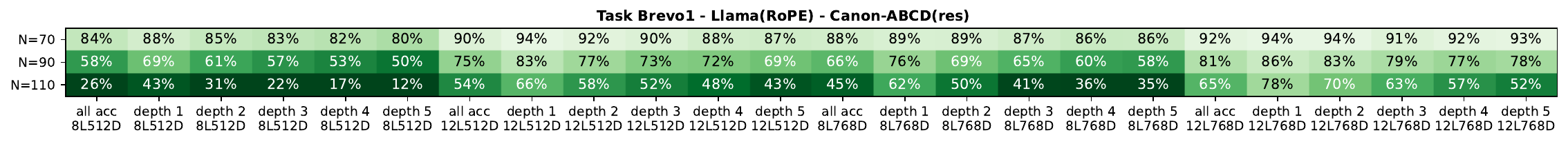}
\\
\includegraphics[page=1,trim={2.5mm 1.5mm 2.5mm 1.5mm},clip,width=\imgwidthBase]{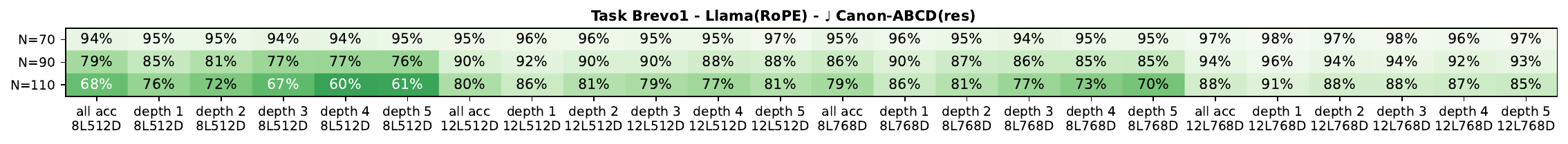}
\caption{\label{fig:breavo-depth}Detailed accuracies for Task \textsc{Brevo1}, shown overall and stratified by dependency graph depths $1, 2, 3, 4, 5$.}
\end{figure}

\parhead{Task Brevo}
On reasoning breadth, we observe a 30\% improvement by introducing Canon-ABCD. Specifically, the accuracies of RoPE to solve \textsc{Brevo1}($N=70$) or \textsc{Brevo2}($N=30$) resemble the performance of RoPE+Canon to solve \textsc{Brevo1}($N=90$) or \textsc{Brevo2}($N=40$). Since input length scales with $N$, this reflects a roughly 30\% increase in reasoning breadth.

To understand the source of this improvement, we analyze the accuracy across tasks stratified by the \emph{depth} of the dependency graph. Recall each query in \textsc{Brevo} requires the model to identify all vertices it recursively depends on, forming a sub-DAG of varying (minimum) depth. In \figureref{fig:breavo-depth}, we plot model accuracy not only overall but also separately for problem instances spanning DAG depths of $1, 2, 3, 4, 5$. The results show that vanilla RoPE struggles with instances involving greater DAG depth, whereas RoPE+Canon improves reasoning performance on deeper structures.
This suggests that Canon-ABCD enhances localized reasoning paths within Transformer blocks, allowing for better handling of recursive dependency, which can be challenging for standard attention alone.

\parhead{Task Capo} On knowledge capacity, prior work \cite{AL2024-knowledgeScaling} found that gated MLP layers in Llama(RoPE) reduce model capacity due to slower and less stable training dynamics. One remedy proposed in that work is to revert gated MLP back to standard MLP; however, this sacrifices reasoning capability (see \sectionref{sec:trans-canon:mlp}). Here, we present an alternative solution: adding Canon layers. Canon layers improve training speed and increase the effective capacity by 10–15\% in the controlled 100-exposure pretraining regime for \textsc{Capo}. On a separate note, GPT2(RoPE) models that originally employ standard MLP exhibit no capacity loss after Canon layers are introduced (\figureref{fig:moe-capo}).

\parhead{Task Mano} On knowledge manipulation, Canon layers increase manipulable length. RoPE+Canon matches the performance of vanilla RoPE on Mano($L=10$) when tested on Mano($L=13$), a 30\% improvement in length. This again stems from Canon layers accelerating hierarchical learning (\sectionref{sec:trans-canon:why}), enabling the model to scale from simpler tasks ($L=1$) to more complex ones ($L=2$, $L=3$, and beyond) faster. For simplicity, we omit the hierarchical learning speed visualization.

\parhead{Task Lano} Canon layers improve RoPE's performance on hierarchical language structure reasoning, though interpreting the gains requires some algorithmic background. For context, dataset \textsf{cfg3k} adds one level of structural complexity above \textsf{cfg3f}, using the same CFG rule distribution (see \appendixref{app:lano}). RoPE+Canon outperforms standard RoPE on \textsf{cfg3k}, but still struggles to fully handle this increased complexity. This is expected, as deeper CFG structures increase sequence length $n$ by 2--3×, and parsing these CFGs with dynamic programming involves worst-case time complexity $O(n^3)$. Consequently, \textsf{cfg3k} poses an arguably more than 8× greater computational challenge compared to \textsf{cfg3f}. Our intermediate dataset \textsf{cfg3j} has difficulty around 4×, suggesting RoPE+Canon can handle roughly twice as challenging structure-learning tasks compared to RoPE.

\parhead{Summary} Canon layers consistently improve performance across reasoning, knowledge and language tasks, all without introducing instability or accuracy trade-offs.

\subsection{Why Canon Helps: Hierarchical Feature Learning}
\label{sec:trans-canon:why}

\begin{figure}[t!]
\centering
{\includegraphics[page=10,trim={0mm 70mm 10mm 0mm},clip,width=0.99\textwidth]{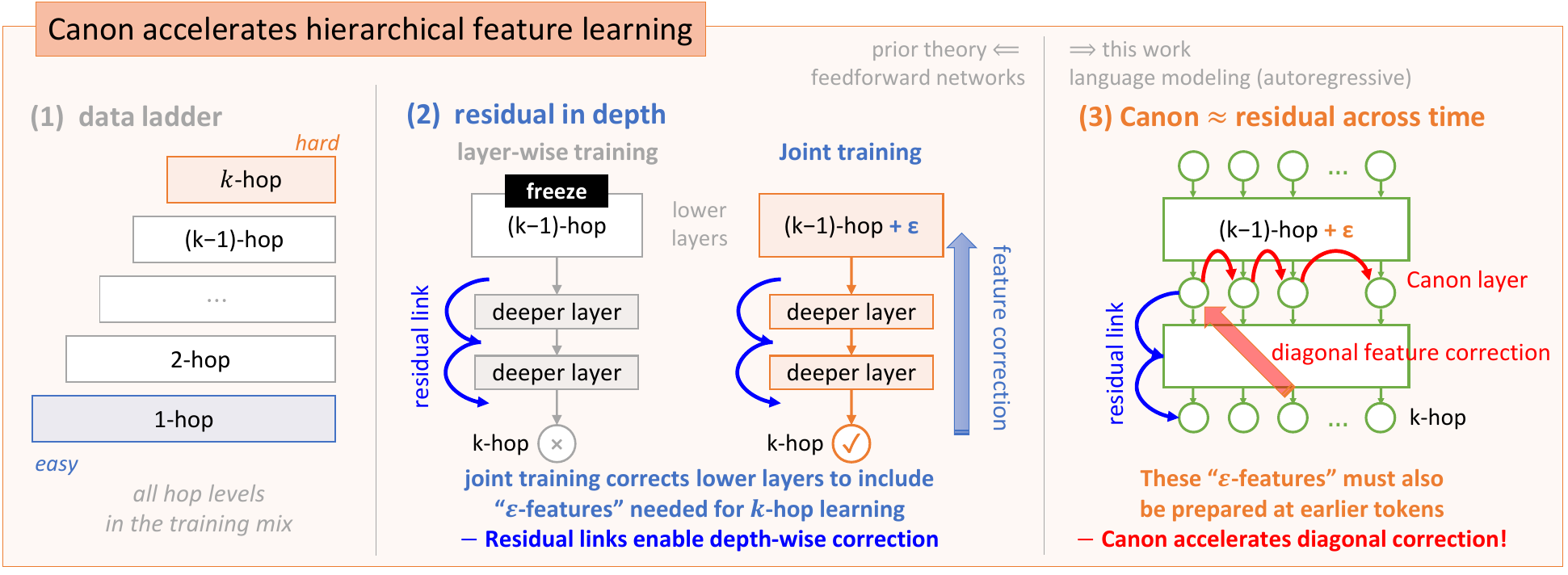}}
\caption{\label{fig:bfc-schematic}Canon accelerates hierarchical feature learning through diagonal backward feature correction.}
\end{figure}

\begin{mdframed}
\begin{sresult}{2.1}[\figureref{fig:depo2-curve}, \ref{fig:bfc-schematic}, \ref{fig:trans-ablation}]\label{res:2.1}
Where Canon's 2--4$\times$ reasoning-depth gain comes from:
\begin{itemize}
    \item \textbf{Not its first layer.} CanonFirst-ABCD recovers only a small fraction of the gain. (Fig.~\ref{fig:trans-ablation})
    \item \textbf{Faster hierarchical learning.} Canon layers act as residual links along the time horizon, letting backward feature correction travel \defem{diagonally} --- down layers and back in time. (Fig.~\ref{fig:bfc-schematic})
\end{itemize}
\end{sresult}
\end{mdframed}

Why is the 2--4$\times$ gain in \resultref{res:2} so large? Representationally, Canon acts only \emph{locally}: by making multi-token segments legible at shallow layers, it lets one attention layer perform associative recall that would otherwise take two (recall \figureref{fig:copy}). But this \emph{cannot explain a 2--4$\times$ increase} in reasoning depth. The answer must lie in \emph{how} deep reasoning is learned, not in what the architecture can represent.

\parhead{Canon is more than its first layer}
One might suspect Canon helps only because its first layer stitches together multi-token units --- say, an English word spelled across several tokens. Two observations refute this:
\begin{itemize}[nolistsep,topsep=2pt]
\item \emph{CanonFirst-ABCD} --- Canon in the first block \emph{only} --- recovers just a small fraction of full Canon-ABCD's gain (\figureref{fig:trans-ablation}). The contrast is sharpest on \textsc{Depo2}($K=16$), our deepest reasoning task (\figureref{fig:app:rope-ablation}), and sharper still for NoPE (\figureref{fig:app:nope-ablation}) --- foreshadowing \sectionref{sec:trans-canon:nope}: Canon can take over most of RoPE's old job of locating nearby tokens.
\item In some cases, Canon \textbf{does not improve} \textsc{Depo} at $k=1$ at all, but strongly improves $k=2,\dots,16$: the 1-hop learning curves are near-identical with and without Canon (\figureref{fig:depo2-curve}).\footnote{This does not contradict \figureref{fig:copy}: there, \emph{depth} was the binding constraint (1 vs.\ 2 layers), whereas an 8-layer budget makes $k=1$ retrieval easy either way.}
\end{itemize}
So Canon may buy us nothing on the easiest retrieval; its advantage appears \emph{only at greater depth} --- a signature of learning dynamics, not of first-layer representation.

\parhead{How deep learning performs hierarchical learning}
Deep tasks are not learned all at once. On \textsc{Depo}, a model first fits 1-hop queries, then 2-hop, then 3-hop, and so on --- easy to hard. This is \emph{hierarchical learning}. \citet{al19-rnngen,allen2020backward} gave the first theory of when this is possible --- for three layers and for $\omega(1)$ layers, respectively. Our one-line summary: ``existence is cheap, learnability is everything'' --- a deep function may be easy to write down, yet impossible to learn except by \emph{properly} climbing a ladder of easier subtasks. Their theory identifies two necessary ingredients: \textbf{(1)} training data that \emph{contains} the ladder ($1,2,\dots,k$ hops), and \textbf{(2)} residual connections along depth.

But why should the two combined suffice --- why do they let a model acquire a deep function by first learning its shallow pieces and then building the deeper ones on top? The answer is a phenomenon they identified, called \textbf{backward feature correction}~\citep{allen2020backward}: the lower layers keep receiving signal, and keep improving, while the deeper ones are still learning. Let us explain.

Here is the thought experiment. Suppose a model has already mastered $(k\!-\!1)$-hop; now freeze it and train fresh layers on top to learn $k$-hop. No representation power is lost --- the frozen part plus the new layers can still express the $k$-hop function --- yet in practice this \textbf{fails completely}.\footnote{That layerwise training does not work is folklore; see~\citep{allen2020backward} for controlled experiments.}
Why? A network fit \emph{only} to $(k\!-\!1)$-hop has no reason to retain, in its last layer, \emph{some} hidden features that hop $k$ may need.\footnote{In \textsc{Depo}: which value was queried, whether a token sits in the query or the answer, where it sits. To fit $(k\!-\!1)$-hop the model need only emit one next token, so it may discard the rest.}
Instead, to truly learn $k$-hop, the lower layers need to prepare ``$(k\!-\!1)$-hop $+\,\epsilon$'' features, where the extra $\epsilon$ is useless for $(k\!-\!1)$-hop yet essential for hop $k$. Such an $\epsilon$ can be learned only when all layers are trained \emph{jointly}: the $k$-hop gradient must reach \emph{back} into the lower layers and correct them. Ingredient (1) supplies the $(k\!-\!1)$-hop stage to build on; ingredient (2) lets the correction signal reach it.

\parhead{From feedforward to language models}
That theory~\citep{al19-rnngen,allen2020backward} was written for feedforward networks. A language model has a second axis beyond depth --- the time horizon. To answer a $k$-hop query, the necessary ``$\epsilon$-features'' must also sit at \emph{earlier tokens}, not merely at lower layers. Backward feature correction must therefore \bblue{\emph{travel diagonally}}: down the layers and back along the token axis (\figureref{fig:bfc-schematic}). This calls for a third ingredient: \textbf{(3)} residual connections along time.

Ordinary residual connections carry correction only vertically; attention can carry it horizontally, but only after it learns where to look --- a slow, global route. Canon layers instead supply a direct, local one at every layer: by analogy, ``residual links in the time horizon.''\footnote{An analogy only; not to be confused with Canon's \emph{own} residual links, Canon(res), whose complementary role is to preserve the \emph{vertical} pathway (\sectionref{sec:trans-canon:ablation}).} This is why Canon-ABCD climbs the reasoning hierarchy so much faster while leaving $k=1$ untouched:
\vspace{-2mm}
\begin{center}
\textbf{Canon layers speed up hierarchical feature learning.}
\end{center}
\vspace{-2mm}

We stress that this is an \emph{efficiency} claim, not a capability one: with unlimited training data and steps, vanilla RoPE could eventually reach comparable accuracy on \textsc{Depo2}($K{=}16$); Canon simply gets there in far fewer steps.

\subsection{NoPE with Canon Layers}
\label{sec:trans-canon:nope}

\begin{figure}[t!]
\centering
\setlength{\imgwidthBase}{0.14\textwidth}
\vspace{-5mm}
\hspace*{-5mm}
\includegraphics[page=1,trim={2.5mm 1.5mm 2.5mm 1.5mm},clip,height=\imgwidthBase]{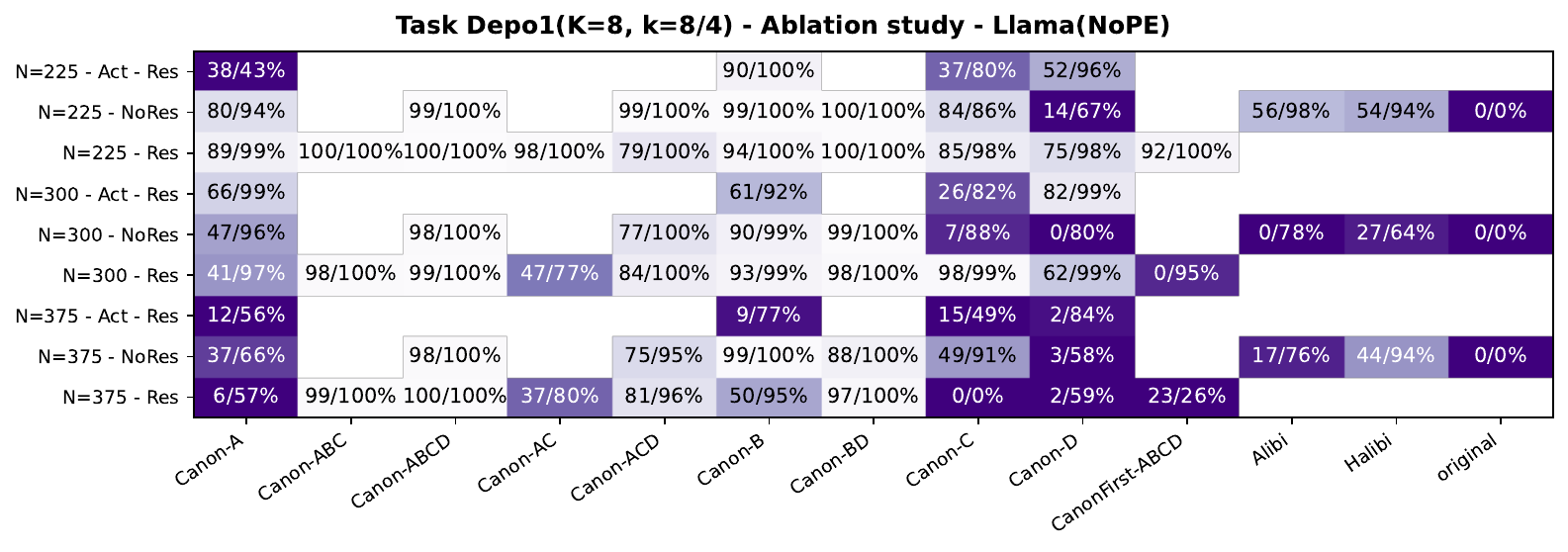}
\includegraphics[page=1,trim={2.5mm 1.5mm 2.5mm 1.5mm},clip,height=\imgwidthBase]{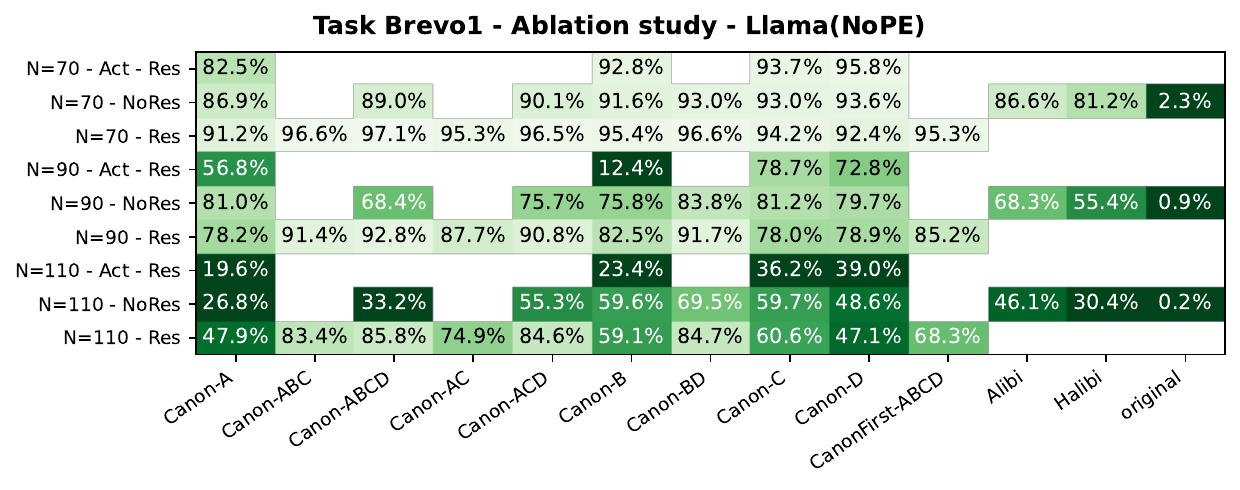}
\includegraphics[page=1,trim={2.5mm 1.5mm 2.5mm 1.5mm},clip,height=\imgwidthBase]{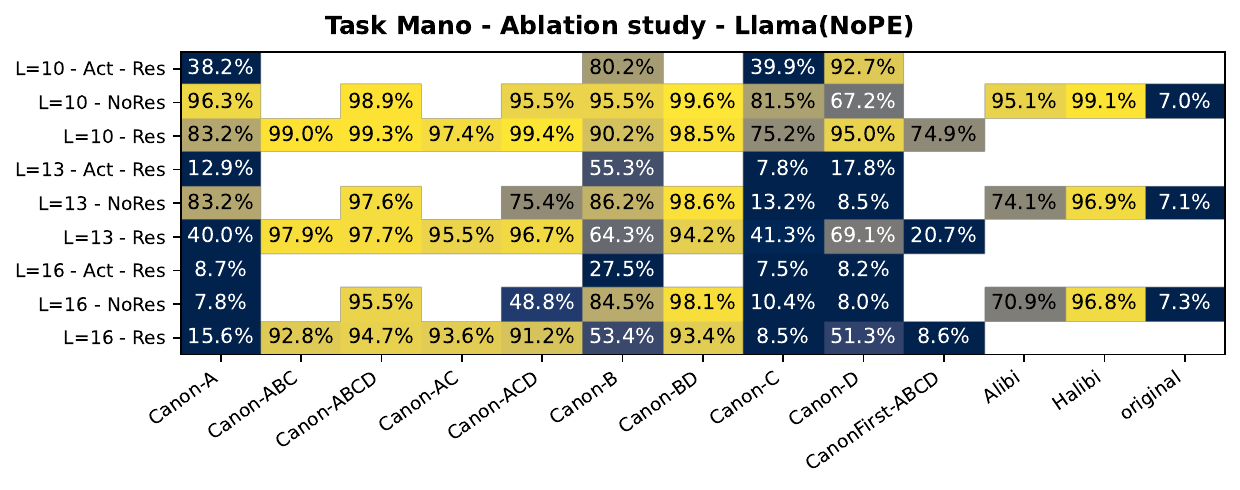}
\hspace*{-5mm}
\\
\hspace*{-5mm}
\includegraphics[page=1,trim={2.5mm 1.5mm 2.5mm 1.5mm},clip,height=\imgwidthBase]{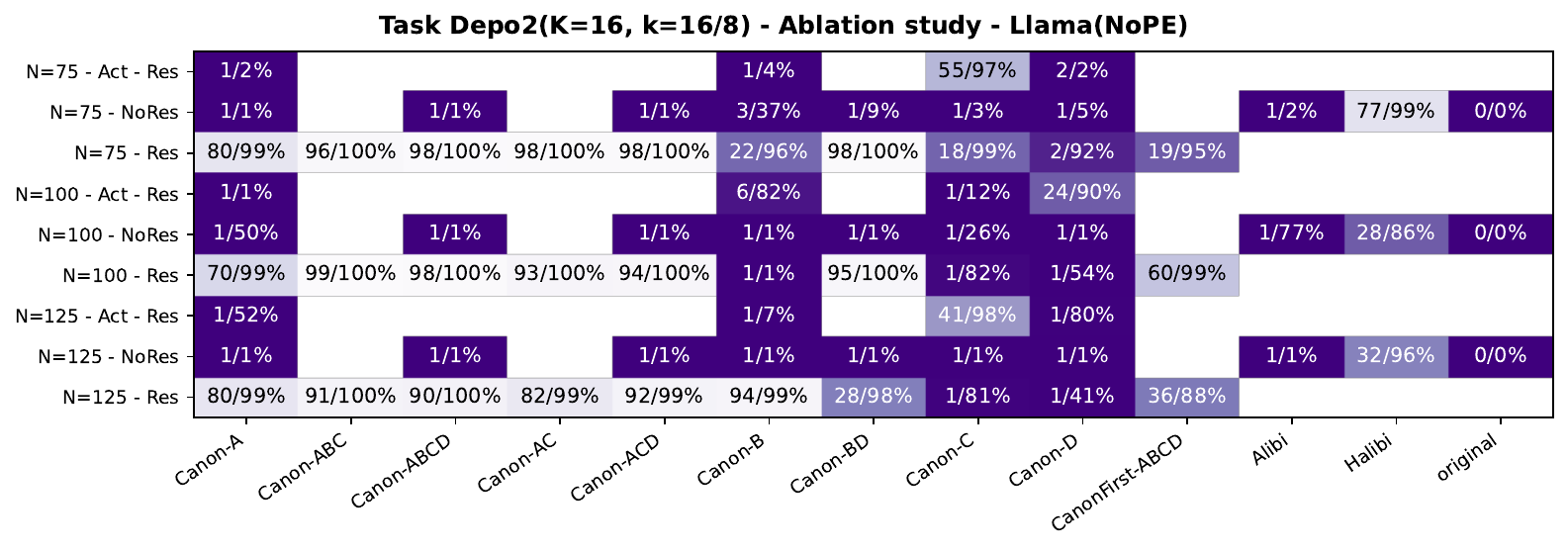}
\includegraphics[page=1,trim={2.5mm 1.5mm 2.5mm 1.5mm},clip,height=\imgwidthBase]{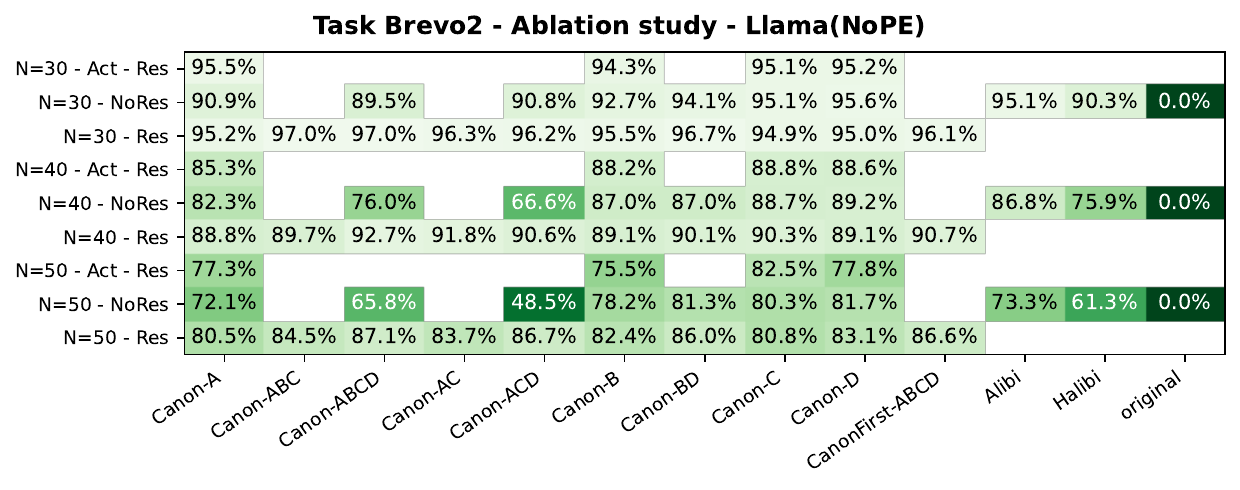}
\includegraphics[page=1,trim={2.5mm 1.5mm 2.5mm 1.5mm},clip,height=\imgwidthBase]{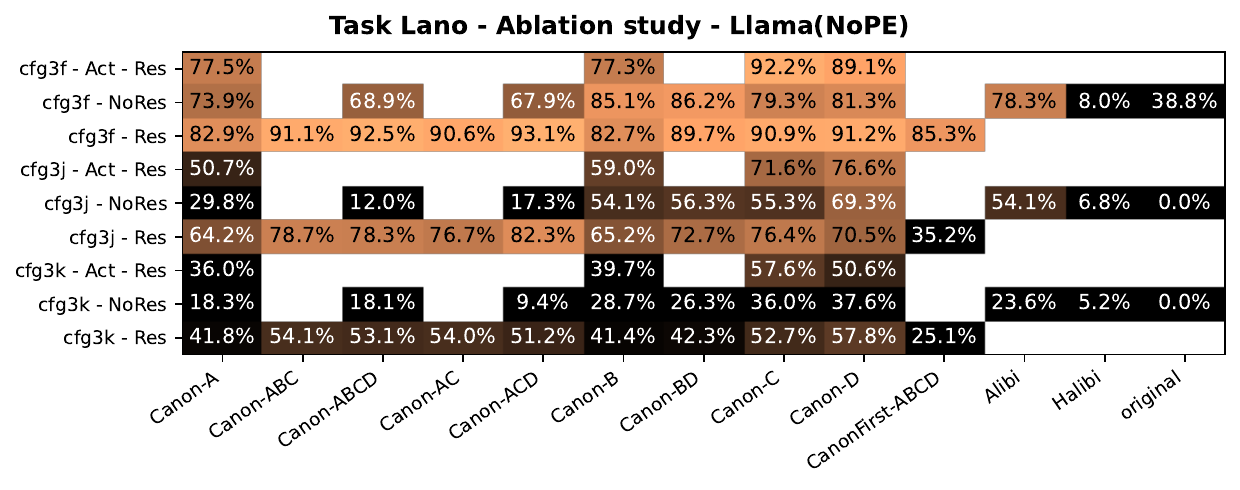}
\hspace*{-5mm}
\\
\vspace{-4mm}
\seplineb
\\
\hspace*{-5mm}
\includegraphics[page=1,trim={2.5mm 1.5mm 2.5mm 1.5mm},clip,height=\imgwidthBase]{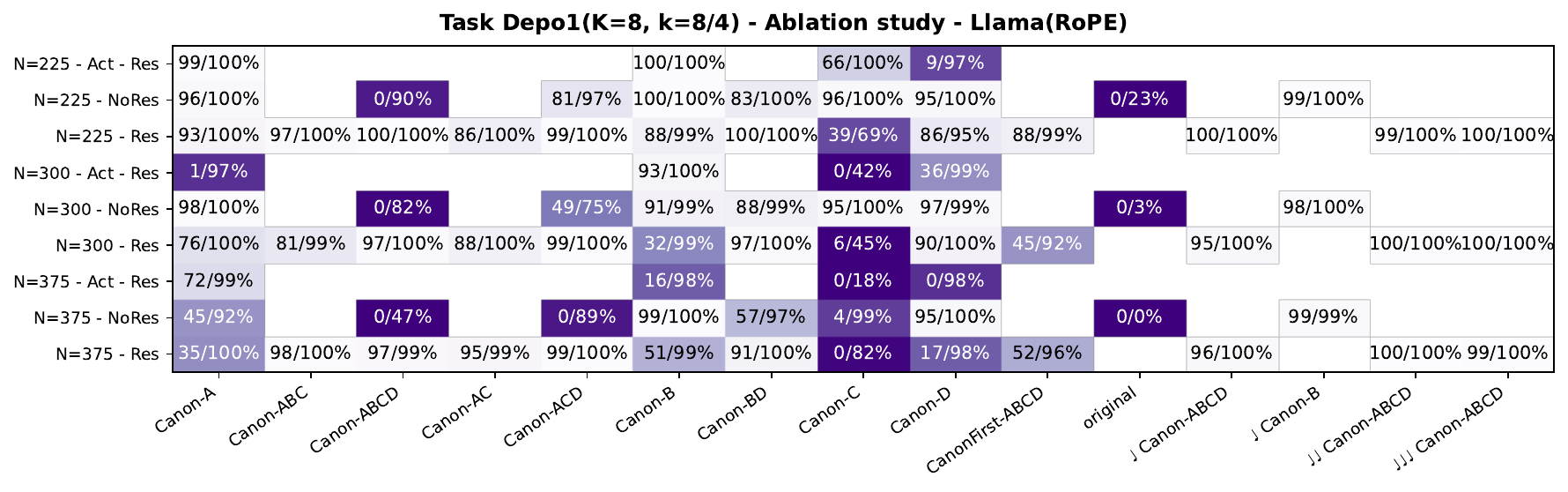}
\includegraphics[page=1,trim={2.5mm 1.5mm 2.5mm 1.5mm},clip,height=\imgwidthBase]{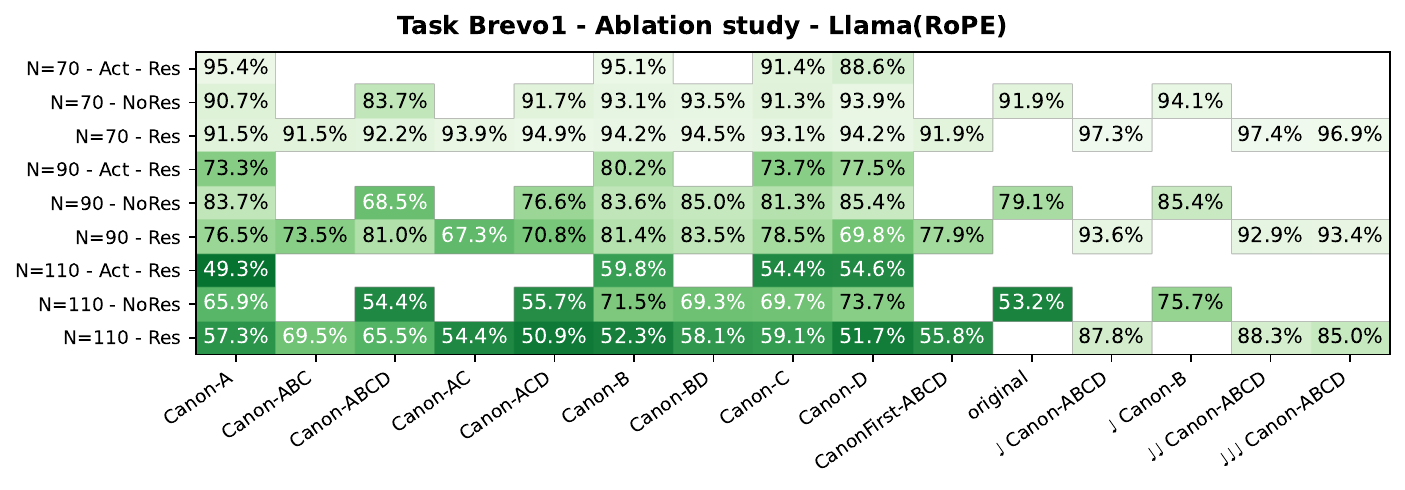}
\hspace*{-5mm}
\\
\hspace*{-5mm}
\includegraphics[page=1,trim={2.5mm 1.5mm 2.5mm 1.5mm},clip,height=\imgwidthBase]{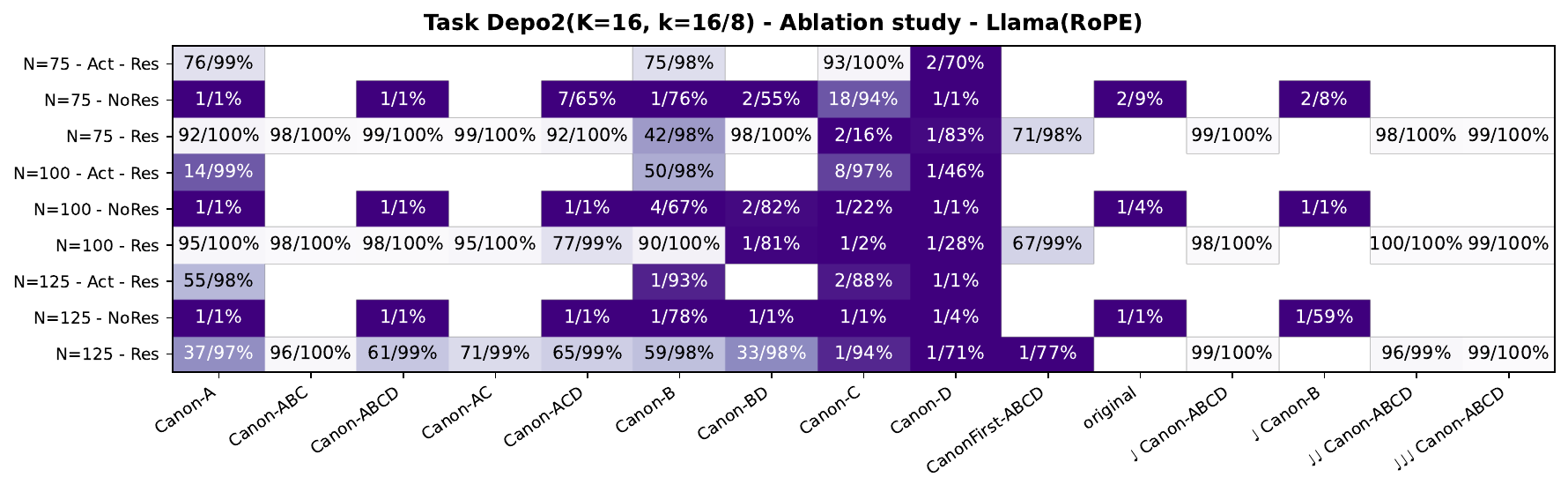}
\includegraphics[page=1,trim={2.5mm 1.5mm 2.5mm 1.5mm},clip,height=\imgwidthBase]{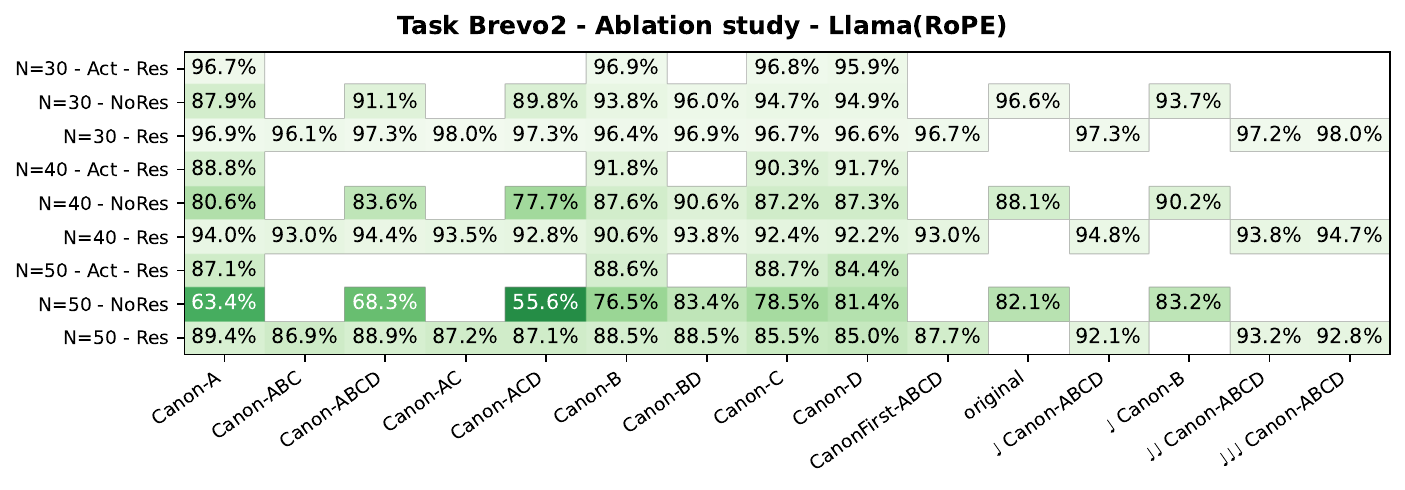}
\hspace*{-5mm}
\\
\hspace*{-5mm}
\includegraphics[page=1,trim={2.5mm 1.5mm 2.5mm 1.5mm},clip,height=\imgwidthBase]{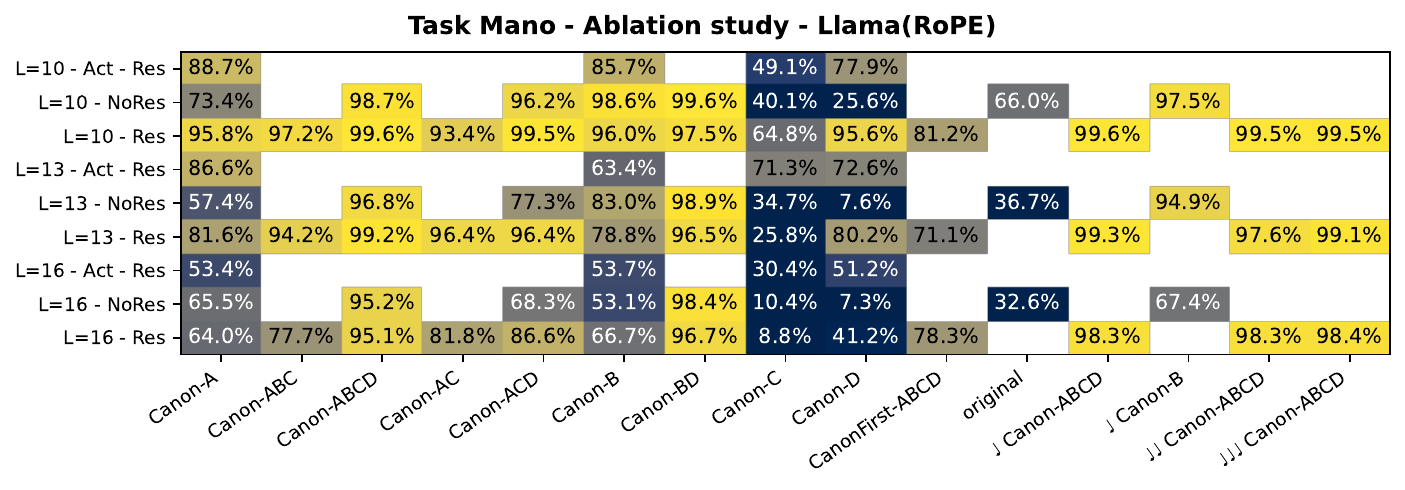}
\includegraphics[page=1,trim={2.5mm 1.5mm 2.5mm 1.5mm},clip,height=\imgwidthBase]{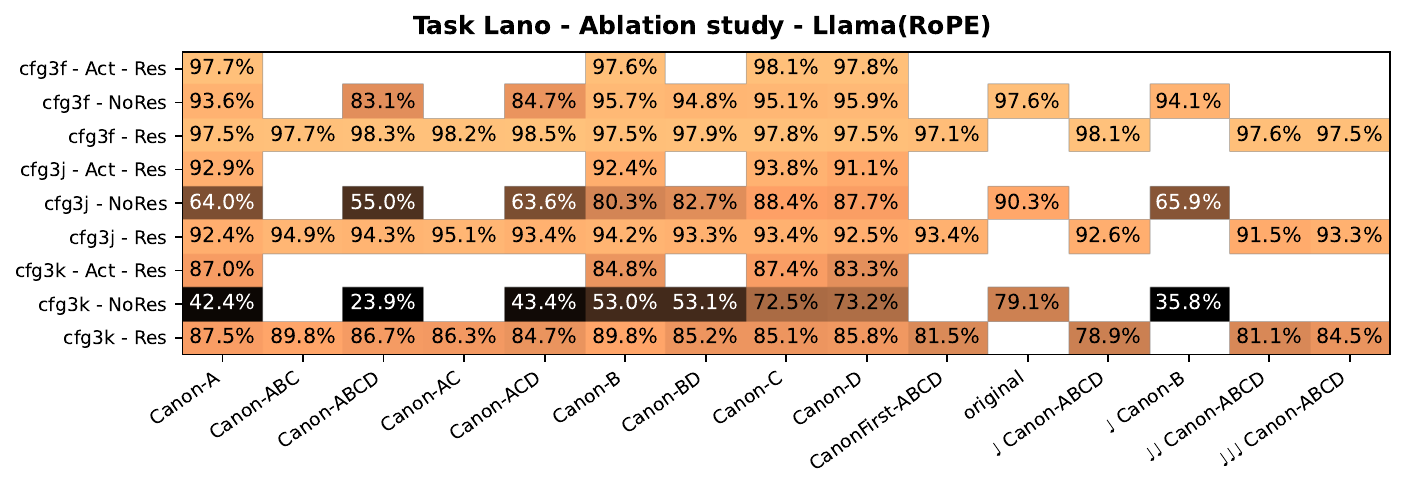}
\hspace*{-5mm}
\caption{\label{fig:trans-ablation}%
\textbf{Ablation study on 12-layer, 768-dim Transformers—NoPE (top) and RoPE (bottom)}—with \bblue{Canon} variants (\bblue{A--D}), \bblue{CanonFirst} (applied only to first block), \bblue{res}idual links, \bblue{act}ivation functions, \bblue{ALiBi}, and \bblue{H-Alibi}. Blank entries indicate untested configs due to resource limits. Additional ablation studies (with more model sizes) are in \figureref{fig:app:rope-ablation} (RoPE), \figureref{fig:app:primer} (RoPE+Primer), and \figureref{fig:app:nope-ablation} (NoPE) in \appendixref{app:more-experiments}.
}
\end{figure}

\begin{mdframed}
\begin{sresult}{3}[\figureref{fig:trans-canon}+\ref{fig:trans-ablation}\footnote{(Sub-results correspond to \figureref{fig:trans-canon} (4th vs 5th column), \figureref{fig:trans-ablation}, and \figureref{fig:trans-canon} (3rd column), respectively.)}]
\label{res:3}
Canon layers transform NoPE. Key findings include:
\begin{itemize}
    \item NoPE+Canon matches RoPE+Canon and even surpasses it on \textsc{Depo} --- a remarkable result given that without Canon, NoPE achieves essentially zero performance on all measures.
    \item NoPE+Canon significantly outperforms existing fixes for NoPE, such as ALiBi and H-Alibi.
    \item With Canon layers, RoPE usage can be greatly reduced: RoPE on only 1/4 dims (denoted RoPE+\musQuarter{}Canon) outperforms both RoPE/NoPE+Canon, great news for length generalization.
\end{itemize}
\end{sresult}
\end{mdframed}

\parhead{Canon layers  skyrocket  NoPE performance} Canon layers dramatically improve NoPE (No Positional Embedding) transformers, lifting them from near-zero accuracy to competitive levels, even slightly surpassing RoPE+Canon on reasoning depth. Evidently, Canon layers take over much of RoPE's job of locating nearby tokens.\footnote{This is why, as previewed in \sectionref{sec:trans-canon:why}, adding Canon in the first block only (CanonFirst-ABCD) recovers the least precisely for NoPE (\figureref{fig:app:nope-ablation}): that job must be redone at \emph{every} layer, not only the first.} NoPE-Canon is only weaker on Task \textsc{Lano}, which involves hierarchical structural learning over long sequences, thus relying more heavily on relative distance between input tokens; yet even there NoPE-Canon remains competitive with alternatives such as Mamba2/GDN.

\parhead{Dominance over existing fixes on NoPE} While NoPE excels at length generalization, its performance on complex reasoning tasks has historically been weak. Fixes like ALiBi~\cite{press2021train} and Hard-Alibi~\cite{jelassi2024repeat} partially address this: ALiBi applies a distance-based penalty to attention weights\footnote{Specifically, adding $-|j-i| \cdot 2^{-8h/H}$ to the logits of head $h$ of $H$ total heads.}, while Hard-Alibi disables attention beyond distance $h$ for the $h$-th head. Although these methods improve NoPE performance (partly mimicking RoPE), Canon layers clearly dominate. As shown in \figureref{fig:trans-ablation} (top), NoPE+Canon significantly outperforms both alternatives.

\parhead{Minimal RoPE usage with Canon layers} Canon layers eliminate the need for heavy RoPE usage, and excessive RoPE can even hurt performance. With Canons, minimal RoPE usage is sufficient—often preferable—for optimal results. For example, enabling RoPE on half of the heads at half of their dimensions (denoted \musQuarter{}Canon) consistently outperforms full RoPE usage or NoPE, as shown in \figureref{fig:trans-canon} (3rd column). \textbf{This is great news for long-context generalization}: RoPE is a known bottleneck for Transformers with longer inputs. As Canon layers allow significantly reduced RoPE without performance loss, they become indispensable for length generalization tasks.%
\footnote{Alternative reduced-RoPE configurations explored in  the appendix  include: \musQuarter\musQuarter{} (1/4 of heads at full dimensions) and \musQuarter\musQuarter\musQuarter{} (all heads at 1/4 dimensions, as in GPT-NeoX~\cite{gpt-neox-20b}). Among these, \musQuarter{} and \musQuarter\musQuarter\musQuarter{} are comparable, slightly better than \musQuarter\musQuarter{} according to \figureref{fig:quarters} on Page~\pageref{fig:quarters}.}

\begin{remark}
Despite their versatility, Canon layers alone cannot fully resolve extremely challenging tasks that require deep hierarchical reasoning over long sequences (e.g., \textsf{cfg3k} in Task \textsc{Lano}). Such tasks, requiring $O(n^3)$ dynamic programming over 1000 tokens, remain computationally demanding. Nevertheless, Canon layers consistently offer huge improvements outside these specialized scenarios.
\end{remark}

\parhead{These findings translate to real-life}
To be shown in \sectionref{sec:real-life}, NoPE+Canon consistently matches or surpasses RoPE+Canon in real-world pretraining; the RoPE+\musQuarter{}Canon variants outperform RoPE+Canon on several reasoning tasks, particularly involving long-context inputs.

\begin{remark}
This paper focuses on architectural differences within computational stages after relevant information is retrieved into manageable contexts (e.g., 4096 tokens). Techniques like DeepSeek's NSA architecture \cite{yuan2025native}, designed for retrieval and compression from extremely long inputs (e.g., 1M tokens), are complementary to Canon layers. Such techniques and Canon layers can thus jointly handle distinct processing phases in long-context models.
\end{remark}

\subsection{Ablation Studies With Canon Layers}
\label{sec:trans-canon:ablation}

This section systematically investigates the design choices in Canon layers via ablation studies.

\parhead{Component-level contributions}
Each Canon component (A/B/C/D) contributes meaningfully to performance, with cumulative benefits from combinations. Adding even a single Canon layer yields notable gains, and stacking multiple Canon layers across sub-layers further amplifies these improvements, especially on weaker architectures like NoPE. Summaries appear in \figureref{fig:trans-ablation} (for model size 12L768D) and additional size experiments in \appendixref{app:more-experiments}.

\parhead{Role of residual connections}
Residual links around Canon layers --- i.e., the ``$h_t+$'' part of \eqref{eqn:canon-res} --- are critical for training stability and effective learning, preserving vertical computational pathways and allowing global representations to selectively incorporate local context. Without residual connections, training becomes slower and less stable (see rows marked ``NoRes'' in \figureref{fig:trans-ablation}).

\parhead{Independence of Attention/MLP}
Prior works (e.g., the GLA~\cite{yang2023gated} codebase and Primer~\cite{so2109primer}) focused solely on convolution operations within attention projections --- Canon-B(no-res). However, we find that Canon-ACD alone already achieves substantial performance improvements, without modifying attention mechanisms. Similarly, Canon-ABC or even Canon-AC perform strongly without adjusting MLP layers. They all strongly outperform Canon-B(no-res) and thus outperform Primer. This highlights Canon layers' general role in enhancing horizontal information flow across architecture sub-layers, \emph{independently complementing} attention or MLP mechanisms.

\parhead{Nonlinear activations and computational simplicity}
Contrary to prior works (e.g., H3/Mamba), adding activation functions such as SiLU after the Canon layers does not yield noticeable benefits. Canon layers effectively inject local context directly into token positions, and nonlinear operations are sufficiently handled by the attention and MLP blocks (see rows marked ``Act'' in \figureref{fig:trans-ablation}).

\begin{mdframed}
\begin{sresult}{4}[\figureref{fig:trans-ablation}]
\label{res:4}
Canon layers are lightweight, versatile, and effective enhancements that integrate seamlessly into Transformers. Key findings:
\begin{itemize}
    \item Canon-A/B/C/D yield meaningful, cumulative improvements when stacked, and can be flexibly applied anywhere independent of attention or MLP modifications.
    \item Residual connections in Canon design are essential for stable, efficient training.
    \item Adding nonlinear activations (e.g., SiLU) provides no measurable benefit, simplifying design.
\end{itemize}
\end{sresult}
\end{mdframed}

\noindent
(This \textbf{differs from prior works}: we show where to insert Canon layers, how to stabilize them, and why they matter.)

\begin{figure}[t!]
\centering
\setlength{\imgwidthBase}{0.32\textwidth}
\vspace{-5mm}\includegraphics[page=1,trim={2.5mm 1.5mm 2.5mm 1.5mm},clip,width=\imgwidthBase]{biocap/Llama-original}
\includegraphics[page=1,trim={2.5mm 1.5mm 2.5mm 1.5mm},clip,width=\imgwidthBase]{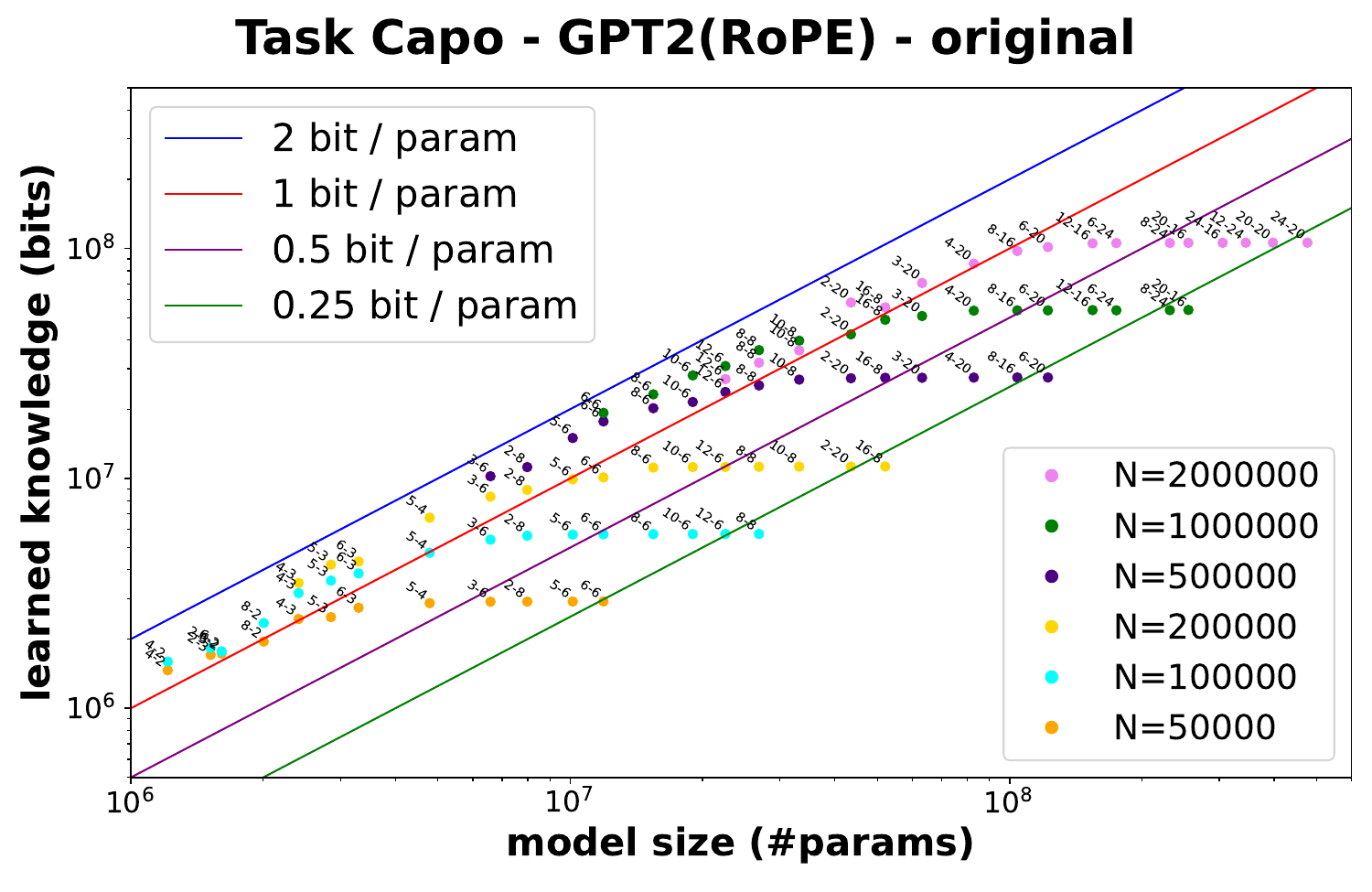}
\includegraphics[page=1,trim={2.5mm 1.5mm 2.5mm 1.5mm},clip,width=\imgwidthBase]{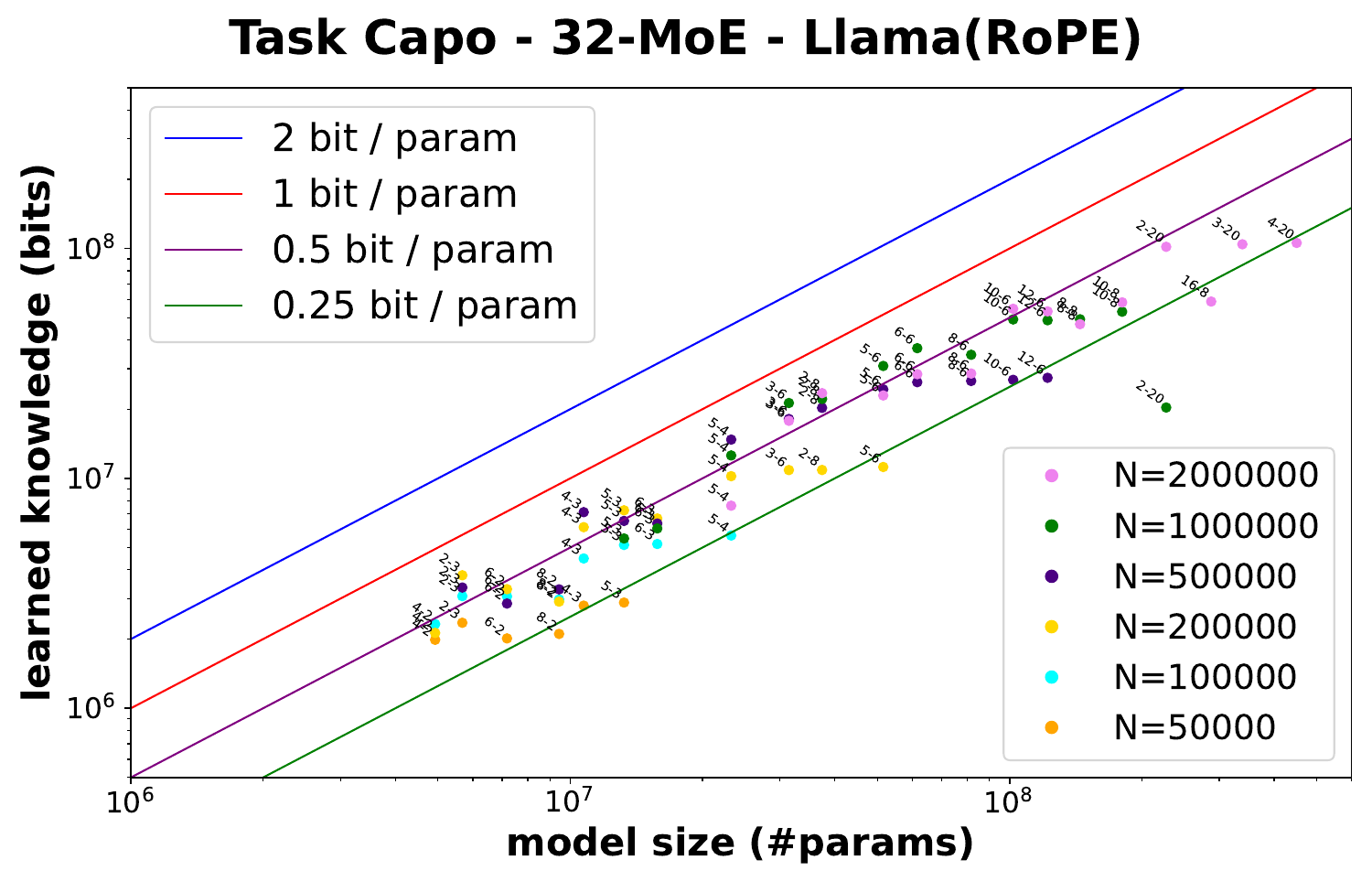}
\includegraphics[page=1,trim={2.5mm 1.5mm 2.5mm 1.5mm},clip,width=\imgwidthBase]{biocap/Llama-ABCD}
\includegraphics[page=1,trim={2.5mm 1.5mm 2.5mm 1.5mm},clip,width=\imgwidthBase]{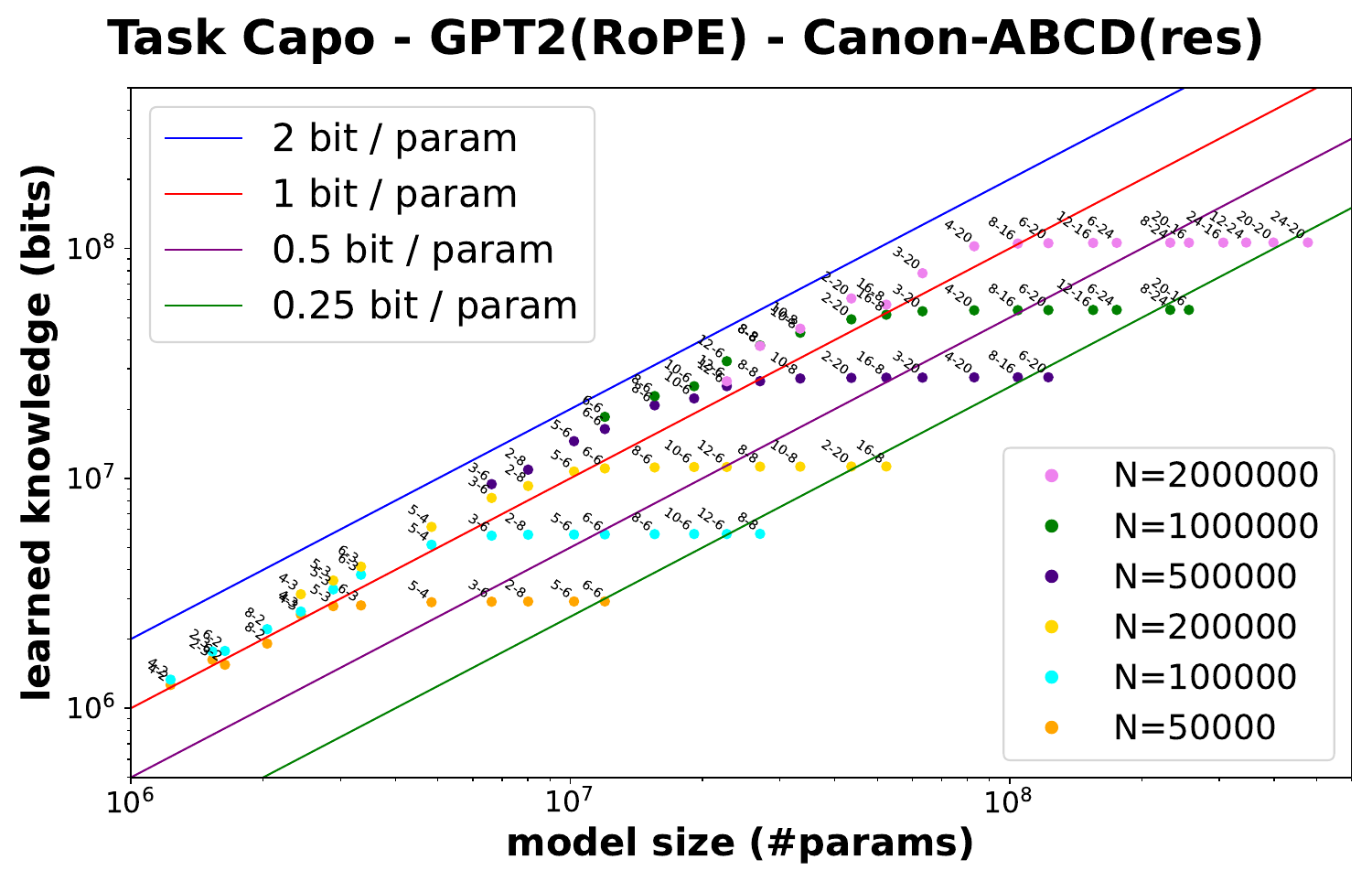}
\includegraphics[page=1,trim={2.5mm 1.5mm 2.5mm 1.5mm},clip,width=\imgwidthBase]{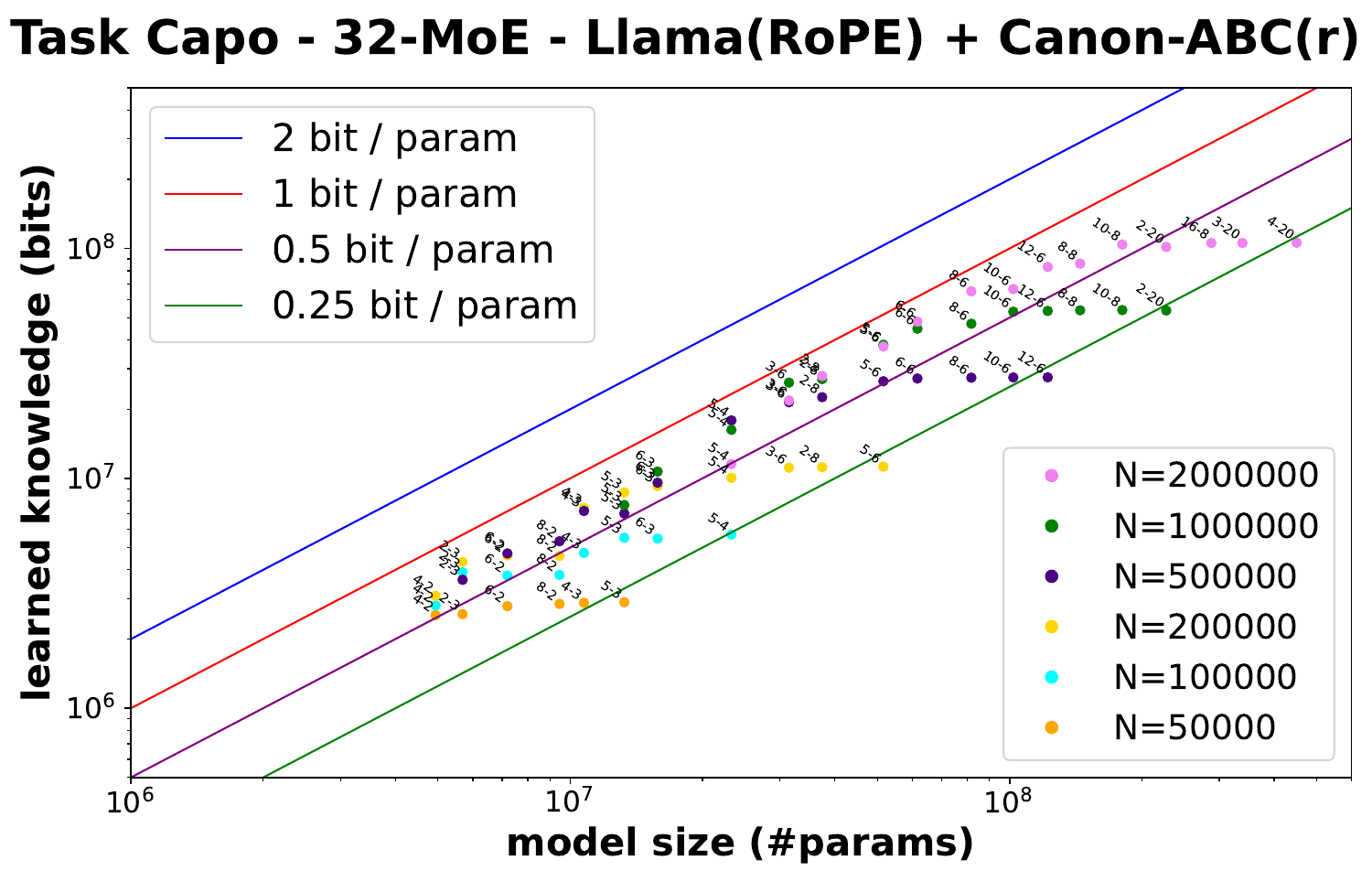}
\caption{\label{fig:moe-capo}%
\textbf{Evaluation of knowledge capacity} (\textsc{Capo}) across architectures, measured as bits per parameter.
The first row represents baseline models, while the second row shows improvements with Canon layers added.
\newline
\textbf{Conclusion:} Canon layers enhance knowledge storage for architectures that are slower to train, such as gated MLP and MoE, mitigating the capacity gap between gated and standard MLP as identified in \cite{AL2024-knowledgeScaling}.
}
\end{figure}

\subsection{MLP and Mixture-of-Experts}
\label{sec:trans-canon:mlp}

Our synthetic playground provides a valuable framework to evaluate \emph{broader architectural choices}.

\parhead{Gated vs. standard MLPs}
Gated MLPs~\cite{shazeer2020glu}, which replace standard MLP operations $V \sigma (Wx)$ by $V (\sigma(W_1 x)\cdot(W_2 x))$, improve expressiveness and parameter efficiency. Widely adopted by large-scale models (e.g., PaLM~\cite{chowdhery2023palm}, Llama~\cite{touvron2023llama,llama2}, Mistral~\cite{jiang2023mistral}), gated MLPs have become standard design choices. However, \cite{AL2024-knowledgeScaling} found that gated MLP reduces knowledge capacity by about 30\% in limited-exposure scenarios (e.g., 100-exposure Task \textsc{Capo}) due to slower convergence.

Thus, what is the best tradeoff? Our experiments (\figureref{fig:gpt} on Page~\pageref{fig:gpt}) confirm gated MLP has a slight advantage over standard MLP (``GPT2-style'') on reasoning-heavy tasks, showing noticeable improvements on knowledge manipulation (\textsc{Mano}) and smaller gains on reasoning breadth (\textsc{Brevo}). Thus, replacing gated MLP with standard MLP may not be the best choice. However, keep in mind that adding Canon layers already partially mitigates gated MLP's capacity loss (recall \resultref{res:2}), due to improving training dynamics and speed, recovering about half of its lost capacity.

\parhead{Mixture-of-Experts}
Mixture-of-Experts (MoE)~\cite{fedus2022switch,shazeer2016outrageously} enhances parameter efficiency by replacing dense MLPs with multiple parallel ``experts,'' selectively routing tokens to fewer active experts. While MoE achieves good scalability (particularly on knowledge capacity) and competitive inference-time performance, it suffers from significantly slower knowledge acquisition speed during training. For example, a 32-expert transformer may acquire 10$\times$ less knowledge in the same 100-exposure regime (mimicking rare knowledge) compared to dense models (\figureref{fig:moe-capo}). Could Canon layers mitigate this due to their improved training dynamics?

Integrating Canon layers with MoE, however, poses a challenge. Canon-D relies on neighboring tokens' hidden states, conflicting with MoE's independent token-wise expert dispatching. Adapting Canon-D to MoE would require complex engineering. To avoid such complexity, we test Canon-ABC layers alone, which already significantly accelerate MoE knowledge acquisition and improve bit-per-parameter efficiency (\figureref{fig:moe-capo}), recovering at least half of the MoE-induced capacity loss.

\parhead{MLP with Squared ReLU}
The Primer~\cite{so2109primer} paper proposes using $\text{ReLU}^2$ as the activation function in standard MLPs, reporting improved performance over gated MLPs (e.g., SwiGLU) on real-world data. They also claim this gain exceeds that of Canon-B(no-res), which they refer to as ``Multi-DConv-Head Attention.'' In our synthetic playground (see \figureref{fig:app:relu2} on Page~\pageref{fig:app:relu2}), we confirm that $\text{ReLU}^2$ slightly improves standard MLPs (though not necessarily outperforming gated MLPs, consistent with recent findings~\cite{zhang2024relu}), while applying $\text{ReLU}^2$ to gated MLPs degrades performance. However, \emph{these effects are \textbf{negligible} compared to the gains provided by Canon layers}.

\begin{mdframed}
\begin{sresult}{5}[\figureref{fig:moe-capo}+\ref{fig:gpt}+\ref{fig:app:relu2}]
\label{res:5}
Key insights for MLP and MoE architectures:
\begin{itemize}
    \item Gated MLP slightly outperforms standard MLP (especially on \textsc{Mano}).
    \item Gated MLP reduces knowledge capacity (\textsc{Capo}); Canon layers partially recover this loss.
    \item $\text{ReLU}^2$ activation slightly improves standard MLP but degrades performance in gated MLP.
    \item Canon-ABC substantially improves MoE knowledge acquisition and bit-per-param capacity.
\end{itemize}
\end{sresult}
\end{mdframed}

\begin{figure}[t!]
\centering
\setlength{\imgwidthBase}{0.2\textwidth}
\vspace{-3mm}\hspace*{-5mm}
\includegraphics[page=1,trim={2.5mm 1.5mm 2.5mm 1.5mm},clip,width=\imgwidthBase]{perm_4/GLA-original}
\includegraphics[page=1,trim={2.5mm 1.5mm 2.5mm 1.5mm},clip,width=\imgwidthBase]{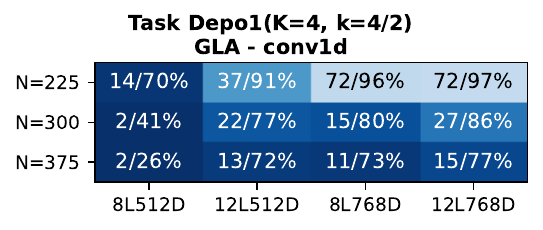}
\includegraphics[page=1,trim={2.5mm 1.5mm 2.5mm 1.5mm},clip,width=\imgwidthBase]{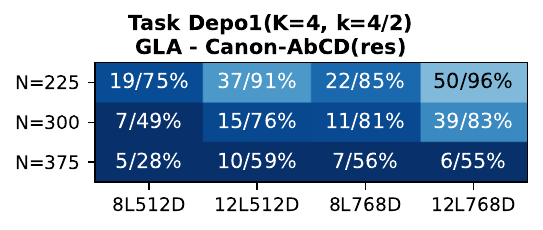}
\includegraphics[page=1,trim={2.5mm 1.5mm 2.5mm 1.5mm},clip,width=\imgwidthBase]{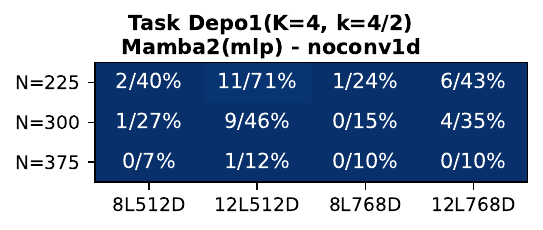}
\includegraphics[page=1,trim={2.5mm 1.5mm 2.5mm 1.5mm},clip,width=\imgwidthBase]{perm_4/Mamba2_mlp_-original_conv1d_}
\hspace*{-5mm}
\\
\hspace*{-5mm}
\includegraphics[page=1,trim={2.5mm 1.5mm 2.5mm 1.5mm},clip,width=\imgwidthBase]{perm_multi_4/GLA-original}
\includegraphics[page=1,trim={2.5mm 1.5mm 2.5mm 1.5mm},clip,width=\imgwidthBase]{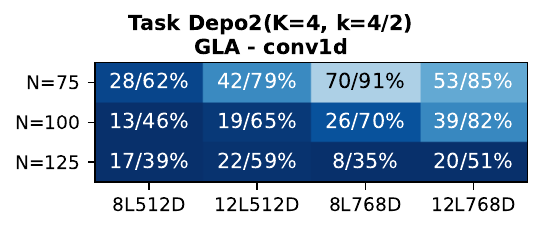}
\includegraphics[page=1,trim={2.5mm 1.5mm 2.5mm 1.5mm},clip,width=\imgwidthBase]{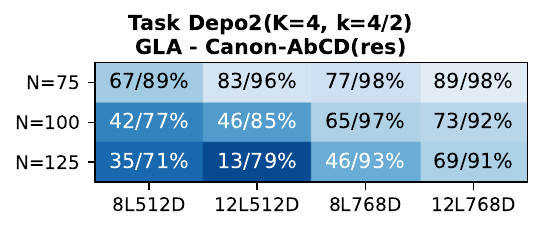}
\includegraphics[page=1,trim={2.5mm 1.5mm 2.5mm 1.5mm},clip,width=\imgwidthBase]{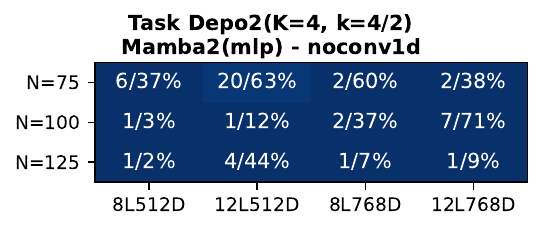}
\includegraphics[page=1,trim={2.5mm 1.5mm 2.5mm 1.5mm},clip,width=\imgwidthBase]{perm_multi_4/Mamba2_mlp_-original_conv1d_}
\hspace*{-5mm}
\\
\hspace*{-5mm}
\includegraphics[page=1,trim={2.5mm 1.5mm 2.5mm 1.5mm},clip,width=\imgwidthBase]{top_sort/GLA-original}
\includegraphics[page=1,trim={2.5mm 1.5mm 2.5mm 1.5mm},clip,width=\imgwidthBase]{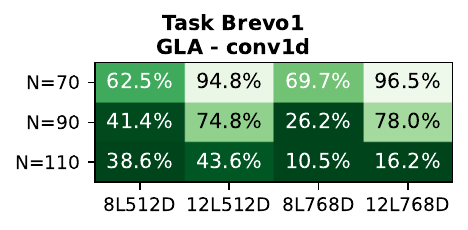}
\includegraphics[page=1,trim={2.5mm 1.5mm 2.5mm 1.5mm},clip,width=\imgwidthBase]{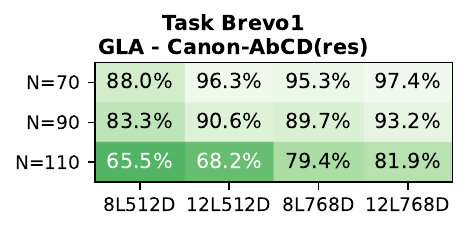}
\includegraphics[page=1,trim={2.5mm 1.5mm 2.5mm 1.5mm},clip,width=\imgwidthBase]{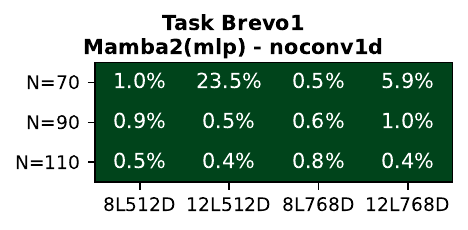}
\includegraphics[page=1,trim={2.5mm 1.5mm 2.5mm 1.5mm},clip,width=\imgwidthBase]{top_sort/Mamba2_mlp_-original_conv1d_}
\hspace*{-5mm}
\\
\hspace*{-5mm}
\includegraphics[page=1,trim={2.5mm 1.5mm 2.5mm 1.5mm},clip,width=\imgwidthBase]{top_sort_multi/GLA-original}
\includegraphics[page=1,trim={2.5mm 1.5mm 2.5mm 1.5mm},clip,width=\imgwidthBase]{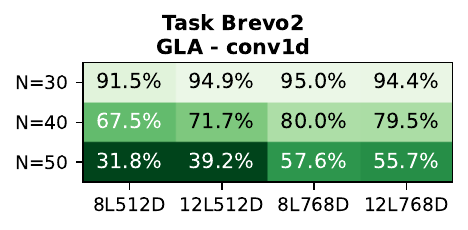}
\includegraphics[page=1,trim={2.5mm 1.5mm 2.5mm 1.5mm},clip,width=\imgwidthBase]{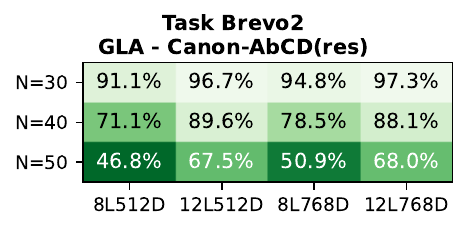}
\includegraphics[page=1,trim={2.5mm 1.5mm 2.5mm 1.5mm},clip,width=\imgwidthBase]{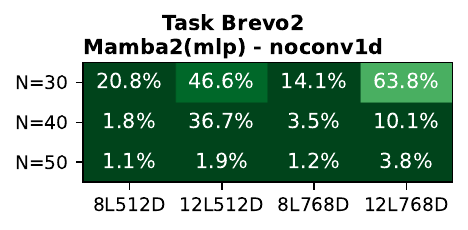}
\includegraphics[page=1,trim={2.5mm 1.5mm 2.5mm 1.5mm},clip,width=\imgwidthBase]{top_sort_multi/Mamba2_mlp_-original_conv1d_}
\hspace*{-5mm}
\\
\hspace*{-5mm}
\includegraphics[page=1,trim={2.5mm 1.5mm 2.5mm 1.5mm},clip,width=\imgwidthBase]{arith/GLA-original}
\includegraphics[page=1,trim={2.5mm 1.5mm 2.5mm 1.5mm},clip,width=\imgwidthBase]{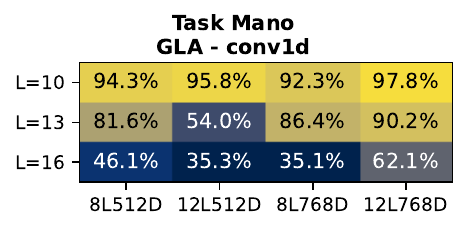}
\includegraphics[page=1,trim={2.5mm 1.5mm 2.5mm 1.5mm},clip,width=\imgwidthBase]{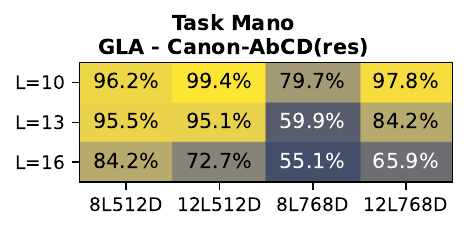}
\includegraphics[page=1,trim={2.5mm 1.5mm 2.5mm 1.5mm},clip,width=\imgwidthBase]{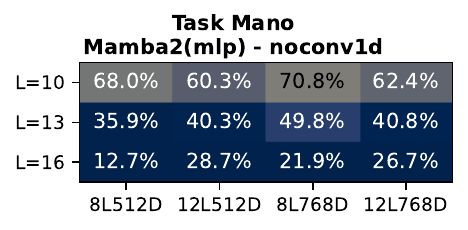}
\includegraphics[page=1,trim={2.5mm 1.5mm 2.5mm 1.5mm},clip,width=\imgwidthBase]{arith/Mamba2_mlp_-original_conv1d_}
\hspace*{-5mm}
\\
\hspace*{-5mm}
\includegraphics[page=1,trim={2.5mm 1.5mm 2.5mm 1.5mm},clip,width=\imgwidthBase]{biocap/gla-original}
\includegraphics[page=1,trim={2.5mm 1.5mm 2.5mm 1.5mm},clip,width=\imgwidthBase]{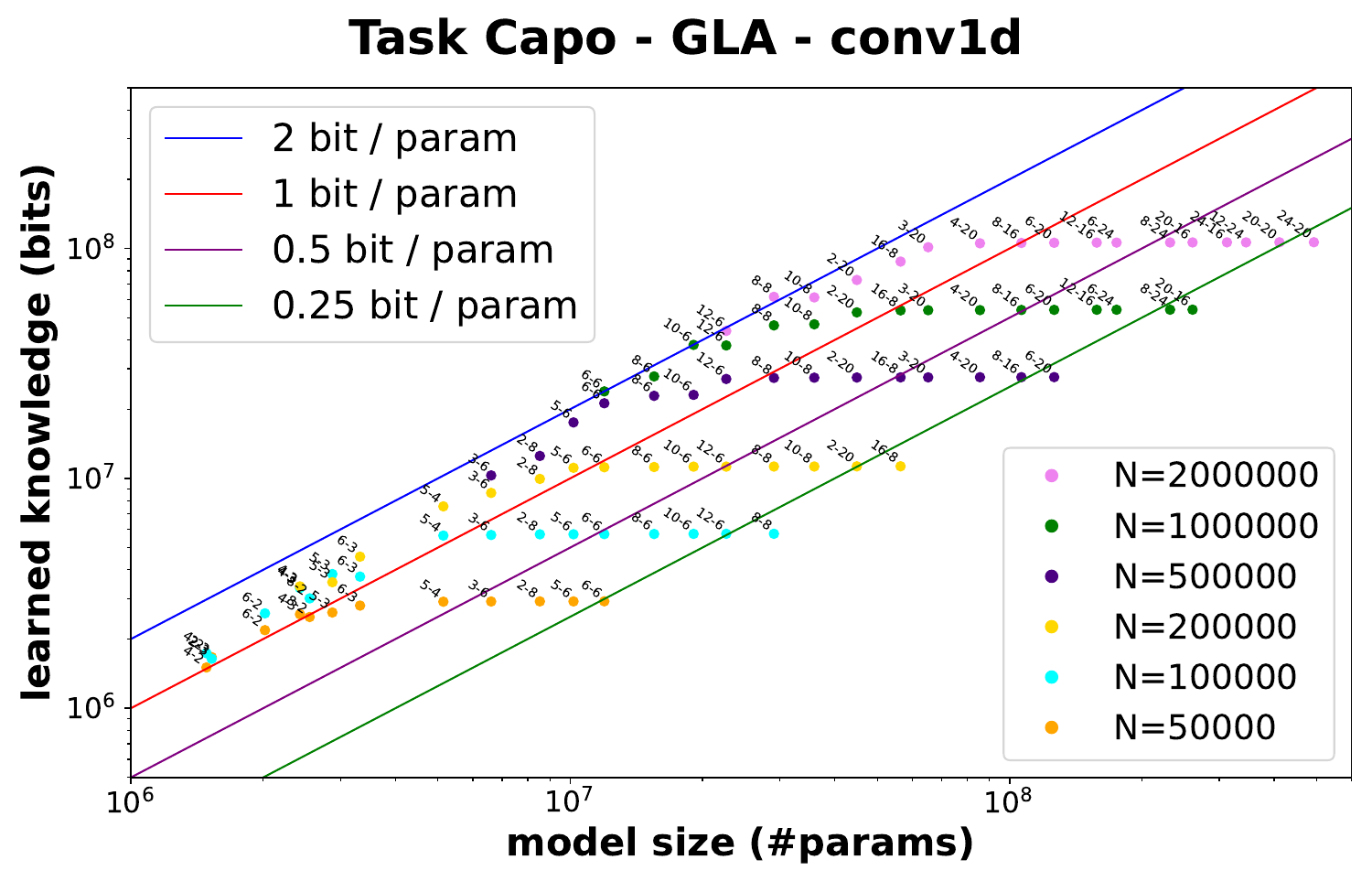}
\includegraphics[page=1,trim={2.5mm 1.5mm 2.5mm 1.5mm},clip,width=\imgwidthBase]{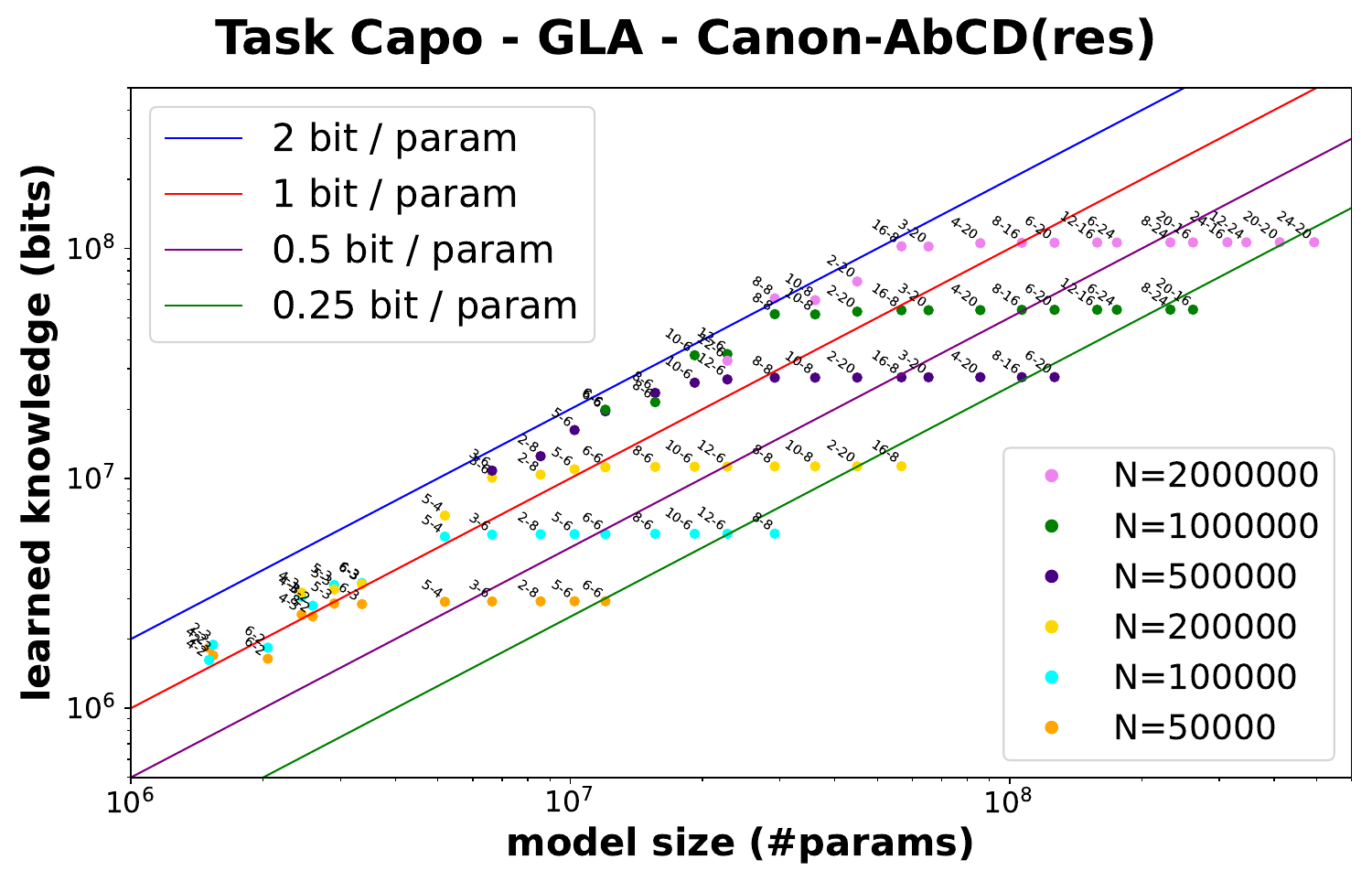}
\includegraphics[page=1,trim={2.5mm 1.5mm 2.5mm 1.5mm},clip,width=\imgwidthBase]{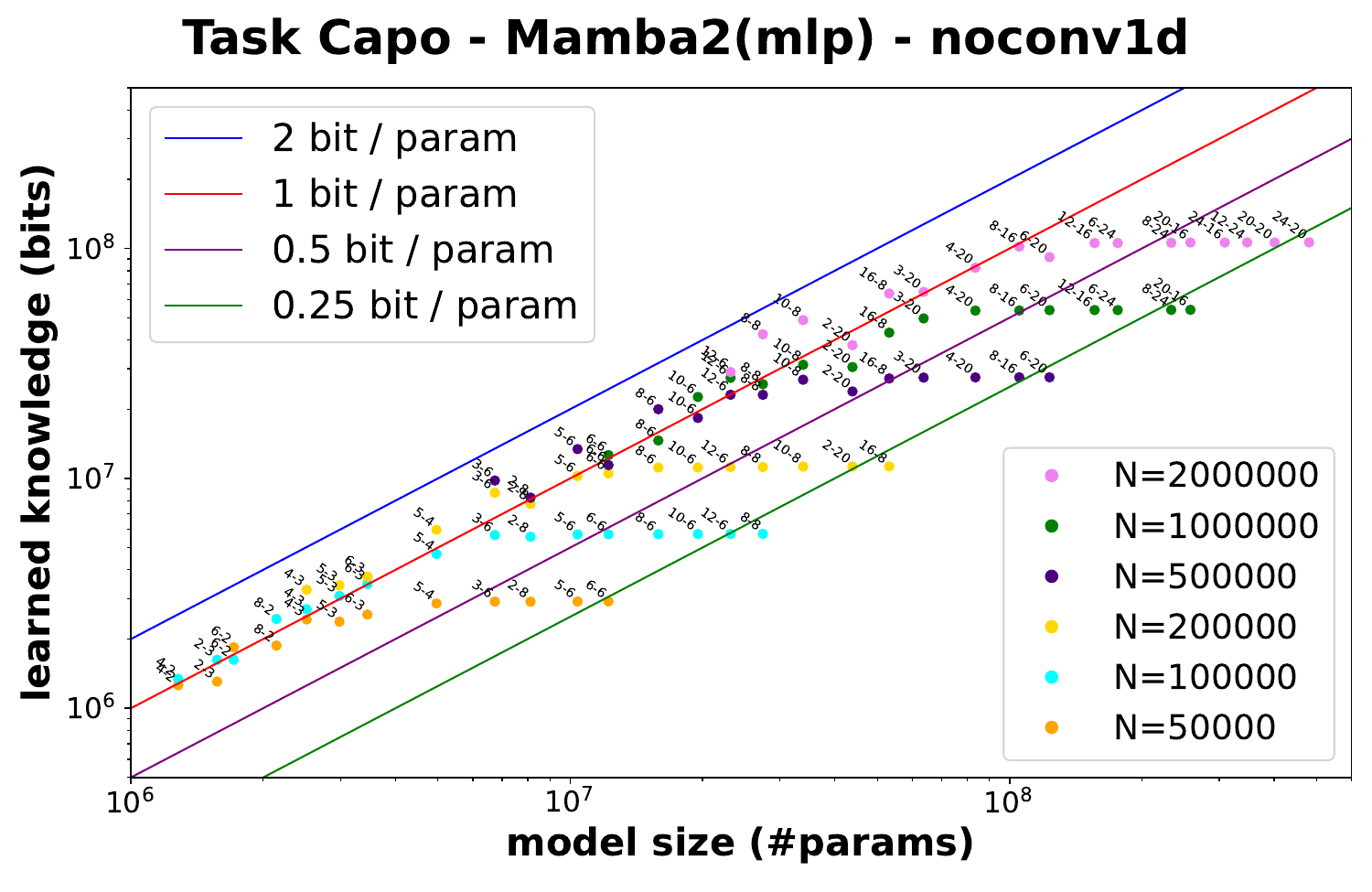}
\includegraphics[page=1,trim={2.5mm 1.5mm 2.5mm 1.5mm},clip,width=\imgwidthBase]{biocap/mamba-mamba-mlp}
\hspace*{-5mm}
\\
\hspace*{-5mm}
\includegraphics[page=1,trim={2.5mm 1.5mm 2.5mm 1.5mm},clip,width=\imgwidthBase]{cfg/GLA-original}
\includegraphics[page=1,trim={2.5mm 1.5mm 2.5mm 1.5mm},clip,width=\imgwidthBase]{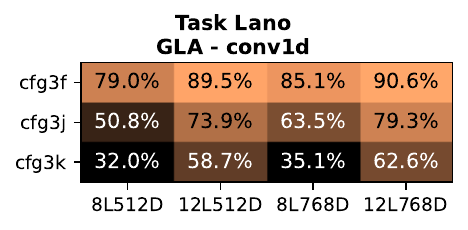}
\includegraphics[page=1,trim={2.5mm 1.5mm 2.5mm 1.5mm},clip,width=\imgwidthBase]{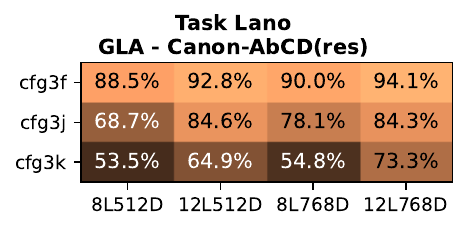}
\includegraphics[page=1,trim={2.5mm 1.5mm 2.5mm 1.5mm},clip,width=\imgwidthBase]{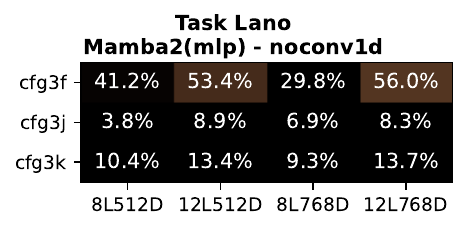}
\includegraphics[page=1,trim={2.5mm 1.5mm 2.5mm 1.5mm},clip,width=\imgwidthBase]{cfg/Mamba2_mlp_-original_conv1d_}
\hspace*{-5mm}
\caption{\label{fig:gla-and-mamba}\textbf{Columns 1, 2, 3, 5}: Canon drastically improves GLA, making it better than Mamba2 (\resultref{res:6.1-gla}).
\newline
\textbf{Columns 1, 4, 5}: Removing \texttt{conv1d} reduces Mamba2's performance back to match GLA (\resultref{res:7.1-mamba}).
\newline
\textit{Remark.} Synthetic results here predict similar trends in real-life experiments (\resultref{res:12} and \figureref{fig:real-life}).
}
\end{figure}

\section{When Linear Models Meet Canon}
\label{sec:linear-models}

The three base linear models we study—GLA, Mamba2(mlp), and GDN—share a block-wise structure where each block consists of a ``linear attention'' layer (GLA, GDN, or Mamba2) followed by an MLP. This design naturally defines four insertion points for Canon layers, analogous to standard Transformers: \textbf{A} before the linear attention, \textbf{B} inside it, \textbf{C} before the MLP, and \textbf{D} inside. In the following subsections, we analyze each architecture separately.

\subsection{When Linear Attention Meets Canon}
\label{sec:gla-canon}

Linear attention models reduce computation by maintaining a compact state instead of attending over all tokens. In Gated Linear Attention (GLA)~\cite{yang2023gated}, the attention map is updated recursively as
$
W_t = \alpha_t W_{t-1} + v_t k_t^\top,
$
where $W_t \in \mathbb{R}^{d_{\text{key}}\times d_{\text{value}}}$ remains fixed in size regardless of context length. This design is efficient but effectively averages over past tokens, weakening the influence of nearby ones—crucial for reasoning. Canon layers restore localized horizontal context flow, alleviating this limitation and improving reasoning fidelity.

Following the original GLA release, its authors added a \texttt{conv1d}-based enhancement in their GitHub repo—corresponding to our Canon-B variant but using SiLU activation and omitting residual connections. We refer to this as \textbf{GLA conv1d} or equivalently \textbf{GLA+Canon-b}. To show the strongest comparison, our \textbf{Canon-AbCD(res)} extends it by adding residual Canon-ACD layers while keeping their \texttt{conv1d}. We also explore the full Canon-ABCD design in the appendix.

As shown in \figureref{fig:gla-and-mamba}, integrating Canon-AbCD substantially boosts GLA’s original (non-conv1d) performance across all benchmarks, transforming it from a weak baseline into a strong competitor. Despite its simplicity, GLA+Canon matches or surpasses Mamba2, particularly on reasoning breadth (\textsc{Brevo}). This upward trend persists in large-scale real-world pretraining (\sectionref{sec:real-life}), improving nearly all standard evaluation metrics.

\begin{mdframed}
\begin{sresult}{6.1}[\figureref{fig:gla-and-mamba}]
\label{res:6.1-gla}
Adding Canon layers:
\begin{itemize}
    \item
    Dramatically improves GLA’s original performance—raising reasoning depth from 1-hop to 4-hop, doubling reasoning breadth, and more than doubling knowledge manipulation length.
    \item
    Brings GLA on par with or beyond Mamba2, significantly outperforming it on \textsc{Brevo}.
    \item
    Yields additional gains even over the stronger GLA conv1d baseline.
\end{itemize}
\end{sresult}
\end{mdframed}

As in the Transformer case, we perform ablations to determine optimal Canon placement. GLA also supports feature-map variants like $W_t = \alpha_t W_{t-1} + v_t \phi(k_t)^\top$, with the popular choice $\phi(x)=1+\text{elu}(x)$~\cite{katharopoulos2020transformers}. We test Canon compatibility both with and without this feature map.

\begin{mdframed}
\begin{sresult}{6.2}[\figureref{fig:gla-ablation}+\ref{fig:full-gla} on Page~\pageref{fig:gla-ablation}]
\label{res:6.2-gla}
Ablation study on GLA:
\begin{itemize}
    \item \textbf{Residualness.}
    Unlike in full Transformers, Canon residualness is less critical: non-residual variants work better for \textsc{Mano}, while residual ones suit \textsc{Lano}/\textsc{Brevo1}.
    \item \textbf{Positioning.}
    Canon design is \defem{not intrinsic} to the attention layer. Canon-ACD (or even Canon-A/C/D alone) can outperform Canon-b/B on many tasks, and combining all is best.
    \item \textbf{Feature maps.}
    Canon works well with $1+\text{elu}(x)$ feature map, though better without it.%
    \footnote{Consistent with~\cite{yang2023gated}, where original GLA (without Canon) also performed better without feature maps.}
\end{itemize}
\end{sresult}
\end{mdframed}

Overall, these ablations highlight the importance of horizontal information flow independent of the architecture sublayers.
Interested readers can find our full ablation results in \figureref{fig:gla-ablation}+\ref{fig:full-gla}, where we for instance carefully compared Canon-AbCD(res/no-res), Canon-ABCD(res/no-res), and many more. We recommend the \textbf{Canon-AbCD(res)} configuration for GLA—keeping the non-residual \texttt{conv1d} from their original codebase while combining it with our residual, activation-free Canon-ACD. This achieves strong gains with \emph{minimal code changes}.

\subsection{When Mamba Meets Canon}
\label{sec:mamba-canon}

While Mamba2 is recognized as a state-space model (SSM), it quietly includes a non-linear \texttt{conv1d} operation in each SSM block.%
\footnote{Mamba1 also contains this component, but since Mamba2 consistently outperforms it, we report only Mamba2.}
Originally introduced in H3~\cite{fu2022hungry} as a \emph{shift-SSM}, this mechanism effectively acts as a partial Canon-B layer—performing horizontal mixing on selected coordinates, applying non-linear activation, and omitting residual connections.

Surprisingly, this built-in \texttt{conv1d} contributes more to Mamba2’s performance than its SSM formulation itself. Disabling it sharply degrades results, reducing Mamba2 to GLA-level performance on both synthetic (\figureref{fig:gla-and-mamba}) and real-world datasets (\sectionref{sec:real-life}). This raises a key question: is Mamba2’s strength primarily due to its Canon-like \texttt{conv1d} rather than the state-space mechanism?

To isolate this effect, we refer to Mamba2’s internal \texttt{conv1d} as \textbf{Canon-b}, and extend it by adding residual Canon-A/C/D layers—denoted \textbf{Mamba2(mlp)+Canon-AbCD}. We also test our own Canon-B design in later ablations and in the appendix.\footnote{For example, with \texttt{ssm\_state\_size=64} and \texttt{num\_heads=16}, our Canon-B applies to all $4d+144$ intermediate coordinates for hidden size $d$, whereas Mamba2’s original \texttt{conv1d} acts only on a subset ($2d+o(d)$) with activation.}
We additionally examine Mamba2 without MLP layers (which exposes Canon-A/B positions), reported in the appendix.\footnote{Such Mamba2 doubles the layer count and recurrent state size compared to Mamba2(mlp). In practice, Mamba2(mlp) is preferred, e.g., in Falcon-H1~\cite{tiifalconh1}.}

\begin{figure}[t!]
\centering
\setlength{\imgwidthBase}{0.2\textwidth}
\vspace{-3mm}\hspace*{-5mm}
\includegraphics[page=1,trim={2.5mm 1.5mm 2.5mm 1.5mm},clip,width=\imgwidthBase]{perm_4/Mamba2_mlp_-original_conv1d_}
\includegraphics[page=1,trim={2.5mm 1.5mm 2.5mm 1.5mm},clip,width=\imgwidthBase]{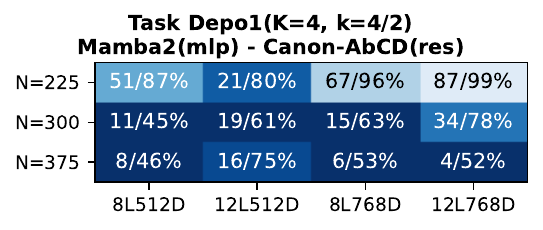}
\includegraphics[page=1,trim={2.5mm 1.5mm 2.5mm 1.5mm},clip,width=\imgwidthBase]{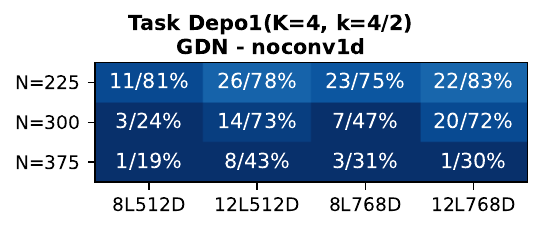}
\includegraphics[page=1,trim={2.5mm 1.5mm 2.5mm 1.5mm},clip,width=\imgwidthBase]{perm_4/GDN-original_conv1d_}
\includegraphics[page=1,trim={2.5mm 1.5mm 2.5mm 1.5mm},clip,width=\imgwidthBase]{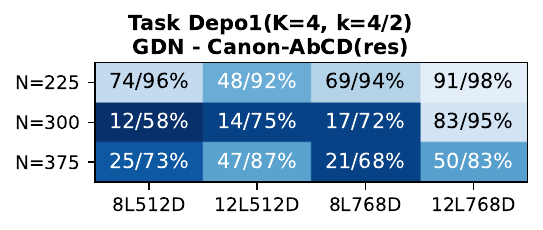}
\hspace*{-5mm}
\\
\hspace*{-5mm}
\includegraphics[page=1,trim={2.5mm 1.5mm 2.5mm 1.5mm},clip,width=\imgwidthBase]{perm_multi_4/Mamba2_mlp_-original_conv1d_}
\includegraphics[page=1,trim={2.5mm 1.5mm 2.5mm 1.5mm},clip,width=\imgwidthBase]{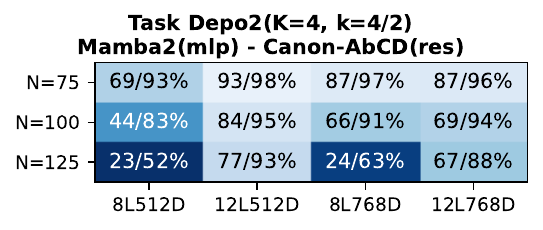}
\includegraphics[page=1,trim={2.5mm 1.5mm 2.5mm 1.5mm},clip,width=\imgwidthBase]{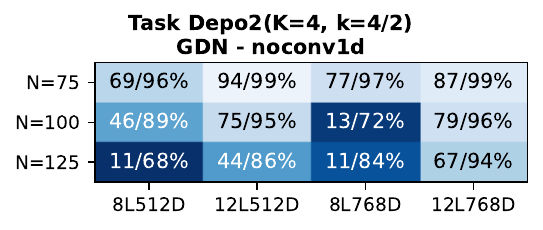}
\includegraphics[page=1,trim={2.5mm 1.5mm 2.5mm 1.5mm},clip,width=\imgwidthBase]{perm_multi_4/GDN-original_conv1d_}
\includegraphics[page=1,trim={2.5mm 1.5mm 2.5mm 1.5mm},clip,width=\imgwidthBase]{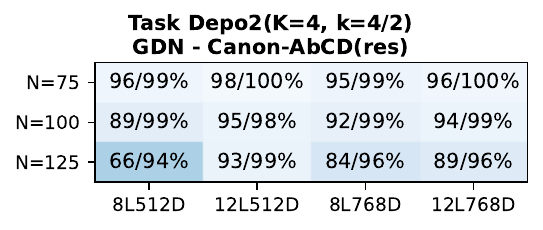}
\hspace*{-5mm}
\\
\hspace*{-5mm}
\includegraphics[page=1,trim={2.5mm 1.5mm 2.5mm 1.5mm},clip,width=\imgwidthBase]{top_sort/Mamba2_mlp_-original_conv1d_}
\includegraphics[page=1,trim={2.5mm 1.5mm 2.5mm 1.5mm},clip,width=\imgwidthBase]{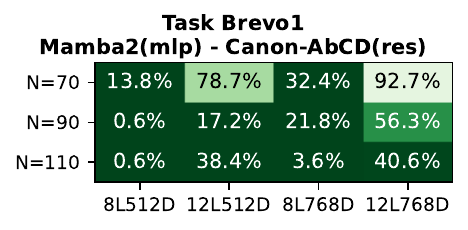}
\includegraphics[page=1,trim={2.5mm 1.5mm 2.5mm 1.5mm},clip,width=\imgwidthBase]{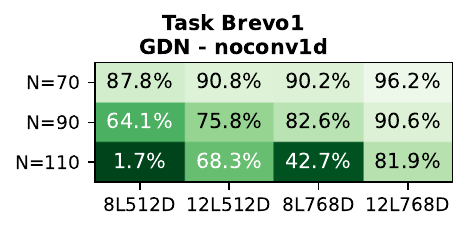}
\includegraphics[page=1,trim={2.5mm 1.5mm 2.5mm 1.5mm},clip,width=\imgwidthBase]{top_sort/GDN-original_conv1d_}
\includegraphics[page=1,trim={2.5mm 1.5mm 2.5mm 1.5mm},clip,width=\imgwidthBase]{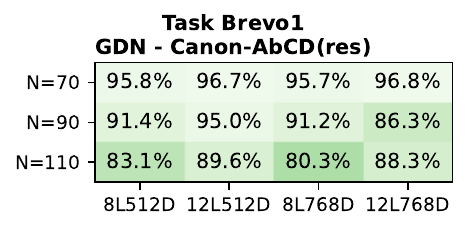}
\hspace*{-5mm}
\\
\hspace*{-5mm}
\includegraphics[page=1,trim={2.5mm 1.5mm 2.5mm 1.5mm},clip,width=\imgwidthBase]{top_sort_multi/Mamba2_mlp_-original_conv1d_}
\includegraphics[page=1,trim={2.5mm 1.5mm 2.5mm 1.5mm},clip,width=\imgwidthBase]{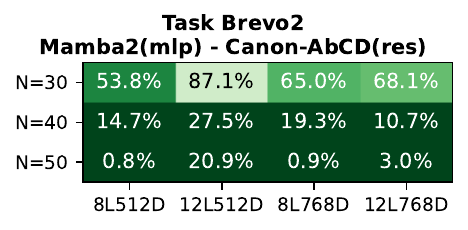}
\includegraphics[page=1,trim={2.5mm 1.5mm 2.5mm 1.5mm},clip,width=\imgwidthBase]{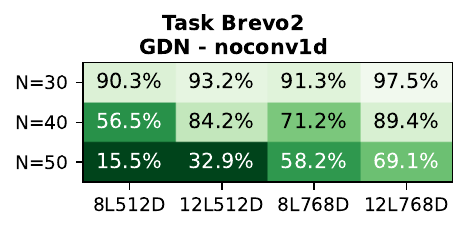}
\includegraphics[page=1,trim={2.5mm 1.5mm 2.5mm 1.5mm},clip,width=\imgwidthBase]{top_sort_multi/GDN-original_conv1d_}
\includegraphics[page=1,trim={2.5mm 1.5mm 2.5mm 1.5mm},clip,width=\imgwidthBase]{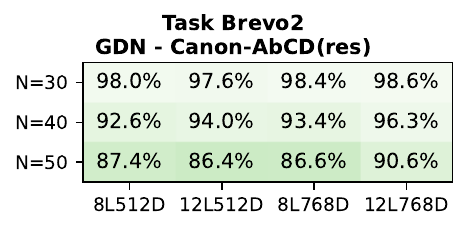}
\hspace*{-5mm}
\\
\hspace*{-5mm}
\includegraphics[page=1,trim={2.5mm 1.5mm 2.5mm 1.5mm},clip,width=\imgwidthBase]{biocap/mamba-mamba-mlp}
\includegraphics[page=1,trim={2.5mm 1.5mm 2.5mm 1.5mm},clip,width=\imgwidthBase]{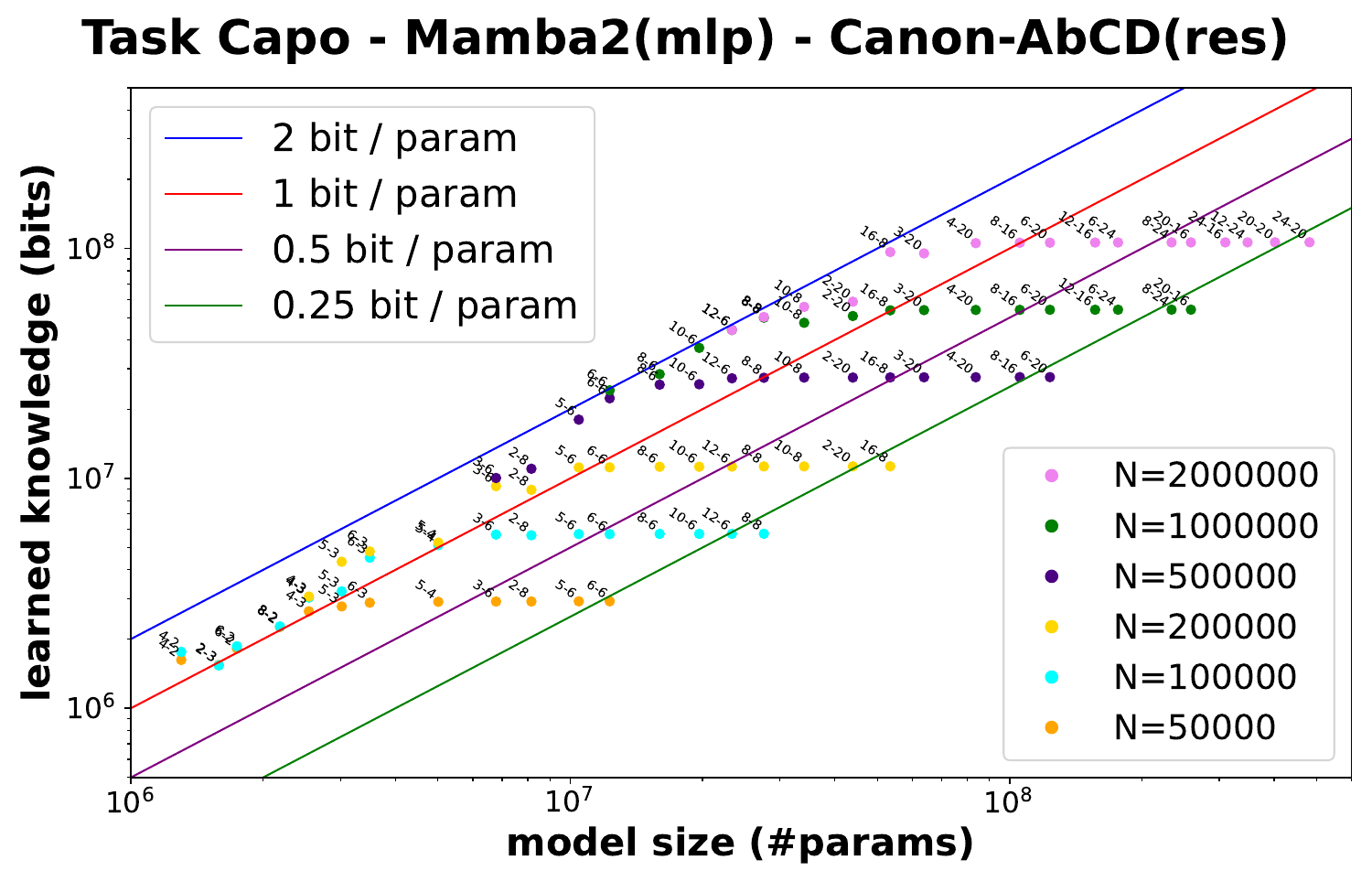}
\includegraphics[page=1,trim={2.5mm 1.5mm 2.5mm 1.5mm},clip,width=\imgwidthBase]{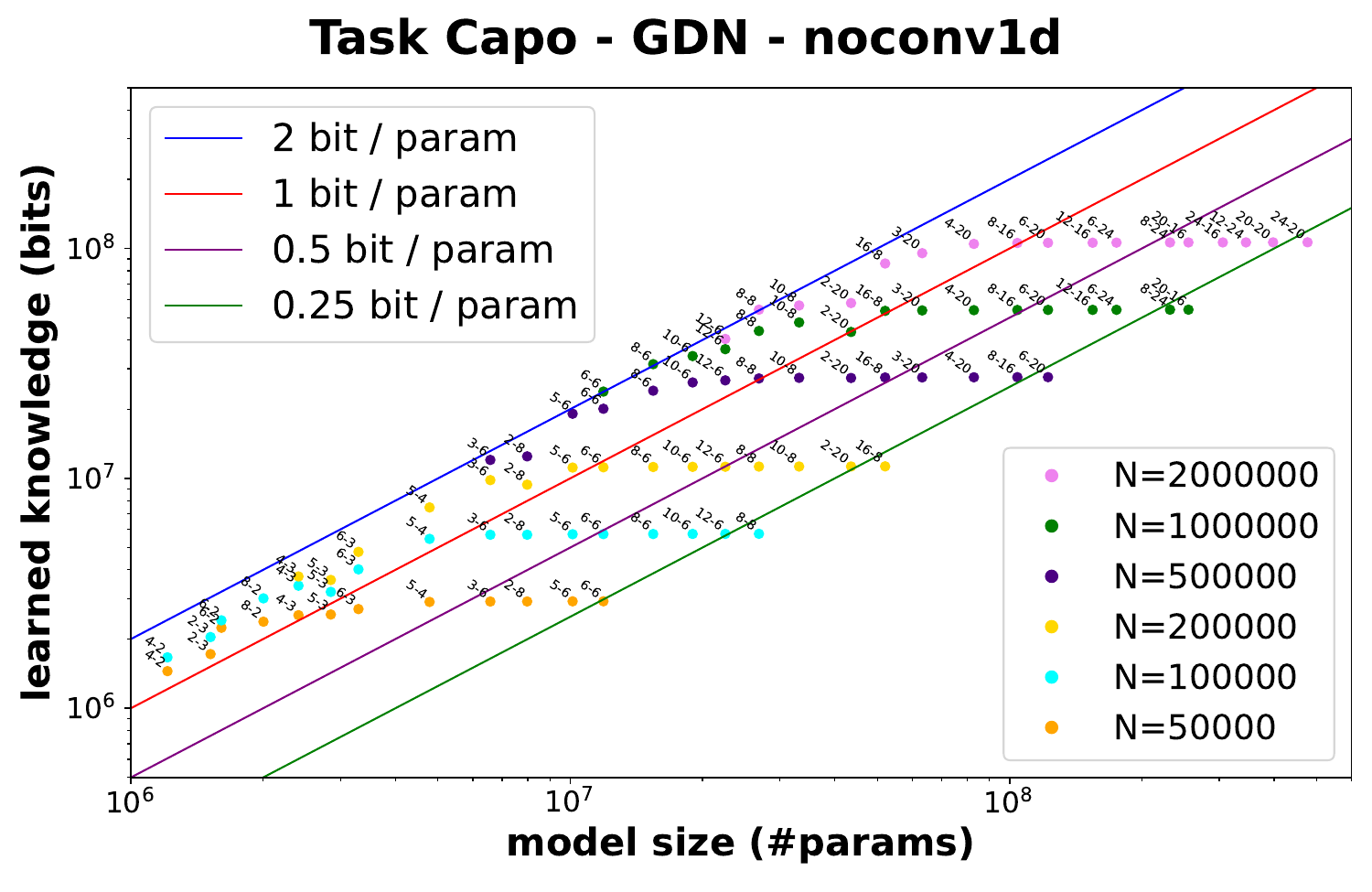}
\includegraphics[page=1,trim={2.5mm 1.5mm 2.5mm 1.5mm},clip,width=\imgwidthBase]{biocap/gdn-original}
\includegraphics[page=1,trim={2.5mm 1.5mm 2.5mm 1.5mm},clip,width=\imgwidthBase]{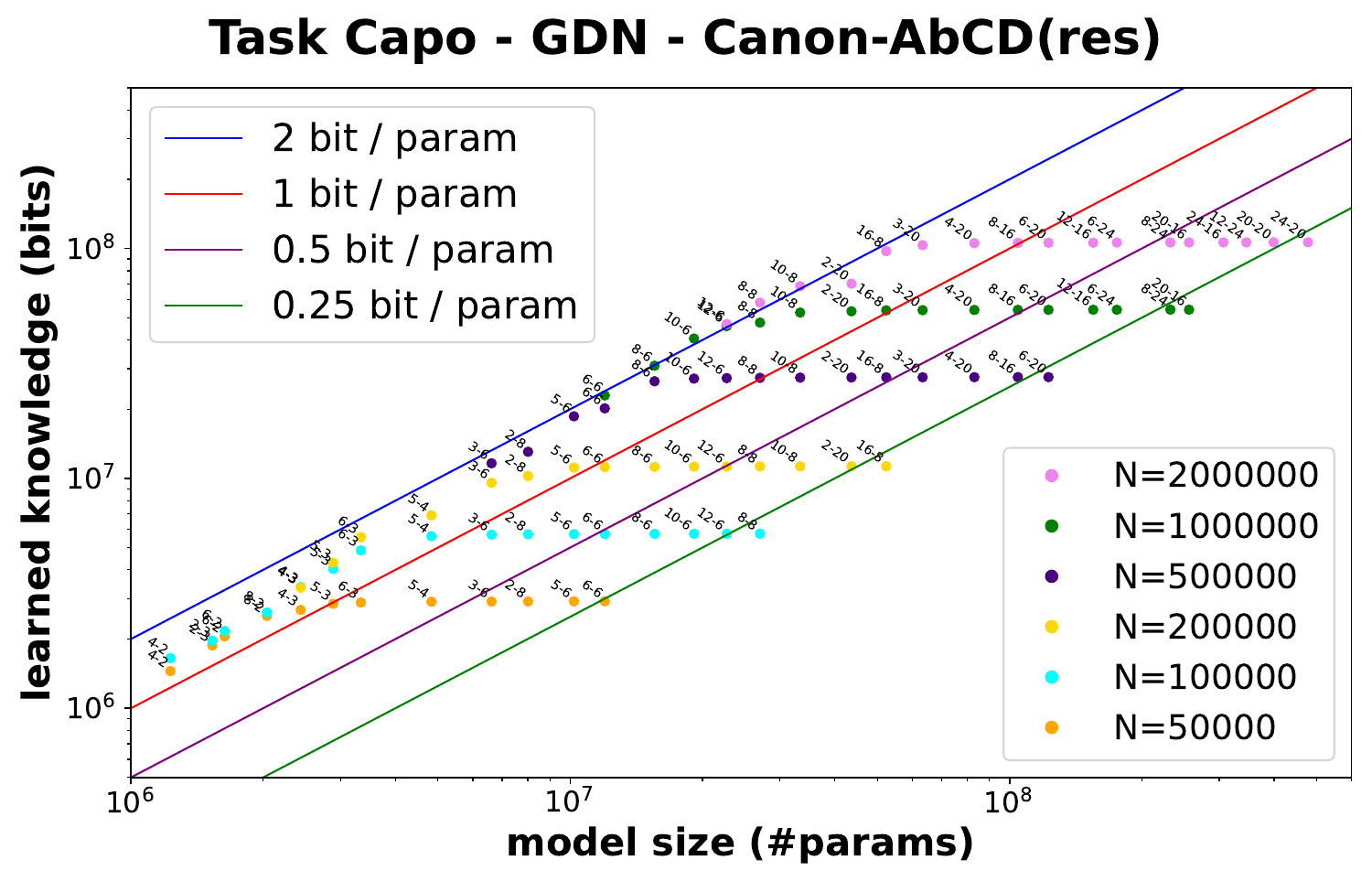}
\hspace*{-5mm}
\\
\hspace*{-5mm}
\includegraphics[page=1,trim={2.5mm 1.5mm 2.5mm 1.5mm},clip,width=\imgwidthBase]{arith/Mamba2_mlp_-original_conv1d_}
\includegraphics[page=1,trim={2.5mm 1.5mm 2.5mm 1.5mm},clip,width=\imgwidthBase]{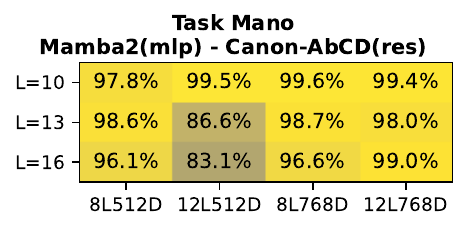}
\includegraphics[page=1,trim={2.5mm 1.5mm 2.5mm 1.5mm},clip,width=\imgwidthBase]{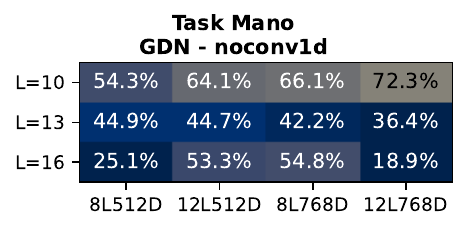}
\includegraphics[page=1,trim={2.5mm 1.5mm 2.5mm 1.5mm},clip,width=\imgwidthBase]{arith/GDN-original_conv1d_}
\includegraphics[page=1,trim={2.5mm 1.5mm 2.5mm 1.5mm},clip,width=\imgwidthBase]{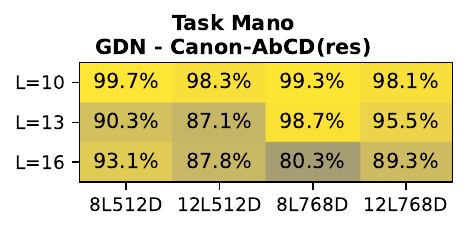}
\hspace*{-5mm}
\\
\hspace*{-5mm}
\includegraphics[page=1,trim={2.5mm 1.5mm 2.5mm 1.5mm},clip,width=\imgwidthBase]{cfg/Mamba2_mlp_-original_conv1d_}
\includegraphics[page=1,trim={2.5mm 1.5mm 2.5mm 1.5mm},clip,width=\imgwidthBase]{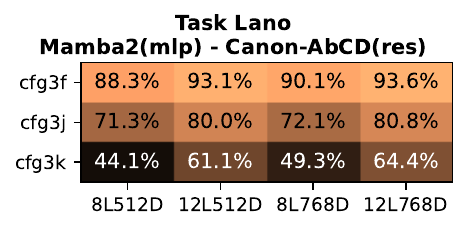}
\includegraphics[page=1,trim={2.5mm 1.5mm 2.5mm 1.5mm},clip,width=\imgwidthBase]{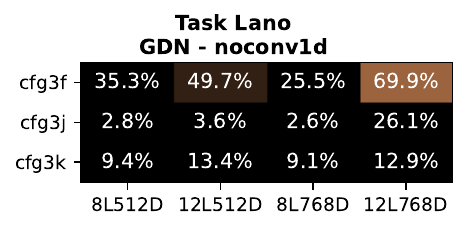}
\includegraphics[page=1,trim={2.5mm 1.5mm 2.5mm 1.5mm},clip,width=\imgwidthBase]{cfg/GDN-original_conv1d_}
\includegraphics[page=1,trim={2.5mm 1.5mm 2.5mm 1.5mm},clip,width=\imgwidthBase]{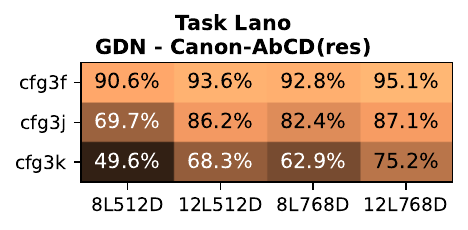}
\hspace*{-5mm}
\caption{\label{fig:mamba-main}%
\textbf{Mamba(mlp) and GDN architectures} with no \texttt{conv1d}, with \texttt{conv1d} (original), and with full Canon.
}
\end{figure}

As shown in \figureref{fig:mamba-main}, adding Canon-AbCD further improves Mamba2(mlp) performance over the built-in \texttt{conv1d} (Canon-b), especially on \textsc{Mano} and \textsc{Lano}.

\begin{mdframed}
\begin{sresult}{7.1}[\figureref{fig:gla-and-mamba}+\ref{fig:mamba-main}]
\label{res:7.1-mamba}
Key observations on Mamba2:
\begin{itemize}
    \item Mamba2 includes an internal non-linear \texttt{conv1d} (partial Canon-B) that contributes more to performance than the SSM itself. Removing it drops performance to GLA levels.
    \item Replacing this with full Canon-AbCD layers further improves, notably on \textsc{Mano}, \textsc{Lano}.
\end{itemize}
\end{sresult}
\end{mdframed}

\noindent
(Mamba1~\cite{gu2023mamba} shows similar trends but is consistently outperformed by Mamba2 in our playground.)

To further understand Canon–Mamba interactions, we perform ablations varying Canon position, residualness, and initialization. Results mirror GLA: Canon layers remain effective even when placed outside the SSM block, showing that horizontal information flow is architecture-independent.

For initialization, we test the recent \emph{mimetic initialization}~\cite{trockman2024mimetic}, proposed to enhance associative recall and length generalization. However, our experiments (\figureref{fig:mamba-ablation}+\ref{fig:full-mamba2}) find no measurable benefit—and often degradation—on other tasks, suggesting that mimetic init may have overfit length generalization at the cost of broader reasoning. These findings highlight the \emph{\bf importance} of evaluating architectural choices over a \emph{{\bf diverse} synthetic playground}.

\begin{mdframed}
\begin{sresult}{7.2}[\figureref{fig:mamba-ablation}+\ref{fig:full-mamba2-mlp} on Page~\pageref{fig:mamba-ablation}]
\label{res:7.2-mamba}
Ablation study on Mamba2(mlp):
\begin{itemize}
    \item Mamba2(mlp) slightly prefers residual Canon for \textsc{Lano}, but non-residual for \textsc{Mano}.
    \item Canon layers stay effective outside the SSM block; e.g., Canon-ACD surpasses Mamba2(conv1d) on \textsc{Depo2}/\textsc{Lano}, highlighting their strength as general horizontal-mixing modules.
    \item Mimetic initialization~\cite{trockman2024mimetic}, designed for length generalization, harms shorter-context performance, reinforcing the need for diverse-task evaluation.
\end{itemize}
\end{sresult}
\end{mdframed}

\noindent
We also evaluate Mamba2 without MLP layers (\figureref{fig:mamba-ablation}+\ref{fig:full-mamba2}); results remain consistent with those above. Interested readers can refer to \figureref{fig:mamba-ablation}+\ref{fig:full-mamba2}+\ref{fig:full-mamba2-mlp} for complete ablation results, including detailed comparisons between Canon-ABCD(res/no-res), Canon-AbCD(res/no-res) and many more. Our overall recommendation remains \textbf{Canon-AbCD(res)} for simplicity.

\subsection{When Gated DeltaNet Meets Canon}
\label{sec:gdn}

Gated DeltaNet (GDN)~\cite{yang2024gated} extends GLA with a gated delta-rule update. Instead of GLA’s $W_t = \alpha_t W_{t-1} + v_t k_t^\top$, GDN adopts $W_t = \alpha_t W_{t-1}(I - \beta_t k_t k_t^\top) + \beta_t v_t k_t^\top$, where $\beta_t$ controls the balance between forgetting and writing. This formulation retains GLA’s efficiency while adaptively suppressing redundant information, allegedly yielding better reasoning and improved gradient flow.

Each GDN block retains the linear-attention–plus–MLP structure but also includes a non-residual, activated \texttt{conv1d} layer within its linear attention sublayer—referred to here as \textbf{conv1d} or \textbf{Canon-b}. This component remains important, though less critical than in GLA or Mamba2. Removing it \emph{destroys} knowledge manipulation (\textsc{Mano}) and hierarchical reasoning (\textsc{Lano}), while in-context reasoning (\textsc{Depo}/\textsc{Brevo}) is largely unaffected. (\sectionref{sec:real-life} later shows such differences may vanish in academic-scale real-life pretraining, \emph{highlighting the importance of a versatile synthetic pretrain playground}.)

Following prior sections, we extend GDN by adding residual Canon-A/C/D layers, forming \textbf{GDN+Canon-AbCD}. We also test our own Canon-B design in later ablations and the appendix. As shown in \figureref{fig:mamba-main}, Canon-AbCD slightly improves GDN+\texttt{conv1d} across benchmarks.

\begin{mdframed}
\begin{sresult}{8.1}[\figureref{fig:mamba-main}]
\label{res:gdn.1}
Key observations on GDN:
\begin{itemize}
    \item GDN is less dependent on its internal \texttt{conv1d} (Canon-b) for strong performance.
    \item Replacing it with full Canon-AbCD layers still yields improvements, albeit marginal.
\end{itemize}
\end{sresult}
\end{mdframed}

We further perform ablation studies on Canon positioning and residualness:
\begin{mdframed}
\begin{sresult}{8.2}[\figureref{fig:gdn-ablation}+\ref{fig:full-gdn} on Page~\pageref{fig:gdn-ablation}]
\label{res:gdn.2}
Ablation studies on GDN:
\begin{itemize}
    \item GDN slightly prefers non-residual Canon on \textsc{Mano}, though overall differences are minor.
    \item Canon layers remain effective even outside the GDN layer; e.g., Canon-ACD performs on par with GDN+\texttt{conv1d}, underscoring their generality as horizontal-mixing components.
\end{itemize}
\end{sresult}
\end{mdframed}

\noindent
Interested readers can refer to \figureref{fig:gdn-ablation}+\ref{fig:full-gdn} for full ablation results, including detailed comparisons among Canon-ABCD(res/no-res), Canon-AbCD(res/no-res), and others. For simplicity and consistency, we recommend \textbf{Canon-AbCD(res)} as the default configuration.

\section{Final Comparisons and Lessons for Architecture Design}
\label{sec:final-compare}

\begin{figure}[t!]
\centering
\setlength{\imgwidthBase}{0.19\textwidth}
\vspace{-3mm}%
\includegraphics[page=1,trim={2.5mm 1.5mm 2.5mm 1.5mm},clip,width=\imgwidthBase]{perm/Llama_RoPE_-Res-______Canon-ABCD}
\includegraphics[page=1,trim={2.5mm 1.5mm 2.5mm 1.5mm},clip,width=\imgwidthBase]{perm/Llama_NoPE_-Res-Canon-ABCD}
\includegraphics[page=1,trim={2.5mm 1.5mm 2.5mm 1.5mm},clip,width=\imgwidthBase]{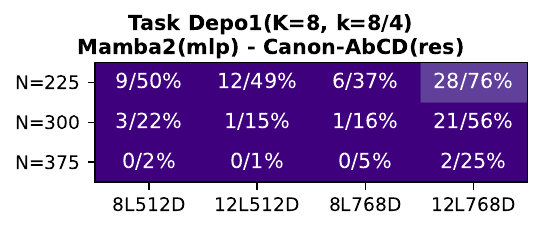}
\includegraphics[page=1,trim={2.5mm 1.5mm 2.5mm 1.5mm},clip,width=\imgwidthBase]{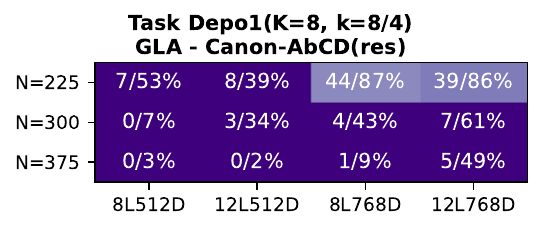}
\includegraphics[page=1,trim={2.5mm 1.5mm 2.5mm 1.5mm},clip,width=\imgwidthBase]{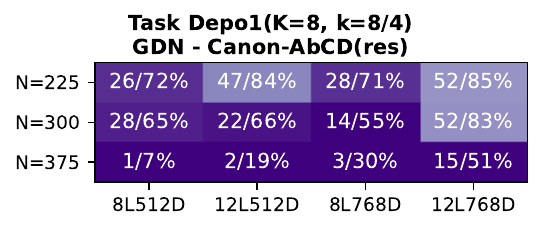}
\\
\hspace{\imgwidthBase}
\hspace{\imgwidthBase}
\includegraphics[page=1,trim={2.5mm 1.5mm 2.5mm 1.5mm},clip,width=\imgwidthBase]{perm_4/Mamba2_mlp_-Res-Canon-AbbCD}
\includegraphics[page=1,trim={2.5mm 1.5mm 2.5mm 1.5mm},clip,width=\imgwidthBase]{perm_4/GLA-Res-Canon-AbbCD}
\includegraphics[page=1,trim={2.5mm 1.5mm 2.5mm 1.5mm},clip,width=\imgwidthBase]{perm_4/GDN-Res-Canon-AbbCD}
\\
\includegraphics[page=1,trim={2.5mm 1.5mm 2.5mm 1.5mm},clip,width=\imgwidthBase]{perm_multi/Llama_RoPE_-Res-______Canon-ABCD}
\includegraphics[page=1,trim={2.5mm 1.5mm 2.5mm 1.5mm},clip,width=\imgwidthBase]{perm_multi/Llama_NoPE_-Res-Canon-ABCD}
\includegraphics[page=1,trim={2.5mm 1.5mm 2.5mm 1.5mm},clip,width=\imgwidthBase]{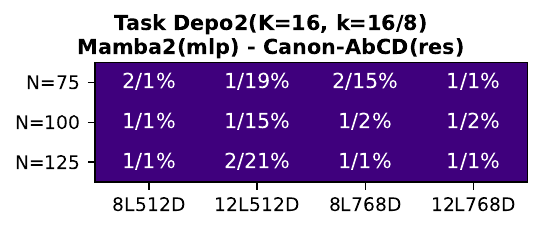}
\includegraphics[page=1,trim={2.5mm 1.5mm 2.5mm 1.5mm},clip,width=\imgwidthBase]{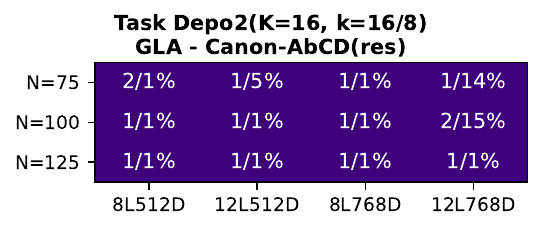}
\includegraphics[page=1,trim={2.5mm 1.5mm 2.5mm 1.5mm},clip,width=\imgwidthBase]{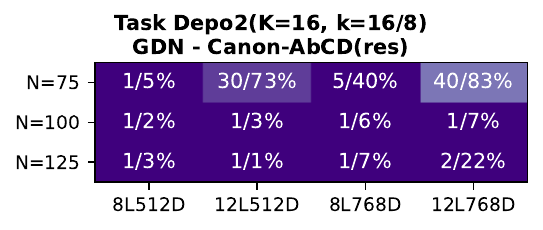}
\\
\hspace{\imgwidthBase}
\hspace{\imgwidthBase}
\includegraphics[page=1,trim={2.5mm 1.5mm 2.5mm 1.5mm},clip,width=\imgwidthBase]{perm_multi_4/Mamba2_mlp_-Res-Canon-AbbCD}
\includegraphics[page=1,trim={2.5mm 1.5mm 2.5mm 1.5mm},clip,width=\imgwidthBase]{perm_multi_4/GLA-Res-Canon-AbbCD}
\includegraphics[page=1,trim={2.5mm 1.5mm 2.5mm 1.5mm},clip,width=\imgwidthBase]{perm_multi_4/GDN-Res-Canon-AbbCD}
\\
\includegraphics[page=1,trim={2.5mm 1.5mm 2.5mm 1.5mm},clip,width=\imgwidthBase]{top_sort/Llama_RoPE_-Res-______Canon-ABCD}
\includegraphics[page=1,trim={2.5mm 1.5mm 2.5mm 1.5mm},clip,width=\imgwidthBase]{top_sort/Llama_NoPE_-Res-Canon-ABCD}
\includegraphics[page=1,trim={2.5mm 1.5mm 2.5mm 1.5mm},clip,width=\imgwidthBase]{top_sort/Mamba2_mlp_-Res-Canon-AbbCD}
\includegraphics[page=1,trim={2.5mm 1.5mm 2.5mm 1.5mm},clip,width=\imgwidthBase]{top_sort/GLA-Res-Canon-AbbCD}
\includegraphics[page=1,trim={2.5mm 1.5mm 2.5mm 1.5mm},clip,width=\imgwidthBase]{top_sort/GDN-Res-Canon-AbbCD}
\\
\includegraphics[page=1,trim={2.5mm 1.5mm 2.5mm 1.5mm},clip,width=\imgwidthBase]{top_sort_multi/Llama_RoPE_-Res-______Canon-ABCD}
\includegraphics[page=1,trim={2.5mm 1.5mm 2.5mm 1.5mm},clip,width=\imgwidthBase]{top_sort_multi/Llama_NoPE_-Res-Canon-ABCD}
\includegraphics[page=1,trim={2.5mm 1.5mm 2.5mm 1.5mm},clip,width=\imgwidthBase]{top_sort_multi/Mamba2_mlp_-Res-Canon-AbbCD}
\includegraphics[page=1,trim={2.5mm 1.5mm 2.5mm 1.5mm},clip,width=\imgwidthBase]{top_sort_multi/GLA-Res-Canon-AbbCD}
\includegraphics[page=1,trim={2.5mm 1.5mm 2.5mm 1.5mm},clip,width=\imgwidthBase]{top_sort_multi/GDN-Res-Canon-AbbCD}
\\
\includegraphics[page=1,trim={2.5mm 1.5mm 2.5mm 1.5mm},clip,width=\imgwidthBase]{biocap/Llama-quarter-ABCD}
\includegraphics[page=1,trim={2.5mm 1.5mm 2.5mm 1.5mm},clip,width=\imgwidthBase]{biocap/nope-ABCD}
\includegraphics[page=1,trim={2.5mm 1.5mm 2.5mm 1.5mm},clip,width=\imgwidthBase]{biocap/mamba-mamba-mlp-AbbCD}
\includegraphics[page=1,trim={2.5mm 1.5mm 2.5mm 1.5mm},clip,width=\imgwidthBase]{biocap/gla-AbbCD}
\includegraphics[page=1,trim={2.5mm 1.5mm 2.5mm 1.5mm},clip,width=\imgwidthBase]{biocap/gdn-AbbCD}
\\
\includegraphics[page=1,trim={2.5mm 1.5mm 2.5mm 1.5mm},clip,width=\imgwidthBase]{arith/Llama_RoPE_-Res-______Canon-ABCD}
\includegraphics[page=1,trim={2.5mm 1.5mm 2.5mm 1.5mm},clip,width=\imgwidthBase]{arith/Llama_NoPE_-Res-Canon-ABCD}
\includegraphics[page=1,trim={2.5mm 1.5mm 2.5mm 1.5mm},clip,width=\imgwidthBase]{arith/Mamba2_mlp_-Res-Canon-AbbCD}
\includegraphics[page=1,trim={2.5mm 1.5mm 2.5mm 1.5mm},clip,width=\imgwidthBase]{arith/GLA-Res-Canon-AbbCD}
\includegraphics[page=1,trim={2.5mm 1.5mm 2.5mm 1.5mm},clip,width=\imgwidthBase]{arith/GDN-Res-Canon-AbbCD}
\\
\includegraphics[page=1,trim={2.5mm 1.5mm 2.5mm 1.5mm},clip,width=\imgwidthBase]{cfg/Llama_RoPE_-Res-______Canon-ABCD}
\includegraphics[page=1,trim={2.5mm 1.5mm 2.5mm 1.5mm},clip,width=\imgwidthBase]{cfg/Llama_NoPE_-Res-Canon-ABCD}
\includegraphics[page=1,trim={2.5mm 1.5mm 2.5mm 1.5mm},clip,width=\imgwidthBase]{cfg/Mamba2_mlp_-Res-Canon-AbbCD}
\includegraphics[page=1,trim={2.5mm 1.5mm 2.5mm 1.5mm},clip,width=\imgwidthBase]{cfg/GLA-Res-Canon-AbbCD}
\includegraphics[page=1,trim={2.5mm 1.5mm 2.5mm 1.5mm},clip,width=\imgwidthBase]{cfg/GDN-Res-Canon-AbbCD}
\caption{\label{fig:final-compare}%
\textbf{Final comparison of base architectures} equipped with full-score Canon layers: RoPE(\musQuarter), NoPE, Mamba2, GLA and GDN. Most notably, with Canon layers added, Mamba2/GLA/GDN still underperform Transformers by $2\times$ in reasoning depth, with meaningful results only for \textsc{Depo}(K=4).
}
\end{figure}

Applying Canon uniformly across all architectures creates a controlled environment—like dropping them from the same height at the Tower of Pisa—revealing their \emph{true} architectural trade-offs. We exclude hybrid models (e.g., Griffin~\cite{de2024griffin}, Samba~\cite{ren2024samba}) to isolate behaviors of the \emph{base} architectures.

\subsection{Summary on Linear Models vs. Canon Layers}
\label{sec:linear-summary}

While many more linear-time architectures remain worth exploring, this study focuses on GLA, Mamba2, and GDN.%
\footnote{GDN results were newly added after the NeurIPS 2025 accepted version (V1.1).}
Despite their structural differences, several consistent insights emerge.

\begin{mdframed}
\begin{sresult}{9}[\sectionref{sec:linear-models}+\figureref{fig:final-compare}]
\label{res:9-linear}
Summary of Canon effects on linear models:
\begin{itemize}
    \item \textbf{Universality.} Canon-ACD already matches internal \texttt{conv1d}, showing that horizontal mixing is \defem{useful across all sublayers}, not limited to linear attention (i.e., the recurrent / SSM layer).
    \item \textbf{Robustness.} Adding Canon layers \defem{never hurts}; the residual design stabilizes training.
    \item \textbf{Sufficiency.} Most performance appears achievable with the simplest \defem{GLA+Canon-AbCD}, suggesting the current direction of linear-model architecture design may warrant re-evaluation.
\end{itemize}
\end{sresult}
\end{mdframed}

To elaborate more on the third bullet, modern models (Mamba2, GDN) show only marginal gains over the simple \textsf{GLA+Canon-AbCD} baseline.
This suggests that many recent architectural innovations may largely \emph{replicate Canon-like horizontal mixing} rather than introduce fundamentally new computation.
While such mechanisms can reduce explicit reliance on Canon layers, their improvements remain limited—raising the question of whether increasing architectural complexity truly expands capability or merely redistributes existing functions.%
\footnote{In our follow-up work~\cite{PhysicsLM42}, we show that Canon layers can lift GLA to match GDN (+ full Canon) even on 1B- to 8B-sized models pretrained using real-life data, further strengthening this point.}

\begin{remark}
We do not claim that ``replicating'' Canon is unworthy—such designs may improve efficiency and reduce GPU memory.
However, it is crucial to understand what the model actually learns: complex module designs \emph{need not realize} complex functions, as optimizers may often converge to simpler functions (e.g., Canon-like solutions in this case).%
\footnote{The same holds broadly in deep learning: although an $\ell$-layer quadratic MLP can represent a $2^{\ell}$-degree parity function, learning it is computationally intractable. Existence rarely implies learnability via training~\cite{allen2020backward}.}
\end{remark}

\subsection{Summary on Transformer vs. Linear Models}
\label{sec:quadratic-summary}

We now compare Transformers and linear models under a \emph{controlled, apple-to-apples} setting with full Canon layers added to all architectures.

\begin{mdframed}
\begin{sresult}{10}[\figureref{fig:final-compare}]
\label{res:10}
With full-score Canon layers added, we find:
\begin{itemize}
\item reasoning depth: RoPE(\musQuarter) $\approx$ NoPE {\color{red} $\gg$} Mamba2 $\approx$ GLA $\approx$ GDN  (e.g., {\color{red}4× deeper reasoning});
\item reasoning breadth: GDN $\geq$ RoPE(\musQuarter{}) $\approx$ NoPE $\approx$ GLA $>$ Mamba2;
\item knowledge capacity: Mamba2 $\approx$ GLA $\approx$ GDN {\color{red} $\gg$} RoPE(\musQuarter) $\approx$ NoPE (e.g., {\color{red}1.4× capacity});
\item knowledge manipulation: Mamba2 $\approx$ RoPE(\musQuarter) $\approx$ NoPE $\approx$ GDN $>$ GLA;
\item hierarchical structure: RoPE(\musQuarter) {\color{red} $>$} NoPE $\approx$ Mamba2 $\approx$ GLA $\approx$ GDN.
\end{itemize}
\end{sresult}
\end{mdframed}

\begin{remark}
The initial comparison (\figureref{fig:initial}) was not controlled: Mamba2 and GDN included internal \texttt{conv1d} layers, whereas GLA and Transformers did not.
By adding full Canon (Canon-ABCD or -AbCD) layers to all, the comparison becomes scientifically meaningful.
\end{remark}

While others may interpret the fine differences across architectures, we focus here on the most pronounced contrasts.
First, linear models—regardless of design—consistently show a $\sim$40\% gain in \textsc{Capo} knowledge capacity compared to full Transformers.
This is intuitive: their recurrent structure better supports associative-memory representations (an existential proof), and more importantly, optimizers can \emph{learn} such representations effectively in practice.

More surprising is the behavior on reasoning depth.
Linear models remain systematically weaker—about 2× on \textsc{Depo1} (8-hop vs.\ 4-hop) and up to 4× on \textsc{Depo2} (16-hop vs.\ 4-hop)—even under identical training conditions.
We next examine this phenomenon in detail.

\begin{figure}[t!]
\centering
\setlength{\imgwidthBase}{0.32\textwidth}
\includegraphics[page=1,trim={2.5mm 1.5mm 2.5mm 1.5mm},clip,width=\imgwidthBase]{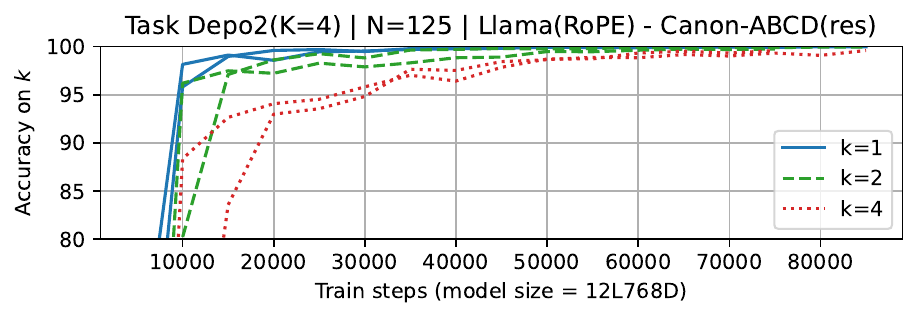}
\includegraphics[page=1,trim={2.5mm 1.5mm 2.5mm 1.5mm},clip,width=\imgwidthBase]{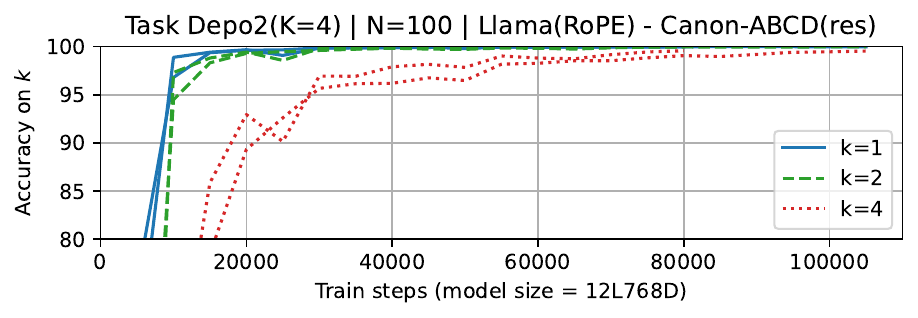}
\includegraphics[page=1,trim={2.5mm 1.5mm 2.5mm 1.5mm},clip,width=\imgwidthBase]{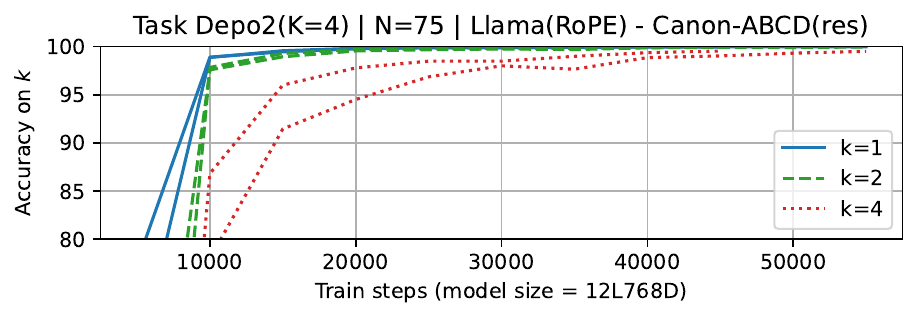}
\\
\includegraphics[page=1,trim={2.5mm 1.5mm 2.5mm 1.5mm},clip,width=\imgwidthBase]{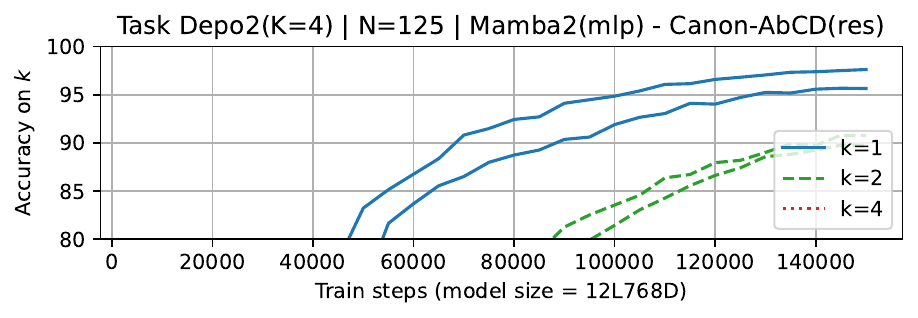}
\includegraphics[page=1,trim={2.5mm 1.5mm 2.5mm 1.5mm},clip,width=\imgwidthBase]{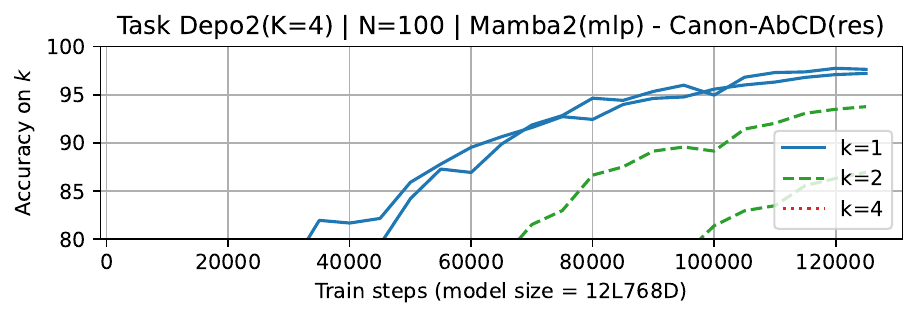}
\includegraphics[page=1,trim={2.5mm 1.5mm 2.5mm 1.5mm},clip,width=\imgwidthBase]{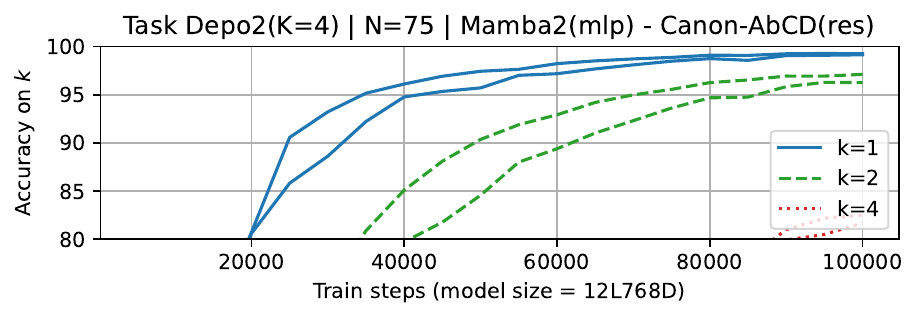}
\\
\includegraphics[page=1,trim={2.5mm 1.5mm 2.5mm 1.5mm},clip,width=\imgwidthBase]{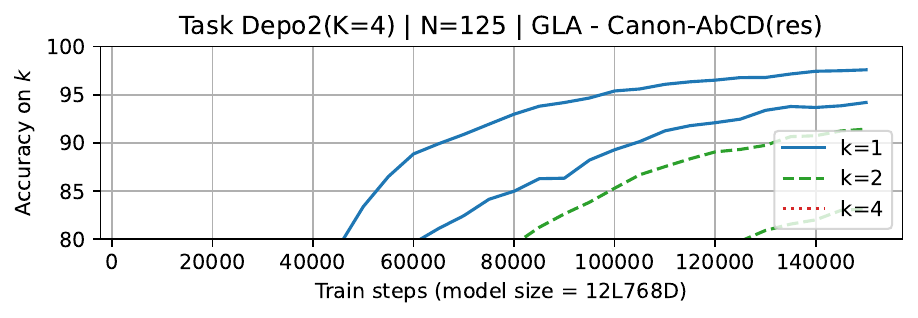}
\includegraphics[page=1,trim={2.5mm 1.5mm 2.5mm 1.5mm},clip,width=\imgwidthBase]{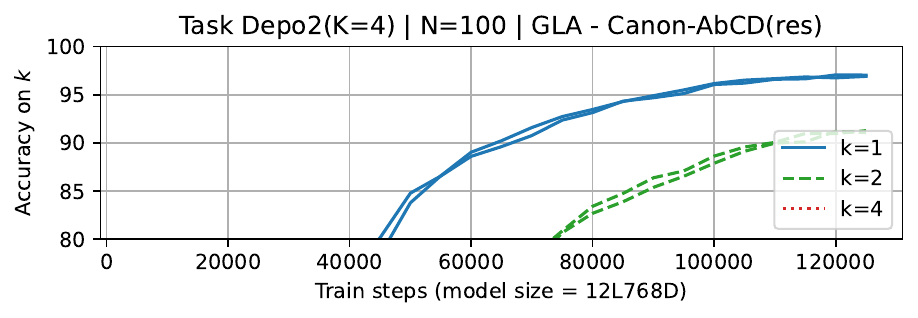}
\includegraphics[page=1,trim={2.5mm 1.5mm 2.5mm 1.5mm},clip,width=\imgwidthBase]{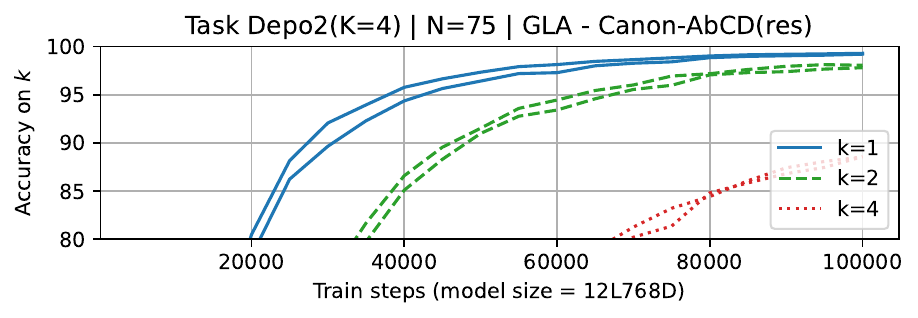}
\\
\includegraphics[page=1,trim={2.5mm 1.5mm 2.5mm 1.5mm},clip,width=\imgwidthBase]{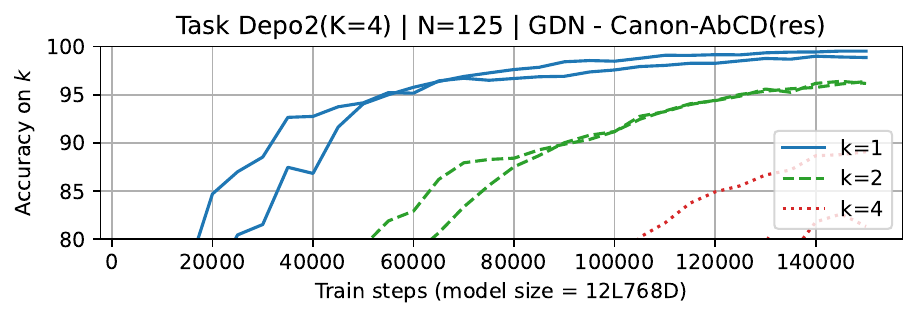}
\includegraphics[page=1,trim={2.5mm 1.5mm 2.5mm 1.5mm},clip,width=\imgwidthBase]{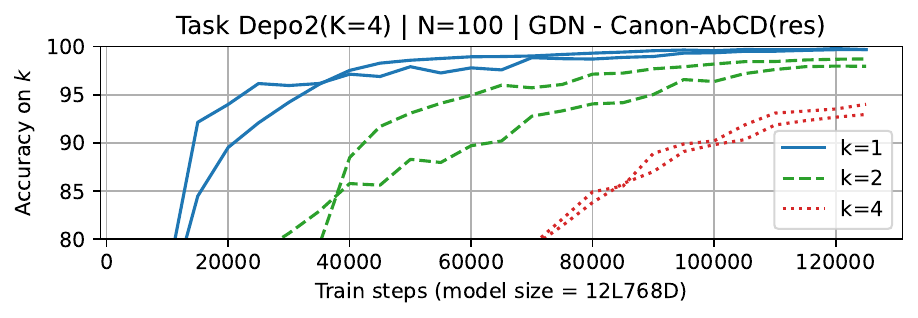}
\includegraphics[page=1,trim={2.5mm 1.5mm 2.5mm 1.5mm},clip,width=\imgwidthBase]{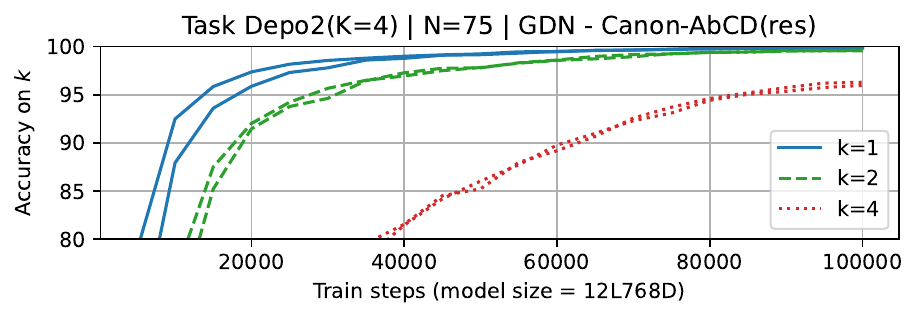}
\caption{\label{fig:depo1-curve-mamba}%
\textbf{Training curves for 12L768D architectures on }\textsc{Depo2}(K=4), evaluated at $k=1, 2, 4$ and $n = N$, with results shown across two best LRs for each $k$.
Results for other data are in \figureref{fig:depo1-curve-mamba:ext} on Page~\pageref{fig:depo1-curve-mamba:ext}.
}
\end{figure}

\parhead{Deep Dive into Deep Reasoning for Linear-Time Models}
We find that, due to compression of in-context knowledge, linear models struggle to reach 99\% accuracy even on simple 1- or 2-hop retrievals (\figureref{fig:depo1-curve-mamba}), despite extended training.
When reasoning depth exceeds 2 hops, early-step errors compound rapidly, preventing successful deep reasoning.
In contrast, Transformers—especially with Canons—achieve near-perfect 1- and 2-hop accuracy very quickly (\figureref{fig:depo1-curve-mamba}).

Importantly, this weakness is \bblue{not due to insufficient recurrent memory}.

For instance, in Mamba2, \textbf{each layer} passes $128d$ parameters (expansion $\times$ \texttt{ssm\_state\_size} $\times$ hidden size $d$)—\textbf{hundreds of times more than sufficient} to store the full input sequence.%
\footnote{In Task \textsc{Depo}, representing $N$ key-value pairs with vocabulary $V$ requires at most $2N\log_2 V$ bits. For \textsc{Depo2}, with $N{=}75$ and $V{\leq}2500$, this is under 1700 bits, compared to Mamba2’s recurrent state of $12\times128\times768{\approx}1.2$M 32-bit floats. This occupies $\sim$0.001 bits per float; in contrast, long-term (factual) memory in weights can reach 2 bits per float (see \cite{AL2024-knowledgeScaling} and our Task~\textsc{Capo}).}
Moreover, Mamba2 performs well on 1-hop tasks ($K{=}1$) even with a single layer, confirming the bottleneck is not information-theoretic (a finding also to be reinforced in \sectionref{sec:real-life}).

The same pattern holds for GLA and GDN, whose per-layer recurrent states (64$d$–144$d$) also provide ample capacity to store entire contexts (see \appendixref{app:arch} for architecture specifications).
Hence, the true limitation lies in \emph{memory dynamics}—how efficiently in-context information is encoded during compression and how reliably it is retrieved for reasoning.
Errors in encoding or retrieval accumulate across hops, severely degrading multi-hop reasoning.

These results expose the \emph{Achilles’ heel} of current linear architectures and point to a concrete direction for future research: improving the fidelity of compressed in-context memory.
Until such limitations are resolved, hybrid approaches that combine sliding-window attention (for deep reasoning) with linear or state-space components (for long-context compression) remain the most practical path forward.

\begin{mdframed}
\begin{sresult}{11}[\figureref{fig:depo1-curve-mamba}]
\label{res:11}
Linear models such as Mamba2/GLA/GDN struggle with deep reasoning—not from lack of memory, but from accumulated errors in compression and retrieval.
Hybrid models combining Transformers and linear layers, equipped with Canon, mitigate these limitations.
\end{sresult}
\end{mdframed}

\section{Real-Life Experiments}
\label{sec:real-life}

\begin{figure}[t!]
\centering
\setlength{\imgwidthBase}{0.98\textwidth}
\vspace{-3mm}%
{\includegraphics[page=7,trim={0mm 92mm 5mm 0mm},clip,width=\imgwidthBase]{plots}}
\includegraphics[page=1,trim={2.5mm 1.5mm 2.5mm 1.5mm},clip,width=\imgwidthBase]{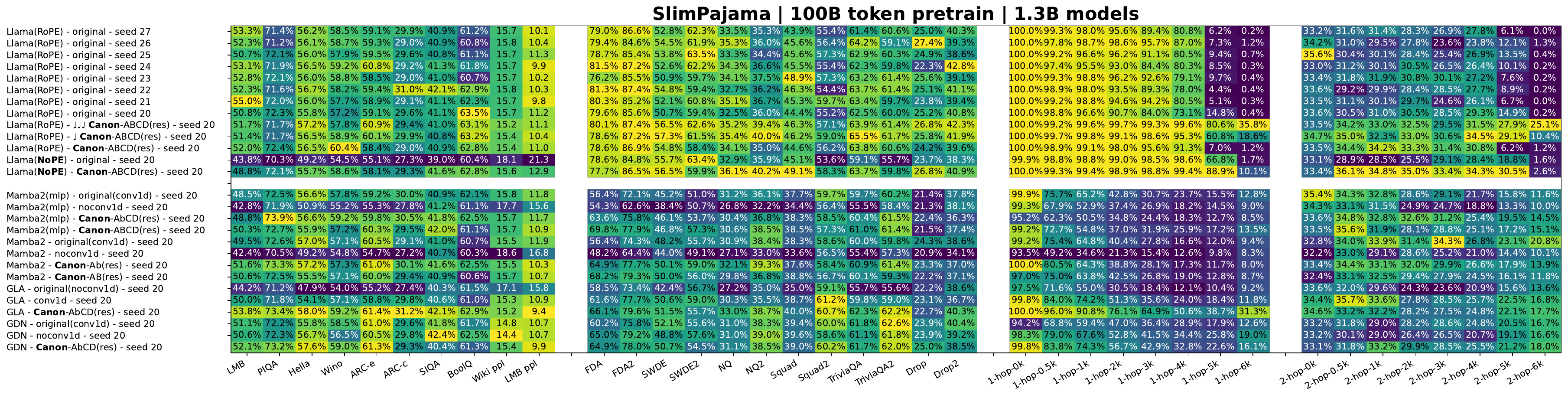}
\includegraphics[page=1,trim={2.5mm 1.5mm 2.5mm 1.5mm},clip,width=\imgwidthBase]{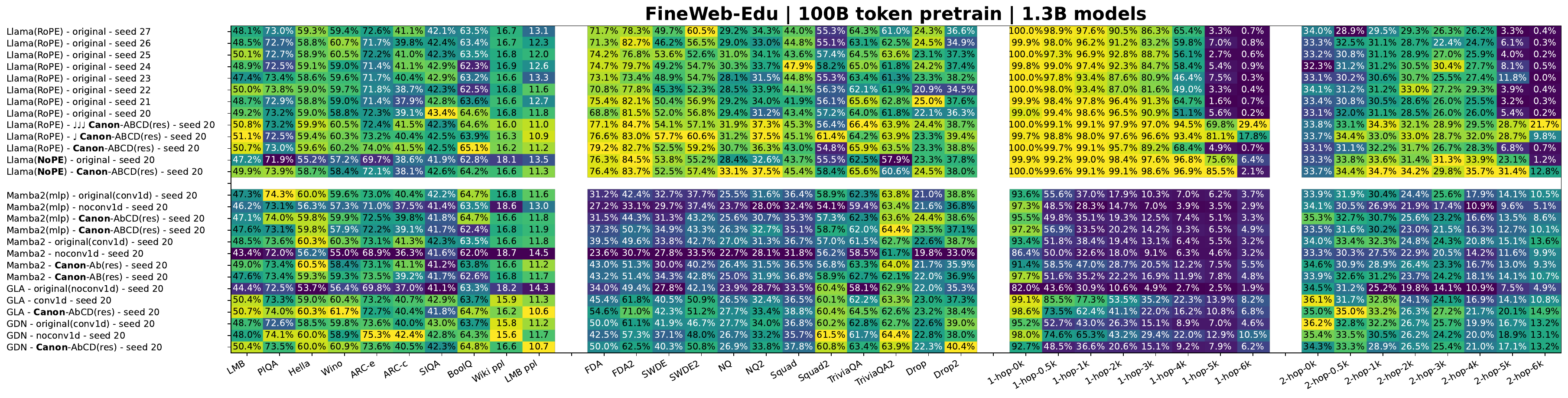}
\caption{\label{fig:real-life}%
\textbf{Performance of 1.3B models pretrained on 100B tokens} across discriminative (left), generative (middle), and 1/2-hop reasoning (right) tasks.
Best of 2 learning rates for Llama; 3 for GLA, Mamba, and GDN.
GPT2 variants (e.g., squared ReLU) shown in \figureref{fig:real-life:include-gpt} on Page~\pageref{fig:real-life:include-gpt}.
}
\end{figure}

We conduct real-life pretraining at the academic scale: 1.3B-parameter models trained on 100B tokens from FineWeb-Edu~\cite{penedo2024fineweb} and SlimPajama~\cite{cerebras2023slimpajama}, using a 4096 context length (details in \appendixref{app:other-tasks}).
This mirrors setups common in recent studies such as Titans~\cite{behrouz2024titans}, GDN~\cite{yang2024gated}, and MTA~\cite{golovneva2025multi}, representing the standard academic pretraining paradigm.

\parhead{Evaluation suites}
We first evaluate all models on two benchmark suites.
The first, based on \texttt{lm-evaluation-harness}~\cite{eval-harness}, covers \bblue{discriminative tasks}: PIQA~\cite{bisk2020piqa}, HellaSwag~\cite{zellers2019hellaswag}, WinoGrande~\cite{sakaguchi2019winogrande}, ARC-easy/challenge~\cite{clark2018think}, SIQA~\cite{sap2019social}, BoolQ~\cite{clark2019boolq}, WikiText, and LAMBADA~\cite{paperno2016lambada}.
Following prior work~\cite{yang2024gated,behrouz2024titans}, we adopt the original accuracy metrics for consistency.%
\footnote{Following tradition \cite{yang2024gated,behrouz2024titans,yang2023gated}, we use (\texttt{acc\_n}) for HellaSwag and ARC-c, but \texttt{acc} for other tasks.}

The second \bblue{generative-task} suite uses the \emph{Just Read Twice (JRT)} protocol~\cite{arora2024just}, designed to reduce noise in generative testing at this scale.%
\footnote{Generative testing can be noisy at this scale, as such models often struggle with prompt comprehension. JRT addresses this by repeating the context and question twice, allowing models to more accurately reveal their intrinsic generative capabilities.}
Tasks include SWDE, FDA, SQuAD(v2)~\cite{rajpurkar-etal-2018-know}, TriviaQA~\cite{joshi-etal-2017-triviaqa}, NQ~\cite{kwiatkowski-etal-2019-natural}, and DROP~\cite{dua-etal-2019-drop}, plus their JRT-enhanced variants (denoted as FDA2, SWDE2, etc.)
We again follow the official JRT codebase for evaluation.

\parhead{Key observations across both suites}
Results show large variance across random seeds:

\begin{itemize}
    \item Benchmark scores fluctuate with random seeds—up to \textbf{4\% on LAMBADA, 3\% on BoolQ}, and 1–3\% elsewhere.
    Generative tasks vary even more (\textbf{9\% on FDA}, \textbf{8\% on SWDE}, 3–5\% on others).
    The same holds even if data shuffling is fixed and model init varies (\appendixref{app:real-life:random}).
\end{itemize}

\noindent
\emph{Hence, only differences beyond these thresholds are statistically meaningful.}
From \figureref{fig:real-life}:

\begin{itemize}
    \item Linear models (Mamba2, GLA, GDN) underperform full Transformers on generative tasks, \textbf{even for contexts shorter than training length}.\footnote{Generative task prompts are capped at 1024–2048 tokens (per original codebase), while training used 4096.}
    Retrieval-heavy tasks (FDA, SWDE) amplify this gap, consistent with \resultref{res:11}.

    \item NoPE, GLA, and Mamba2 (w/o \texttt{conv1d}) perform poorly in base form but improve markedly with full Canon.
    GLA+Canon surpasses Mamba2 and matches GDN (even with Canon); NoPE+Canon performs on par with RoPE.
    GDN is least sensitive to Canon yet not clearly stronger than GLA+Canon—consistent with \resultref{res:3}, \ref{res:6.1-gla}, \ref{res:7.1-mamba}, \ref{res:gdn.1}, and \ref{res:9-linear}.

    \item At this scale (1.3B/100B), RoPE, RoPE+Canon, and NoPE+Canon perform comparably, and most linear+Canon variants cluster together.
    \textbf{Academic-scale pretraining cannot reliably distinguish finer architectural differences.}
\end{itemize}

\parhead{Needle, Babilong, and our Multi-Hop Reasoning Tasks}
The Needle-in-a-Haystack (\bblue{NIAH}) task from RULER~\cite{hsieh2024ruler} tests recall of a ``needle'' value (e.g., a magic number) in long text.
This makes it \emph{too easy}: models—especially linear ones—may appear accurate while failing at the most basic short-context retrieval (see later).
For completeness, results are shown in \figureref{fig:reallife-babilong} (Page~\pageref{fig:reallife-babilong}).

The \bblue{Babilong} dataset~\cite{kuratov2024babilong} embeds bAbi~\cite{weston2015towards} tasks in long junk-filled passages to test multi-hop reasoning but proves overly difficult at this scale.%
\footnote{For instance, in \texttt{babilong.qa2}, ``Charlie got a bottle ... Charlie moved to the balcony.'' → ``Where is the bottle?''—models score $<37\%$ even without junk, i.e., random guessing.}
As shown in \figureref{fig:reallife-babilong}, Babilong results are mostly indistinguishable; only trends are clear:
\begin{itemize}
    \item Linear models underperform Transformers even on short contexts, confirming their weakness stems from \emph{inefficient compression and retrieval}, not memory size (\resultref{res:11}).
    \item Transformers gain on longer contexts when RoPE is reduced (RoPE\musQuarter) or removed (NoPE), particularly in 4k-token junk passages (c.f. \resultref{res:3}).
\end{itemize}

To balance NIAH’s simplicity and Babilong’s difficulty, we introduce \textbf{our own multi-hop reasoning tasks}.
\bblue{1-hop-L} embeds \emph{five} birth-year statements within Wikipedia passages of length $L$, requiring direct recall of one of the birth years.
\bblue{2-hop-L} embeds \emph{three} birth-year statements plus three equivalence links (e.g., ``XXX was born in the same year as YYY''), requiring inference of the linked names’ birth years. Details are in \appendixref{app:other-tasks}.
Results (\figureref{fig:real-life}) show:

\begin{itemize}
    \item All models struggle with 2-hop-L, achieving only 30–36\% (near random) even with $L=0$.
    \item \textbf{1-hop-L separates architectures}: Transformers beat linear models \textbf{even at very short context lengths ($L\le500$)}, while NoPE/RoPE\musQuarter{} generalize better as $L$ increases.
\end{itemize}

To summarize:

\begin{mdframed}
\begin{sresult}{12}[\figureref{fig:real-life}+\ref{fig:reallife-babilong}]
\label{res:12}
Academic-scale pretraining (1.3B params, 100B tokens, 4k context) shows \defem{high noise and limited resolution}, making most architectural differences statistically insignificant. Yet several \defem{consistent findings} hold:
\begin{itemize}[leftmargin=5mm]
    \item Linear models (Mamba2, GLA, GDN) underperform full Transformers \textbf{even on short-context retrieval tasks} (FDA, SWDE, or 1-hop-L with $L\le500$), even with Canon (\resultref{res:11}).
    \item Canon elevates NoPE to RoPE-level (\resultref{res:3}), GLA to Mamba2/GDN-level (\resultref{res:6.1-gla}, \ref{res:9-linear}); removing \texttt{conv1d} downgrades Mamba2 to GLA (\resultref{res:7.1-mamba}) but hardly affects GDN (\resultref{res:gdn.1}).
    \item \defem{All models fail 2-hop reasoning, even within 100 tokens}, revealing the limits of academic-scale pretraining.
    \item Reducing or removing RoPE (NoPE, RoPE\musQuarter) improves long-context generalization (\resultref{res:3}).
\end{itemize}
\end{sresult}
\end{mdframed}

\section{Conclusion and Future Direction}
\label{sec:conclusion}

Academic-scale pretraining suffers from high noise and failed multi-hop reasoning, hindering reliable architectural comparison. Our controlled synthetic playground offers a \textbf{cost-effective, principled alternative}: by decomposing intelligence into atomic tasks, we discover and optimize \emph{Canon layers}—lightweight constructs that enhance reasoning depth and breadth, knowledge capacity and manipulation, and structural reasoning across diverse architectures.

Canon layers revive weaker models (e.g., NoPE, GLA) to match or surpass stronger baselines (e.g., RoPE, Mamba2), reduce reliance on RoPE to improve length generalization, and pinpoint that linear models' depth limitations arise from compression/retrieval inefficiencies rather than memory. Like residual connections or LoRA—simple yet powerful—Canon layers may become a minimal yet broadly applicable architectural primitive.

While our academic-scale real-world experiments align with synthetic findings, industrial-scale validation remains crucial; we hope our systematic, economical methodology \textbf{encourages future investigations} at larger scales. We plan to open-source our playground and evaluation suite to support rigorous, reproducible architecture research.

\parhead{Future Directions}
Several interesting directions arise from this work:
\begin{itemize}

\item \textsc{Alternative Canon Implementations.}
We focused on simple linear convolutional (kernel size~4) Canon layers for their simplicity and efficient CUDA kernels. Future work should explore dynamic, adaptive convolutions—with weights conditioned on hidden states to enable gating—to assess whether performance gains justify the added computational overhead.

\item \textsc{Fine-grained Canon Design.}
We briefly explored selective application (e.g., early layers) and cross-layer connections—e.g., $h'_{\ell+1} = h_{\ell+1} + \mathrm{Canon}(h_{\ell})$—which can fuse multiple intra-layer Canon operations into a single step, improving efficiency. A systematic evaluation within our synthetic framework could identify optimal Canon configurations. We are open to exploring this direction further, especially if the community expresses significant interest.

\item \textsc{Evaluating Emergent Architectures.}
We selected one representative per architecture family to ensure controlled comparisons and consistent inclusion of Canon layers. Without this rigor, results may misleadingly attribute Canon's gains to inherent architectural differences (e.g., Mamba2's built-in conv1d).
With controlled comparisons in mind, future work can fairly evaluate emergent architectures, potentially discovering new components with statistically significant improvements.

\item \textsc{Enriching the Synthetic Playground.}
Our five synthetic tasks are only a starting point. Designing additional tasks that isolate other architectural capabilities \emph{beyond those revealed here}—while remaining as atomic as possible—is crucial for finer-grained characterization of model strengths and weaknesses.

\item \textsc{Interpretability and Probing.}
We omitted interpretability and probing analyses here for clarity, despite existing frameworks for most tasks (e.g., \textsc{Lano}~\cite{AL2023-cfg}, \textsc{Capo}~\cite{AL2023-knowledge}, \textsc{Mano}~\cite{AL2023-knowledgeUB}, \textsc{Brevo}~\cite{YXLZ2024-gsm1,YXLZ2024-gsm2}). We have conducted preliminary probing for \textsc{Depo}, revealing internal model strategies such as positional parsing (even/odd position encoding ``$\to a$'' or ``$a\to$'') and preprocessing of permutations before the first query (analogous to \textsc{Brevo}~\cite{YXLZ2024-gsm1}). We choose not to include them for clarity, as this paper focuses on architectural comparison.

\item \textsc{Sparking New Architecture Designs.}
By pinpointing specific weaknesses (e.g., linear models' reasoning depth limits and compression inefficiencies), our framework provides targeted signals for improved future designs. We hope synthetic benchmarking informs and inspires the next generation of architecture innovations.

\end{itemize}

\appendix

\clearpage

\bigskip
\bigskip
\begin{center}
{\Huge
\textsc{Appendix}
}
\end{center}

This  appendix contains full technical specifications and implementation details for all experiments presented in the main paper. It is intended to support reproduction and in-depth inspection. We provide complete training protocols and evaluation procedures for all five synthetic tasks (Depo, Brevo, Capo, Mano, Lano), real-life experiments (1-hop-L, 2-hop-L, Babilong), and 100B-token SlimPajama/FineWeb-Edu pretraining. We also document the architectural configurations for all models, including Transformers, GLA, Mamba, GDN variants, and MoEs. Additional ablation figures, KL-divergence evaluations, and variant comparisons are included for readers interested in deeper technical insights or replication of results.

\section{Details on Synthetic Pretraining Tasks}
\label{app:all-tasks}

We intend to release the code for generating all synthetic pretraining datasets used in this paper, though this may require additional time. To make this paper fully self-contained, we provide detailed specifications below.

\begin{remark}
Throughout this paper, we utilize combinations of A100, H100, and H200 GPUs with bf16 mixed-precision training. While we report the total batch size used in our experiments, we do not specify the exact number of GPUs, as this does not materially affect the results.%
\footnote{For instance, training with a single GPU and a batch size of 128 is equivalent to training with 64 GPUs where each GPU processes a batch size of 2. Our codebase supports dynamic GPU allocation, ensuring the total batch size is fixed across training runs while the number and type of GPUs may vary.}
\end{remark}

\subsection{Details on Task \textsc{Depo}: Mental Reasoning Depth}
\label{app:depo}

The synthetic pretraining task \textsc{Depo} is designed to evaluate mental reasoning depth by requiring multi-step traversal over directed permutations. The dataset is defined by two parameters: the maximum permutation size $N$ and the reasoning depth $K$. Each problem instance is generated as follows:

First, a permutation length $n$ is sampled uniformly from $\{3, 4, \dots, N\}$. A directed permutation of $n$ nodes is then created, representing a cycle where each node points to its successor: $x_1 \rightarrow x_2 \rightarrow \dots \rightarrow x_n \rightarrow x_1$. The permutation is presented as edges in the form of ordered pairs $(x_i, x_{i+1})$, but these edges are shuffled randomly into a sequence of $2n$ tokens. This random ordering ensures that the original cycle structure is not immediately apparent, which would otherwise make the task trivial. The final data format is:
\begin{center}
\vspace{-5mm}\verb|<bos> x1 y1 x2 y2 ... xn yn <query k_1> q_1 <ans> a_1 ... <query k_t> q_t <ans> a_t |
\end{center}

Here, $x_i \rightarrow y_i$ represents shuffled edges of the permutation. For each query $q_j$, a node is randomly chosen from $\{x_1, \dots, x_n\}$, and its $k_j$-th successor in the permutation is computed based on the reasoning depth $k_j \in [K]$, sampled uniformly. The correct answer $a_j$ is the $k_j$-th successor of node $q_j$. The number of queries $t$ is set as $\min(10, n)$ to balance computational feasibility while ensuring smaller graphs remain interpretable.

Two variants of \textsc{Depo} are used:
\begin{itemize}
    \item \textsc{Depo1}: Each node name is encoded as 1–2 tokens, with a vocabulary size of 50.
    \item \textsc{Depo2}: Each node name spans 5–7 tokens using a small vocabulary size of 4, introducing ambiguity that challenges the model’s disambiguation capabilities.%
    \footnote{\label{footnote:word} Multi-token names are generated such that the first $\ell-1$ tokens are chosen from $[1, V]$, while the final token is selected from $[V+1, 2V]$. This creates implicit word boundaries similar to those handled by BPE-based tokenization strategies, such as GPT2Tokenizer.}
\end{itemize}
In addition to node names, special tokens are used: \verb|<bos>|, \verb|<ans>|, and \verb|<query k>| for $k \in \{1, \dots, K\}$. The total number of special tokens is $K+2$.

\parhead{Sampling distribution} To ensure controlled task difficulty progression, $n$ is sampled proportionally to $\frac{1}{\sqrt{N} + n}$. This distribution biases training toward simpler cases early on, allowing the model to gradually build foundational reasoning skills before encountering harder examples. Although this distribution is not perfect, it is both simple and effective, enabling clean comparisons between architectural designs without introducing unnecessary hyperparameter complexity. More sophisticated curriculum-based approaches, such as scheduled difficulty~\cite{lee2025selfimproving}, may provide an alternative solution but could introduce significant noise, thereby complicating controlled comparisons.

\begin{remark}
This distribution was proposed and tested thoroughly by ZA in 2023 in a number of settings, and subsequently tested (via private communication) by Alfarano in modular arithmetic pretraining~\cite{saxena2024teaching}, where it was benchmarked against other options and shown to also perform well.
While synthetic data like this cannot fully replicate the intricacies of real-world distributions, it allows us to simulate an ideal training regime. This forward-looking approach anticipates future improvements in pretraining data—such as higher-quality datasets or RL-based post-training—and evaluates model architectures based on their scalability under such optimal conditions.
\end{remark}

\parhead{Training protocol} To reduce computational cost, we employ label masking: cross-entropy loss is computed only on tokens associated with \verb|<ans>| and $a_j$. This optimization halves training duration without affecting architectural comparisons. Problem instances are generated online, concatenated, and aligned into 2048-token context windows. Left alignment ensures that the first problem instance in each context is never truncated, as truncation leads to incomplete edges and unusable data.

\parhead{Evaluation protocol} During evaluation, the permutation size is fixed at $n = N$, and reasoning depth is tested at both $k = K$ (maximum depth) and $k = K/2$ (intermediate depth). The protocol mirrors training by generating and concatenating evaluation samples online into 2048-token windows. Accuracy is reported over all answer tokens in the window, ensuring that results are stable regardless of whether answers appear early or late in the sequence.

\parhead{Data splits and hyperparameters} For \textsc{Depo1}, we use $N = 375$, $300$, $225$ and primarily $K = 8$, while testing $K = 4$ for weaker models. Models are trained from scratch with fresh data while using a fixed random seed to ensure data consistency across architectures. Training uses a batch size of 128, AdamW optimizer ($\beta = 0.9,0.98$ and $\veps=10^{-6}$), weight decay of 0.03, learning rate warmup for the first 1000 steps, and cosine decay to 10\%. Training steps are set to 112.5k, 100k, or 87.5k, adjusted for the problem lengths $N = 375$, $300$, $225$. The best accuracy is reported across four runs using learning rates $\{0.0003, 0.0005, 0.001, 0.002\}$.

Similarly, in \textsc{Depo2}, we use $N = 125$, $100$, $75$ and $K = 16$ (or $K = 4$ for weaker models). Training steps are set to 150k, 125k, and 100k, respectively.

\subsection{Details on Task \textsc{Brevo}: Mental Reasoning Breadth}
\label{app:brevo}

Our pretraining synthetic task \textsc{Brevo} is designed to test mental reasoning breadth by requiring a subgraph topological sort from a given directed acyclic graph (DAG). The dataset is defined by a maximum graph (node) size $N$. For each problem instance, we first sample a graph of size $n \in \{3, 4, \dots, N\}$ using the same sampling distribution $\propto \frac{1}{\sqrt{N} + n}$ as employed in \textsc{Depo}, and generate data in the following format:
\begin{center}
\verb|<bos> x1 y1 x2 y2 ... xm ym <query> q <ans> a1 a2 ... ap <eos>|
\end{center}
Here, the $2m$ tokens define $m$ directed edges $x_i \rightarrow y_i$ spanning $n$ nodes, meaning that $y_i$ depends on $x_i$. Given a query vertex $q$, the model must return all vertices it recursively depends on, in topological order starting from the leaves. Specifically, if $u \rightarrow v \rightarrow q$, the model must output $u$ before $v$.

\parhead{DAG generation protocol} After sampling $n$, we generate the random DAG as follows. First, we randomly shuffle all the vertices and begin inserting edges. We select a random number $L \in \{1, \dots, \lceil \frac{n-1}{4} \rceil + 1\}$, designating the first $L$ vertices as leaves (no incoming edges). Starting from vertex $L + 1$, we iteratively process each vertex by selecting all preceding vertices that have an out-degree of at most 3. From this set, we randomly pick a subset of between 1 and 4 vertices and connect them to the current vertex. This process continues until all vertices are traversed, yielding a DAG with a maximum in-degree and out-degree of 4.%
\footnote{Constraining the maximum in-degree and out-degree to 4 prevents the dependency graph from becoming too shallow, which would make the task trivial.}

At this point, the vertices naturally form a topological order from left to right. We then select a random query vertex from the last quarter of the vertices. Choosing vertices closer to the right increases the depth of the dependency graph while avoiding degenerate cases where all nodes are reachable (such as if the query were the last vertex). Finally, we reshuffle all the vertices and assign random names to them. Vertex names are uniquely selected, as described below.

\parhead{Vertex names} In \textsc{Brevo1}, each vertex name consists of a single unique token, randomly selected from $\{1, \dots, N\}$. In \textsc{Brevo2}, each vertex name spans 2--4 tokens using a vocabulary of size 4, which introduces ambiguity (e.g., multiple token combinations can encode unique vertex names). See \footnoteref{footnote:word} for the method used to generate multi-token words. Aside from vertex names, we use 4 distinct special tokens: \verb|<bos>|, \verb|<query>|, \verb|<ans>|, and \verb|<eos>|.

\parhead{Training protocol} To reduce computational costs, we enable label masking, where the cross-entropy loss is computed only on \verb|<ans>|, \verb|<eos>|, and $a_j$ tokens. Selective testing showed that this technique saves training time without affecting architectural comparisons. Instances are generated online, concatenated, and left-aligned into context windows. By left-aligned, we mean that the first instance in each context window is never truncated. Without left alignment, truncation of the first instance would render it incomplete (e.g., missing edges in the graph), making the instance a useless training example.

\parhead{Evaluation protocol} During evaluation, we fix $n = N$ and test only the largest graph. The model is prompted with a random DAG of size $n$ and query vertex $q$, and tasked to generate the answer sequence $a_1, \dots, a_p$. The generated sequence is then parsed and validated against the following criteria:
\begin{itemize}
    \item The answer sequence must contain all reachable vertices from $q$ and no non-reachable vertices.
    \item The vertices in the answer sequence must appear in a valid topological order. Since topological orderings are not unique, any valid ordering is accepted.
\end{itemize}
Invalid tokens, duplicate outputs, or missing vertices are not accepted, and no partial credit is given.

\parhead{Training details} In \textsc{Brevo1}, we use $N = 110, 90, 70$ with vertex names consisting of one token, and each problem fits within 1024 tokens. Models are trained from scratch with fresh data but a fixed seed (ensuring pretraining data consistency across model architectures). Training uses a context length of 1024, a total batch size of 256, AdamW optimizer ($\beta = 0.9,0.98$ and $\veps=10^{-6}$), weight decay of 0.03, learning rate warmup over the first 1000 steps, and cosine decay to 10\%. Pretraining lasts 150k, 125k, or 100k steps respectively for $N = 110, 90, 70$, accounting for the varying problem lengths. We report the best performance out of four runs using learning rates $\{0.0003, 0.0005, 0.001, 0.002\}$.

In \textsc{Brevo2}, we use $N = 50, 40, 30$, with vertex names spanning 2--4 tokens, and each problem fits within 1536 tokens. Models are trained in the same manner as \textsc{Brevo1}, except that we use a context length of 1536, a total batch size of 192, and pretraining lasts 250k, 225k, or 200k steps respectively for $N = 50, 40, 30$.

The comparison between \textsc{Brevo1} and \textsc{Brevo2} demonstrates that the ambiguity introduced by multi-token vertex names does not noticeably impact architectural comparisons, which is the focus of this paper.

\subsection{Details on Task \textsc{Capo}: Knowledge Capacity}
\label{app:capo}

The synthetic pretraining task \textsc{Capo} borrows directly from \citet{AL2024-knowledgeScaling}, where the authors introduced the $\bioS(N)$ dataset. This dataset contains $N$ biographies of randomly generated individuals, each described by six attributes: birth date, birth city, university, major, employer, and working city.%
\footnote{The working city is derived from the employer’s headquarters, while all other attributes are sampled uniformly and independently. Possible attribute domains include $N_0 = 400 \times 400 \times 1000$ person names, $12 \times 28 \times 200$ birth dates, $200$ birth cities, $300$ universities, $100$ majors, $263$ employers, and two pronouns.\label{footnote:bio-domain}}

To represent these biographies in natural language, each individual is described via randomly selected English sentences for every \emph{exposure} to the pretraining data. Sentence templates correspond to the individual’s attributes, ensuring diverse paraphrasing across exposures. For example:
\begin{equation*}
\text{
\begin{varwidth}{\linewidth}
\scriptsize
\underline{Anya Briar Forger} was born on \underline{October 2, 1996}. She spent her early years in \underline{Princeton, NJ}. She received mentorship and guidance from faculty members at \underline{Massachusetts Institute of Technology}. She completed her education with a focus on \underline{Communications}. She had a professional role at \underline{Meta Platforms}. She was employed in \underline{Menlo Park, CA}.
\end{varwidth}
}
\end{equation*}

The diversity in writing ensures that models learn to store explicit knowledge about an individual’s attributes, rather than merely memorizing surface-level patterns in specific templates~\cite{AL2023-knowledge,AL2023-knowledgeUB}. Following the recommendations of \cite{AL2024-knowledgeScaling}, we pretrain models over 100 \emph{exposures} per individual, which provides a controlled environment for comparing architectural differences. Training beyond 100 exposures diminishes architectural differences, as longer training typically allows all models to converge toward similar levels of performance~\cite{AL2024-knowledgeScaling}.

\parhead{Knowledge format independence} Previous experimental evidence suggests that a model’s knowledge capacity does not heavily depend on the specific format in which the knowledge is stored. For example, one could consider synthetic alternatives such as longer word lengths, different vocabulary sizes, or even abstract encoding formats. Importantly, any such synthetic configuration remains a reliable discriminator for comparing model architectures. For simplicity and interpretability, however, we adhere to the more English-like biography format in $\bioS(N)$, aligned with \cite{AL2024-knowledgeScaling}.

\parhead{Clean experimental comparisons} Models could alternatively be pretrained on exposures distributed according to power-law dynamics or incorporating infrequent ``junk data.'' While such approaches might better mimic real-life datasets, they introduce subtle stochastic effects that can depend heavily on the formatting of rare samples. To avoid confounding factors, we adopt the cleaner 100-exposure baseline for pretraining individual biographies, as it allows for clearer isolation of architectural capabilities.

\parhead{Evaluation protocol} After pretraining on $\bioS(N)$ data, knowledge capacity is measured based on the number of \emph{bits} a model reliably stores. This quantity is further normalized to \textbf{bits per parameter} to account for model scale. Partial correctness (e.g., recalling the year but not the full date of birth) is accounted for in the bit computation to ensure fine-grained evaluation of knowledge storage. For detailed computation, we direct readers to \cite{AL2024-knowledgeScaling}. Unlike other tasks presented in this paper, measurement of bits per parameter requires varying both data sizes $N$ and model sizes to compute the Pareto frontier of knowledge capacity versus parameter count. For this reason, we vary $N$ between $50\text{K}$ and $2\text{M}$ while testing models ranging from $1\text{M}$ to $500\text{M}$ parameters.

\parhead{Pretraining setup} To ensure consistency across all architectures, pretraining uses the GPT-2 tokenizer and ties weights for embedding and output layers. Tying weights ensures consistent learning dynamics across model families (e.g., GPT, Llama, Mamba, GLA), while limiting the vocabulary size to 3275 tokens (from GPT-2’s original 50257 tokens), as the $\bioS(N)$ dataset does not utilize the entire vocabulary.

Batch size, learning rate decay, and other hyperparameters strictly follow the 100-exposure baseline outlined in \cite{AL2024-knowledgeScaling}, with only minor modifications. Specifically, we test \emph{two} learning rates per configuration (selected from their three choices) and report the best results. As a result, our reported knowledge capacity in \figureref{fig:initial} may slightly deviate from the original results, though introducing Canon layers restores capacity without adding hyperparameter choices.

\parhead{Hyperparameters for dense models} The following hyperparameters were used for dense models in the 100-exposure setup:
\begin{itemize}[nolistsep]
\item For $N=50\text{K}$: weight decay $wd=0.01$, learning rates $lr=0.001/0.0005$, batch size 12.
\item For $N=100\text{K}$: $wd=0.01$, $lr=0.001/0.0005$, batch size 24.
\item For $N=200\text{K}$: $wd=0.01$, $lr=0.001/0.0005$, batch size 48.
\item For $N=500\text{K}$: $wd=0.01$, $lr=0.001/0.0005$, batch size 96.
\item For $N=1\text{M}$: $wd=0.01$, $lr=0.001/0.0005$, batch size 192.
\item For $N=2\text{M}$: $wd=0.01$, $lr=0.0005/0.0003$, batch size 384.
\end{itemize}

\parhead{Hyperparameters for MoE models} Mixture-of-Experts (MoE) training was conducted using the \texttt{tutel\_moe} package~\cite{tutel}, consistent with \cite{AL2024-knowledgeScaling}. MoE training uses 32 experts with $topk=1$ and $cap\_factor=2$. Due to the larger learning rates required for MoE-based pretraining, we use the following hyperparameters:
\begin{itemize}[nolistsep]
\item For $N=50\text{K}$: $wd=0.01$, $lr=0.005/0.002/0.001$, batch size 12.
\item For $N=100\text{K}$: $wd=0.01$, $lr=0.005/0.002/0.001$, batch size 24.
\item For $N=200\text{K}$: $wd=0.01$, $lr=0.005/0.002/0.001$, batch size 48.
\item For $N=500\text{K}$: $wd=0.01$, $lr=0.002/0.001$, batch size 96.
\item For $N=1\text{M}$: $wd=0.01$, $lr=0.002/0.001$, batch size 192.
\item For $N=2\text{M}$: $wd=0.01$, $lr=0.001/0.0005$, batch size 384.
\end{itemize}

\subsection{Details on Task \textsc{Mano}: Knowledge Manipulation}
\label{app:mano}

The synthetic pretraining task \textsc{Mano} evaluates a model's ability to manipulate stored knowledge mentally without relying on explicit intermediate cues (e.g., Chain-of-Thought reasoning). Unlike memorization tasks, \textsc{Mano} requires multi-step internal computation, testing the model’s capacity for hierarchical manipulation.

\parhead{Task format and setup} The dataset is defined by a maximum length $L$, with each instance consisting of arithmetic expressions of $\ell$ operations, where $\ell$ is sampled uniformly from $[1, L]$. Expressions are presented in prefix (pre-order) notation to eliminate ambiguities in parentheses and operator precedence. For example, a length-$\ell = 3$ instance is:
\begin{center}
\verb|<bos> <len_3> + * a b - c d <ans> ans |
\end{center}
This corresponds to the expression $((a \times b) + (c - d)) \bmod 23$, where operands $a$, $b$, $c$, and $d$ are integers sampled uniformly from $[0, 22]$. The task involves three operations (+, -, *), each represented as distinct tokens, with all computations performed modulo 23.

The factual base consists of three $23 \times 23$ arithmetic tables (addition, subtraction, and multiplication), which models learn implicitly during pretraining. Operands are encoded as single tokens from $[0, 22]$, while special tokens (\verb|<bos>|, \verb|<ans>|, and \verb|<query_|$\ell$\verb|>| for $\ell \in [L]$) structure the sequence.

Expressions are generated recursively: the first operator is sampled uniformly from the three available options, and its operands are split into sub-lengths $\ell'$ and $\ell - 1 - \ell'$ (with $\ell'$ chosen uniformly), recursively generating sub-expressions.

\parheadno{Why modular arithmetic?} Modular arithmetic (mod 23) ensures manageable knowledge size while introducing sufficient diversity in intermediate and final results. Similarly, limiting operations to addition, subtraction, and multiplication simplifies task design while retaining depth, enabling models to focus on hierarchical manipulation instead of memorizing surface-level patterns.

\parhead{Training protocol} Models are pretrained on three datasets corresponding to difficulty levels $L = 16$, $L = 13$, and $L = 10$. The cross-entropy loss is applied over all tokens (problem description and answer), without label masking, since hierarchical manipulation requires attention across the full sequence. Instances are generated online, concatenated, and left-aligned into context windows of length 1024.

Models are trained from scratch using fixed random seeds for consistency across architectures. Training lasts 110k, 95k, and 80k steps for $L = 16$, $L = 13$, and $L = 10$, respectively. Hyperparameters include a batch size of 64, AdamW optimizer ($\beta = 0.9,0.98$ and $\veps=10^{-6}$), weight decay of 0.1, learning rate warmup for 1000 steps, and cosine decay to 10\% of the initial learning rate. Results are reported based on eight training runs with learning rates $\{0.0001, 0.0002, 0.0003, 0.0005\}$ and two seeds.

\parhead{Evaluation protocol} During evaluation, expressions are sampled from the same distribution used for training, with $\ell$ fixed at $L$ (maximum difficulty). Accuracy is computed over all problem instances within 1024-token context windows, including non-first instances. Since outputs are single tokens representing exact modular arithmetic results, partial correctness is not applied.

\begin{figure}[t!]
\centering
\vspace{-3mm}
{\includegraphics[page=1,trim={0mm 62mm 45mm 0mm},clip,width=0.9\textwidth]{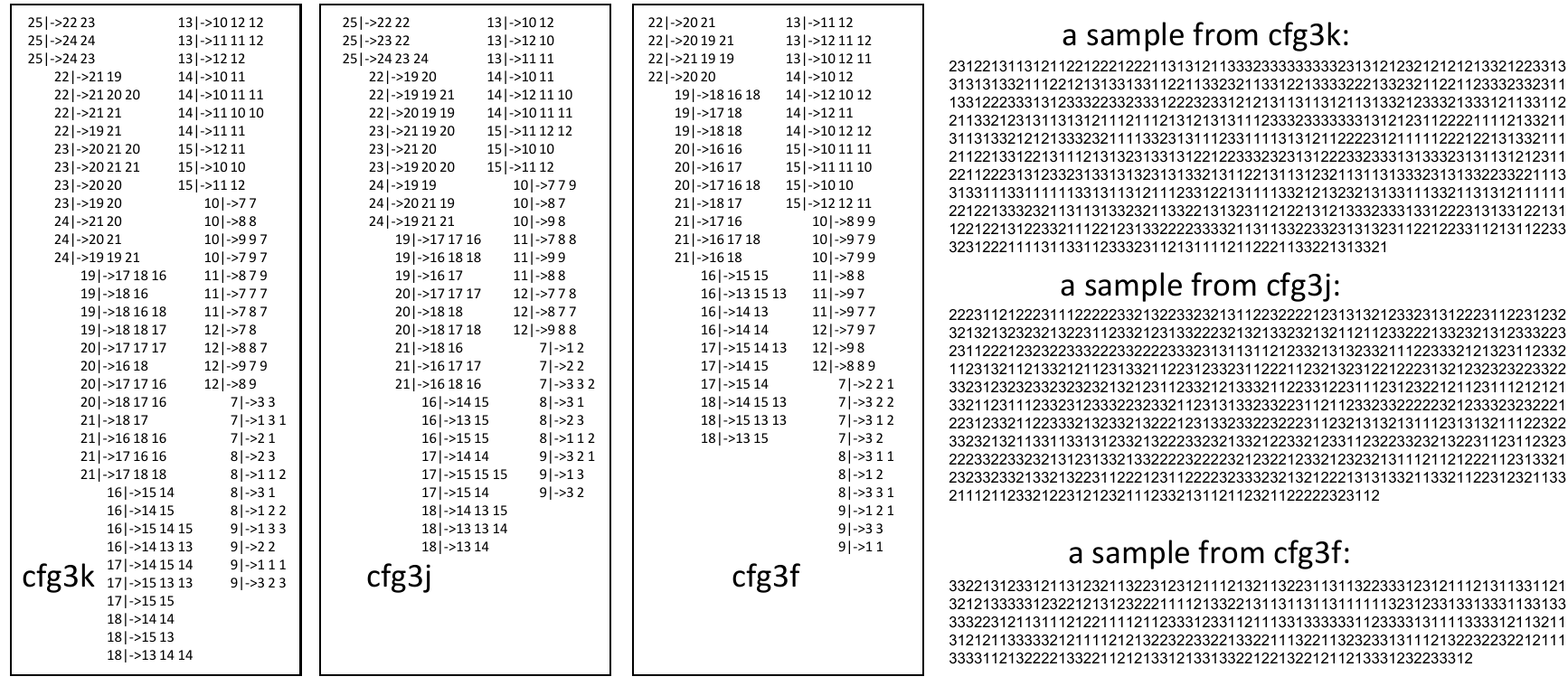}}
\caption{\label{fig:cfg}Task \textsc{Lano}: \textbf{our constructed dataset} \textsf{cfg3k}, \textsf{cfg3j} against the \textsf{cfg3f} dataset from \cite{AL2023-cfg}.}
\end{figure}

\subsection{Details on Task \textsc{Lano}: Hierarchical Language Structure}
\label{app:lano}

The synthetic pretraining task \textsc{Lano} evaluates a language model's ability to perform structural reasoning, specifically long-range structural planning that requires dynamic programming to resolve ambiguity. Unlike in-context reasoning tasks (e.g., \textsc{Depo}, \textsc{Brevo}) or knowledge reasoning tasks (e.g., \textsc{Mano}), \textsc{Lano} challenges models to learn hierarchical structures governed by probabilistic context-free rules and process sequences that cannot be resolved locally.

\parhead{Task format and setup} Sentences are generated probabilistically using context-free rules. The \textsf{cfg3f} dataset \citep{AL2024-knowledgeScaling} starts with the root non-terminal (NT) symbol $22$, which uniformly expands into one of four rules:
\[
22 \mapsto 20\ 21,\quad 22 \mapsto 20\ 19\ 21,\quad 22 \mapsto 21\ 19\ 19,\quad 22 \mapsto 20\ 20.
\]
Each rule is chosen with probability $1/4$, ensuring uniform randomness. Rules are applied recursively and probabilistically to NT symbols (e.g., $19$, $20$, $21$), replacing all NT symbols with terminal (T) symbols $1$, $2$, or $3$. The process generates sentences composed entirely of terminal symbols based on probabilistic expansions.

Pretraining involves predicting next tokens in CFG-generated sequences without access to the underlying rules, requiring models to learn structural reasoning implicitly. During evaluation, models are prompted with a single \verb|<bos>| token and tasked to generate CFG-compliant sentences using temperature 1. Accuracy is assigned only for fully valid sentences, with no partial credit applied.%

\parhead{Parsing difficulty and ambiguity} Parsing CFG-generated sequences is uniquely challenging because resolving derivation chains requires global reasoning. For example, parsing ``221213133'' requires resolving structural ambiguity between terminal symbols that cannot be inferred from local patterns alone. Instead, parsing requires an $O(n^3)$ dynamic programming algorithm to globally reconstruct relationships across the sequence, even when CFG rules (from \figureref{fig:cfg}) are explicitly available. During pretraining, models face additional difficulty as they must learn these relationships without direct access to the probabilistic rules.

Building upon \textsf{cfg3f} as a baseline, we introduce two extended datasets in this paper:
\begin{itemize}[nolistsep]
    \item \textsf{cfg3k}: Retains the probabilistic framework of \textsf{cfg3f} but increases depth by one level, doubling sequence length and increasing parsing complexity by eight times due to the cubic nature of dynamic programming ($O(n^3)$).
    \item \textsf{cfg3j}: Extends \textsf{cfg3f} by one level but reduces the number of rules, creating shorter sequences with intermediate difficulty between \textsf{cfg3f} and \textsf{cfg3k}.
\end{itemize}
Both datasets use the same probabilistic generation process and are detailed in \figureref{fig:cfg}.

\parhead{Training details} Pretraining uses cross-entropy loss computed over all tokens without label masking. Sentences are generated online, concatenated, and aligned into context windows. For \textsf{cfg3f}, we use a context length of 512 as in \citep{AL2024-knowledgeScaling}, while longer datasets \textsf{cfg3j} and \textsf{cfg3k} require extended context lengths of 1536.

Models are trained from scratch using fixed seeds for consistency across architectures. Training uses a batch size of 96, AdamW optimizer ($\beta = 0.9,0.98$ and $\veps=10^{-6}$), weight decay of 0.1, no learning rate warmup, and linear decay to 0. Pretraining lasts 100k steps, and results are reported from four training runs using learning rates $\{0.0002, 0.0003, 0.0005, 0.001\}$.

\parhead{Evaluation details} During evaluation, models generate sentences from a \verb|<bos>| prompt using temperature 1 and beam width 1.%
\footnote{This is crucial to ensure that the model is generating the genuine probabilistic distribution of sentences; if using temperature 0 for instance, the generation is always a fixed string, and accuracy would be either 0 or 100\% forever.}
Generated sentences are validated using an $O(n^3 m)$ dynamic programming parser ($n$: sequence length, $m$: CFG rules) to confirm compliance. An alternative evaluation computes KL divergence between the model’s next-token prediction distribution and the ground-truth CFG predictions. Both methods yield consistent architecture comparisons.

\section{Details on Other + Real-Life Experiments}
\label{app:other-tasks}

This section provides a brief description of additional tasks used in the paper.

\parhead{Full Copy}
In \figureref{fig:copy}, we evaluated the performance of models with 1 or 2 layers on a trivial pretraining task. This task involves choosing $N=500$ and generating a sequence starting with \verb|<bos>|, followed by a random permutation of $N$ tokens between 1 and $N$, then appending \verb|<query>| and an identical copy of the sequence.
The task uses label masking, where the loss is computed only on the $N$ answer tokens. Models are pretrained with a context length of 1024, a total batch size of 32, AdamW optimizer ($\beta = 0.9,0.98$ and $\veps=10^{-6}$), weight decay of 0.03, learning rate warmup for the first 1000 steps, and cosine decay to 10\%. Training duration is set to 50k steps, and the best results are reported across learning rates $\{0.0005, 0.001, 0.002, 0.005\}$.
For a cleaner comparison, in this experiment only we disabled the value matrix (\texttt{v\_proj}) in the attention layers; the MLP layers are unchanged.

\begin{figure}[H]
\centering
\subfigure[Evaluated with $t=1$]
{\includegraphics[page=1,trim={2.5mm 1.5mm 2.5mm 1.5mm},clip,width=0.325\textwidth]{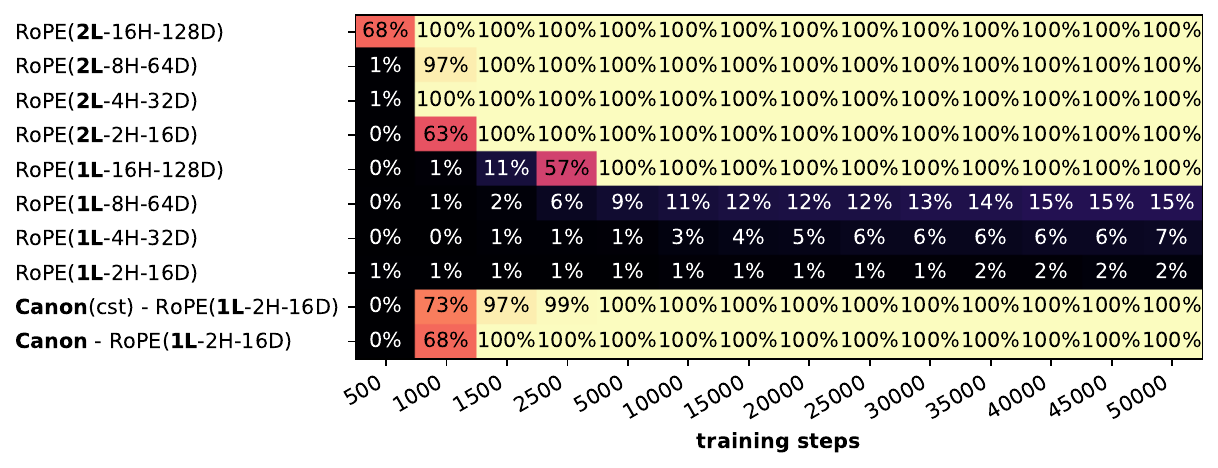}}
\subfigure[Evaluated with $t=2$]
{\includegraphics[page=1,trim={2.5mm 1.5mm 2.5mm 1.5mm},clip,width=0.325\textwidth]{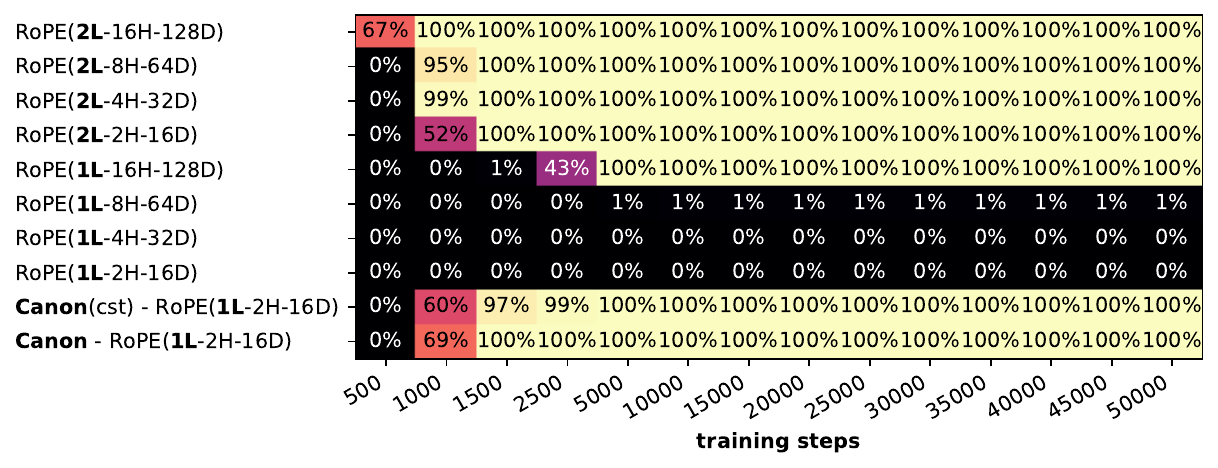}}
\subfigure[Evaluated with $t=4$]
{\includegraphics[page=1,trim={2.5mm 1.5mm 2.5mm 1.5mm},clip,width=0.325\textwidth]{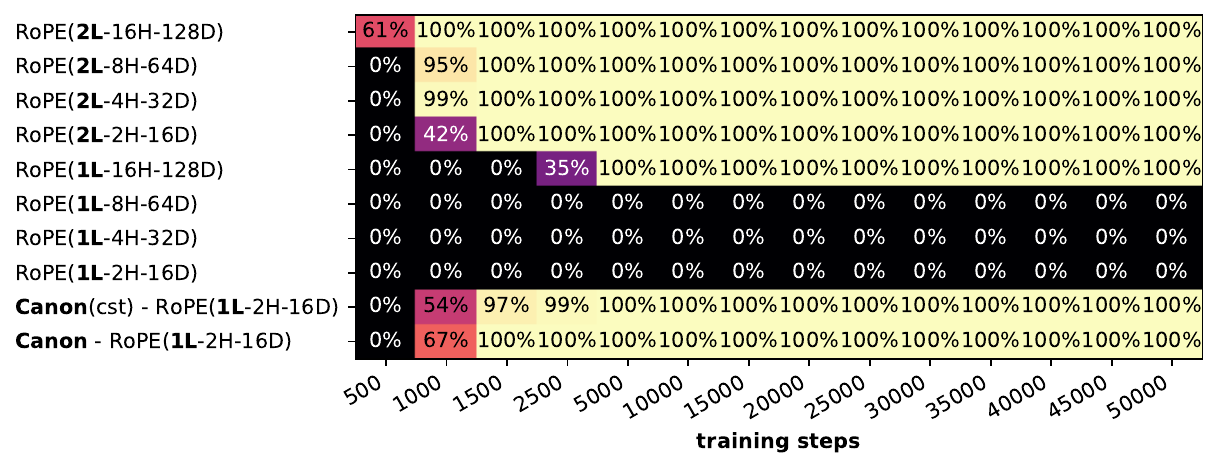}}
\caption{\label{fig:copy2}A trivial experiment for copying 500 tokens, evaluated only on correctly copying the first $t$ tokens.}
\end{figure}

For this task, we also assessed the models' ability to correctly copy the first $t=1,2,4$ tokens within the sequence. As shown in \figureref{fig:copy2}, these results are nearly identical to those in \figureref{fig:copy}.

\parhead{Task 1-hop-L and 2-hop-L}
In the real-life experiment (\sectionref{sec:real-life}), we evaluated models' performance on extremely simple 1-hop and 2-hop information retrieval tasks.

For the \textsc{1-hop-L} task, we prepared five random birth year statements of the form ``[name] was born in the year of [year],'' where names are generated as random combinations of first, middle, and last names, and years are sampled uniformly from 1950 to 2009. The five sentences were embedded into random Wikipedia documents of length $L$ tokens, with each statement inserted between sentences at up to five randomly chosen positions. Finally, the model was prompted with ``\verb|\n\n|Answer me: {name} was born in the year of'' to test its ability to retrieve the birth year. This setup closely replicates the needle-in-a-haystack task~\cite{hsieh2024ruler}, but we intentionally made the task more ``natural English'' by using birth years (commonly found in pretraining datasets like Wikipedia) instead of abstract multi-digit numbers.

For the \textsc{2-hop-L} task, three random birth year statements were prepared in the same format as above. This was followed by three equivalence statements of the form ``[name1] was born in the same year as [name2],'' where random names were generated such that the equivalences formed a bijection between the two sets of three random names. To simplify the task, we did not shuffle the ordering of the statements; the three equivalence statements always followed the three original ones. These six sentences were then embedded into random Wikipedia documents of length $L$ tokens, inserted at up to six different positions between sentences, respecting their original order. At the end, the model was prompted with ``\verb|\n\n|Answer me: {name} was born in the year of'' to test its ability to infer and retrieve the correct birth year. To further assist the model, an instructional statement was added at the beginning of the context.
\footnote{``You will be asked questions about people's birth years, and the birth year descriptions are hidden in some random text. Some people's birth years are directly given, while others are given in the form that `name1' was born in the same year as `name2'.''}
This design represents arguably the simplest possible and most natural 2-hop in-context reasoning task, yet even with $L=0$, models largely failed to perform, as demonstrated in \figureref{fig:real-life}.

\parhead{Babilong}
For the Babilong experiments, we found the default few-shot prompts (qa1–qa5) slightly suboptimal and replaced them with improved ones, which are released in our GitHub repo~\cite{PhysicsLM42}.

\parhead{SlimPajama and FineWeb-Edu 100B}
The SlimPajama dataset is taken from HuggingFace (\texttt{cerebras/SlimPajama-627B}), using the first 100M samples (more than 100B tokens).
FineWeb-Edu~\cite{penedo2024fineweb} is obtained from \texttt{HuggingFaceFW/fineweb-edu}, using its predefined 100B split.
Both datasets provide sufficient scale for our 1.3B-model pretraining experiments.

Following standard practice, all data are tokenized in order, concatenated into a continuous text stream, and sampled into random 4096-token windows for pretraining across architectures.
We train with total batch size 48 using AdamW ($\beta_1{=}0.9$, $\beta_2{=}0.98$, $\epsilon{=}10^{-6}$, weight decay 0.03).
Llama and GPT models use learning rates $\{0.001, 0.002\}$, while linear models (Mamba, GLA, GDN) use $\{0.0005, 0.001, 0.002\}$ for stronger baselines.
Each model is trained for $510{,}000$ steps, processing $4096\times48\times510{,}000{\approx}100.2$B tokens per run.
For each evaluation task, we report the \emph{best accuracy} across the tested learning rates.

To ensure fairness, all architectures share the same random seed, guaranteeing identical data order and content—even if runs are interrupted and resumed.
This setup minimizes variability from data differences and isolates architectural effects.
For Llama(RoPE), we additionally test eight random seeds to measure variance, shown in \figureref{fig:real-life} and detailed in \appendixref{app:real-life:random}, including both joint (data + model init) and model-init-only random seed variations.
Architecture specifications appear in \appendixref{app:arch:real-life}.

\parhead{Beyond 100B-1.3B}
We find that academic-scale pretraining (100B tokens, 1.3B models) is too noisy to reveal subtle architectural gaps (e.g., Llama vs.\ Llama+Canon).
Larger-scale experiments (1–8B models pretrained on 1–2T tokens) are therefore reported in our follow-up work~\cite{PhysicsLM42}.

\section{Details on Architectures Used}
\label{app:arch}

\parhead{Transformer Models (Llama/GPT)}
In this paper, ``Llama(RoPE)'' refers to the HuggingFace (HF) implementation LlamaForCausalLM, which employs rotary embeddings across all hidden dimensions and utilizes gated MLP layers. We did not enable group-query attention, as this study focuses on smaller-scale models. The intermediate size is set to $\frac{8d}{3}$, ensuring that each MLP layer contains $8d^2$ trainable parameters, consistent with standard MLP layers. ``Llama(NoPE)'' refers to the same architecture with rotary embedding completely disabled. ``Llama(RoPE)\musQuarter'' refers to the version where rotary embeddings are applied to only a quarter of the dimensions. The variants \musQuarter, \musQuarter\musQuarter, and \musQuarter\musQuarter\musQuarter represent differing fractions of dimensionality on which RoPE is enabled, as described in the main paper.

For direct comparisons, ``GPT2(RoPE)'' refers to the Llama architecture with gated MLP layers replaced by standard MLP layers. The intermediate size in these models is set to $4d$, ensuring that each MLP layer contains $8d^2$ trainable parameters.%
\footnote{The original GPT2 architecture differs from Llama in other minor ways, such as using GeLU activation and slightly different initialization. We do not investigate these small architectural differences in this paper.}

We denote ``GPT2(RoPE,R2)'' as the GPT2(RoPE) model with its \texttt{silu} activation replaced by $\text{ReLU}^2$, following the design proposed in Primer~\cite{so2109primer}. Similarly, ``Llama(RoPE,R2)'' refers to Llama(RoPE) with $\text{ReLU}^2$ in place of \texttt{silu}.

\parheadsc{ALiBi and H-Alibi}
For ALiBi~\cite{press2021train}, we follow the original recommendation of using a geometric sequence $2^{-8/n}$ for an $n$-head Transformer, which determines how each head is biased toward local context. For H-Alibi~\cite{jelassi2024repeat}, we use their proposed strategy of restricting the $h$-th head to attend only to the nearest $h$ tokens, and applied to half of the heads. (We briefly tested applying this to one-third of the heads instead, but observed slightly worse performance.)

\parhead{Mamba Models}
For ``Mamba2,'' we use the HF implementation Mamba2ForCausalLM, with recommended configuration parameters (2 means expansion factor):
$$
\text{\texttt{ssm\_state\_size=64}, \texttt{num\_heads=16}, and \texttt{head\_dim=hidden\_size * 2 / num\_heads}.}
$$
This setup ensures each Mamba layer has $6d^2 + o(d^2)$ trainable parameters.
The recurrent state size (per layer) is therefore $2d \times \texttt{ssm\_state\_size}=128d$ plus conv1d.
We briefly tested \texttt{num\_heads=8} but observed worse results, so did not include it. The model initialization follows the HF default (which uses PyTorch default init as opposed to a fixed 0.02 std init).%
\footnote{We briefly tested the 0.02 init and did not observe a significant difference.}

For ``Mamba2(mlp),'' we use the same HF implementation but alternate between Mamba SSM layers and gated MLP layers. The intermediate size for gated MLP is $2d$, ensuring each MLP layer contains $6d^2$ trainable parameters. This ensures that $\ell$-layer $d$-dimensional Llama(RoPE) and Mamba2(mlp), as well as $2\ell$-layer $d$-dimensional Mamba2, have comparable parameter counts.

\parheadsc{Mamba1}
We briefly tested Mamba1 and found it consistently outperformed by Mamba2 in our pretraining playground, so we excluded it from main results. Notably, removing its \texttt{conv1d} layer also degrades Mamba1 to GLA-level performance.

\parheadsc{Mimetic initialization}
Following~\cite{trockman2024mimetic}, we enabled $A \approx 1$ (via $c{=}8$), $\Delta \approx 1$ (via $b_\Delta{=}0.54$), $W_C^\top W_B \approx I$, and $\mathrm{conv1d} \approx I$. We also tested $c{=}4$ and $c{=}2$ but observed no improvement.

\parhead{GLA Models}
For Gated Linear Attention (GLA)~\cite{yang2023gated}, we use the official \texttt{fla-org} implementation~\cite{yang2024fla}.%
\footnote{We use the default $\texttt{expand\_k}=0.5$ and $\texttt{expand\_v}=1$.
From March to May 2025, the repo authors updated \texttt{initializer\_range} to 0.006 (from the previously popular 0.02), which we found to negatively affect performance. We reverted it to 0.02; the authors also restored this value on May 3, 2025. We further disabled \texttt{rescale\_prenorm\_residual} for fair comparison. This option, inherited from GPT-2~\cite{radford2019language}, scales down the output projection (e.g., \texttt{o\_proj}) initialization by $1/\sqrt{N}$, where $N$ is the number of residual layers. The default HF implementations of Llama and Mamba2 both have this disabled, whereas the \texttt{fla-org} implementations enable it by default. We find that disabling this slightly improves model performance, and after communicating with \citet{yang2024fla}, they also disabled it on June 24, 2025.\label{footnote:gla} Some of these were introduced after V2.0 of this paper, leading to small diffs in experimental results compared to V1.1.}
We use 4 linear attention heads (their default configuration; also suggested by their first author).
With $d=512$ or $768$, this corresponds to $\texttt{headdim}=128$ or $192$, thus the recurrent state size (per layer) is $\frac{d}{2} \times \texttt{headdim} = 64d$ or $96d$ --- \emph{both smaller than the Mamba2 models we tested}.
We briefly tested 8 attention heads but found that these consistently degraded performance.
Each linear attention layer contains about $4d^2$ trainable parameters; the (gated) MLP has an intermediate size of $\frac{8d}{3}$, contributing roughly $8d^2$ parameters, matching Llama.

The default GLA implementation has disabled \texttt{conv1d} (the functionality was not part of the original publication~\cite{yang2023gated}), although their codebase supports \texttt{conv1d}, which we explicitly tested in this paper. They used 0.02 as initializer std for such \texttt{conv1d} layers with SiLU activation.

For GLA(elu) experiments in the ablation studies, we replaced the default feature map with $elu(x) + 1$, and conducted evaluations with and without \texttt{conv1d} and Canon layers.

\parhead{GDN Models}
For Gated DeltaNet (GDN)~\cite{yang2024gated}, we use the official \texttt{fla-org} implementation~\cite{yang2024fla}.%
\footnote{Similar to GLA (see \footnoteref{footnote:gla}), we adopt \texttt{initializer\_range}=0.02 and disable \texttt{rescale\_prenorm\_residual}. Note their default $\texttt{expand\_k}=0.75$ and $\texttt{expand\_v}=1.5$.}
We use 4 or 6 heads for $d=512$ or $d=768$, respectively (as suggested by their first author).
This corresponds to key/value headdim of $(96,192)$, giving a recurrent state size (per layer) of $144d$, comparable to Mamba2.
Each GDN layer contains about $6d^2$ trainable parameters, so we set the (gated) MLP intermediate size to yield another $6d^2$ parameters, matching Llama per layer block.

\parhead{Weight tying, tokenizer}
Unless otherwise stated (i.e., in Task \textsc{Capo}), we do not tie weights between the embedding and output layers in any of the architectures (e.g., Llama, Mamba, GLA, GDN). Additionally, no tokenizers are used during pretraining except for Task \textsc{Capo}.

\parhead{Task Capo}
The knowledge-capacity task pretrains on synthetic biographies following~\cite{AL2024-knowledgeScaling}.
For consistency, we use \texttt{GPT2Tokenizer} and tie embedding/output weights to minimize capacity loss in small models (though the effect is minor).

Since \textsc{Capo} measures bit-per-parameter knowledge capacity, both model and data scales are increased to assess scaling behavior.
Following~\cite{AL2024-knowledgeScaling}, we adopt the \textbf{$\ell$-$h$ notation} for model size, where Llama($\ell$-$h$) has $\ell$ layers, hidden size $64h$, and $h$ heads, and extend this convention to GLA, Mamba2, and GDN for comparability.%
\footnote{GLA: $\ell$ layers, hidden size $64h$, 4 fixed attention heads.
Mamba2: $2\ell$ layers, hidden size $64h$ (ssm state size 64 and num heads 16).
Mamba2(mlp): $\ell$ layers, hidden size $64h$ (ssm state size 64 and num heads 16).
GDN: $\ell$ layers, hidden size $64d$, and $\max\{4, 64d/128\}$ heads.
This ensures comparable parameter counts across architectures.}

GPT2 experiments in \figureref{fig:moe-capo} use the original GPT2 architecture augmented with RoPE, as in~\cite{AL2024-knowledgeScaling}.
Mixture-of-Experts (MoE) experiments employ \texttt{tutel}~\citep{tutel} with 32 standard MLP experts ($topk{=}1$, $cap\_factor{=}2$).

\subsection{Real-Life Experiments}
\label{app:arch:real-life}
For pretraining experiments on SlimPajama and FineWeb-Edu, we use all the architectures listed above alongside the Llama2 tokenizer (with vocab size 32,000)~\cite{llama2}. Weight tying is disabled to maintain consistency with prior works (e.g., \cite{behrouz2024titans,yang2024gated} and references therein).

The architectural configurations used in the real-life experiments are summarized below. They follow the setups described in \sectionref{app:arch}, except that we increase both width and depth to yield approximately 1.35B parameters per model:
\begin{itemize}
    \item \textbf{Llama (RoPE/NoPE):} 24 layers, 32 heads, hidden size $d=2048$.
    \item \textbf{GLA:} 24 layers, 4 heads, hidden size $d=2048$.
    \item \textbf{Mamba2:} 48 layers, 16 heads, hidden size $d=2048$.
    \item \textbf{Mamba2(mlp):} 24 layers, 16 heads, hidden size $d=2048$.
    \item \textbf{GDN:} 24 layers, 12 heads, hidden size $d=2048$.
\end{itemize}

\noindent
For linear models (excluding \texttt{conv1d}), the per-layer recurrent state sizes are $256d$ for GLA, $128d$ for Mamba2 (with twice the layers), and $192d$ for GDN. These are of the same order of magnitude, while remaining close to the original authors' recommended settings. Each model contains roughly $12d^2$ trainable parameters per layer (except Mamba2, which has $6d^2$ per layer but twice as many layers), ensuring a fair comparison across architectures.

\subsection{Canon Implementations}
Canon layers (i.e., types A, B, C, and D) in this paper are implemented using PyTorch's \texttt{nn.Conv1D} with kernel size 4, zero padding, and default initialization (i.e., \texttt{kaiming\_uniform\_} with $a=\sqrt{5}$). Unlike in GLA/GDN and in most linear models, this choice of the ``default initialization'' makes their weights initialized at $O(1)$ instead of 0.02. Based on our testing, this, combined with Canon's residual link, usually gives a \bblue{very stable performance improvement, without ever hurting}.

We use \texttt{causal\_conv1d}~\cite{fu2022hungry} for its fast CUDA implementation.
Canon layers are applied after layer normalization (if present, e.g., Canon-A/C) and before any non-linearity (if present, e.g., Canon-B/D).

We refer to the original conv1d implementations inside GLA/GDN/Mamba2 as Canon-b, and we leave its configuration identical to what was proposed by the original authors. In particular:
\begin{itemize}
\item conv1d in GLA/GDN has 0.02 initialization, with activation, without residual;
\item conv1d in Mamba2 has $O(1)$ initialization, with activation, without residual.
\end{itemize}

We refer to \texttt{cst-Canon} as the constant, untrained version of Canon(res), where the convolution weights are fixed to PyTorch’s default initialization.

Our implementation of Canon-ABCD for Llama, as well as Canon-AbCD for GLA, GDN and Mamba2, has been open-sourced on GitHub~\cite{PhysicsLM42} (up-to-date links at \href{https://physics.allen-zhu.com}{\texttt{physics.allen-zhu.com}}).

\clearpage
\section{Extensions of Figures \ref{fig:depo2-curve} and \ref{fig:depo1-curve-mamba}}
\label{app:figure-extensions}

\begin{figure}[H]
\centering
\setlength{\imgwidthBase}{0.32\textwidth}
\vspace{-3mm}
\includegraphics[page=1,trim={2.5mm 1.5mm 2.5mm 1.5mm},clip,width=\imgwidthBase]{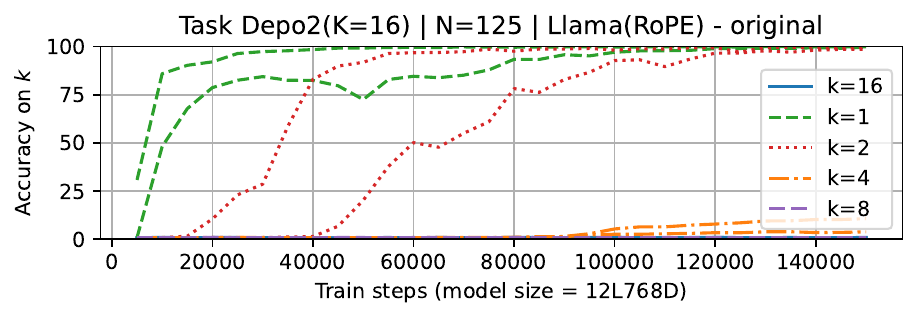}
\includegraphics[page=1,trim={2.5mm 1.5mm 2.5mm 1.5mm},clip,width=\imgwidthBase]{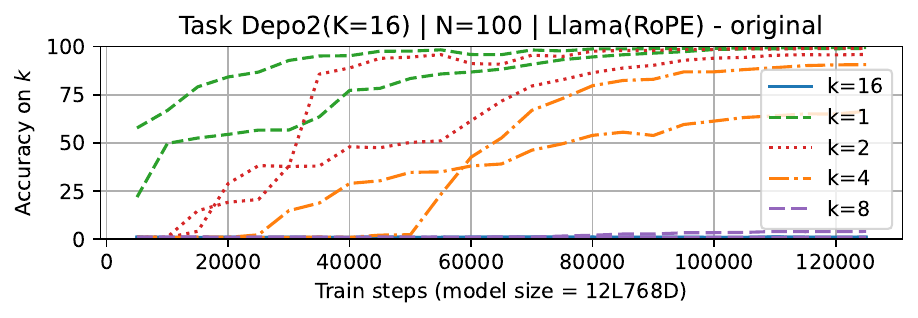}
\includegraphics[page=1,trim={2.5mm 1.5mm 2.5mm 1.5mm},clip,width=\imgwidthBase]{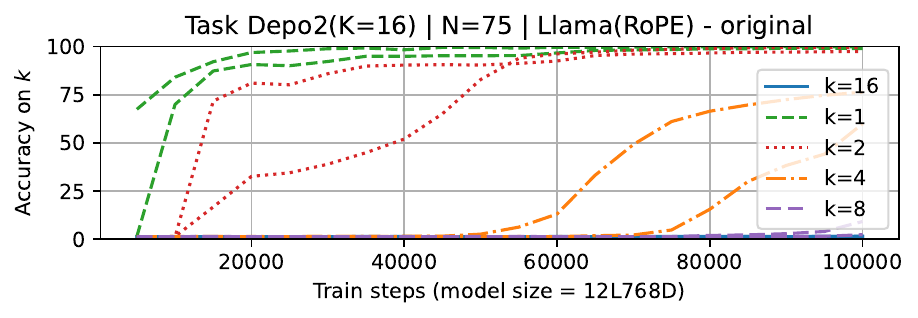}
\\
\includegraphics[page=1,trim={2.5mm 1.5mm 2.5mm 1.5mm},clip,width=\imgwidthBase]{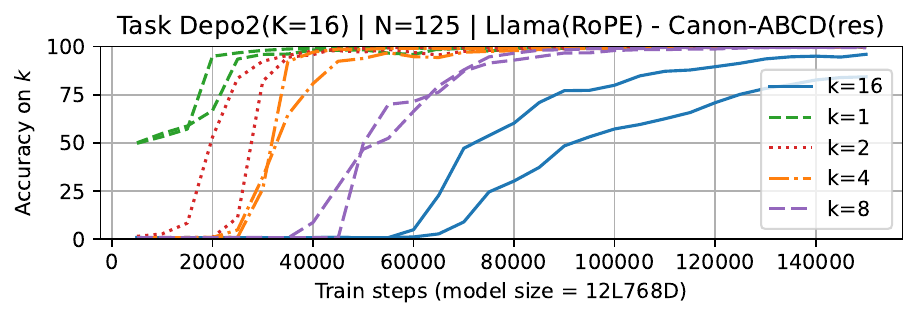}
\includegraphics[page=1,trim={2.5mm 1.5mm 2.5mm 1.5mm},clip,width=\imgwidthBase]{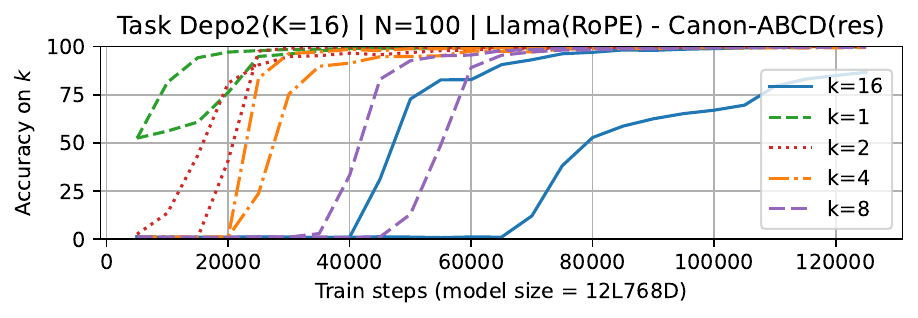}
\includegraphics[page=1,trim={2.5mm 1.5mm 2.5mm 1.5mm},clip,width=\imgwidthBase]{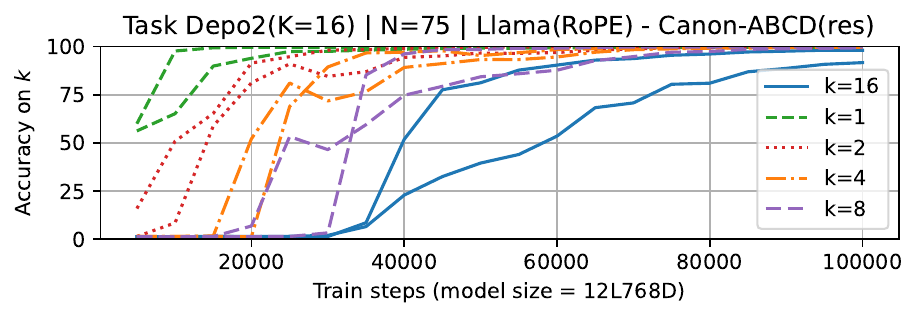}
\\
\vspace{-2mm}
\seplineb
\vspace{2mm}
\\
\includegraphics[page=1,trim={2.5mm 1.5mm 2.5mm 1.5mm},clip,width=\imgwidthBase]{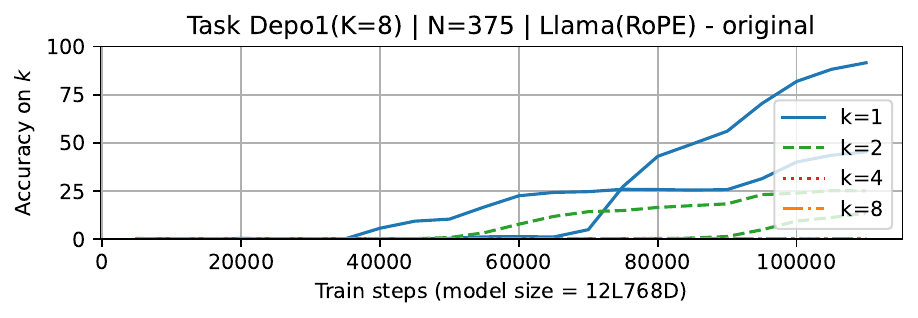}
\includegraphics[page=1,trim={2.5mm 1.5mm 2.5mm 1.5mm},clip,width=\imgwidthBase]{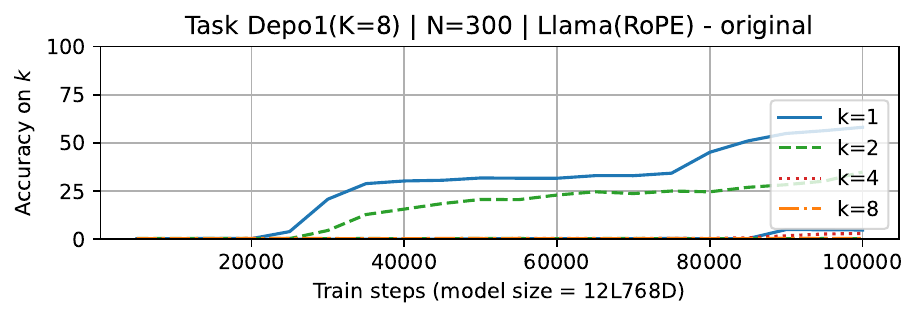}
\includegraphics[page=1,trim={2.5mm 1.5mm 2.5mm 1.5mm},clip,width=\imgwidthBase]{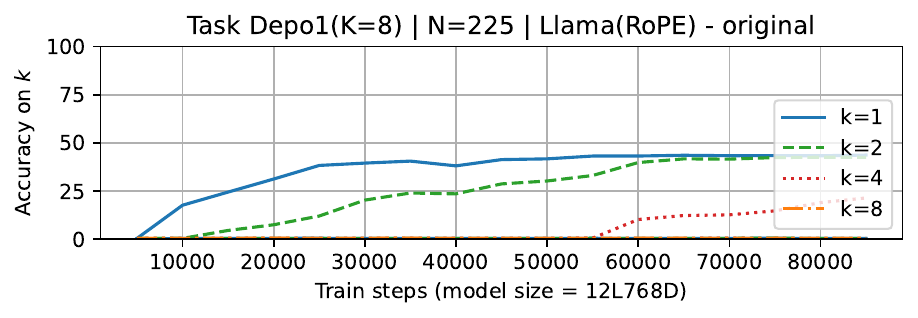}
\\
\includegraphics[page=1,trim={2.5mm 1.5mm 2.5mm 1.5mm},clip,width=\imgwidthBase]{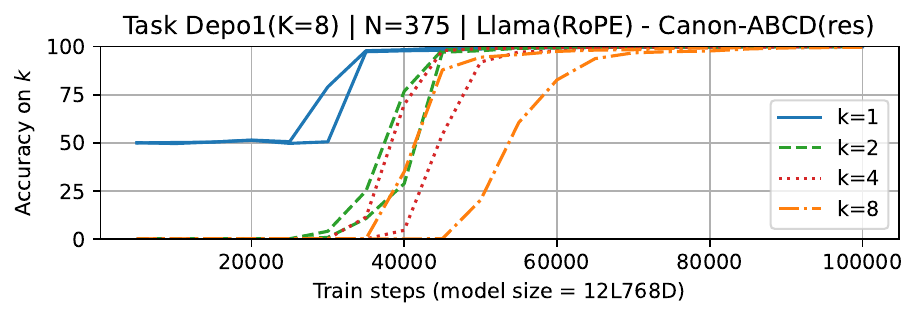}
\includegraphics[page=1,trim={2.5mm 1.5mm 2.5mm 1.5mm},clip,width=\imgwidthBase]{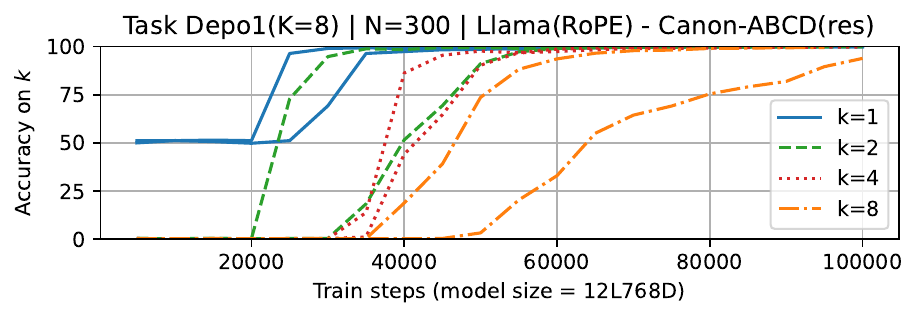}
\includegraphics[page=1,trim={2.5mm 1.5mm 2.5mm 1.5mm},clip,width=\imgwidthBase]{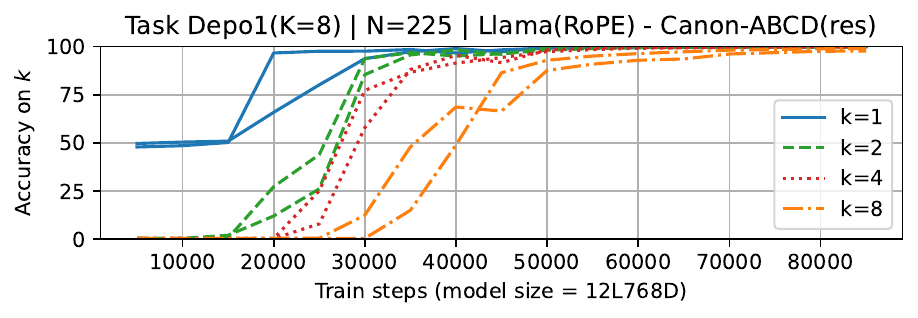}
\\
\vspace{-2mm}
\seplineb
\vspace{2mm}
\\
\includegraphics[page=1,trim={2.5mm 1.5mm 2.5mm 1.5mm},clip,width=\imgwidthBase]{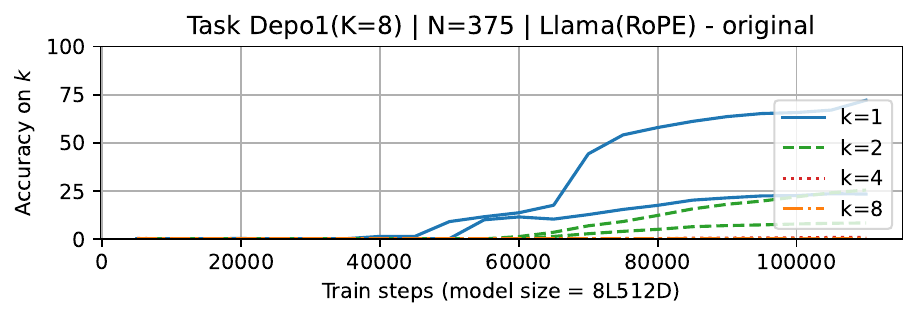}
\includegraphics[page=1,trim={2.5mm 1.5mm 2.5mm 1.5mm},clip,width=\imgwidthBase]{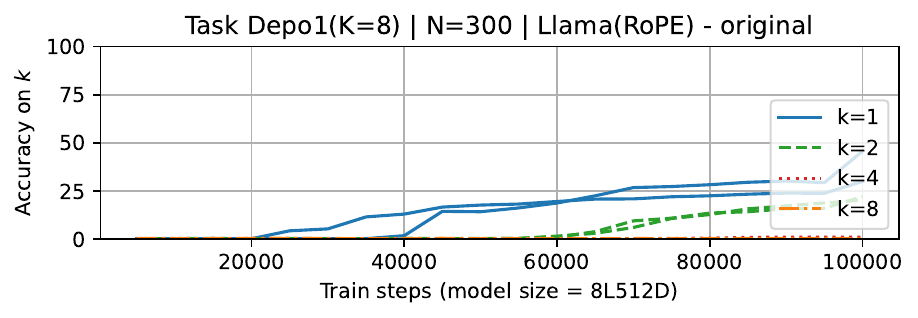}
\includegraphics[page=1,trim={2.5mm 1.5mm 2.5mm 1.5mm},clip,width=\imgwidthBase]{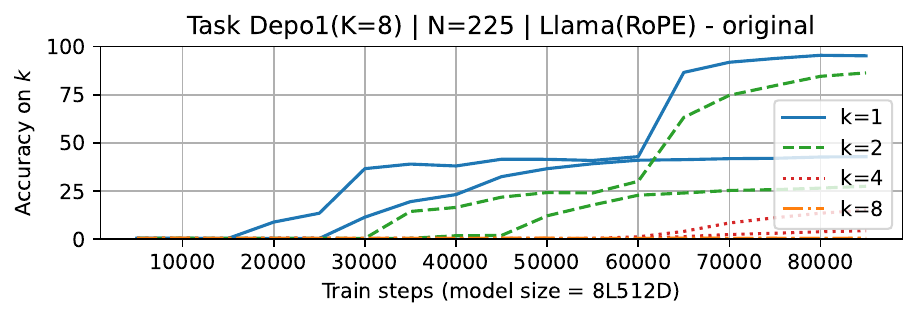}
\\
\includegraphics[page=1,trim={2.5mm 1.5mm 2.5mm 1.5mm},clip,width=\imgwidthBase]{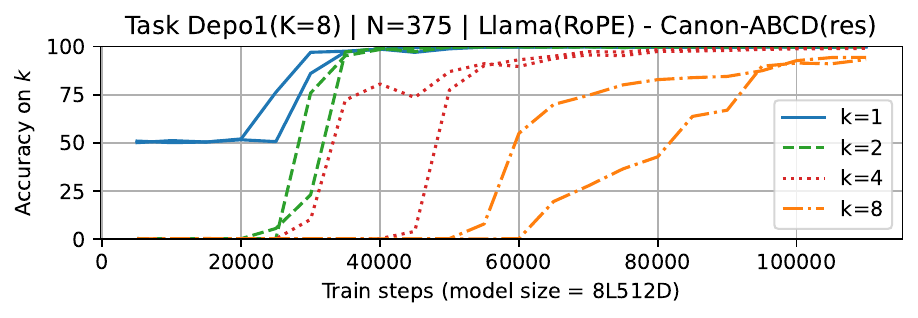}
\includegraphics[page=1,trim={2.5mm 1.5mm 2.5mm 1.5mm},clip,width=\imgwidthBase]{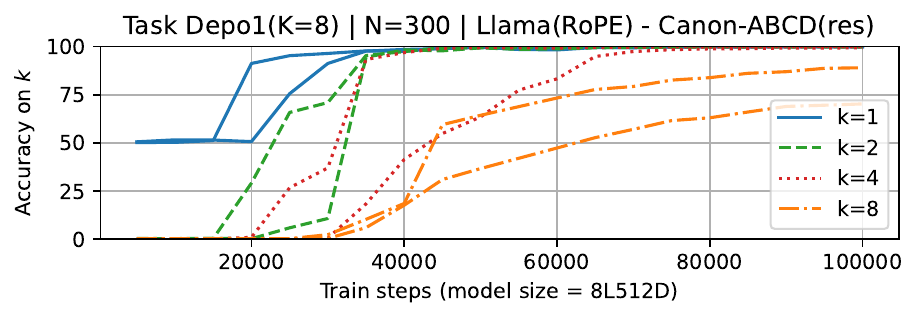}
\includegraphics[page=1,trim={2.5mm 1.5mm 2.5mm 1.5mm},clip,width=\imgwidthBase]{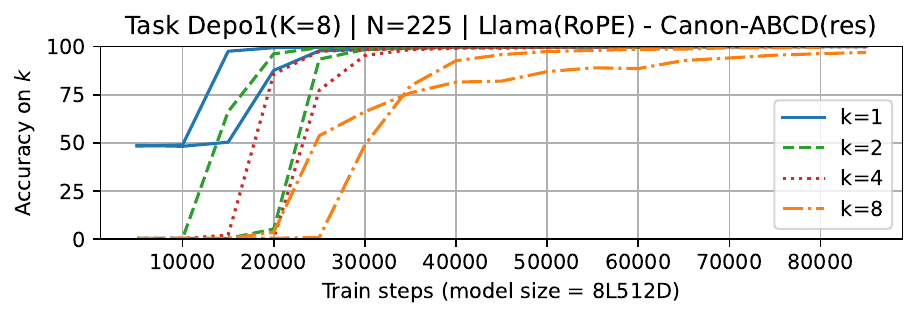}
\\
\vspace{-2mm}
\seplineb
\vspace{2mm}
\\
\includegraphics[page=1,trim={2.5mm 1.5mm 2.5mm 1.5mm},clip,width=\imgwidthBase]{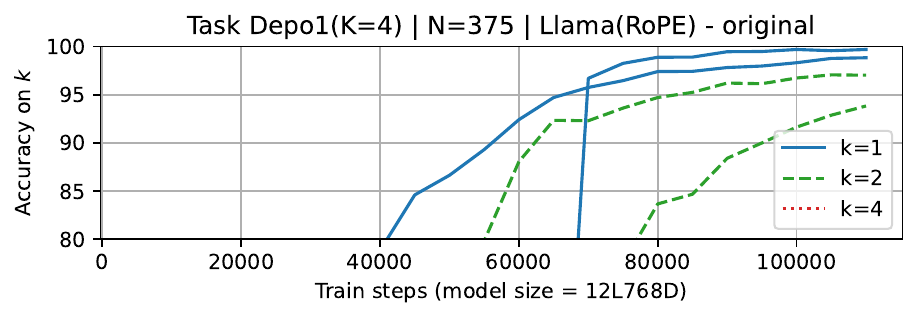}
\includegraphics[page=1,trim={2.5mm 1.5mm 2.5mm 1.5mm},clip,width=\imgwidthBase]{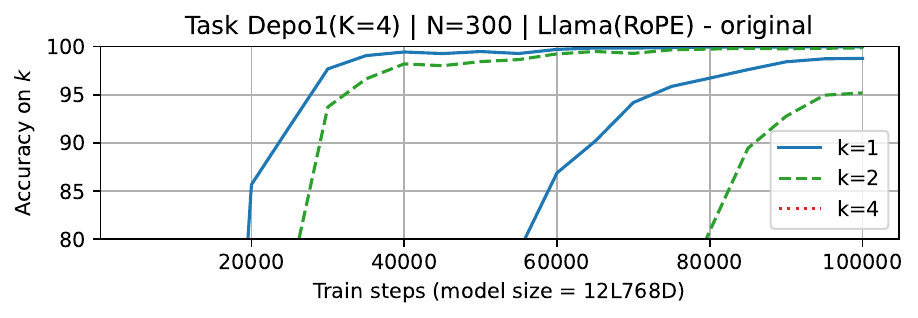}
\includegraphics[page=1,trim={2.5mm 1.5mm 2.5mm 1.5mm},clip,width=\imgwidthBase]{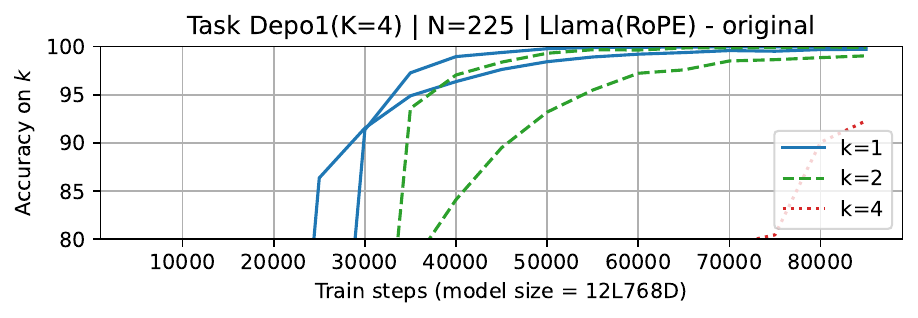}
\\
\includegraphics[page=1,trim={2.5mm 1.5mm 2.5mm 1.5mm},clip,width=\imgwidthBase]{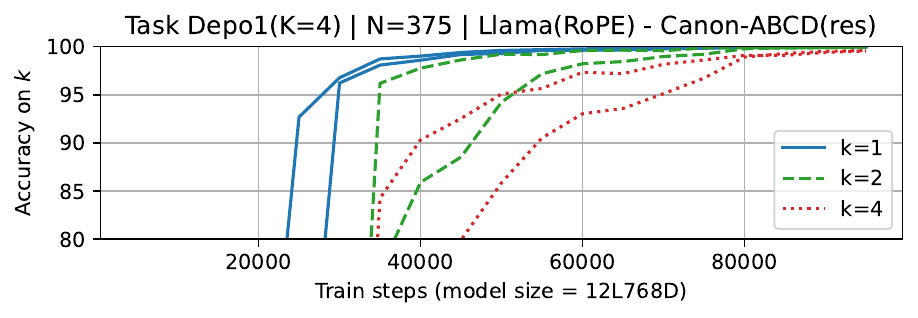}
\includegraphics[page=1,trim={2.5mm 1.5mm 2.5mm 1.5mm},clip,width=\imgwidthBase]{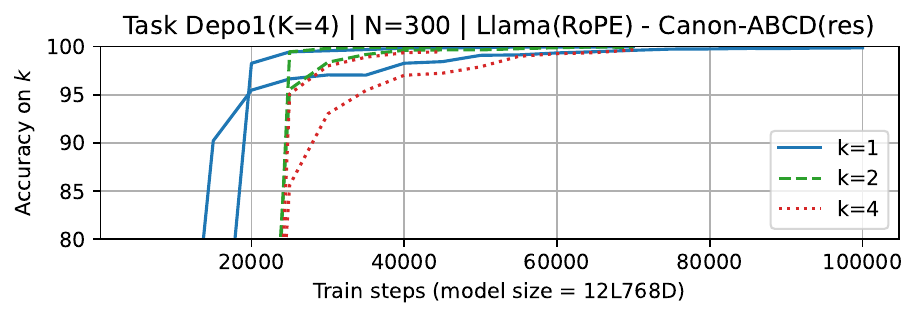}
\includegraphics[page=1,trim={2.5mm 1.5mm 2.5mm 1.5mm},clip,width=\imgwidthBase]{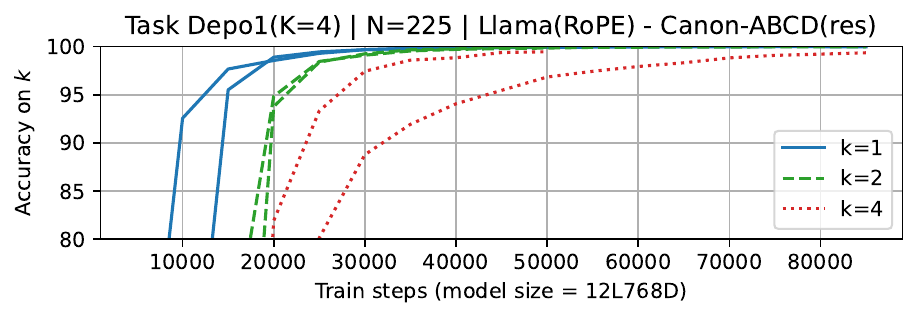}
\\
\vspace{-2mm}
\seplineb
\vspace{2mm}
\\
\includegraphics[page=1,trim={2.5mm 1.5mm 2.5mm 1.5mm},clip,width=\imgwidthBase]{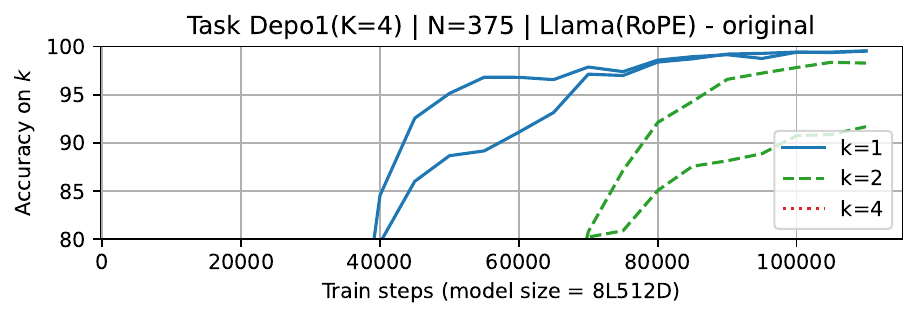}
\includegraphics[page=1,trim={2.5mm 1.5mm 2.5mm 1.5mm},clip,width=\imgwidthBase]{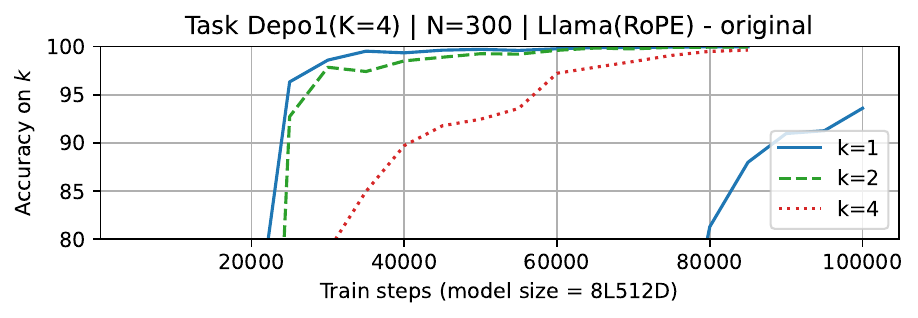}
\includegraphics[page=1,trim={2.5mm 1.5mm 2.5mm 1.5mm},clip,width=\imgwidthBase]{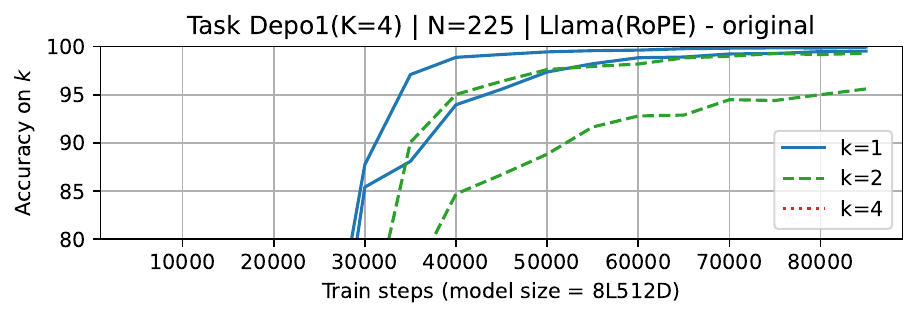}
\\
\includegraphics[page=1,trim={2.5mm 1.5mm 2.5mm 1.5mm},clip,width=\imgwidthBase]{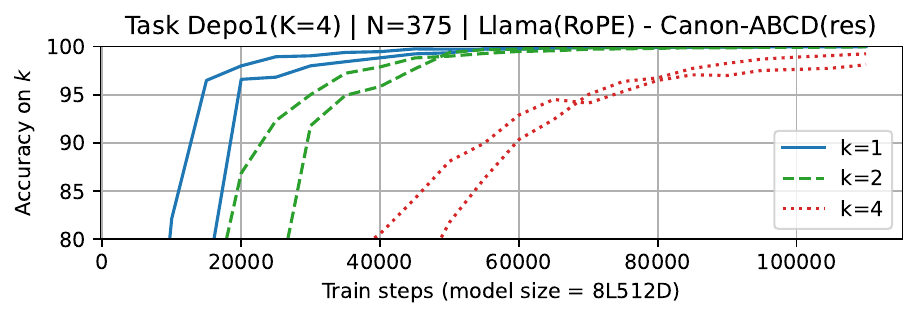}
\includegraphics[page=1,trim={2.5mm 1.5mm 2.5mm 1.5mm},clip,width=\imgwidthBase]{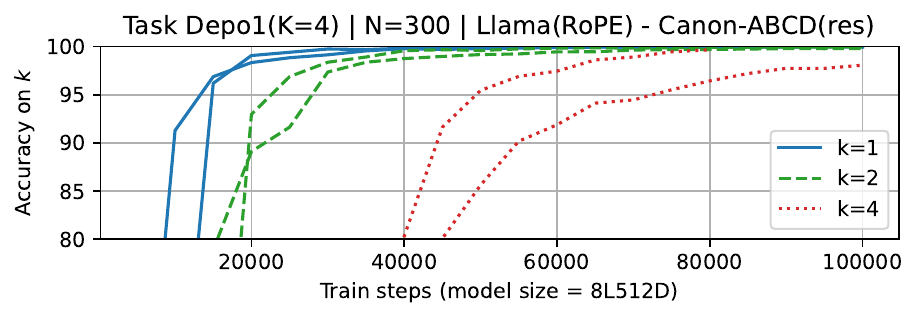}
\includegraphics[page=1,trim={2.5mm 1.5mm 2.5mm 1.5mm},clip,width=\imgwidthBase]{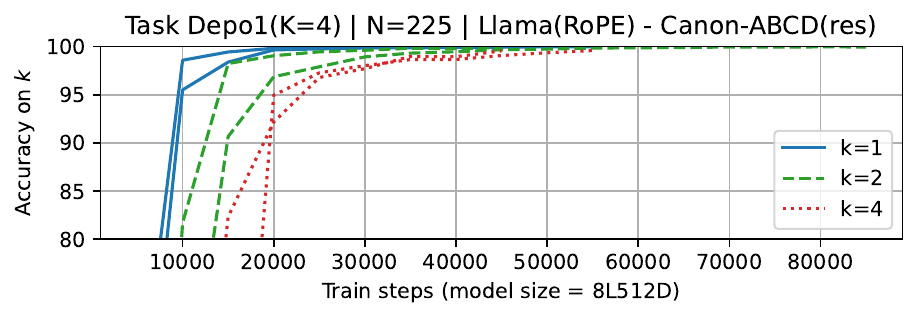}
\caption{\label{fig:depo2-curve:ext}\bblue{This is an extension of \figureref{fig:depo2-curve}}: Training curves for 12L768 and 8L512D RoPE models, with and without Canon layers, on Task \textsc{Depo2}($K=16$), \textsc{Depo1}($K=8$), \textsc{Depo1}($K=4$), evaluated at varied depths and maximum graph size $n=N$, shown at the two best learning rates.}
\end{figure}

\begin{figure}[H]
\centering
\setlength{\imgwidthBase}{0.31\textwidth}
\includegraphics[page=1,trim={2.5mm 1.5mm 2.5mm 1.5mm},clip,width=\imgwidthBase]{curve-perm_4/N=375/Llama_RoPE_-Canon-ABCD_res_}
\includegraphics[page=1,trim={2.5mm 1.5mm 2.5mm 1.5mm},clip,width=\imgwidthBase]{curve-perm_4/N=300/Llama_RoPE_-Canon-ABCD_res_}
\includegraphics[page=1,trim={2.5mm 1.5mm 2.5mm 1.5mm},clip,width=\imgwidthBase]{curve-perm_4/N=225/Llama_RoPE_-Canon-ABCD_res_}
\\
\includegraphics[page=1,trim={2.5mm 1.5mm 2.5mm 1.5mm},clip,width=\imgwidthBase]{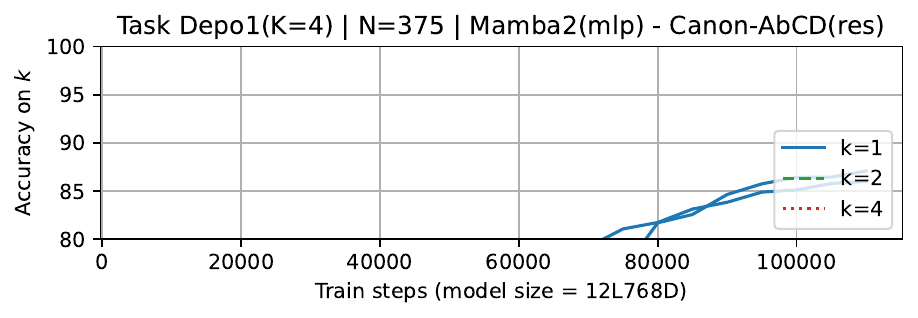}
\includegraphics[page=1,trim={2.5mm 1.5mm 2.5mm 1.5mm},clip,width=\imgwidthBase]{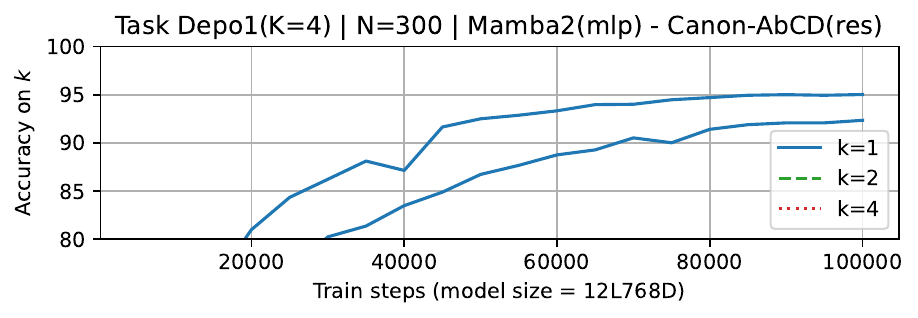}
\includegraphics[page=1,trim={2.5mm 1.5mm 2.5mm 1.5mm},clip,width=\imgwidthBase]{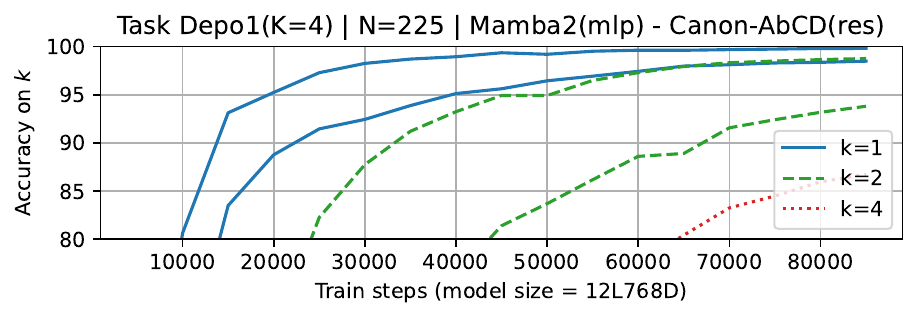}
\\
\includegraphics[page=1,trim={2.5mm 1.5mm 2.5mm 1.5mm},clip,width=\imgwidthBase]{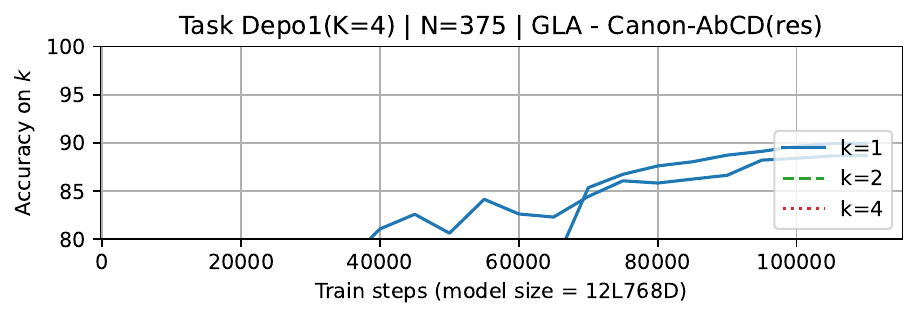}
\includegraphics[page=1,trim={2.5mm 1.5mm 2.5mm 1.5mm},clip,width=\imgwidthBase]{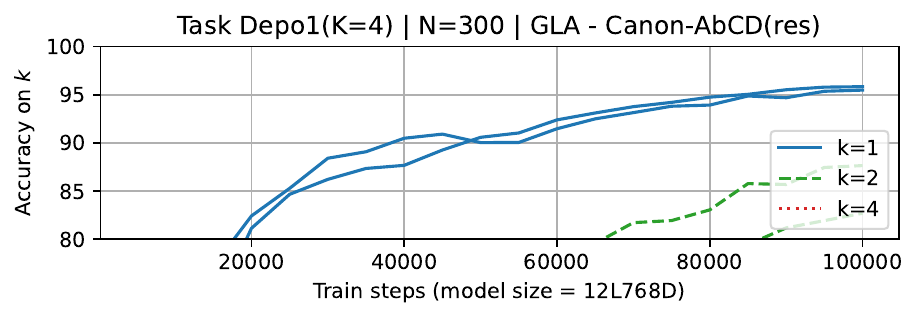}
\includegraphics[page=1,trim={2.5mm 1.5mm 2.5mm 1.5mm},clip,width=\imgwidthBase]{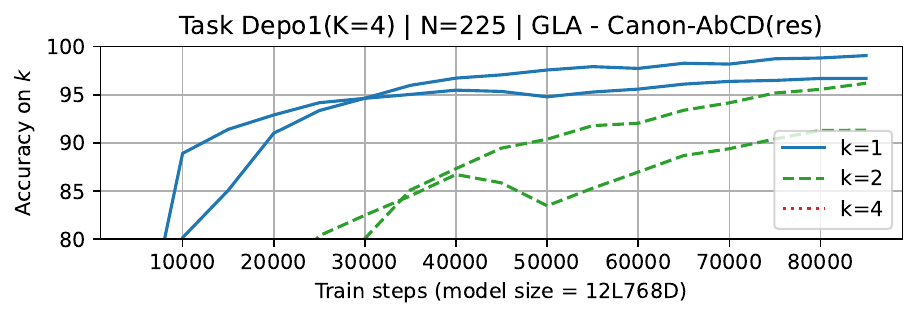}
\\
\includegraphics[page=1,trim={2.5mm 1.5mm 2.5mm 1.5mm},clip,width=\imgwidthBase]{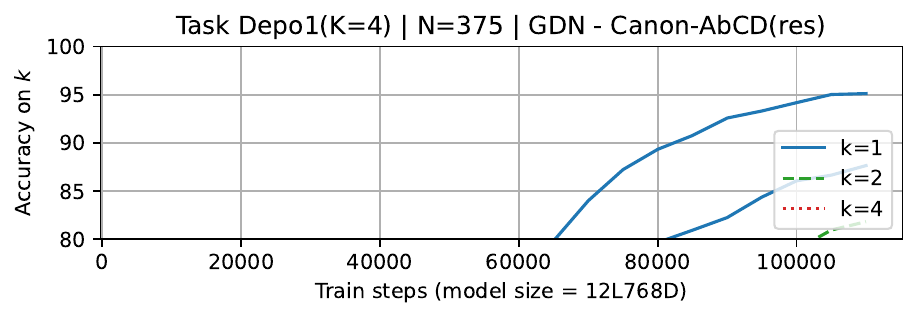}
\includegraphics[page=1,trim={2.5mm 1.5mm 2.5mm 1.5mm},clip,width=\imgwidthBase]{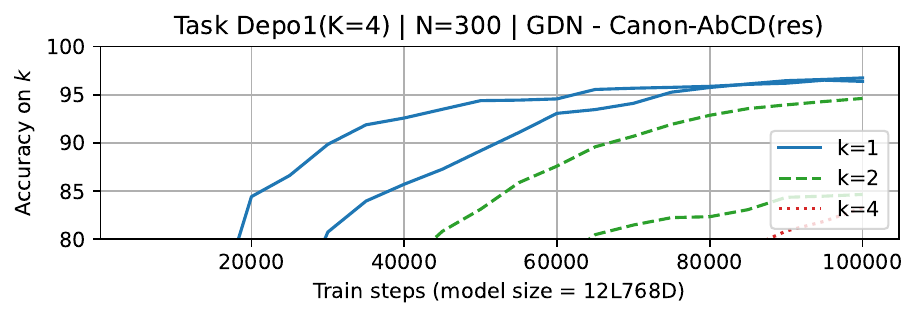}
\includegraphics[page=1,trim={2.5mm 1.5mm 2.5mm 1.5mm},clip,width=\imgwidthBase]{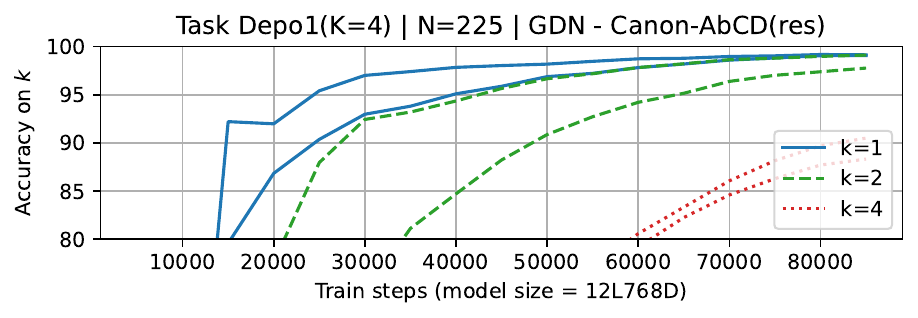}
\\
\vspace{-2mm}
\seplineb
\vspace{2mm}
\\
\includegraphics[page=1,trim={2.5mm 1.5mm 2.5mm 1.5mm},clip,width=\imgwidthBase]{curve-perm/N=375/Llama_RoPE_-Canon-ABCD_res_}
\includegraphics[page=1,trim={2.5mm 1.5mm 2.5mm 1.5mm},clip,width=\imgwidthBase]{curve-perm/N=300/Llama_RoPE_-Canon-ABCD_res_}
\includegraphics[page=1,trim={2.5mm 1.5mm 2.5mm 1.5mm},clip,width=\imgwidthBase]{curve-perm/N=225/Llama_RoPE_-Canon-ABCD_res_}
\\
\includegraphics[page=1,trim={2.5mm 1.5mm 2.5mm 1.5mm},clip,width=\imgwidthBase]{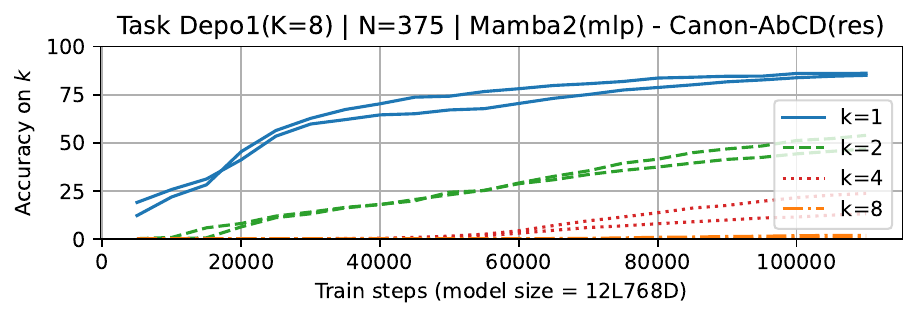}
\includegraphics[page=1,trim={2.5mm 1.5mm 2.5mm 1.5mm},clip,width=\imgwidthBase]{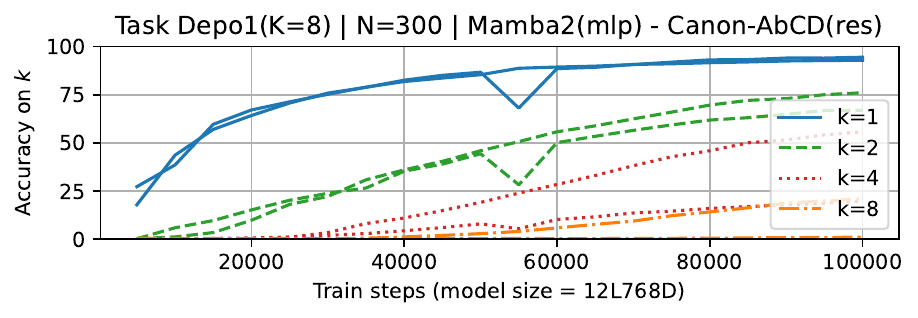}
\includegraphics[page=1,trim={2.5mm 1.5mm 2.5mm 1.5mm},clip,width=\imgwidthBase]{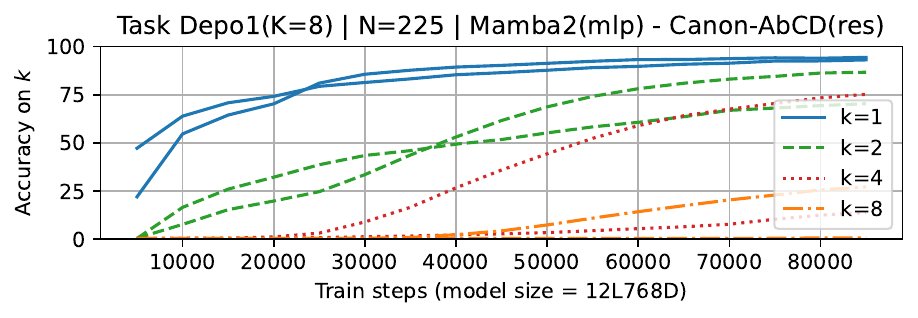}
\\
\includegraphics[page=1,trim={2.5mm 1.5mm 2.5mm 1.5mm},clip,width=\imgwidthBase]{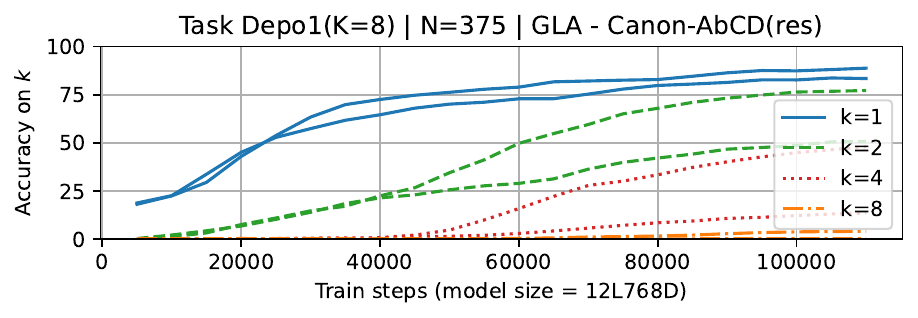}
\includegraphics[page=1,trim={2.5mm 1.5mm 2.5mm 1.5mm},clip,width=\imgwidthBase]{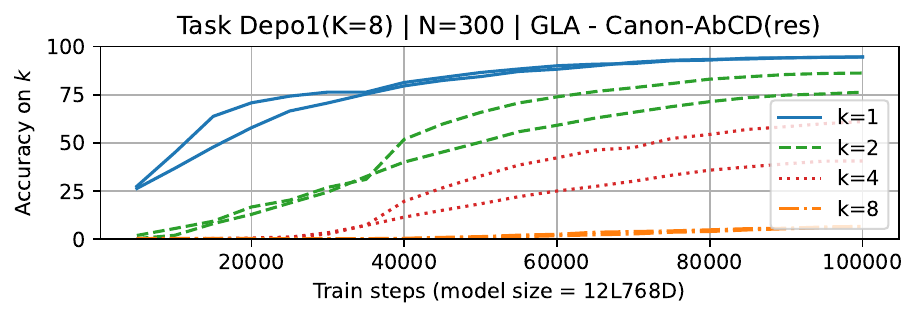}
\includegraphics[page=1,trim={2.5mm 1.5mm 2.5mm 1.5mm},clip,width=\imgwidthBase]{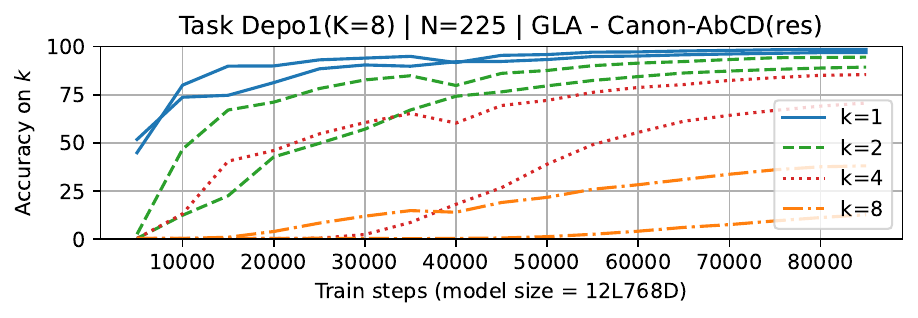}
\\
\includegraphics[page=1,trim={2.5mm 1.5mm 2.5mm 1.5mm},clip,width=\imgwidthBase]{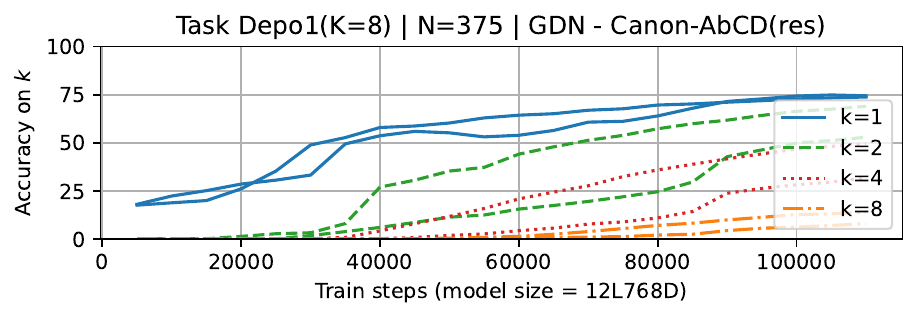}
\includegraphics[page=1,trim={2.5mm 1.5mm 2.5mm 1.5mm},clip,width=\imgwidthBase]{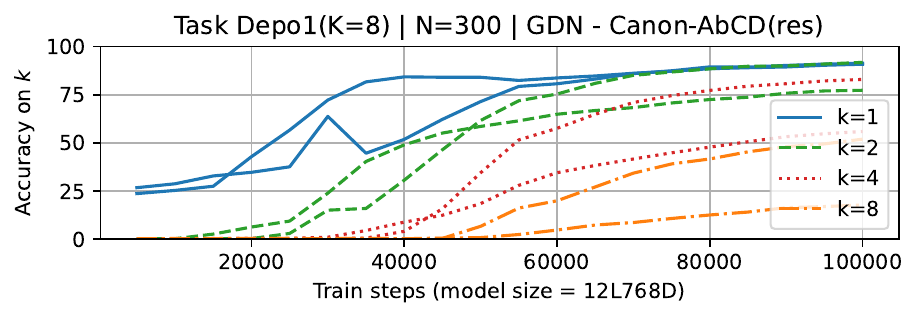}
\includegraphics[page=1,trim={2.5mm 1.5mm 2.5mm 1.5mm},clip,width=\imgwidthBase]{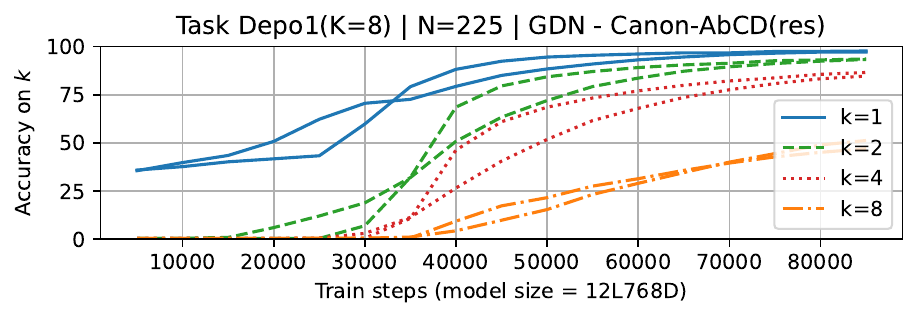}
\\
\vspace{-2mm}
\seplineb
\vspace{2mm}
\\
\includegraphics[page=1,trim={2.5mm 1.5mm 2.5mm 1.5mm},clip,width=\imgwidthBase]{curve-perm_multi/N=125/Llama_RoPE_-Canon-ABCD_res_}
\includegraphics[page=1,trim={2.5mm 1.5mm 2.5mm 1.5mm},clip,width=\imgwidthBase]{curve-perm_multi/N=100/Llama_RoPE_-Canon-ABCD_res_}
\includegraphics[page=1,trim={2.5mm 1.5mm 2.5mm 1.5mm},clip,width=\imgwidthBase]{curve-perm_multi/N=75/Llama_RoPE_-Canon-ABCD_res_}
\\
\includegraphics[page=1,trim={2.5mm 1.5mm 2.5mm 1.5mm},clip,width=\imgwidthBase]{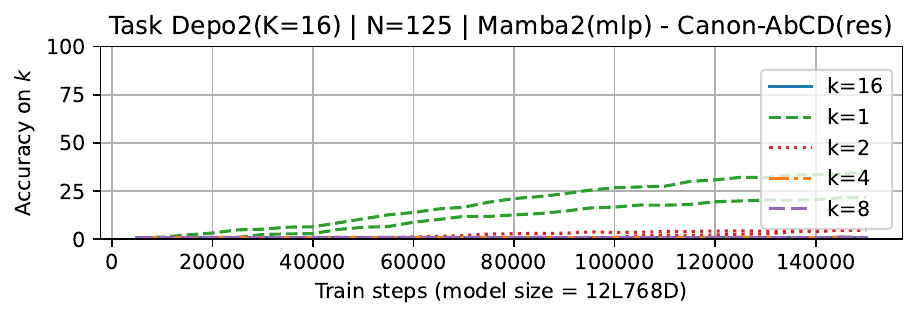}
\includegraphics[page=1,trim={2.5mm 1.5mm 2.5mm 1.5mm},clip,width=\imgwidthBase]{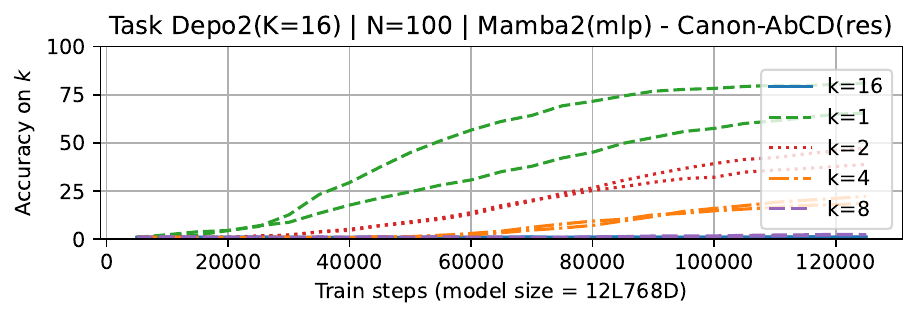}
\includegraphics[page=1,trim={2.5mm 1.5mm 2.5mm 1.5mm},clip,width=\imgwidthBase]{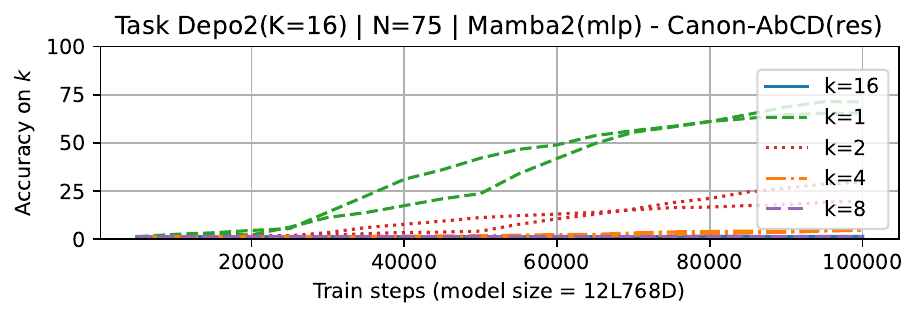}
\\
\includegraphics[page=1,trim={2.5mm 1.5mm 2.5mm 1.5mm},clip,width=\imgwidthBase]{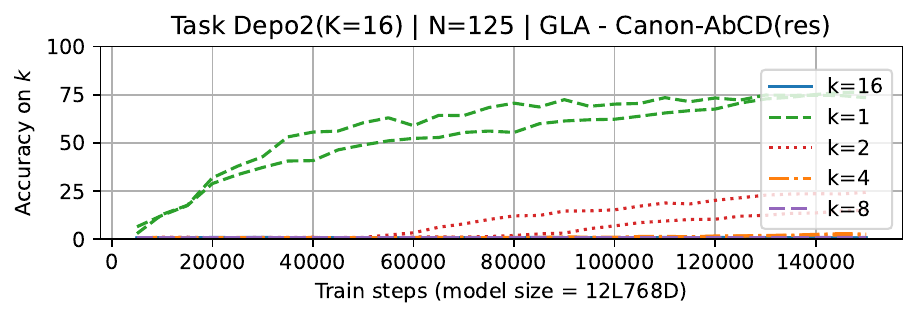}
\includegraphics[page=1,trim={2.5mm 1.5mm 2.5mm 1.5mm},clip,width=\imgwidthBase]{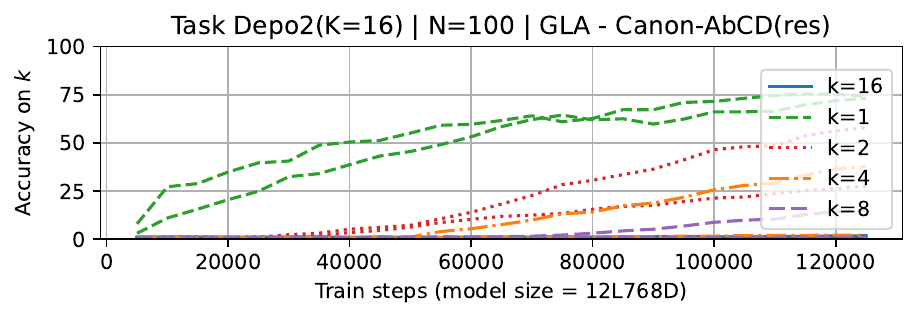}
\includegraphics[page=1,trim={2.5mm 1.5mm 2.5mm 1.5mm},clip,width=\imgwidthBase]{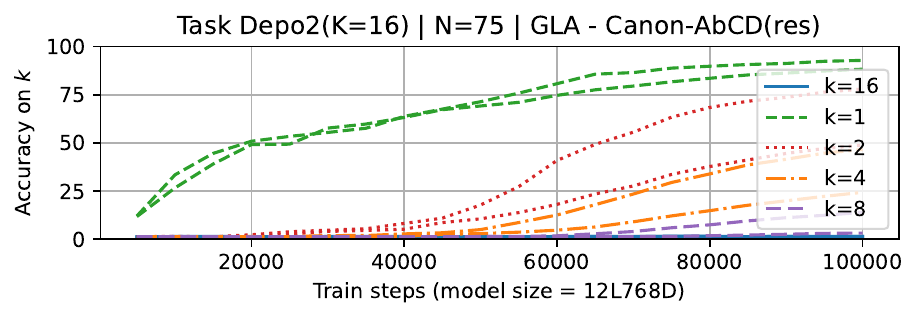}
\\
\includegraphics[page=1,trim={2.5mm 1.5mm 2.5mm 1.5mm},clip,width=\imgwidthBase]{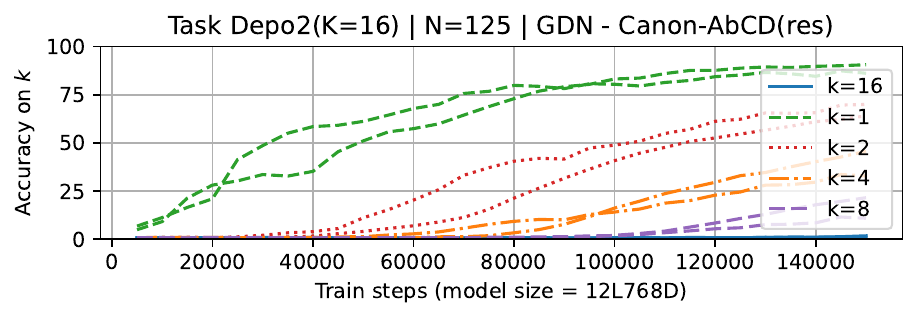}
\includegraphics[page=1,trim={2.5mm 1.5mm 2.5mm 1.5mm},clip,width=\imgwidthBase]{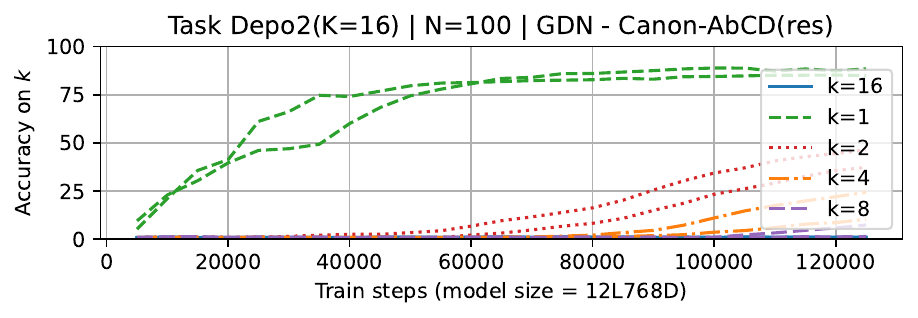}
\includegraphics[page=1,trim={2.5mm 1.5mm 2.5mm 1.5mm},clip,width=\imgwidthBase]{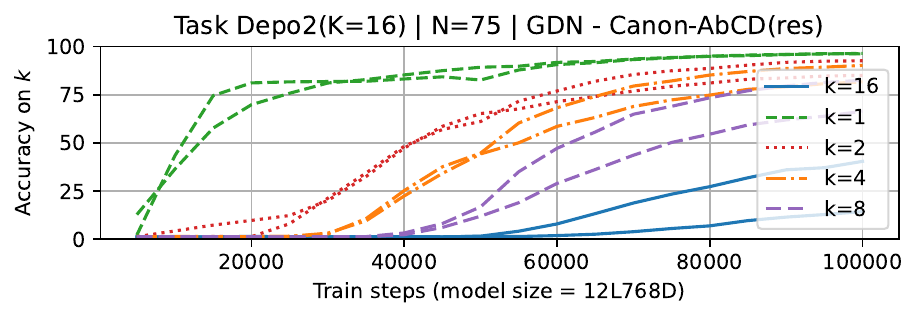}
\caption{\label{fig:depo1-curve-mamba:ext}%
\bblue{This is an extension of \figureref{fig:depo1-curve-mamba}}: Training curves for 12L768D architectures on Task \textsc{Depo1}(K=4 or 8), \textsc{Depo2}(K=16), evaluated at varied $k$ and maximum $n = N$; two best learning rates for each $k$.
}
\end{figure}

\clearpage
\section{More Real-Life Experiments}

\subsection{Insufficiencies of Real-Life Pretraining at Academic Scale}
\label{app:real-life:random}

\begin{figure}[H]
\centering
\setlength{\imgwidthBase}{0.245\textwidth}
\includegraphics[page=1,trim={2.5mm 1.5mm 2.5mm 1.5mm},clip,height=\imgwidthBase]{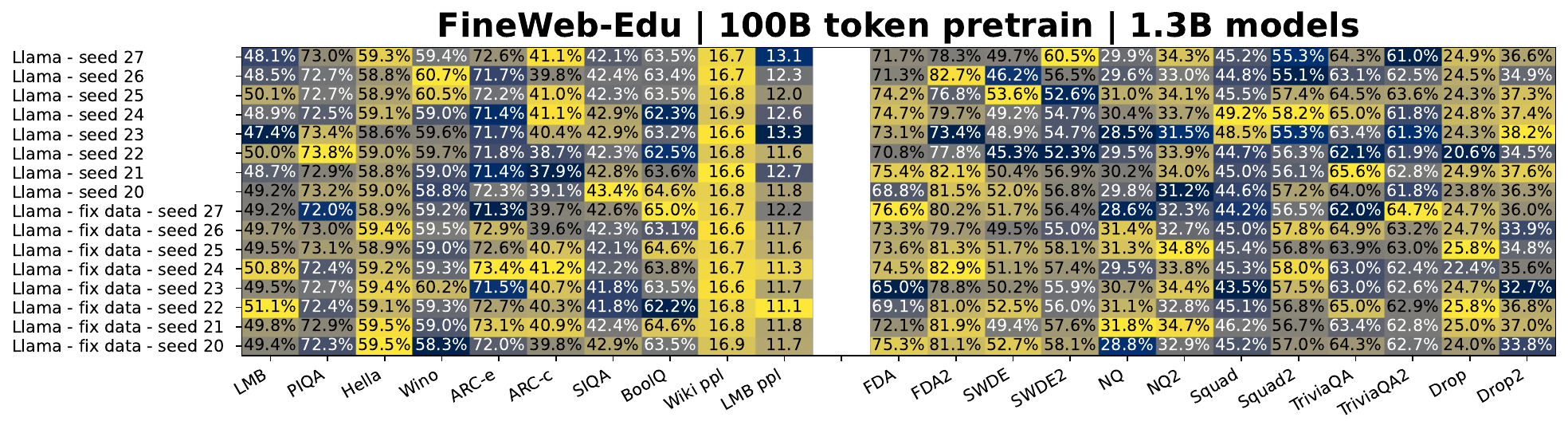}
\includegraphics[page=1,trim={2.5mm 1.5mm 2.5mm 1.5mm},clip,height=\imgwidthBase]{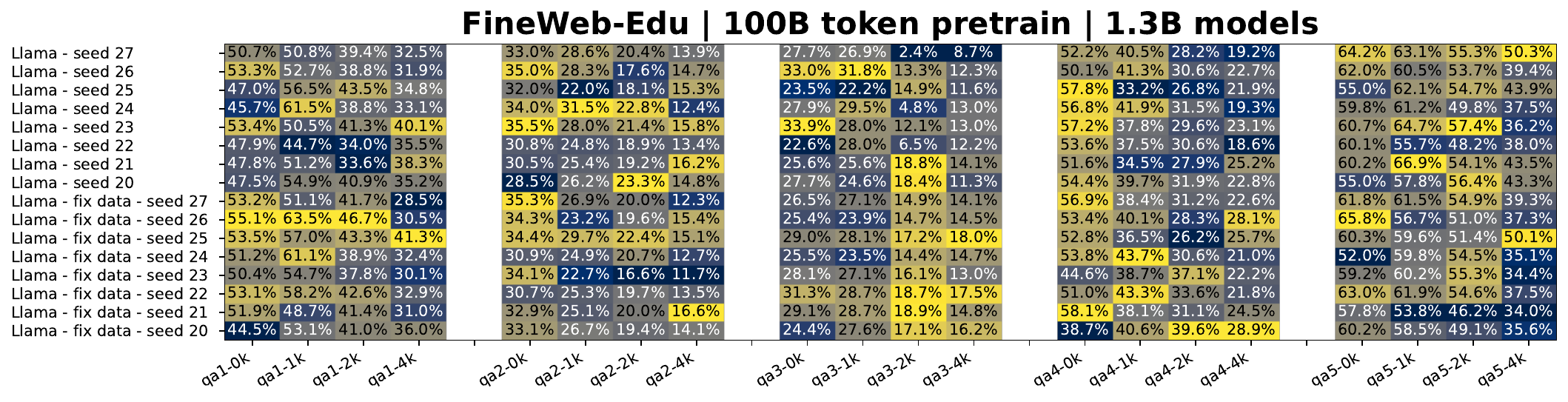}
\includegraphics[page=1,trim={2.5mm 1.5mm 2.5mm 1.5mm},clip,height=\imgwidthBase]{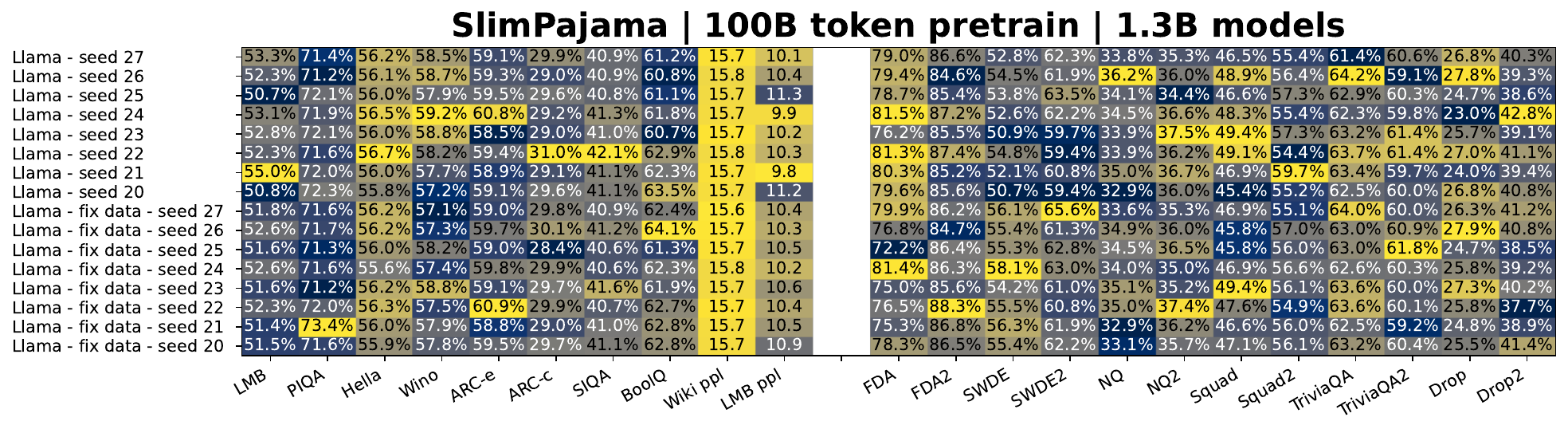}
\includegraphics[page=1,trim={2.5mm 1.5mm 2.5mm 1.5mm},clip,height=\imgwidthBase]{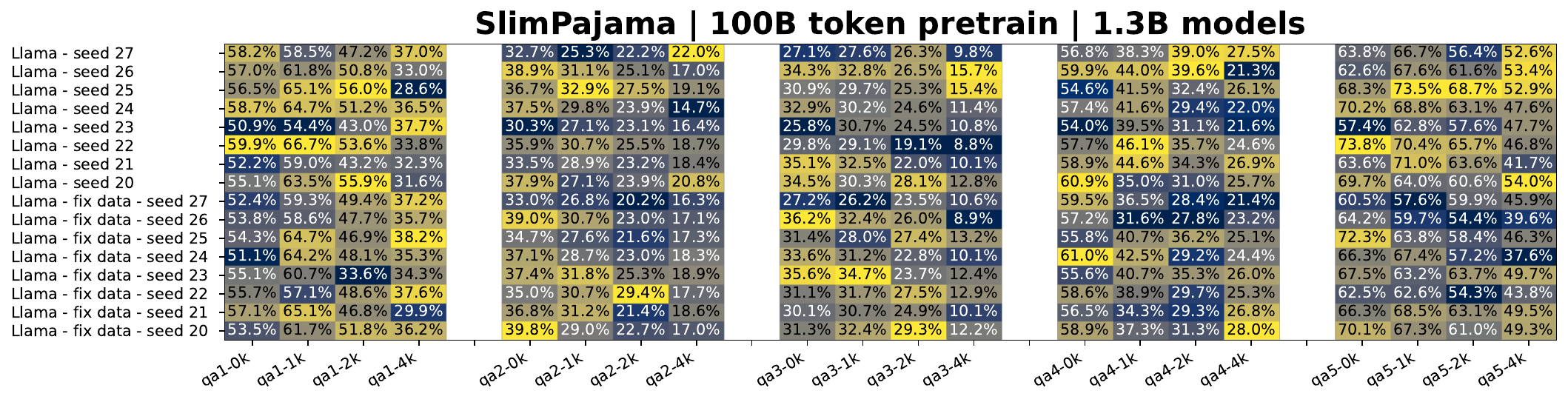}
\captionsetup{singlelinecheck=false}
\caption{\label{fig:real-life:random}%
\textbf{Extended results for \figureref{fig:illustrate1}} showing strong variance in benchmark accuracies for academic-scale real-life pretraining (1.3B models trained for 100B tokens).
\newline
\bblue{Observations:} Accuracy varies greatly across random seeds—both when changing data and initialization, and even when fixing data but varying initialization.
HellaSwag~\cite{zellers2019hellaswag} and wiki-ppl are relatively stable, though perplexity alone is an unreliable indicator of model capability.
}
\end{figure}

As shown earlier in \figureref{fig:illustrate1}, real-life pretrained models (FineWeb-Edu or SlimPajama) display large performance variance across random seeds.
Here, we expand those results in \figureref{fig:real-life:random}, including full experiments over eight seeds.
Following feedback from an anonymous NeurIPS~2025 reviewer, we further test a controlled setup where data order is fixed and only model initialization varies—yet substantial benchmark variance remains.

\subsection{Complete Real-Life Experiments}

\begin{figure}[H]
\centering
\setlength{\imgwidthBase}{1.02\textwidth}
\vspace{-5mm}
\hspace*{-5mm}
\includegraphics[page=1,trim={2.5mm 1.5mm 2.5mm 1.5mm},clip,width=\imgwidthBase]{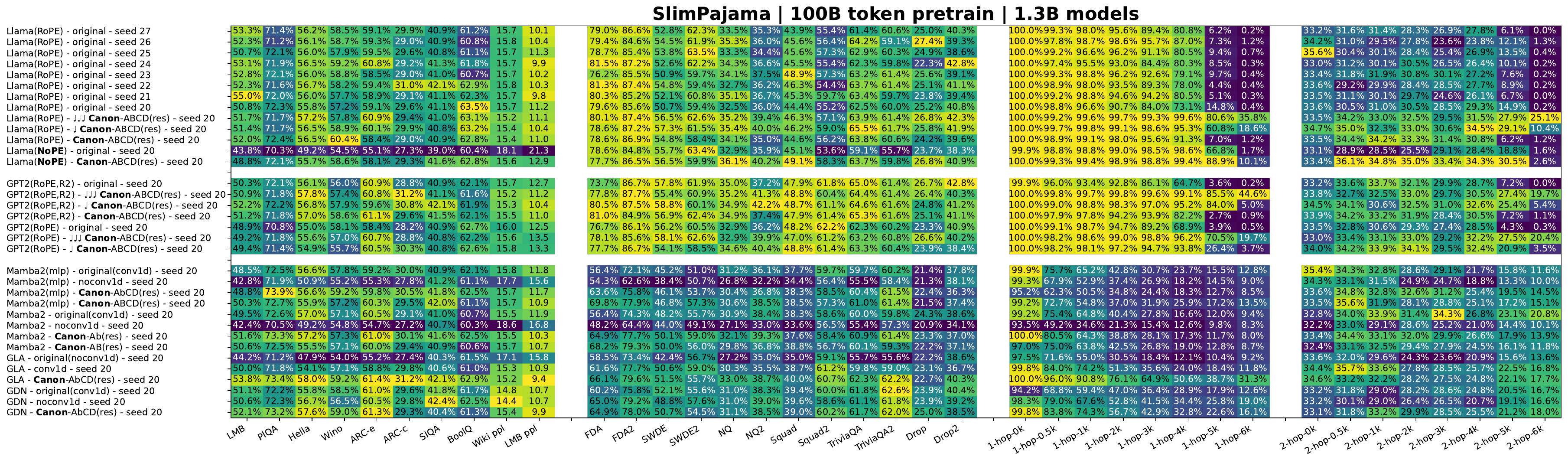}
\hspace*{-5mm}
\\
\hspace*{-5mm}
\includegraphics[page=1,trim={2.5mm 1.5mm 2.5mm 1.5mm},clip,width=\imgwidthBase]{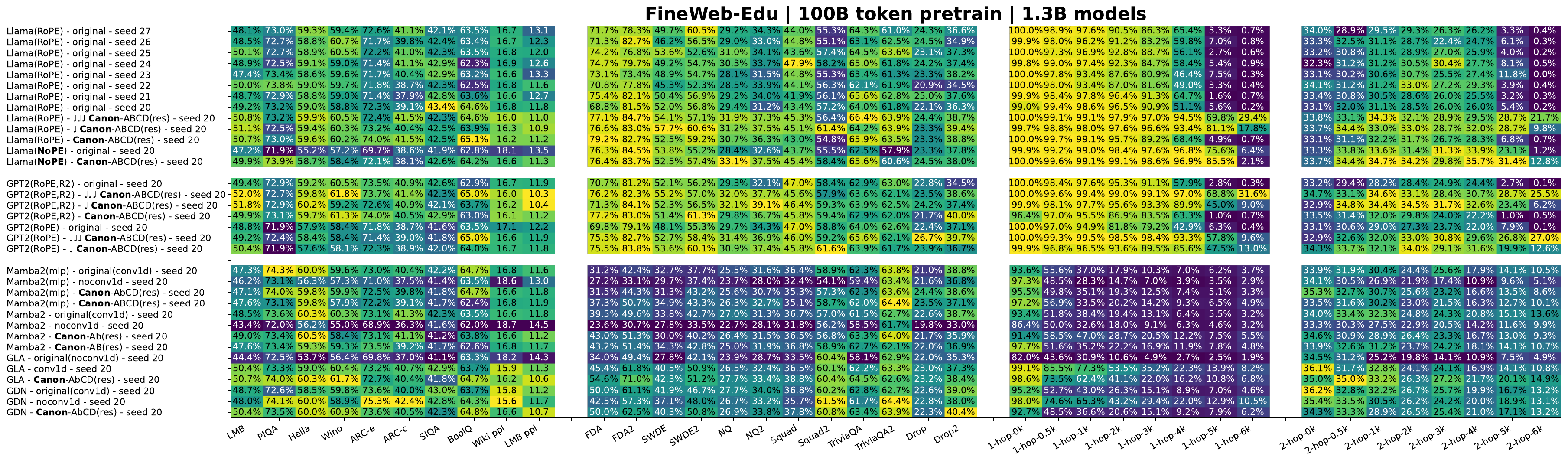}
\hspace*{-5mm}
\caption{\label{fig:real-life:include-gpt}%
This is identical to \figureref{fig:real-life} but \textbf{additionally includes GPT2(RoPE)} models—identical to Llama(RoPE) but using standard MLPs—\textbf{and GPT2(RoPE,R2)}, which uses $\text{ReLU}^2$ activation~\cite{so2109primer}. \bblue{Key conclusions remain unchanged}: reducing RoPE improves length generalization, and many architectural differences (e.g., standard vs.\ gated MLP, \texttt{SiLU} vs.\ $\texttt{ReLU}^2$) are \textbf{buried in noise}.
}
\end{figure}

\begin{figure}[H]
\centering
\setlength{\imgwidthBase}{1.02\textwidth}
\hspace*{-5mm}
\includegraphics[page=1,trim={2.5mm 1.5mm 2.5mm 1.5mm},clip,width=\imgwidthBase]{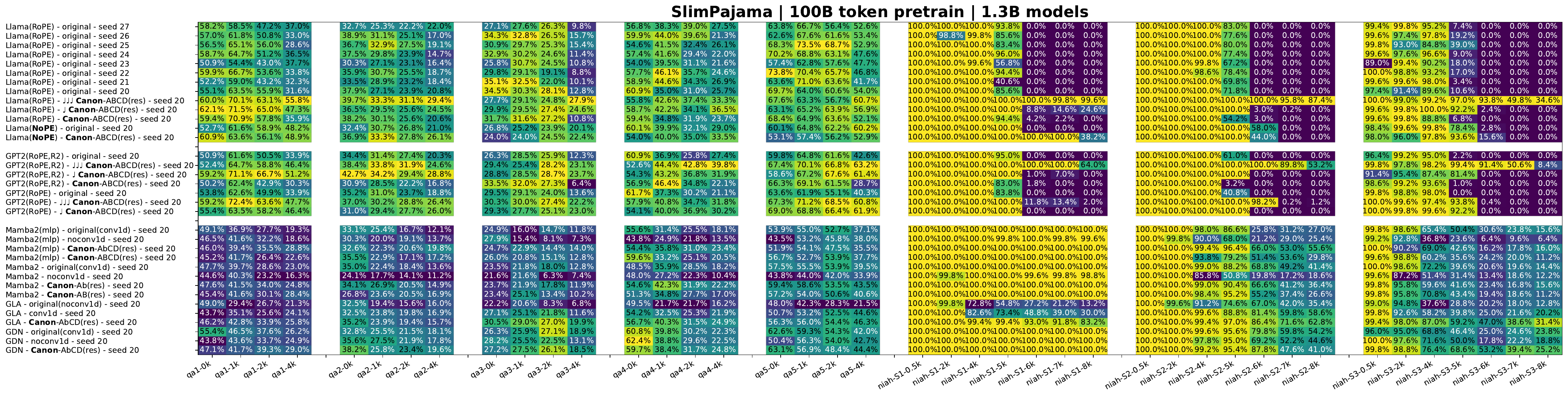}
\hspace*{-5mm}
\\
\hspace*{-5mm}
\includegraphics[page=1,trim={2.5mm 1.5mm 2.5mm 1.5mm},clip,width=\imgwidthBase]{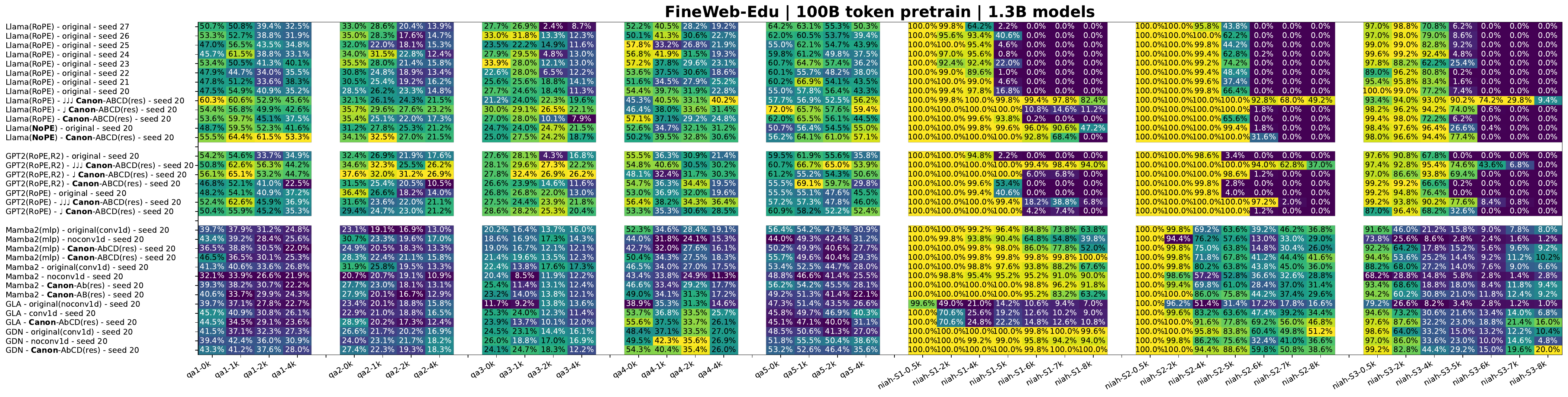}
\hspace*{-5mm}
\caption{\label{fig:reallife-babilong}%
Results on the \textbf{Babilong + S-NIAH} dataset evaluating multi-hop reasoning across varied junk context lengths. Most architectural comparisons are statistically insignificant. \bblue{Key findings include:}
\newline
\textbf{1.} Linear models consistently underperform Transformers, even in short contexts without junk.
\newline
\textbf{2.} Models with reduced RoPE (NoPE, RoPE\musQuarter) achieve  notable  improvements in long-context accuracy.
\newline
\textbf{3.} S-NIAH is \emph{too easy}: linear models appear accurate but fail at short-context 1-hop retrieval (\figureref{fig:real-life:include-gpt}).
}
\end{figure}

\section{More Synthetic Experiments}
\label{app:missing-experiments}

We present missing figures that were intentionally omitted from  the main body  of the paper for the sake of clarity and conciseness.

\begin{figure}[H]
\centering
\setlength{\imgwidthBase}{0.19\textwidth}

\includegraphics[page=1,trim={2.5mm 1.5mm 2.5mm 1.5mm},clip,width=\imgwidthBase]{perm/Llama_RoPE_-original}
\includegraphics[page=1,trim={2.5mm 1.5mm 2.5mm 1.5mm},clip,width=\imgwidthBase]{perm/Llama_RoPE_-Res-______Canon-ABCD}
\includegraphics[page=1,trim={2.5mm 1.5mm 2.5mm 1.5mm},clip,width=\imgwidthBase]{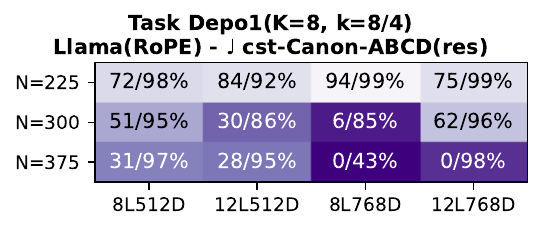}
\includegraphics[page=1,trim={2.5mm 1.5mm 2.5mm 1.5mm},clip,width=\imgwidthBase]{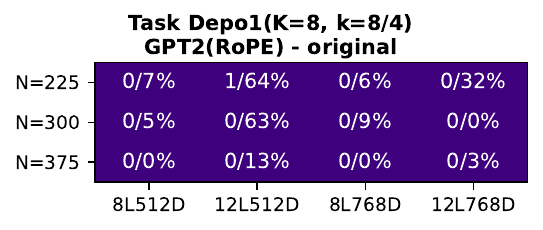}
\includegraphics[page=1,trim={2.5mm 1.5mm 2.5mm 1.5mm},clip,width=\imgwidthBase]{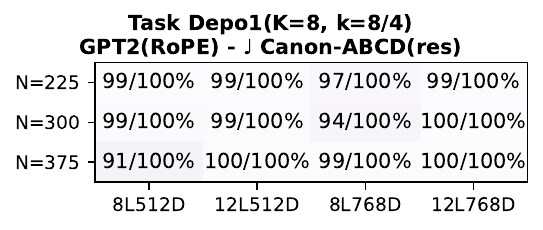}
\\
\includegraphics[page=1,trim={2.5mm 1.5mm 2.5mm 1.5mm},clip,width=\imgwidthBase]{perm_multi/Llama_RoPE_-original}
\includegraphics[page=1,trim={2.5mm 1.5mm 2.5mm 1.5mm},clip,width=\imgwidthBase]{perm_multi/Llama_RoPE_-Res-______Canon-ABCD}
\includegraphics[page=1,trim={2.5mm 1.5mm 2.5mm 1.5mm},clip,width=\imgwidthBase]{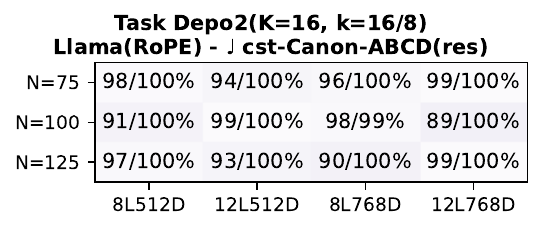}
\includegraphics[page=1,trim={2.5mm 1.5mm 2.5mm 1.5mm},clip,width=\imgwidthBase]{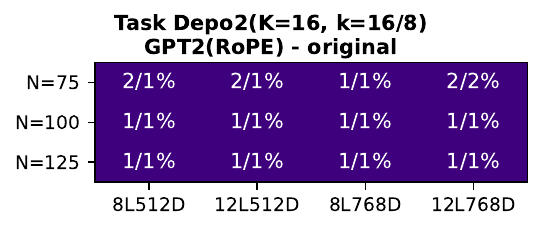}
\includegraphics[page=1,trim={2.5mm 1.5mm 2.5mm 1.5mm},clip,width=\imgwidthBase]{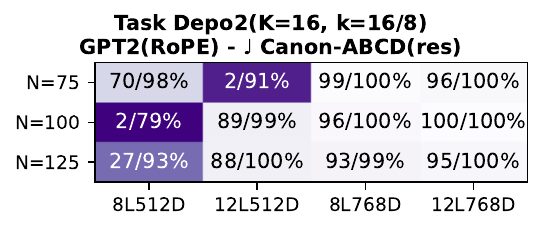}
\\
\includegraphics[page=1,trim={2.5mm 1.5mm 2.5mm 1.5mm},clip,width=\imgwidthBase]{top_sort/Llama_RoPE_-original}
\includegraphics[page=1,trim={2.5mm 1.5mm 2.5mm 1.5mm},clip,width=\imgwidthBase]{top_sort/Llama_RoPE_-Res-______Canon-ABCD}
\includegraphics[page=1,trim={2.5mm 1.5mm 2.5mm 1.5mm},clip,width=\imgwidthBase]{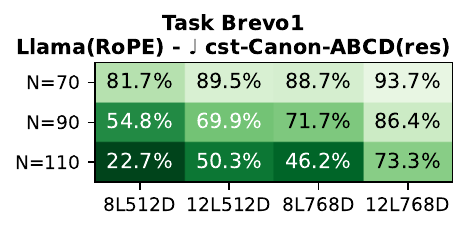}
\includegraphics[page=1,trim={2.5mm 1.5mm 2.5mm 1.5mm},clip,width=\imgwidthBase]{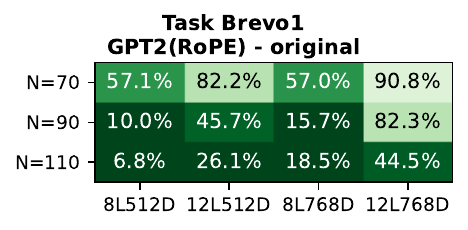}
\includegraphics[page=1,trim={2.5mm 1.5mm 2.5mm 1.5mm},clip,width=\imgwidthBase]{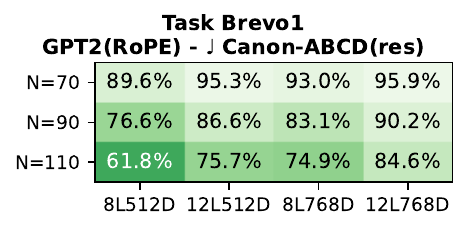}
\\
\includegraphics[page=1,trim={2.5mm 1.5mm 2.5mm 1.5mm},clip,width=\imgwidthBase]{top_sort_multi/Llama_RoPE_-original}
\includegraphics[page=1,trim={2.5mm 1.5mm 2.5mm 1.5mm},clip,width=\imgwidthBase]{top_sort_multi/Llama_RoPE_-Res-______Canon-ABCD}
\includegraphics[page=1,trim={2.5mm 1.5mm 2.5mm 1.5mm},clip,width=\imgwidthBase]{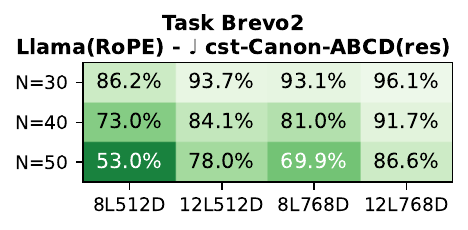}
\includegraphics[page=1,trim={2.5mm 1.5mm 2.5mm 1.5mm},clip,width=\imgwidthBase]{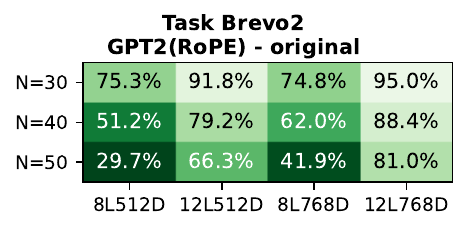}
\includegraphics[page=1,trim={2.5mm 1.5mm 2.5mm 1.5mm},clip,width=\imgwidthBase]{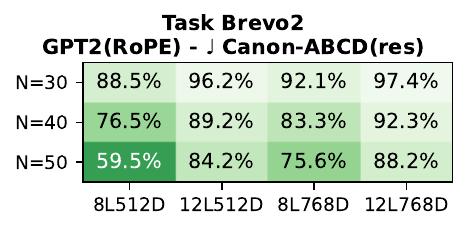}
\\
\includegraphics[page=1,trim={2.5mm 1.5mm 2.5mm 1.5mm},clip,width=\imgwidthBase]{arith/Llama_RoPE_-original}
\includegraphics[page=1,trim={2.5mm 1.5mm 2.5mm 1.5mm},clip,width=\imgwidthBase]{arith/Llama_RoPE_-Res-______Canon-ABCD}
\includegraphics[page=1,trim={2.5mm 1.5mm 2.5mm 1.5mm},clip,width=\imgwidthBase]{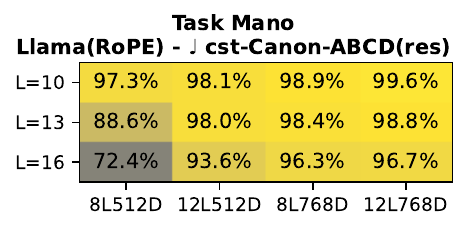}
\includegraphics[page=1,trim={2.5mm 1.5mm 2.5mm 1.5mm},clip,width=\imgwidthBase]{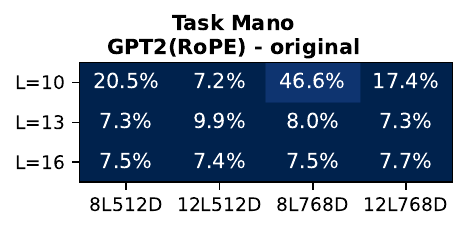}
\includegraphics[page=1,trim={2.5mm 1.5mm 2.5mm 1.5mm},clip,width=\imgwidthBase]{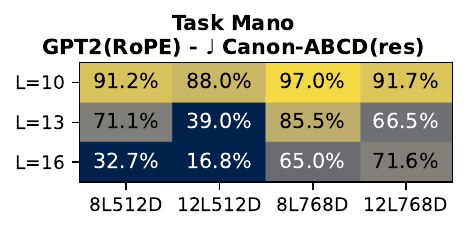}
\\
\includegraphics[page=1,trim={2.5mm 1.5mm 2.5mm 1.5mm},clip,width=\imgwidthBase]{cfg/Llama_RoPE_-original}
\includegraphics[page=1,trim={2.5mm 1.5mm 2.5mm 1.5mm},clip,width=\imgwidthBase]{cfg/Llama_RoPE_-Res-______Canon-ABCD}
\includegraphics[page=1,trim={2.5mm 1.5mm 2.5mm 1.5mm},clip,width=\imgwidthBase]{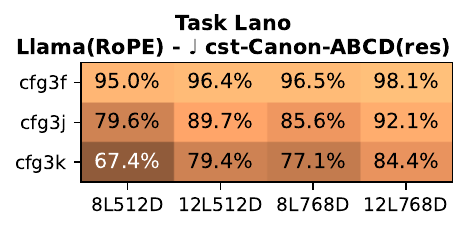}
\includegraphics[page=1,trim={2.5mm 1.5mm 2.5mm 1.5mm},clip,width=\imgwidthBase]{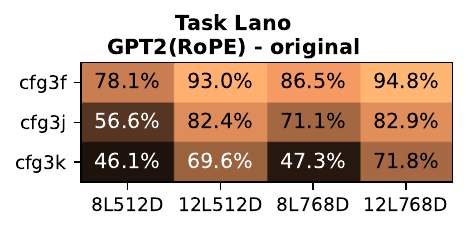}
\includegraphics[page=1,trim={2.5mm 1.5mm 2.5mm 1.5mm},clip,width=\imgwidthBase]{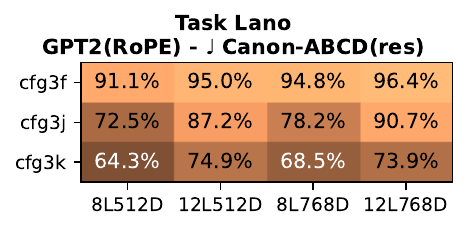}
\\
\includegraphics[page=1,trim={2.5mm 1.5mm 2.5mm 1.5mm},clip,width=\imgwidthBase]{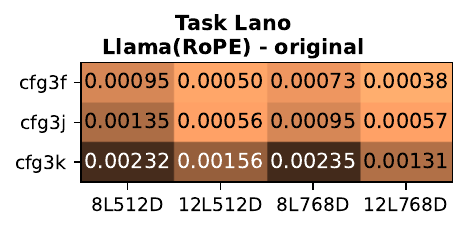}
\includegraphics[page=1,trim={2.5mm 1.5mm 2.5mm 1.5mm},clip,width=\imgwidthBase]{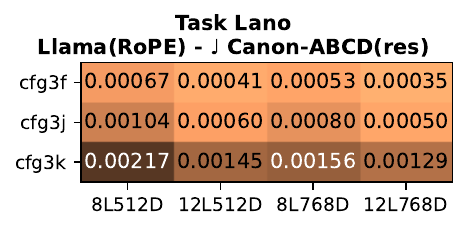}
\includegraphics[page=1,trim={2.5mm 1.5mm 2.5mm 1.5mm},clip,width=\imgwidthBase]{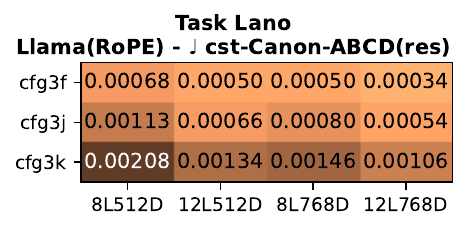}
\includegraphics[page=1,trim={2.5mm 1.5mm 2.5mm 1.5mm},clip,width=\imgwidthBase]{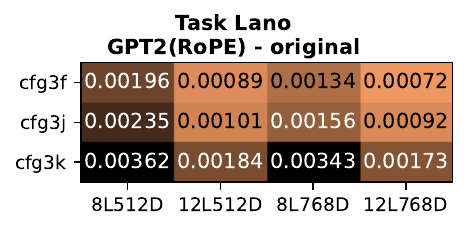}
\includegraphics[page=1,trim={2.5mm 1.5mm 2.5mm 1.5mm},clip,width=\imgwidthBase]{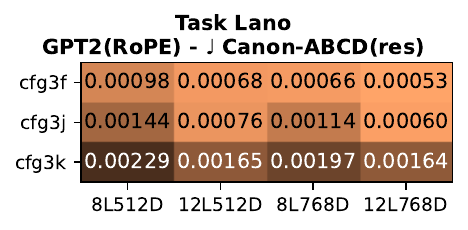}
\caption{\label{fig:gpt}%
\textbf{Columns 1,2,3:} \bblue{Constant Canon} implementation (random, \emph{non-trained} average of the past 3 tokens, denoted \textbf{cst-Canon}) already achieves strong performance, clearly outperforming vanilla Llama.
\newline
\textbf{Columns 2,4,5:} Canon layers also perform strongly on \bblue{GPT2 models (with standard MLP)}. Our playground reveals standard MLP is slightly weaker than gated MLP, especially in knowledge manipulation (cf.\ \resultref{res:5}).
}
\end{figure}

\begin{figure}[H]
\setlength{\imgwidthBase}{0.124\textwidth}
\centering
\hspace*{-7mm}
\includegraphics[page=1,trim={2.5mm 1.5mm 2.5mm 1.5mm},clip,width=\imgwidthBase]{perm/Llama_RoPE_-original}
\includegraphics[page=1,trim={2.5mm 1.5mm 2.5mm 1.5mm},clip,width=\imgwidthBase]{perm/GPT2_RoPE_-original}
\includegraphics[page=1,trim={2.5mm 1.5mm 2.5mm 1.5mm},clip,width=\imgwidthBase]{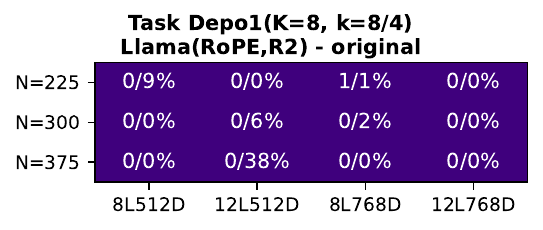}
\includegraphics[page=1,trim={2.5mm 1.5mm 2.5mm 1.5mm},clip,width=\imgwidthBase]{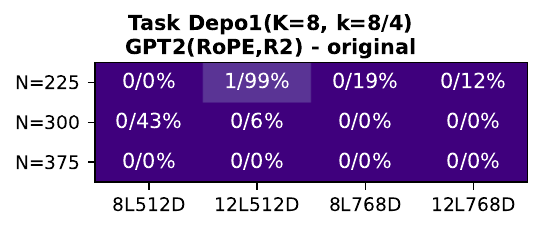}
\includegraphics[page=1,trim={2.5mm 1.5mm 2.5mm 1.5mm},clip,width=\imgwidthBase]{perm/Llama_RoPE_-Res-______Canon-ABCD}
\includegraphics[page=1,trim={2.5mm 1.5mm 2.5mm 1.5mm},clip,width=\imgwidthBase]{perm/GPT2_RoPE_-Res-______Canon-ABCD}
\includegraphics[page=1,trim={2.5mm 1.5mm 2.5mm 1.5mm},clip,width=\imgwidthBase]{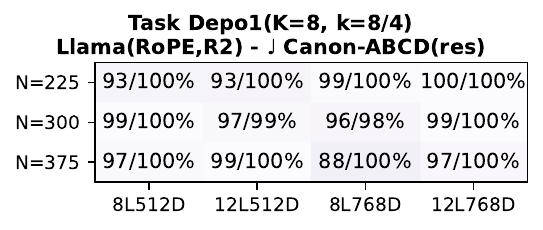}
\includegraphics[page=1,trim={2.5mm 1.5mm 2.5mm 1.5mm},clip,width=\imgwidthBase]{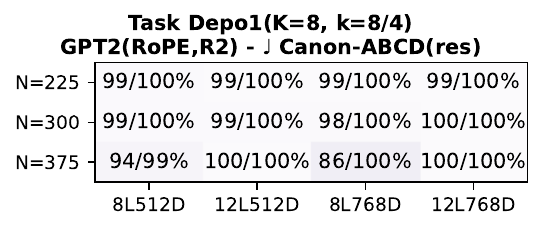}
\hspace*{-7mm}
\\
\hspace*{-7mm}
\includegraphics[page=1,trim={2.5mm 1.5mm 2.5mm 1.5mm},clip,width=\imgwidthBase]{perm_multi/Llama_RoPE_-original}
\includegraphics[page=1,trim={2.5mm 1.5mm 2.5mm 1.5mm},clip,width=\imgwidthBase]{perm_multi/GPT2_RoPE_-original}
\includegraphics[page=1,trim={2.5mm 1.5mm 2.5mm 1.5mm},clip,width=\imgwidthBase]{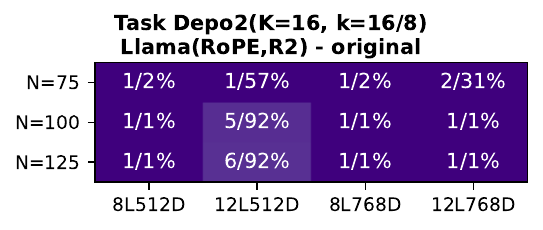}
\includegraphics[page=1,trim={2.5mm 1.5mm 2.5mm 1.5mm},clip,width=\imgwidthBase]{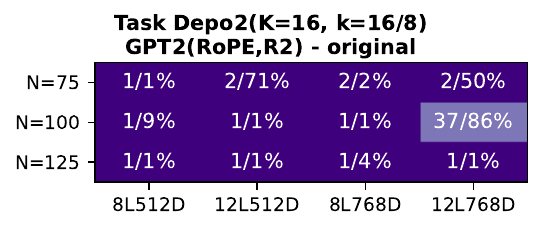}
\includegraphics[page=1,trim={2.5mm 1.5mm 2.5mm 1.5mm},clip,width=\imgwidthBase]{perm_multi/Llama_RoPE_-Res-______Canon-ABCD}
\includegraphics[page=1,trim={2.5mm 1.5mm 2.5mm 1.5mm},clip,width=\imgwidthBase]{perm_multi/GPT2_RoPE_-Res-______Canon-ABCD}
\includegraphics[page=1,trim={2.5mm 1.5mm 2.5mm 1.5mm},clip,width=\imgwidthBase]{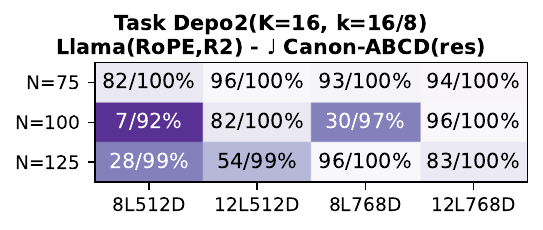}
\includegraphics[page=1,trim={2.5mm 1.5mm 2.5mm 1.5mm},clip,width=\imgwidthBase]{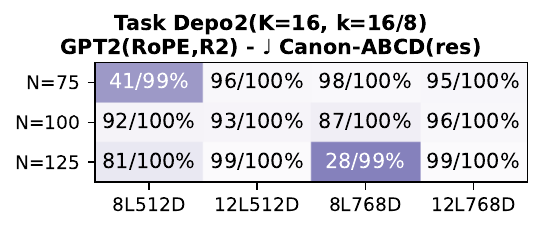}
\hspace*{-7mm}
\\
\hspace*{-7mm}
\includegraphics[page=1,trim={2.5mm 1.5mm 2.5mm 1.5mm},clip,width=\imgwidthBase]{top_sort/Llama_RoPE_-original}
\includegraphics[page=1,trim={2.5mm 1.5mm 2.5mm 1.5mm},clip,width=\imgwidthBase]{top_sort/GPT2_RoPE_-original}
\includegraphics[page=1,trim={2.5mm 1.5mm 2.5mm 1.5mm},clip,width=\imgwidthBase]{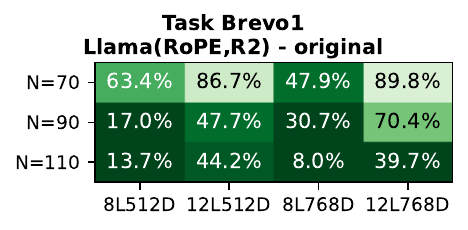}
\includegraphics[page=1,trim={2.5mm 1.5mm 2.5mm 1.5mm},clip,width=\imgwidthBase]{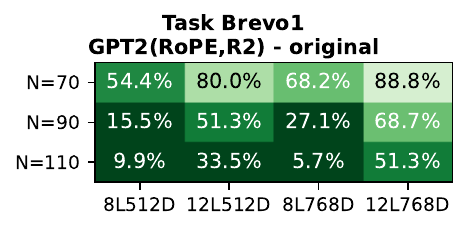}
\includegraphics[page=1,trim={2.5mm 1.5mm 2.5mm 1.5mm},clip,width=\imgwidthBase]{top_sort/Llama_RoPE_-Res-______Canon-ABCD}
\includegraphics[page=1,trim={2.5mm 1.5mm 2.5mm 1.5mm},clip,width=\imgwidthBase]{top_sort/GPT2_RoPE_-Res-______Canon-ABCD}
\includegraphics[page=1,trim={2.5mm 1.5mm 2.5mm 1.5mm},clip,width=\imgwidthBase]{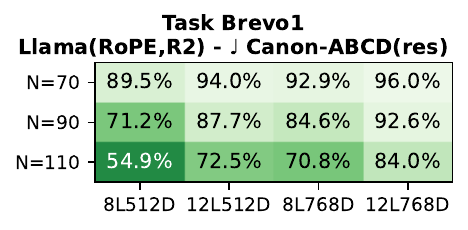}
\includegraphics[page=1,trim={2.5mm 1.5mm 2.5mm 1.5mm},clip,width=\imgwidthBase]{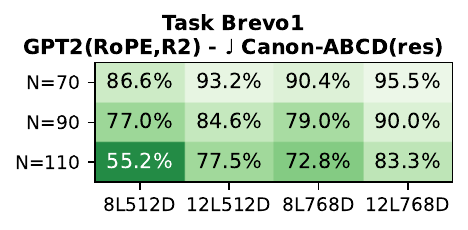}
\hspace*{-7mm}
\\
\hspace*{-7mm}
\includegraphics[page=1,trim={2.5mm 1.5mm 2.5mm 1.5mm},clip,width=\imgwidthBase]{top_sort_multi/Llama_RoPE_-original}
\includegraphics[page=1,trim={2.5mm 1.5mm 2.5mm 1.5mm},clip,width=\imgwidthBase]{top_sort_multi/GPT2_RoPE_-original}
\includegraphics[page=1,trim={2.5mm 1.5mm 2.5mm 1.5mm},clip,width=\imgwidthBase]{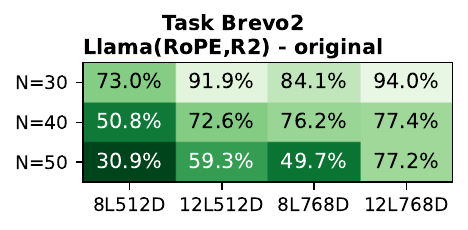}
\includegraphics[page=1,trim={2.5mm 1.5mm 2.5mm 1.5mm},clip,width=\imgwidthBase]{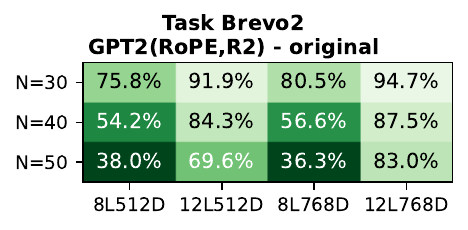}
\includegraphics[page=1,trim={2.5mm 1.5mm 2.5mm 1.5mm},clip,width=\imgwidthBase]{top_sort_multi/Llama_RoPE_-Res-______Canon-ABCD}
\includegraphics[page=1,trim={2.5mm 1.5mm 2.5mm 1.5mm},clip,width=\imgwidthBase]{top_sort_multi/GPT2_RoPE_-Res-______Canon-ABCD}
\includegraphics[page=1,trim={2.5mm 1.5mm 2.5mm 1.5mm},clip,width=\imgwidthBase]{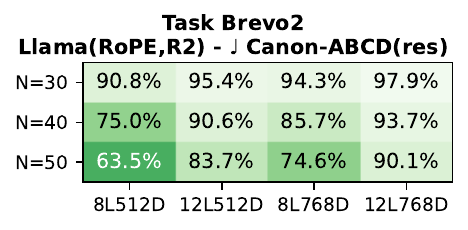}
\includegraphics[page=1,trim={2.5mm 1.5mm 2.5mm 1.5mm},clip,width=\imgwidthBase]{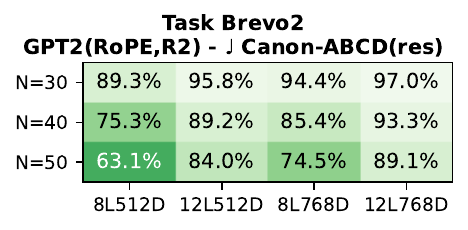}
\hspace*{-7mm}
\\
\hspace*{-7mm}
\includegraphics[page=1,trim={2.5mm 1.5mm 2.5mm 1.5mm},clip,width=\imgwidthBase]{arith/Llama_RoPE_-original}
\includegraphics[page=1,trim={2.5mm 1.5mm 2.5mm 1.5mm},clip,width=\imgwidthBase]{arith/GPT2_RoPE_-original}
\includegraphics[page=1,trim={2.5mm 1.5mm 2.5mm 1.5mm},clip,width=\imgwidthBase]{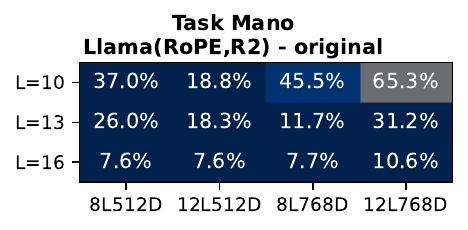}
\includegraphics[page=1,trim={2.5mm 1.5mm 2.5mm 1.5mm},clip,width=\imgwidthBase]{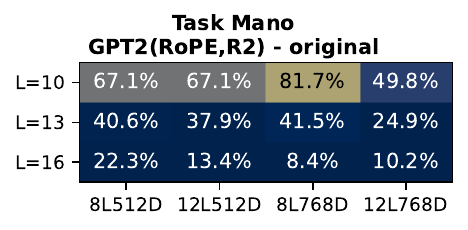}
\includegraphics[page=1,trim={2.5mm 1.5mm 2.5mm 1.5mm},clip,width=\imgwidthBase]{arith/Llama_RoPE_-Res-______Canon-ABCD}
\includegraphics[page=1,trim={2.5mm 1.5mm 2.5mm 1.5mm},clip,width=\imgwidthBase]{arith/GPT2_RoPE_-Res-______Canon-ABCD}
\includegraphics[page=1,trim={2.5mm 1.5mm 2.5mm 1.5mm},clip,width=\imgwidthBase]{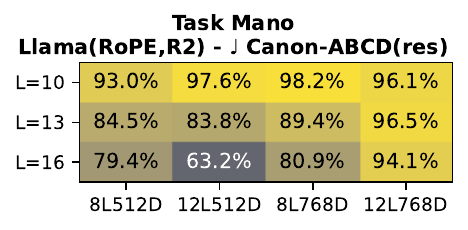}
\includegraphics[page=1,trim={2.5mm 1.5mm 2.5mm 1.5mm},clip,width=\imgwidthBase]{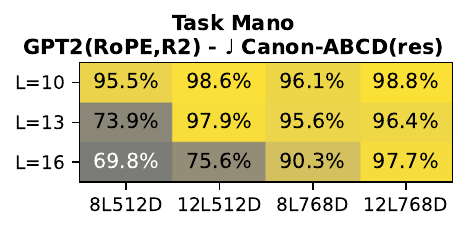}
\hspace*{-7mm}
\\
\hspace*{-7mm}
\includegraphics[page=1,trim={2.5mm 1.5mm 2.5mm 1.5mm},clip,width=\imgwidthBase]{cfg/Llama_RoPE_-original}
\includegraphics[page=1,trim={2.5mm 1.5mm 2.5mm 1.5mm},clip,width=\imgwidthBase]{cfg/GPT2_RoPE_-original}
\includegraphics[page=1,trim={2.5mm 1.5mm 2.5mm 1.5mm},clip,width=\imgwidthBase]{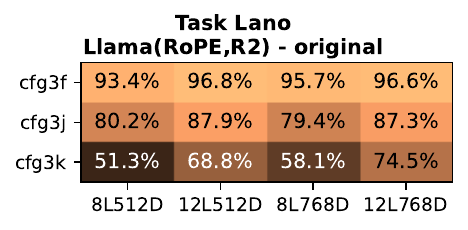}
\includegraphics[page=1,trim={2.5mm 1.5mm 2.5mm 1.5mm},clip,width=\imgwidthBase]{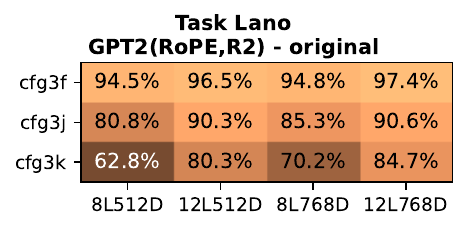}
\includegraphics[page=1,trim={2.5mm 1.5mm 2.5mm 1.5mm},clip,width=\imgwidthBase]{cfg/Llama_RoPE_-Res-______Canon-ABCD}
\includegraphics[page=1,trim={2.5mm 1.5mm 2.5mm 1.5mm},clip,width=\imgwidthBase]{cfg/GPT2_RoPE_-Res-______Canon-ABCD}
\includegraphics[page=1,trim={2.5mm 1.5mm 2.5mm 1.5mm},clip,width=\imgwidthBase]{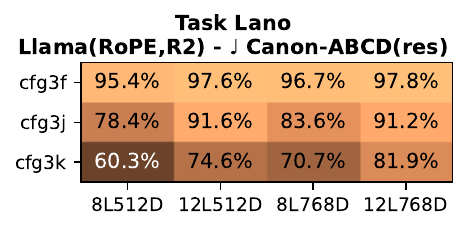}
\includegraphics[page=1,trim={2.5mm 1.5mm 2.5mm 1.5mm},clip,width=\imgwidthBase]{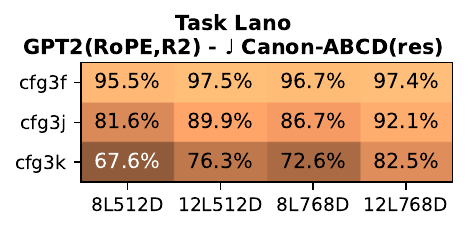}
\hspace*{-7mm}
\\
\hspace*{-7mm}
\includegraphics[page=1,trim={2.5mm 1.5mm 2.5mm 1.5mm},clip,width=\imgwidthBase]{cfg-prob/Llama_RoPE_-original}
\includegraphics[page=1,trim={2.5mm 1.5mm 2.5mm 1.5mm},clip,width=\imgwidthBase]{cfg-prob/GPT2_RoPE_-original}
\includegraphics[page=1,trim={2.5mm 1.5mm 2.5mm 1.5mm},clip,width=\imgwidthBase]{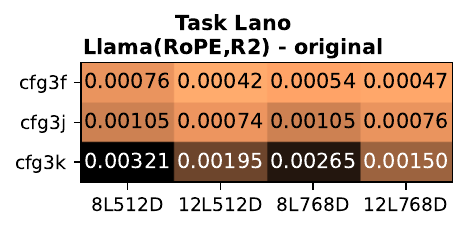}
\includegraphics[page=1,trim={2.5mm 1.5mm 2.5mm 1.5mm},clip,width=\imgwidthBase]{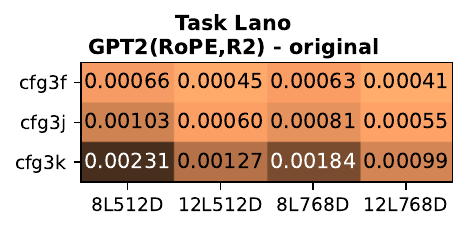}
\includegraphics[page=1,trim={2.5mm 1.5mm 2.5mm 1.5mm},clip,width=\imgwidthBase]{cfg-prob/Llama_RoPE_-Res-______Canon-ABCD}
\includegraphics[page=1,trim={2.5mm 1.5mm 2.5mm 1.5mm},clip,width=\imgwidthBase]{cfg-prob/GPT2_RoPE_-Res-______Canon-ABCD}
\includegraphics[page=1,trim={2.5mm 1.5mm 2.5mm 1.5mm},clip,width=\imgwidthBase]{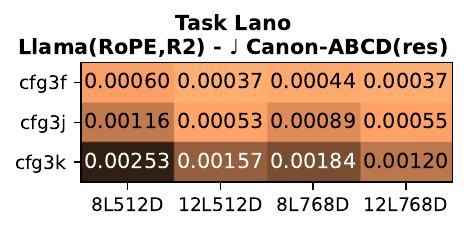}
\includegraphics[page=1,trim={2.5mm 1.5mm 2.5mm 1.5mm},clip,width=\imgwidthBase]{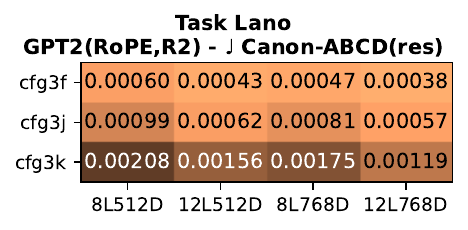}
\hspace*{-7mm}
\\
\setlength{\imgwidthBase}{0.131\textwidth}
\centering
\hspace{-9mm}
\foreach \txt in {%
  gated MLP \\ (silu),%
  standard MLP \\ (silu),%
  gated MLP \\ (relu$^2$),%
  standard MLP \\ (relu$^2$),%
  gated MLP \\ (silu) + Canon,%
  standard MLP \\ (silu) + Canon,%
  gated MLP \\ (relu$^2$) + Canon,%
  standard MLP \\ (relu$^2$) + Canon%
}{%
\begin{minipage}[t]{\imgwidthBase}
  \centering
  \noindent
    {    \parbox{0.99\imgwidthBase}{\centering\tiny \txt}
  }
\end{minipage}%
}
\hspace*{-12mm}
\caption{\label{fig:app:relu2}%
\bblue{Effect of $\text{ReLU}^2$ activation on standard vs. gated MLP.}
\textbf{Columns 1→2, 5→6}: gated MLP outperforms standard MLP with \texttt{silu}.
\textbf{Columns 2→4, 6→8}: adding $\text{ReLU}^2$ to standard MLP yields slight gains.
\textbf{Columns 1→3, 5→7}: adding $\text{ReLU}^2$ to gated MLP hurts performance.
}
\end{figure}

\begin{figure}[H]
\centering
\setlength{\imgwidthBase}{0.19\textwidth}

\includegraphics[page=1,trim={2.5mm 1.5mm 2.5mm 1.5mm},clip,width=\imgwidthBase]{perm/Llama_RoPE_-Res-Canon-ABCD}
\includegraphics[page=1,trim={2.5mm 1.5mm 2.5mm 1.5mm},clip,width=\imgwidthBase]{perm/Llama_RoPE_-Res-______Canon-ABCD}
\includegraphics[page=1,trim={2.5mm 1.5mm 2.5mm 1.5mm},clip,width=\imgwidthBase]{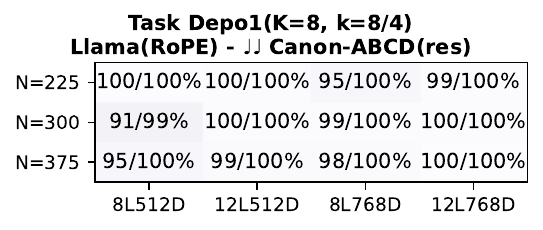}
\includegraphics[page=1,trim={2.5mm 1.5mm 2.5mm 1.5mm},clip,width=\imgwidthBase]{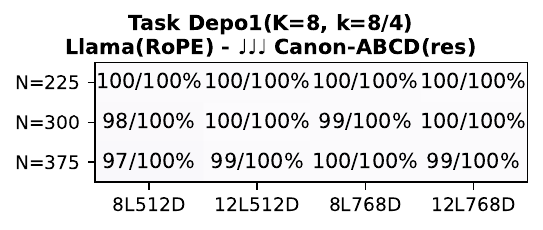}
\includegraphics[page=1,trim={2.5mm 1.5mm 2.5mm 1.5mm},clip,width=\imgwidthBase]{perm/Llama_NoPE_-Res-Canon-ABCD}
\\
\includegraphics[page=1,trim={2.5mm 1.5mm 2.5mm 1.5mm},clip,width=\imgwidthBase]{perm_multi/Llama_RoPE_-Res-Canon-ABCD}
\includegraphics[page=1,trim={2.5mm 1.5mm 2.5mm 1.5mm},clip,width=\imgwidthBase]{perm_multi/Llama_RoPE_-Res-______Canon-ABCD}
\includegraphics[page=1,trim={2.5mm 1.5mm 2.5mm 1.5mm},clip,width=\imgwidthBase]{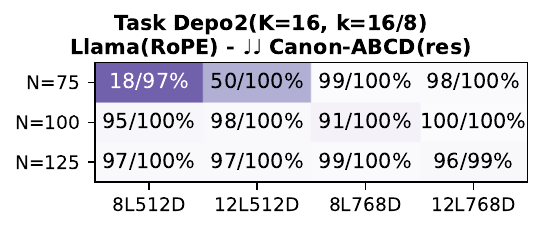}
\includegraphics[page=1,trim={2.5mm 1.5mm 2.5mm 1.5mm},clip,width=\imgwidthBase]{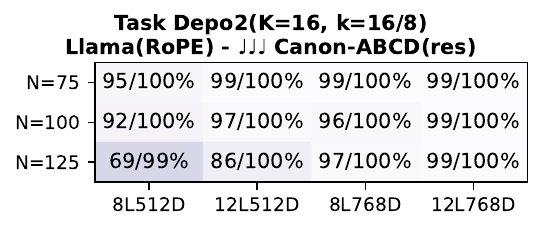}
\includegraphics[page=1,trim={2.5mm 1.5mm 2.5mm 1.5mm},clip,width=\imgwidthBase]{perm_multi/Llama_NoPE_-Res-Canon-ABCD}
\\
\includegraphics[page=1,trim={2.5mm 1.5mm 2.5mm 1.5mm},clip,width=\imgwidthBase]{top_sort/Llama_RoPE_-Res-Canon-ABCD}
\includegraphics[page=1,trim={2.5mm 1.5mm 2.5mm 1.5mm},clip,width=\imgwidthBase]{top_sort/Llama_RoPE_-Res-______Canon-ABCD}
\includegraphics[page=1,trim={2.5mm 1.5mm 2.5mm 1.5mm},clip,width=\imgwidthBase]{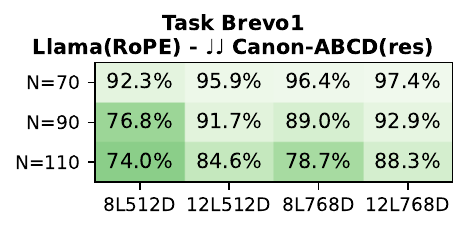}
\includegraphics[page=1,trim={2.5mm 1.5mm 2.5mm 1.5mm},clip,width=\imgwidthBase]{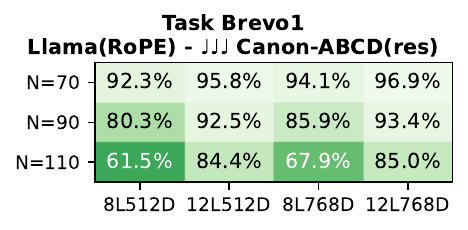}
\includegraphics[page=1,trim={2.5mm 1.5mm 2.5mm 1.5mm},clip,width=\imgwidthBase]{top_sort/Llama_NoPE_-Res-Canon-ABCD}
\\
\includegraphics[page=1,trim={2.5mm 1.5mm 2.5mm 1.5mm},clip,width=\imgwidthBase]{top_sort_multi/Llama_RoPE_-Res-Canon-ABCD}
\includegraphics[page=1,trim={2.5mm 1.5mm 2.5mm 1.5mm},clip,width=\imgwidthBase]{top_sort_multi/Llama_RoPE_-Res-______Canon-ABCD}
\includegraphics[page=1,trim={2.5mm 1.5mm 2.5mm 1.5mm},clip,width=\imgwidthBase]{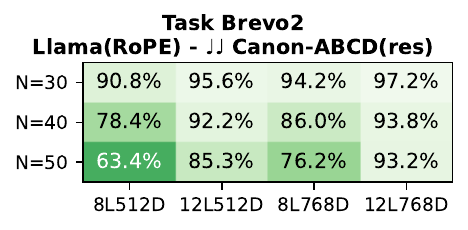}
\includegraphics[page=1,trim={2.5mm 1.5mm 2.5mm 1.5mm},clip,width=\imgwidthBase]{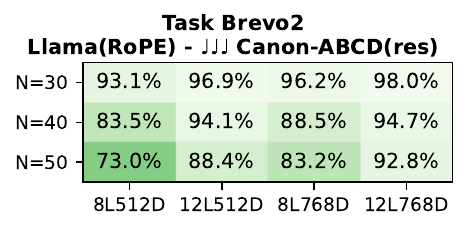}
\includegraphics[page=1,trim={2.5mm 1.5mm 2.5mm 1.5mm},clip,width=\imgwidthBase]{top_sort_multi/Llama_NoPE_-Res-Canon-ABCD}
\\
\includegraphics[page=1,trim={2.5mm 1.5mm 2.5mm 1.5mm},clip,width=\imgwidthBase]{arith/Llama_RoPE_-Res-Canon-ABCD}
\includegraphics[page=1,trim={2.5mm 1.5mm 2.5mm 1.5mm},clip,width=\imgwidthBase]{arith/Llama_RoPE_-Res-______Canon-ABCD}
\includegraphics[page=1,trim={2.5mm 1.5mm 2.5mm 1.5mm},clip,width=\imgwidthBase]{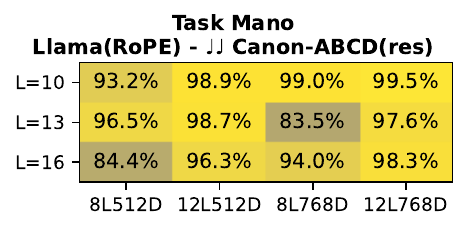}
\includegraphics[page=1,trim={2.5mm 1.5mm 2.5mm 1.5mm},clip,width=\imgwidthBase]{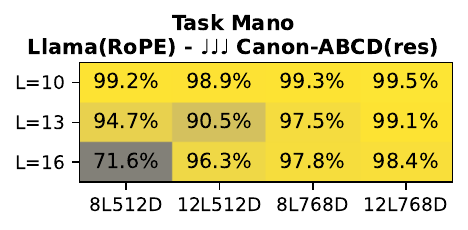}
\includegraphics[page=1,trim={2.5mm 1.5mm 2.5mm 1.5mm},clip,width=\imgwidthBase]{arith/Llama_NoPE_-Res-Canon-ABCD}
\\
\includegraphics[page=1,trim={2.5mm 1.5mm 2.5mm 1.5mm},clip,width=\imgwidthBase]{cfg/Llama_RoPE_-Res-Canon-ABCD}
\includegraphics[page=1,trim={2.5mm 1.5mm 2.5mm 1.5mm},clip,width=\imgwidthBase]{cfg/Llama_RoPE_-Res-______Canon-ABCD}
\includegraphics[page=1,trim={2.5mm 1.5mm 2.5mm 1.5mm},clip,width=\imgwidthBase]{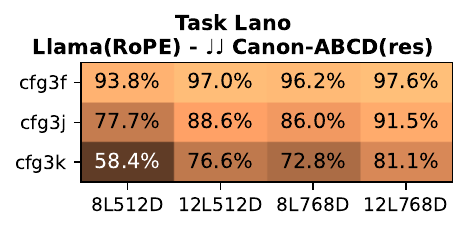}
\includegraphics[page=1,trim={2.5mm 1.5mm 2.5mm 1.5mm},clip,width=\imgwidthBase]{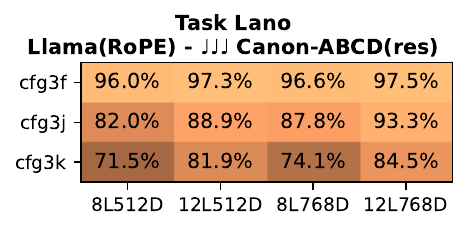}
\includegraphics[page=1,trim={2.5mm 1.5mm 2.5mm 1.5mm},clip,width=\imgwidthBase]{cfg/Llama_NoPE_-Res-Canon-ABCD}
\\
\includegraphics[page=1,trim={2.5mm 1.5mm 2.5mm 1.5mm},clip,width=\imgwidthBase]{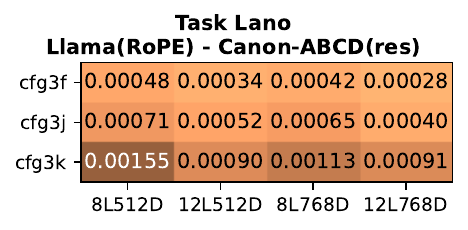}
\includegraphics[page=1,trim={2.5mm 1.5mm 2.5mm 1.5mm},clip,width=\imgwidthBase]{cfg-prob/Llama_RoPE_-Res-______Canon-ABCD}
\includegraphics[page=1,trim={2.5mm 1.5mm 2.5mm 1.5mm},clip,width=\imgwidthBase]{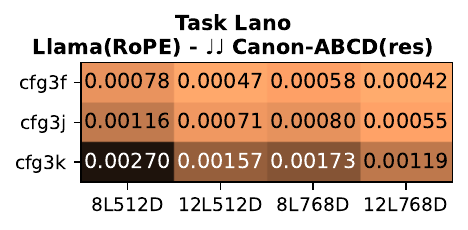}
\includegraphics[page=1,trim={2.5mm 1.5mm 2.5mm 1.5mm},clip,width=\imgwidthBase]{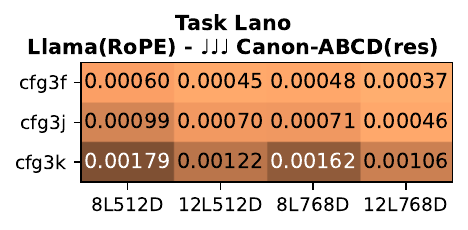}
\includegraphics[page=1,trim={2.5mm 1.5mm 2.5mm 1.5mm},clip,width=\imgwidthBase]{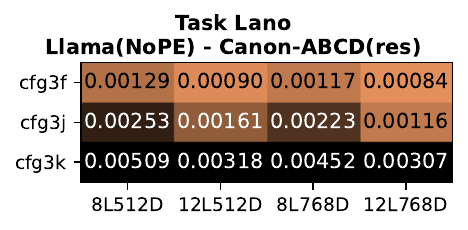}
\caption{\label{fig:quarters}%
Transformer+Canon with \textbf{\color{blue} varying RoPE configurations}. From left to right: (1) RoPE; (2) RoPE\musQuarter{}: half of the heads, each with half the RoPE dimensions; (3) RoPE\musQuarter\musQuarter{}: a quarter of heads with full RoPE dimensions; (4) RoPE\musQuarter\musQuarter\musQuarter{}: all heads each with quarter RoPE dimensions; (5) NoPE.
\newline
\sepline
\newline
\textbf{Conclusion:} Canon layers eliminate the need for extensive RoPE usage, and reducing RoPE usage to 1/4 is even preferable, outperforming both full RoPE and NoPE setups. Among these reduced RoPE variants, RoPE\musQuarter{} achieves slightly better overall performance.
}
\end{figure}

\section{Complete Ablation Studies}
\label{app:more-experiments}

This section presents full ablation results, including KL-divergence evaluations for Task \textsc{Mano}.
These details were omitted from the main text for clarity but are included here for completeness and for readers seeking deeper experimental insight.

\subsection{Llama(RoPE) family}

\begin{figure}[H]
\centering
\setlength{\imgwidthBase}{0.1245\textwidth}
\hspace*{-7mm}
\includegraphics[page=1,trim={2.5mm 1.5mm 2.5mm 1.5mm},clip,width=\imgwidthBase]{perm/Llama_RoPE_-original}
\includegraphics[page=1,trim={2.5mm 1.5mm 2.5mm 1.5mm},clip,width=\imgwidthBase]{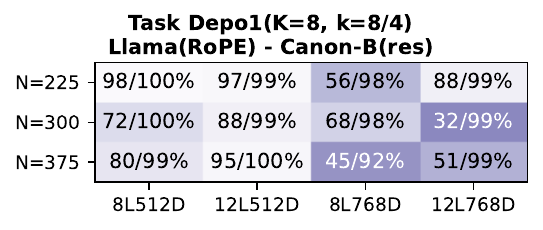}
\includegraphics[page=1,trim={2.5mm 1.5mm 2.5mm 1.5mm},clip,width=\imgwidthBase]{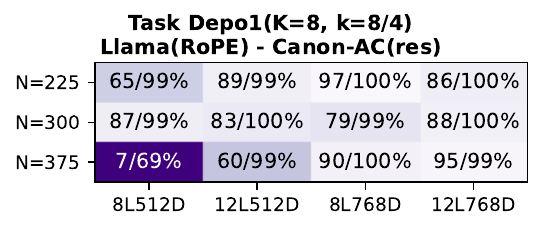}
\includegraphics[page=1,trim={2.5mm 1.5mm 2.5mm 1.5mm},clip,width=\imgwidthBase]{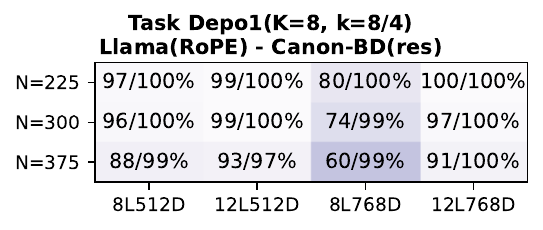}
\includegraphics[page=1,trim={2.5mm 1.5mm 2.5mm 1.5mm},clip,width=\imgwidthBase]{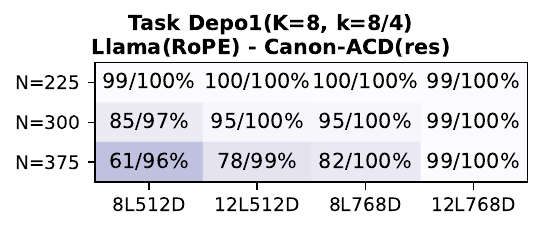}
\includegraphics[page=1,trim={2.5mm 1.5mm 2.5mm 1.5mm},clip,width=\imgwidthBase]{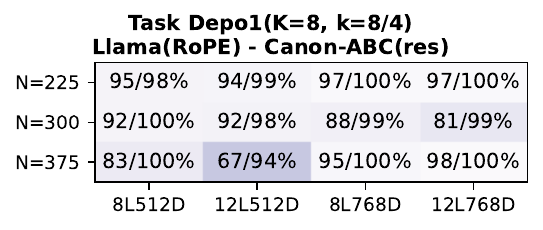}
\includegraphics[page=1,trim={2.5mm 1.5mm 2.5mm 1.5mm},clip,width=\imgwidthBase]{perm/Llama_RoPE_-Res-Canon-ABCD}
\includegraphics[page=1,trim={2.5mm 1.5mm 2.5mm 1.5mm},clip,width=\imgwidthBase]{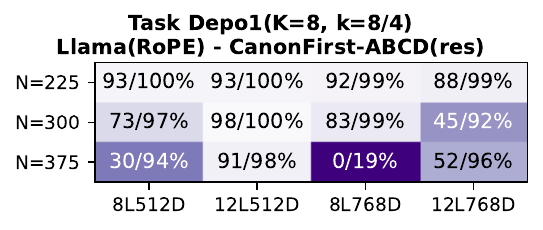}
\hspace*{-7mm}
\\
\hspace*{-7mm}
\includegraphics[page=1,trim={2.5mm 1.5mm 2.5mm 1.5mm},clip,width=\imgwidthBase]{perm_multi/Llama_RoPE_-original}
\includegraphics[page=1,trim={2.5mm 1.5mm 2.5mm 1.5mm},clip,width=\imgwidthBase]{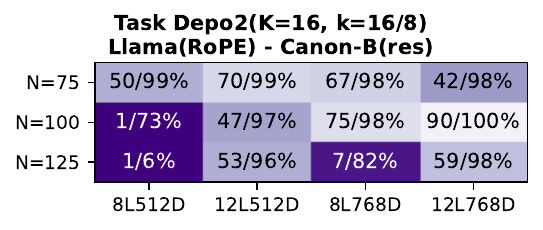}
\includegraphics[page=1,trim={2.5mm 1.5mm 2.5mm 1.5mm},clip,width=\imgwidthBase]{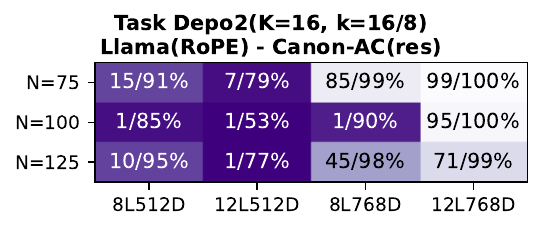}
\includegraphics[page=1,trim={2.5mm 1.5mm 2.5mm 1.5mm},clip,width=\imgwidthBase]{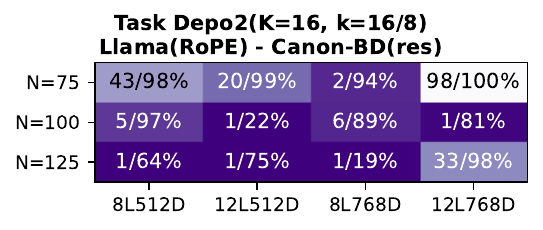}
\includegraphics[page=1,trim={2.5mm 1.5mm 2.5mm 1.5mm},clip,width=\imgwidthBase]{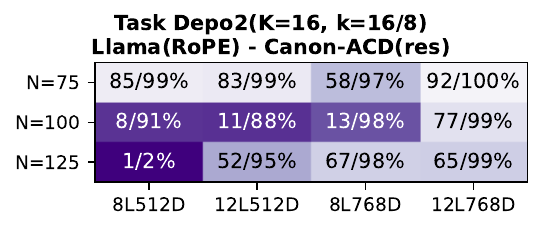}
\includegraphics[page=1,trim={2.5mm 1.5mm 2.5mm 1.5mm},clip,width=\imgwidthBase]{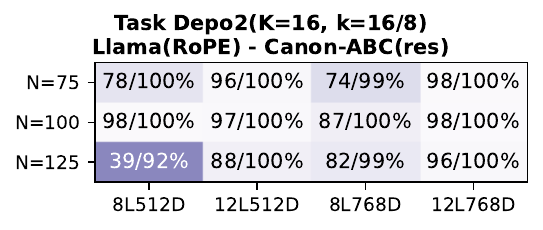}
\includegraphics[page=1,trim={2.5mm 1.5mm 2.5mm 1.5mm},clip,width=\imgwidthBase]{perm_multi/Llama_RoPE_-Res-Canon-ABCD}
\includegraphics[page=1,trim={2.5mm 1.5mm 2.5mm 1.5mm},clip,width=\imgwidthBase]{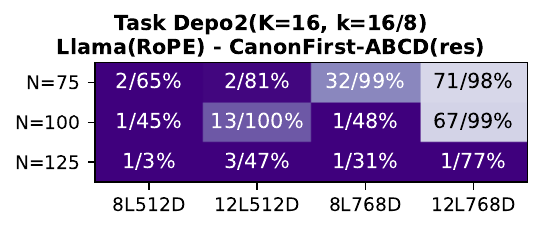}
\hspace*{-7mm}
\\
\hspace*{-7mm}
\includegraphics[page=1,trim={2.5mm 1.5mm 2.5mm 1.5mm},clip,width=\imgwidthBase]{top_sort/Llama_RoPE_-original}
\includegraphics[page=1,trim={2.5mm 1.5mm 2.5mm 1.5mm},clip,width=\imgwidthBase]{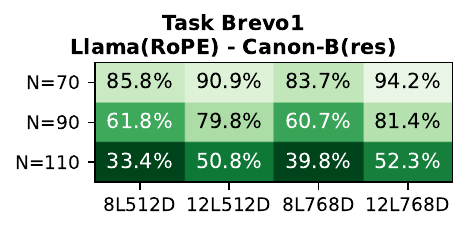}
\includegraphics[page=1,trim={2.5mm 1.5mm 2.5mm 1.5mm},clip,width=\imgwidthBase]{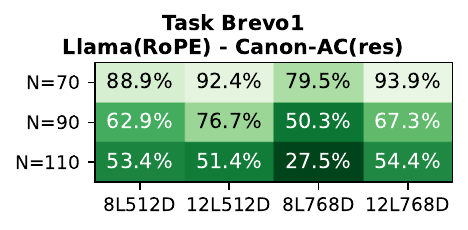}
\includegraphics[page=1,trim={2.5mm 1.5mm 2.5mm 1.5mm},clip,width=\imgwidthBase]{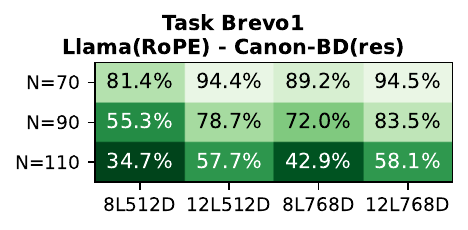}
\includegraphics[page=1,trim={2.5mm 1.5mm 2.5mm 1.5mm},clip,width=\imgwidthBase]{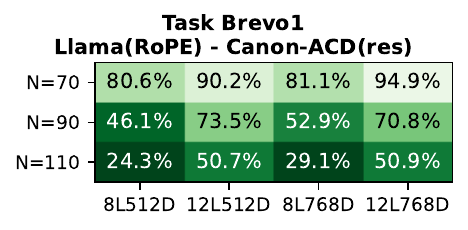}
\includegraphics[page=1,trim={2.5mm 1.5mm 2.5mm 1.5mm},clip,width=\imgwidthBase]{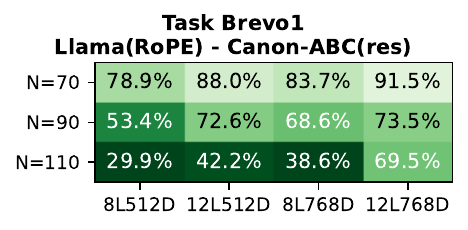}
\includegraphics[page=1,trim={2.5mm 1.5mm 2.5mm 1.5mm},clip,width=\imgwidthBase]{top_sort/Llama_RoPE_-Res-Canon-ABCD}
\includegraphics[page=1,trim={2.5mm 1.5mm 2.5mm 1.5mm},clip,width=\imgwidthBase]{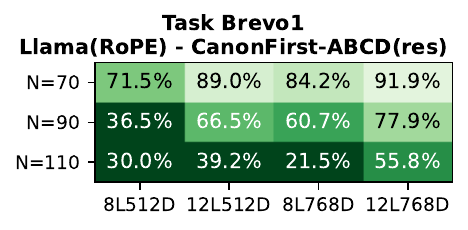}
\hspace*{-7mm}
\\
\hspace*{-7mm}
\includegraphics[page=1,trim={2.5mm 1.5mm 2.5mm 1.5mm},clip,width=\imgwidthBase]{top_sort_multi/Llama_RoPE_-original}
\includegraphics[page=1,trim={2.5mm 1.5mm 2.5mm 1.5mm},clip,width=\imgwidthBase]{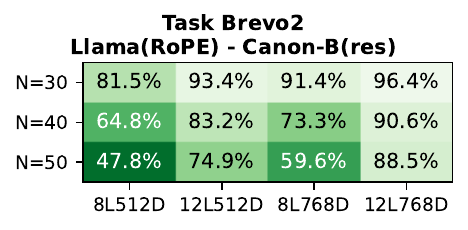}
\includegraphics[page=1,trim={2.5mm 1.5mm 2.5mm 1.5mm},clip,width=\imgwidthBase]{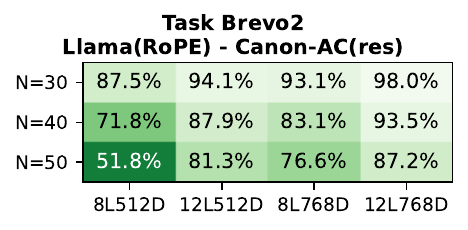}
\includegraphics[page=1,trim={2.5mm 1.5mm 2.5mm 1.5mm},clip,width=\imgwidthBase]{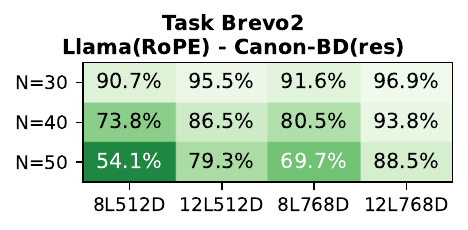}
\includegraphics[page=1,trim={2.5mm 1.5mm 2.5mm 1.5mm},clip,width=\imgwidthBase]{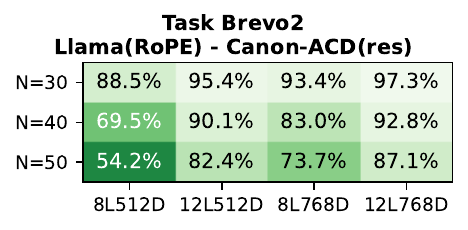}
\includegraphics[page=1,trim={2.5mm 1.5mm 2.5mm 1.5mm},clip,width=\imgwidthBase]{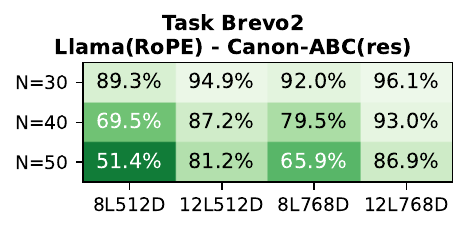}
\includegraphics[page=1,trim={2.5mm 1.5mm 2.5mm 1.5mm},clip,width=\imgwidthBase]{top_sort_multi/Llama_RoPE_-Res-Canon-ABCD}
\includegraphics[page=1,trim={2.5mm 1.5mm 2.5mm 1.5mm},clip,width=\imgwidthBase]{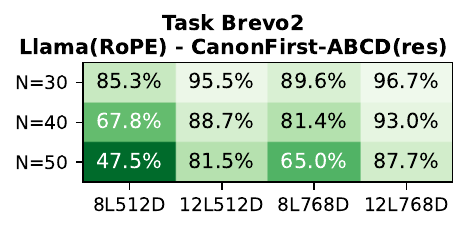}
\hspace*{-7mm}
\\
\hspace*{-7mm}
\includegraphics[page=1,trim={2.5mm 1.5mm 2.5mm 1.5mm},clip,width=\imgwidthBase]{arith/Llama_RoPE_-original}
\includegraphics[page=1,trim={2.5mm 1.5mm 2.5mm 1.5mm},clip,width=\imgwidthBase]{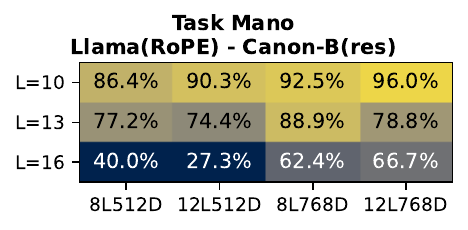}
\includegraphics[page=1,trim={2.5mm 1.5mm 2.5mm 1.5mm},clip,width=\imgwidthBase]{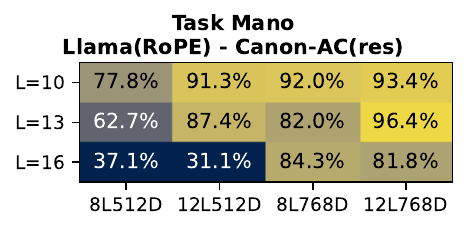}
\includegraphics[page=1,trim={2.5mm 1.5mm 2.5mm 1.5mm},clip,width=\imgwidthBase]{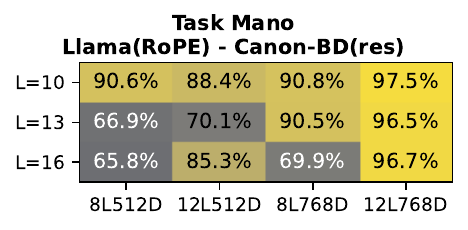}
\includegraphics[page=1,trim={2.5mm 1.5mm 2.5mm 1.5mm},clip,width=\imgwidthBase]{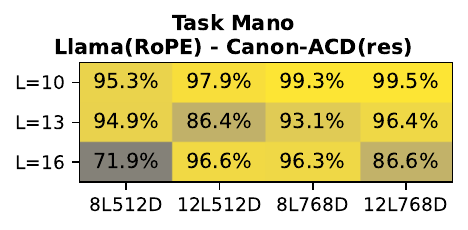}
\includegraphics[page=1,trim={2.5mm 1.5mm 2.5mm 1.5mm},clip,width=\imgwidthBase]{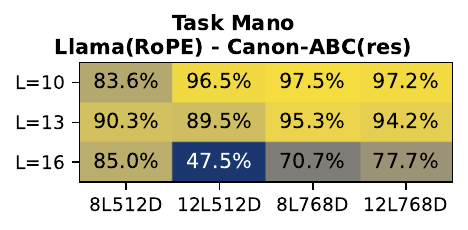}
\includegraphics[page=1,trim={2.5mm 1.5mm 2.5mm 1.5mm},clip,width=\imgwidthBase]{arith/Llama_RoPE_-Res-Canon-ABCD}
\includegraphics[page=1,trim={2.5mm 1.5mm 2.5mm 1.5mm},clip,width=\imgwidthBase]{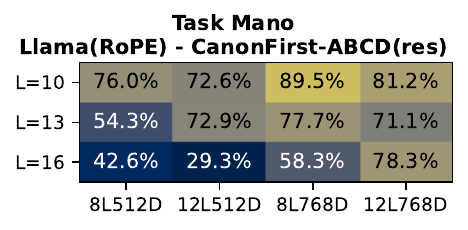}
\hspace*{-7mm}
\\
\hspace*{-7mm}
\includegraphics[page=1,trim={2.5mm 1.5mm 2.5mm 1.5mm},clip,width=\imgwidthBase]{cfg/Llama_RoPE_-original}
\includegraphics[page=1,trim={2.5mm 1.5mm 2.5mm 1.5mm},clip,width=\imgwidthBase]{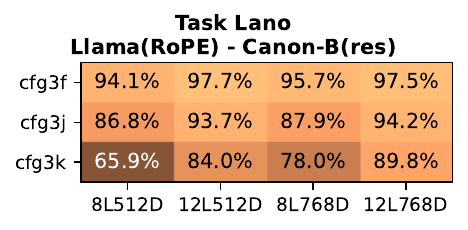}
\includegraphics[page=1,trim={2.5mm 1.5mm 2.5mm 1.5mm},clip,width=\imgwidthBase]{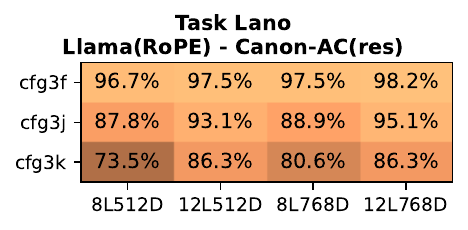}
\includegraphics[page=1,trim={2.5mm 1.5mm 2.5mm 1.5mm},clip,width=\imgwidthBase]{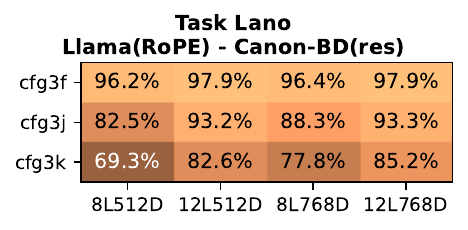}
\includegraphics[page=1,trim={2.5mm 1.5mm 2.5mm 1.5mm},clip,width=\imgwidthBase]{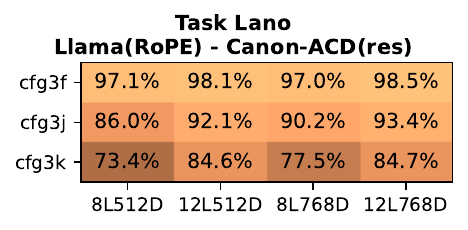}
\includegraphics[page=1,trim={2.5mm 1.5mm 2.5mm 1.5mm},clip,width=\imgwidthBase]{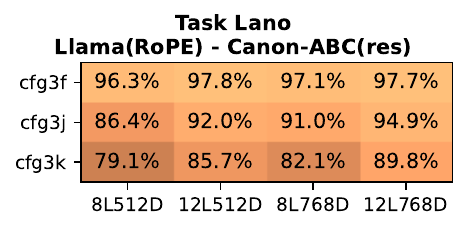}
\includegraphics[page=1,trim={2.5mm 1.5mm 2.5mm 1.5mm},clip,width=\imgwidthBase]{cfg/Llama_RoPE_-Res-Canon-ABCD}
\includegraphics[page=1,trim={2.5mm 1.5mm 2.5mm 1.5mm},clip,width=\imgwidthBase]{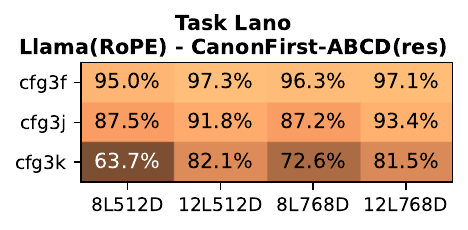}
\hspace*{-7mm}
\\
\hspace*{-7mm}
\includegraphics[page=1,trim={2.5mm 1.5mm 2.5mm 1.5mm},clip,width=\imgwidthBase]{cfg-prob/Llama_RoPE_-original}
\includegraphics[page=1,trim={2.5mm 1.5mm 2.5mm 1.5mm},clip,width=\imgwidthBase]{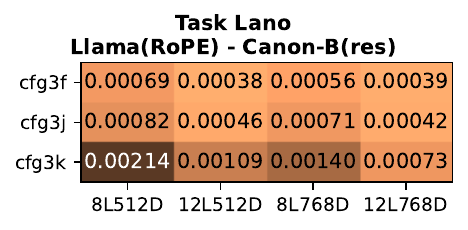}
\includegraphics[page=1,trim={2.5mm 1.5mm 2.5mm 1.5mm},clip,width=\imgwidthBase]{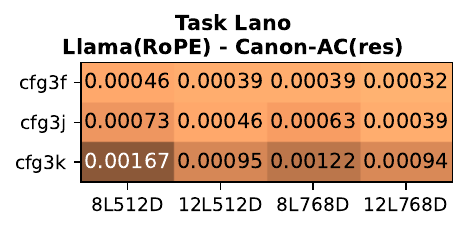}
\includegraphics[page=1,trim={2.5mm 1.5mm 2.5mm 1.5mm},clip,width=\imgwidthBase]{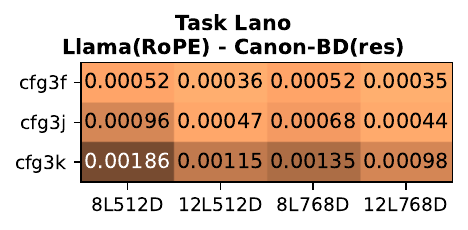}
\includegraphics[page=1,trim={2.5mm 1.5mm 2.5mm 1.5mm},clip,width=\imgwidthBase]{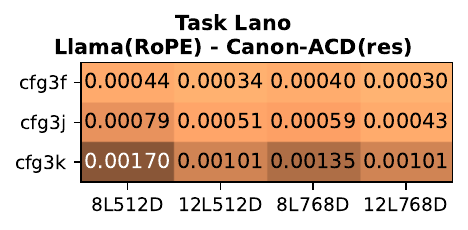}
\includegraphics[page=1,trim={2.5mm 1.5mm 2.5mm 1.5mm},clip,width=\imgwidthBase]{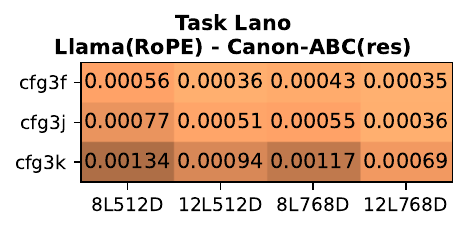}
\includegraphics[page=1,trim={2.5mm 1.5mm 2.5mm 1.5mm},clip,width=\imgwidthBase]{cfg-prob/Llama_RoPE_-Res-Canon-ABCD}
\includegraphics[page=1,trim={2.5mm 1.5mm 2.5mm 1.5mm},clip,width=\imgwidthBase]{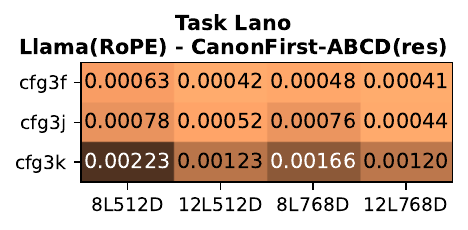}
\hspace*{-7mm}
\caption{\label{fig:app:rope-ablation}\textbf{Llama(\bblue{RoPE}) family}: (left to right) original; Canon-B, -AC, -BD, -ACD, -ABC, -ABCD; and CanonFirst-ABCD, where Canon is applied only to the first layer.
\newline
This figure complements \figureref{fig:trans-ablation} and gives more technical details.}
\end{figure}

\begin{figure}[H]
\centering
\setlength{\imgwidthBase}{0.19\textwidth}
\includegraphics[page=1,trim={2.5mm 1.5mm 2.5mm 1.5mm},clip,width=\imgwidthBase]{perm/Llama_RoPE_-original}
\includegraphics[page=1,trim={2.5mm 1.5mm 2.5mm 1.5mm},clip,width=\imgwidthBase]{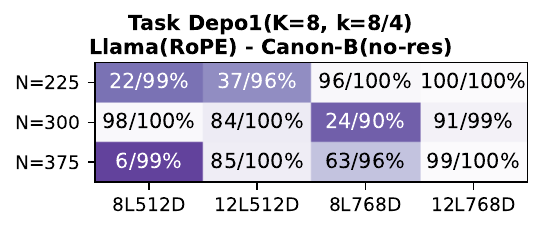}
\includegraphics[page=1,trim={2.5mm 1.5mm 2.5mm 1.5mm},clip,width=\imgwidthBase]{perm/Llama_RoPE_-Res-Canon-ABCD}
\includegraphics[page=1,trim={2.5mm 1.5mm 2.5mm 1.5mm},clip,width=\imgwidthBase]{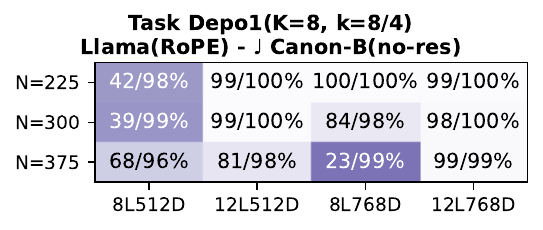}
\includegraphics[page=1,trim={2.5mm 1.5mm 2.5mm 1.5mm},clip,width=\imgwidthBase]{perm/Llama_RoPE_-Res-______Canon-ABCD}
\\
\includegraphics[page=1,trim={2.5mm 1.5mm 2.5mm 1.5mm},clip,width=\imgwidthBase]{perm_multi/Llama_RoPE_-original}
\includegraphics[page=1,trim={2.5mm 1.5mm 2.5mm 1.5mm},clip,width=\imgwidthBase]{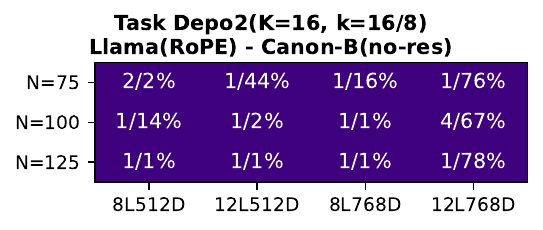}
\includegraphics[page=1,trim={2.5mm 1.5mm 2.5mm 1.5mm},clip,width=\imgwidthBase]{perm_multi/Llama_RoPE_-Res-Canon-ABCD}
\includegraphics[page=1,trim={2.5mm 1.5mm 2.5mm 1.5mm},clip,width=\imgwidthBase]{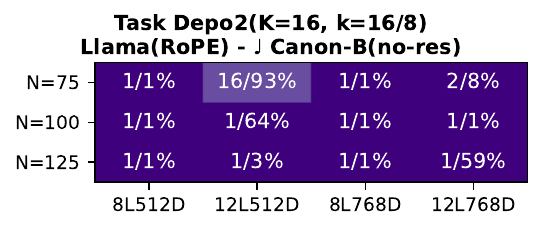}
\includegraphics[page=1,trim={2.5mm 1.5mm 2.5mm 1.5mm},clip,width=\imgwidthBase]{perm_multi/Llama_RoPE_-Res-______Canon-ABCD}
\\
\includegraphics[page=1,trim={2.5mm 1.5mm 2.5mm 1.5mm},clip,width=\imgwidthBase]{top_sort/Llama_RoPE_-original}
\includegraphics[page=1,trim={2.5mm 1.5mm 2.5mm 1.5mm},clip,width=\imgwidthBase]{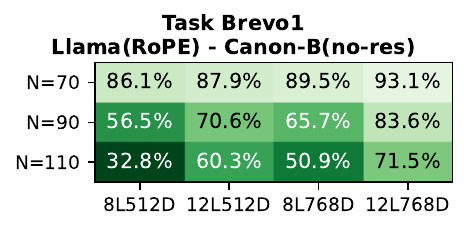}
\includegraphics[page=1,trim={2.5mm 1.5mm 2.5mm 1.5mm},clip,width=\imgwidthBase]{top_sort/Llama_RoPE_-Res-Canon-ABCD}
\includegraphics[page=1,trim={2.5mm 1.5mm 2.5mm 1.5mm},clip,width=\imgwidthBase]{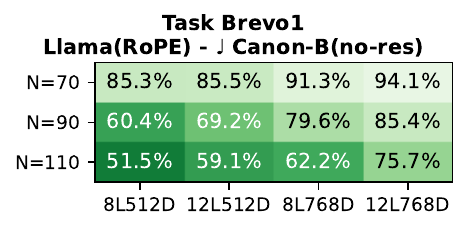}
\includegraphics[page=1,trim={2.5mm 1.5mm 2.5mm 1.5mm},clip,width=\imgwidthBase]{top_sort/Llama_RoPE_-Res-______Canon-ABCD}
\\
\includegraphics[page=1,trim={2.5mm 1.5mm 2.5mm 1.5mm},clip,width=\imgwidthBase]{top_sort_multi/Llama_RoPE_-original}
\includegraphics[page=1,trim={2.5mm 1.5mm 2.5mm 1.5mm},clip,width=\imgwidthBase]{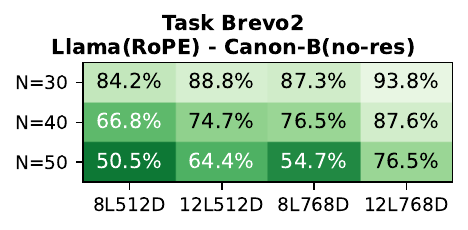}
\includegraphics[page=1,trim={2.5mm 1.5mm 2.5mm 1.5mm},clip,width=\imgwidthBase]{top_sort_multi/Llama_RoPE_-Res-Canon-ABCD}
\includegraphics[page=1,trim={2.5mm 1.5mm 2.5mm 1.5mm},clip,width=\imgwidthBase]{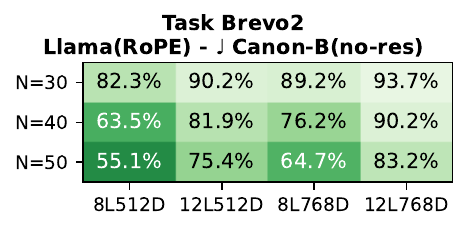}
\includegraphics[page=1,trim={2.5mm 1.5mm 2.5mm 1.5mm},clip,width=\imgwidthBase]{top_sort_multi/Llama_RoPE_-Res-______Canon-ABCD}
\\
\includegraphics[page=1,trim={2.5mm 1.5mm 2.5mm 1.5mm},clip,width=\imgwidthBase]{arith/Llama_RoPE_-original}
\includegraphics[page=1,trim={2.5mm 1.5mm 2.5mm 1.5mm},clip,width=\imgwidthBase]{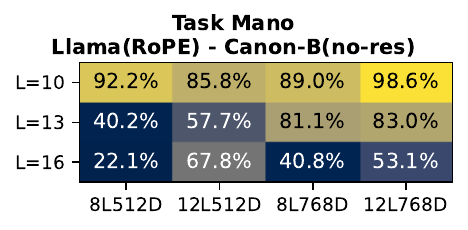}
\includegraphics[page=1,trim={2.5mm 1.5mm 2.5mm 1.5mm},clip,width=\imgwidthBase]{arith/Llama_RoPE_-Res-Canon-ABCD}
\includegraphics[page=1,trim={2.5mm 1.5mm 2.5mm 1.5mm},clip,width=\imgwidthBase]{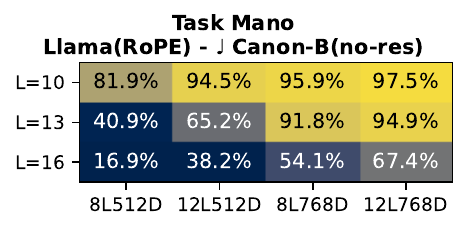}
\includegraphics[page=1,trim={2.5mm 1.5mm 2.5mm 1.5mm},clip,width=\imgwidthBase]{arith/Llama_RoPE_-Res-______Canon-ABCD}
\\
\includegraphics[page=1,trim={2.5mm 1.5mm 2.5mm 1.5mm},clip,width=\imgwidthBase]{cfg/Llama_RoPE_-original}
\includegraphics[page=1,trim={2.5mm 1.5mm 2.5mm 1.5mm},clip,width=\imgwidthBase]{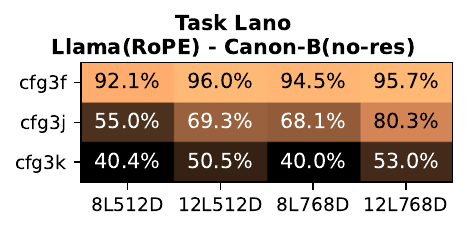}
\includegraphics[page=1,trim={2.5mm 1.5mm 2.5mm 1.5mm},clip,width=\imgwidthBase]{cfg/Llama_RoPE_-Res-Canon-ABCD}
\includegraphics[page=1,trim={2.5mm 1.5mm 2.5mm 1.5mm},clip,width=\imgwidthBase]{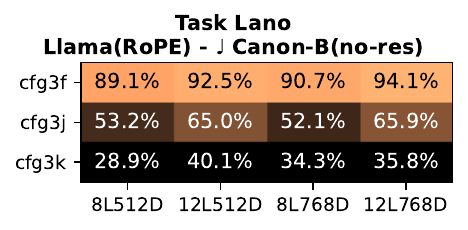}
\includegraphics[page=1,trim={2.5mm 1.5mm 2.5mm 1.5mm},clip,width=\imgwidthBase]{cfg/Llama_RoPE_-Res-______Canon-ABCD}
\\
\includegraphics[page=1,trim={2.5mm 1.5mm 2.5mm 1.5mm},clip,width=\imgwidthBase]{cfg-prob/Llama_RoPE_-original}
\includegraphics[page=1,trim={2.5mm 1.5mm 2.5mm 1.5mm},clip,width=\imgwidthBase]{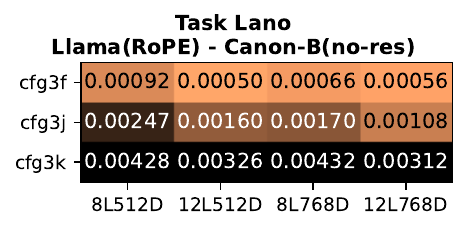}
\includegraphics[page=1,trim={2.5mm 1.5mm 2.5mm 1.5mm},clip,width=\imgwidthBase]{cfg-prob/Llama_RoPE_-Res-Canon-ABCD}
\includegraphics[page=1,trim={2.5mm 1.5mm 2.5mm 1.5mm},clip,width=\imgwidthBase]{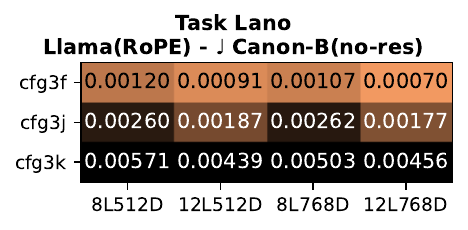}
\includegraphics[page=1,trim={2.5mm 1.5mm 2.5mm 1.5mm},clip,width=\imgwidthBase]{cfg-prob/Llama_RoPE_-Res-______Canon-ABCD}
\caption{\label{fig:app:primer}%
\textbf{Llama(\bblue{RoPE}) family vs. Primer}: (left to right) original, Canon-B(no-res), Canon-ABCD(res), \musQuarter{}Canon-B(no-res), \musQuarter{}Canon-ABCD(res).
\bblue{This figure} complements \figureref{fig:trans-ablation} and \bblue{directly compares to Primer~\cite{so2109primer} (i.e., Canon-B(no-res))}, showing its inefficiency and highlighting: (1) Canon layers are \emph{not tied} to Attention; (2) Canon(res) at multiple points is safe and more effective.
}
\end{figure}

\subsection{Llama(NoPE) family}

\begin{figure}[H]
\centering
\setlength{\imgwidthBase}{0.1245\textwidth}
\hspace*{-7mm}
\includegraphics[page=1,trim={2.5mm 1.5mm 2.5mm 1.5mm},clip,width=\imgwidthBase]{perm/Llama_NoPE_-original}
\includegraphics[page=1,trim={2.5mm 1.5mm 2.5mm 1.5mm},clip,width=\imgwidthBase]{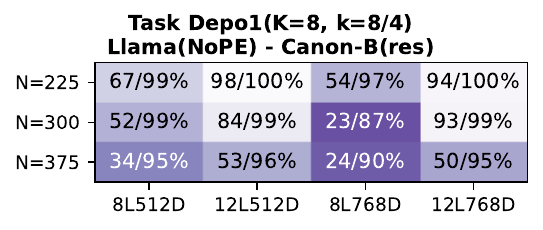}
\includegraphics[page=1,trim={2.5mm 1.5mm 2.5mm 1.5mm},clip,width=\imgwidthBase]{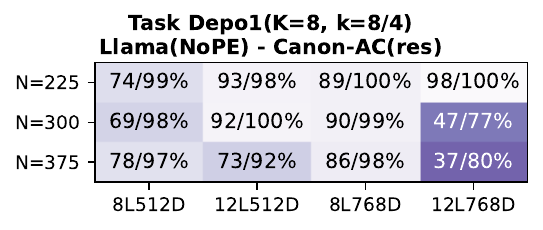}
\includegraphics[page=1,trim={2.5mm 1.5mm 2.5mm 1.5mm},clip,width=\imgwidthBase]{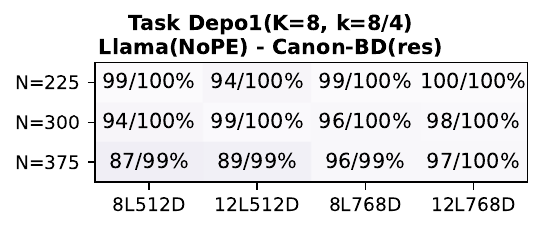}
\includegraphics[page=1,trim={2.5mm 1.5mm 2.5mm 1.5mm},clip,width=\imgwidthBase]{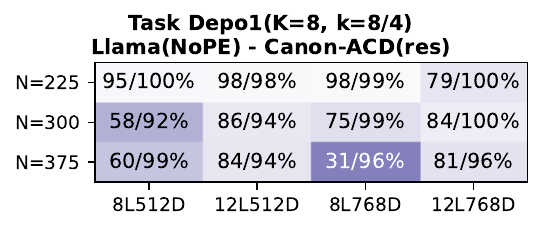}
\includegraphics[page=1,trim={2.5mm 1.5mm 2.5mm 1.5mm},clip,width=\imgwidthBase]{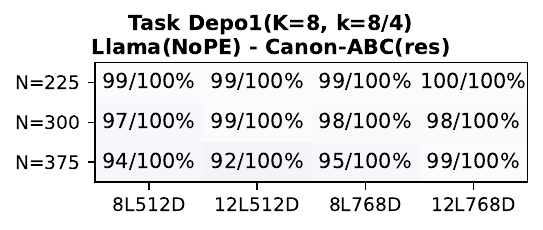}
\includegraphics[page=1,trim={2.5mm 1.5mm 2.5mm 1.5mm},clip,width=\imgwidthBase]{perm/Llama_NoPE_-Res-Canon-ABCD}
\includegraphics[page=1,trim={2.5mm 1.5mm 2.5mm 1.5mm},clip,width=\imgwidthBase]{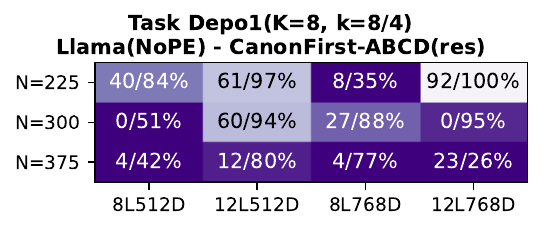}
\hspace*{-7mm}
\\
\hspace*{-7mm}
\includegraphics[page=1,trim={2.5mm 1.5mm 2.5mm 1.5mm},clip,width=\imgwidthBase]{perm_multi/Llama_NoPE_-original}
\includegraphics[page=1,trim={2.5mm 1.5mm 2.5mm 1.5mm},clip,width=\imgwidthBase]{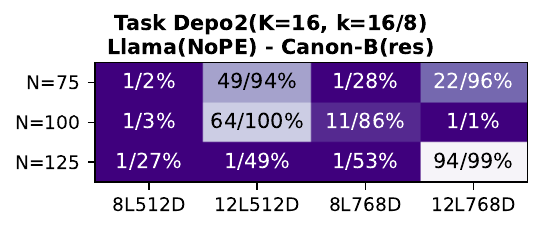}
\includegraphics[page=1,trim={2.5mm 1.5mm 2.5mm 1.5mm},clip,width=\imgwidthBase]{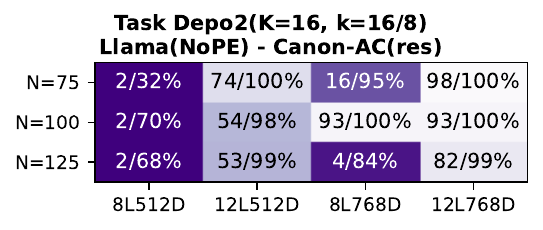}
\includegraphics[page=1,trim={2.5mm 1.5mm 2.5mm 1.5mm},clip,width=\imgwidthBase]{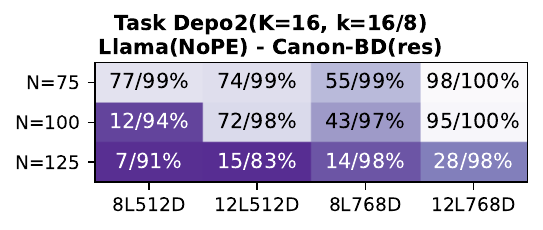}
\includegraphics[page=1,trim={2.5mm 1.5mm 2.5mm 1.5mm},clip,width=\imgwidthBase]{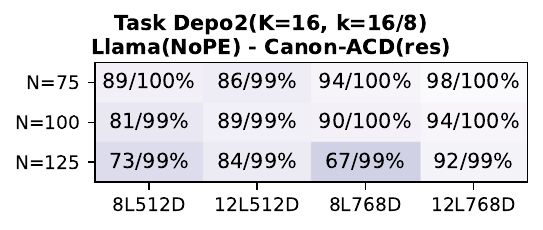}
\includegraphics[page=1,trim={2.5mm 1.5mm 2.5mm 1.5mm},clip,width=\imgwidthBase]{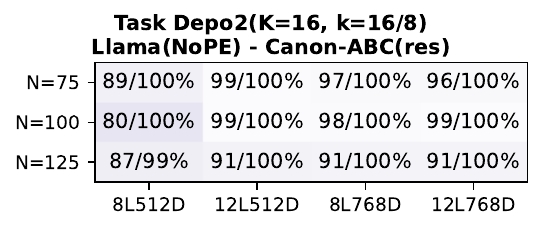}
\includegraphics[page=1,trim={2.5mm 1.5mm 2.5mm 1.5mm},clip,width=\imgwidthBase]{perm_multi/Llama_NoPE_-Res-Canon-ABCD}
\includegraphics[page=1,trim={2.5mm 1.5mm 2.5mm 1.5mm},clip,width=\imgwidthBase]{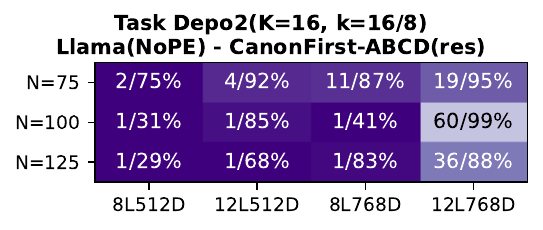}
\hspace*{-7mm}
\\
\hspace*{-7mm}
\includegraphics[page=1,trim={2.5mm 1.5mm 2.5mm 1.5mm},clip,width=\imgwidthBase]{top_sort/Llama_NoPE_-original}
\includegraphics[page=1,trim={2.5mm 1.5mm 2.5mm 1.5mm},clip,width=\imgwidthBase]{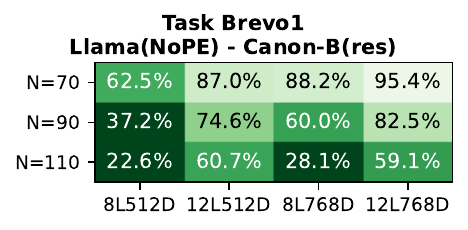}
\includegraphics[page=1,trim={2.5mm 1.5mm 2.5mm 1.5mm},clip,width=\imgwidthBase]{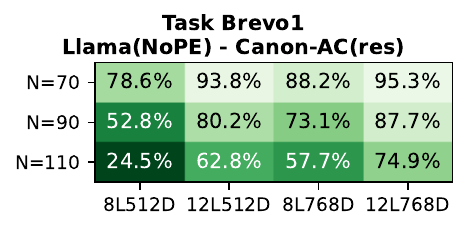}
\includegraphics[page=1,trim={2.5mm 1.5mm 2.5mm 1.5mm},clip,width=\imgwidthBase]{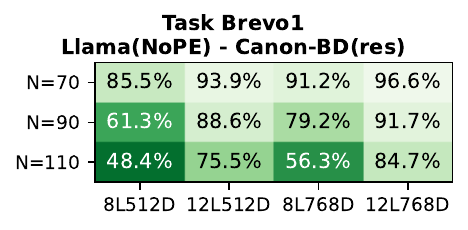}
\includegraphics[page=1,trim={2.5mm 1.5mm 2.5mm 1.5mm},clip,width=\imgwidthBase]{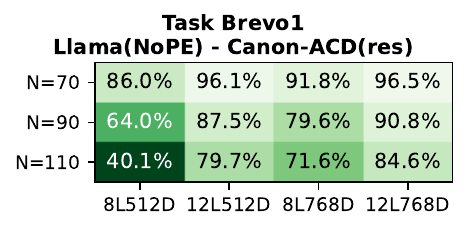}
\includegraphics[page=1,trim={2.5mm 1.5mm 2.5mm 1.5mm},clip,width=\imgwidthBase]{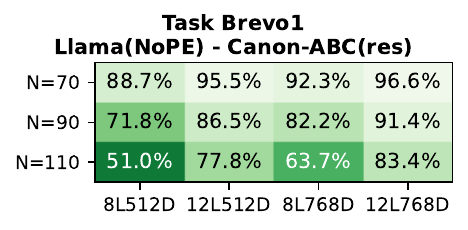}
\includegraphics[page=1,trim={2.5mm 1.5mm 2.5mm 1.5mm},clip,width=\imgwidthBase]{top_sort/Llama_NoPE_-Res-Canon-ABCD}
\includegraphics[page=1,trim={2.5mm 1.5mm 2.5mm 1.5mm},clip,width=\imgwidthBase]{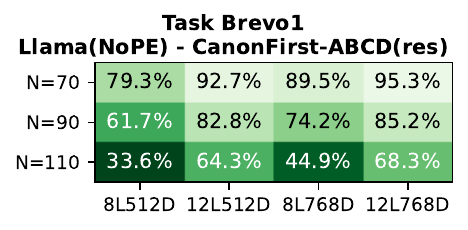}
\hspace*{-7mm}
\\
\hspace*{-7mm}
\includegraphics[page=1,trim={2.5mm 1.5mm 2.5mm 1.5mm},clip,width=\imgwidthBase]{top_sort_multi/Llama_NoPE_-original}
\includegraphics[page=1,trim={2.5mm 1.5mm 2.5mm 1.5mm},clip,width=\imgwidthBase]{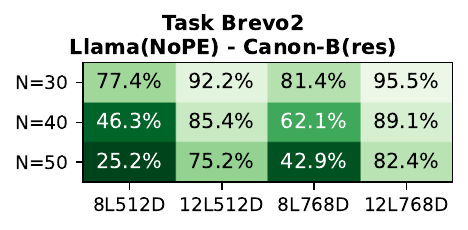}
\includegraphics[page=1,trim={2.5mm 1.5mm 2.5mm 1.5mm},clip,width=\imgwidthBase]{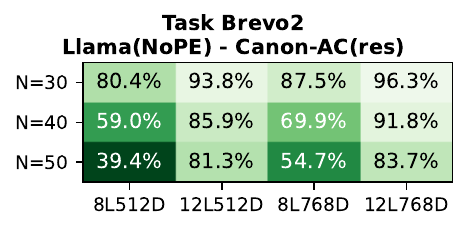}
\includegraphics[page=1,trim={2.5mm 1.5mm 2.5mm 1.5mm},clip,width=\imgwidthBase]{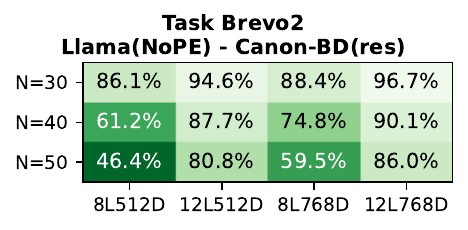}
\includegraphics[page=1,trim={2.5mm 1.5mm 2.5mm 1.5mm},clip,width=\imgwidthBase]{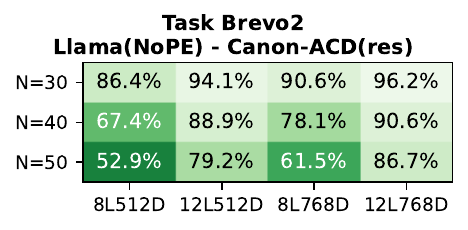}
\includegraphics[page=1,trim={2.5mm 1.5mm 2.5mm 1.5mm},clip,width=\imgwidthBase]{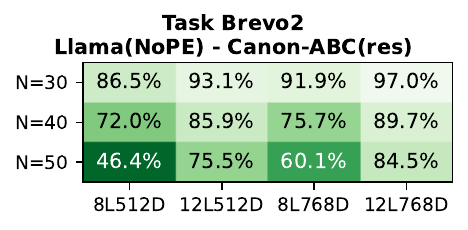}
\includegraphics[page=1,trim={2.5mm 1.5mm 2.5mm 1.5mm},clip,width=\imgwidthBase]{top_sort_multi/Llama_NoPE_-Res-Canon-ABCD}
\includegraphics[page=1,trim={2.5mm 1.5mm 2.5mm 1.5mm},clip,width=\imgwidthBase]{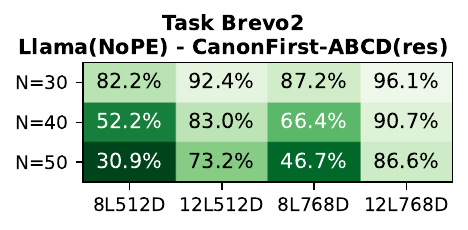}
\hspace*{-7mm}
\\
\hspace*{-7mm}
\includegraphics[page=1,trim={2.5mm 1.5mm 2.5mm 1.5mm},clip,width=\imgwidthBase]{arith/Llama_NoPE_-original}
\includegraphics[page=1,trim={2.5mm 1.5mm 2.5mm 1.5mm},clip,width=\imgwidthBase]{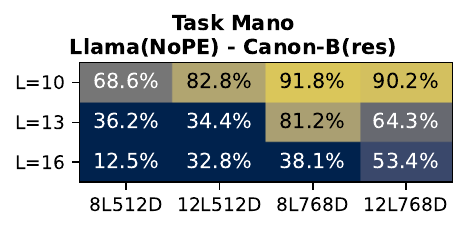}
\includegraphics[page=1,trim={2.5mm 1.5mm 2.5mm 1.5mm},clip,width=\imgwidthBase]{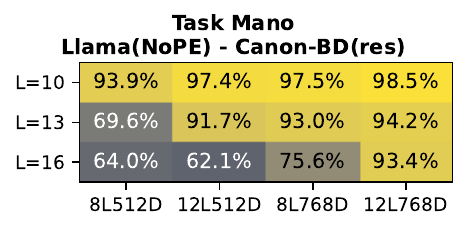}
\includegraphics[page=1,trim={2.5mm 1.5mm 2.5mm 1.5mm},clip,width=\imgwidthBase]{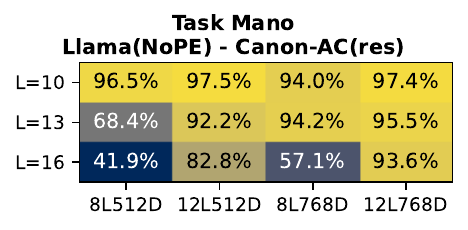}
\includegraphics[page=1,trim={2.5mm 1.5mm 2.5mm 1.5mm},clip,width=\imgwidthBase]{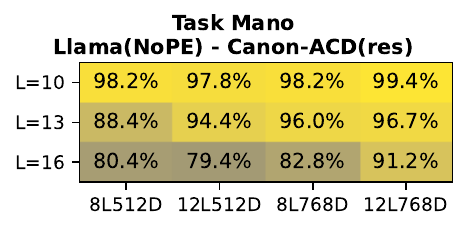}
\includegraphics[page=1,trim={2.5mm 1.5mm 2.5mm 1.5mm},clip,width=\imgwidthBase]{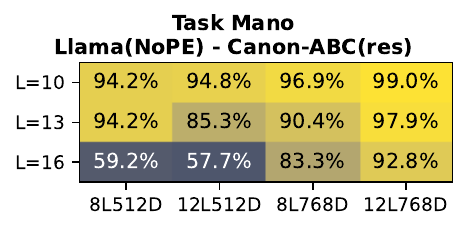}
\includegraphics[page=1,trim={2.5mm 1.5mm 2.5mm 1.5mm},clip,width=\imgwidthBase]{arith/Llama_NoPE_-Res-Canon-ABCD}
\includegraphics[page=1,trim={2.5mm 1.5mm 2.5mm 1.5mm},clip,width=\imgwidthBase]{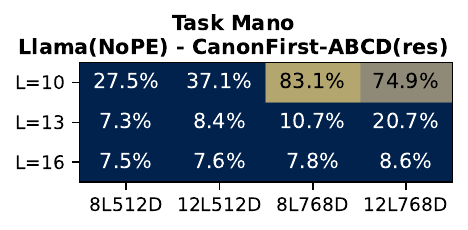}
\hspace*{-7mm}
\\
\hspace*{-7mm}
\includegraphics[page=1,trim={2.5mm 1.5mm 2.5mm 1.5mm},clip,width=\imgwidthBase]{cfg/Llama_NoPE_-original}
\includegraphics[page=1,trim={2.5mm 1.5mm 2.5mm 1.5mm},clip,width=\imgwidthBase]{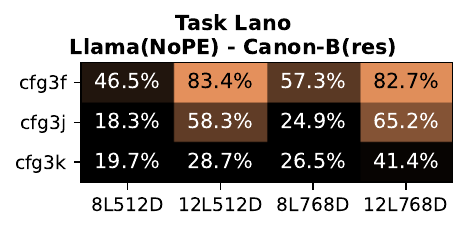}
\includegraphics[page=1,trim={2.5mm 1.5mm 2.5mm 1.5mm},clip,width=\imgwidthBase]{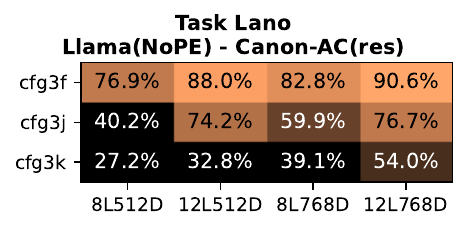}
\includegraphics[page=1,trim={2.5mm 1.5mm 2.5mm 1.5mm},clip,width=\imgwidthBase]{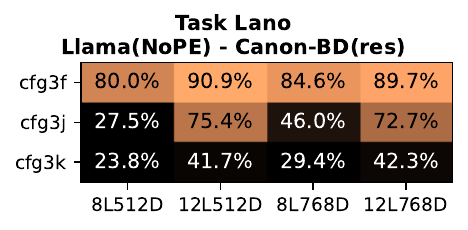}
\includegraphics[page=1,trim={2.5mm 1.5mm 2.5mm 1.5mm},clip,width=\imgwidthBase]{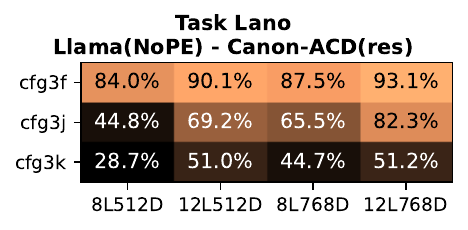}
\includegraphics[page=1,trim={2.5mm 1.5mm 2.5mm 1.5mm},clip,width=\imgwidthBase]{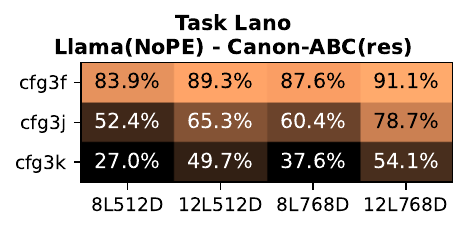}
\includegraphics[page=1,trim={2.5mm 1.5mm 2.5mm 1.5mm},clip,width=\imgwidthBase]{cfg/Llama_NoPE_-Res-Canon-ABCD}
\includegraphics[page=1,trim={2.5mm 1.5mm 2.5mm 1.5mm},clip,width=\imgwidthBase]{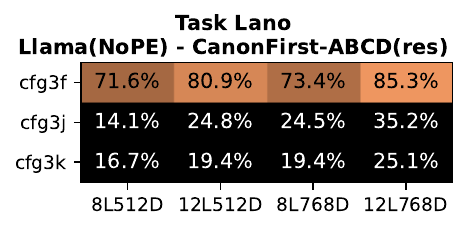}
\hspace*{-7mm}
\\
\hspace*{-7mm}
\includegraphics[page=1,trim={2.5mm 1.5mm 2.5mm 1.5mm},clip,width=\imgwidthBase]{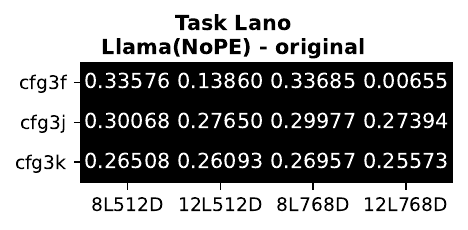}
\includegraphics[page=1,trim={2.5mm 1.5mm 2.5mm 1.5mm},clip,width=\imgwidthBase]{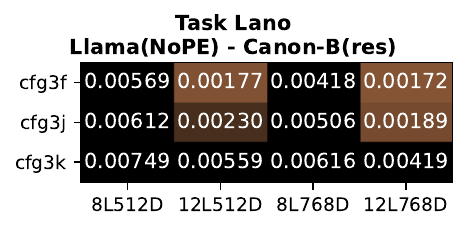}
\includegraphics[page=1,trim={2.5mm 1.5mm 2.5mm 1.5mm},clip,width=\imgwidthBase]{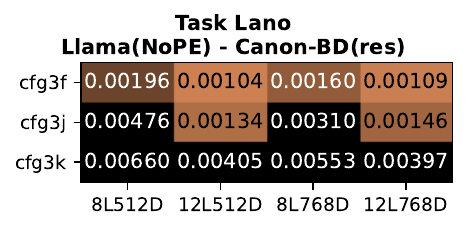}
\includegraphics[page=1,trim={2.5mm 1.5mm 2.5mm 1.5mm},clip,width=\imgwidthBase]{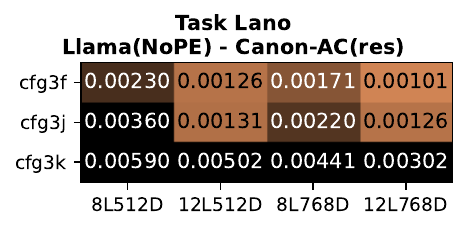}
\includegraphics[page=1,trim={2.5mm 1.5mm 2.5mm 1.5mm},clip,width=\imgwidthBase]{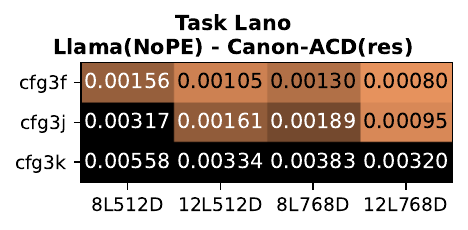}
\includegraphics[page=1,trim={2.5mm 1.5mm 2.5mm 1.5mm},clip,width=\imgwidthBase]{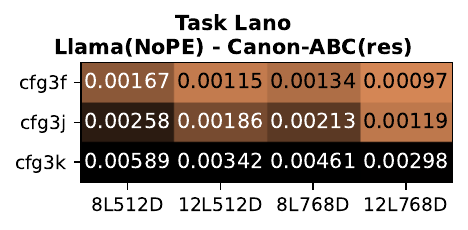}
\includegraphics[page=1,trim={2.5mm 1.5mm 2.5mm 1.5mm},clip,width=\imgwidthBase]{cfg-prob/Llama_NoPE_-Res-Canon-ABCD}
\includegraphics[page=1,trim={2.5mm 1.5mm 2.5mm 1.5mm},clip,width=\imgwidthBase]{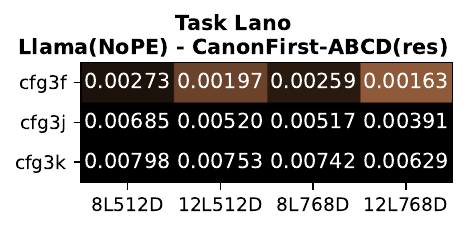}
\hspace*{-7mm}
\caption{\label{fig:app:nope-ablation}\textbf{Llama(\bblue{NoPE}) family}: (left to right) original, Canon-B, -AC, -BD, -ACD, -ABC, -ABCD; and CanonFirst-ABCD, where Canon is applied only to the first layer.
\newline
This figure complements \figureref{fig:trans-ablation} and gives more technical details.}
\end{figure}

\subsection{Mamba2 family}

\begin{figure}[H]
\centering
\vspace{-3mm}\setlength{\imgwidthBase}{0.175\textwidth}
\includegraphics[page=1,trim={2.5mm 1.5mm 2.5mm 1.5mm},clip,height=\imgwidthBase]{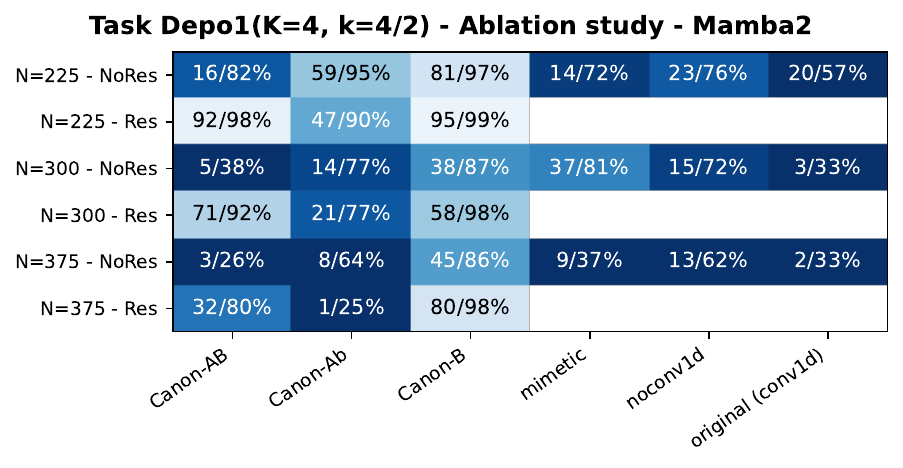}
\includegraphics[page=1,trim={2.5mm 1.5mm 2.5mm 1.5mm},clip,height=\imgwidthBase]{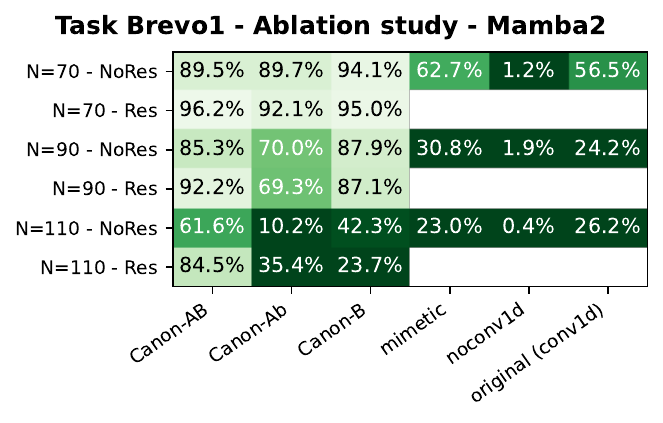}
\includegraphics[page=1,trim={2.5mm 1.5mm 2.5mm 1.5mm},clip,height=\imgwidthBase]{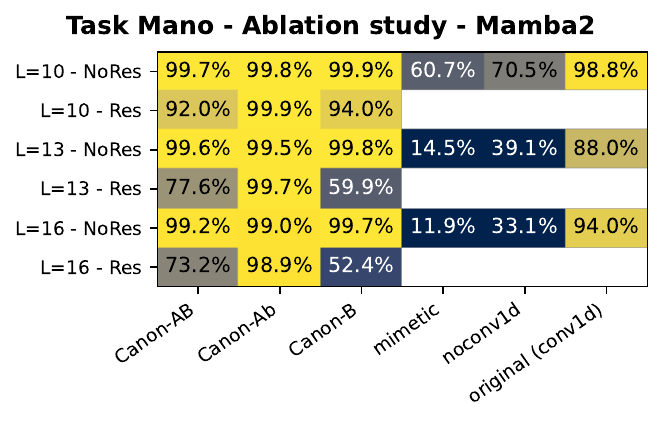}
\\
\includegraphics[page=1,trim={2.5mm 1.5mm 2.5mm 1.5mm},clip,height=\imgwidthBase]{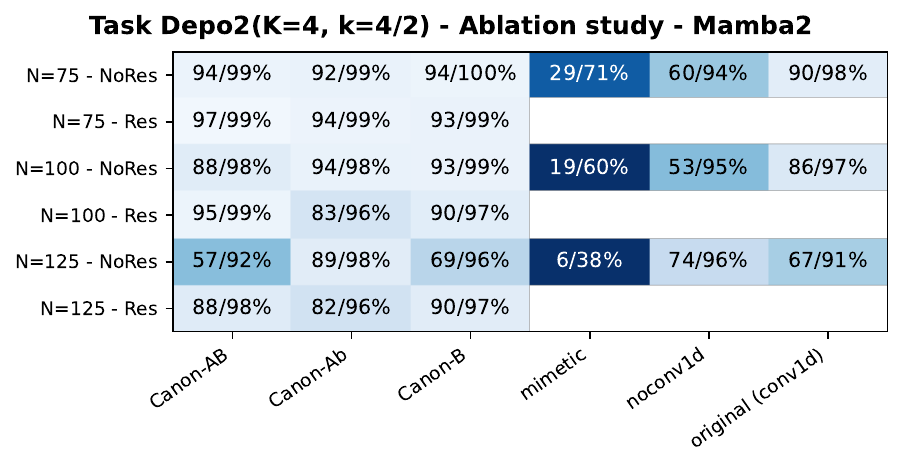}
\includegraphics[page=1,trim={2.5mm 1.5mm 2.5mm 1.5mm},clip,height=\imgwidthBase]{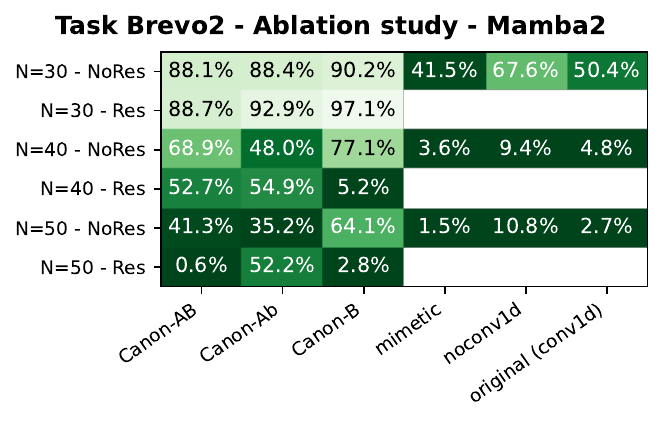}
\includegraphics[page=1,trim={2.5mm 1.5mm 2.5mm 1.5mm},clip,height=\imgwidthBase]{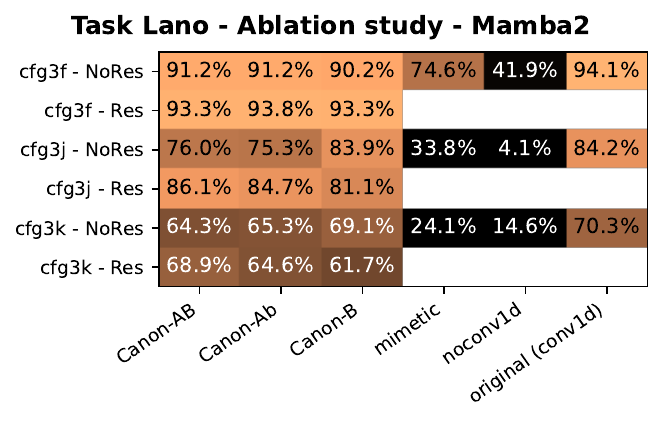}
\\
\vspace{-4mm}
\seplineb
\\
\includegraphics[page=1,trim={2.5mm 1.5mm 2.5mm 1.5mm},clip,height=\imgwidthBase]{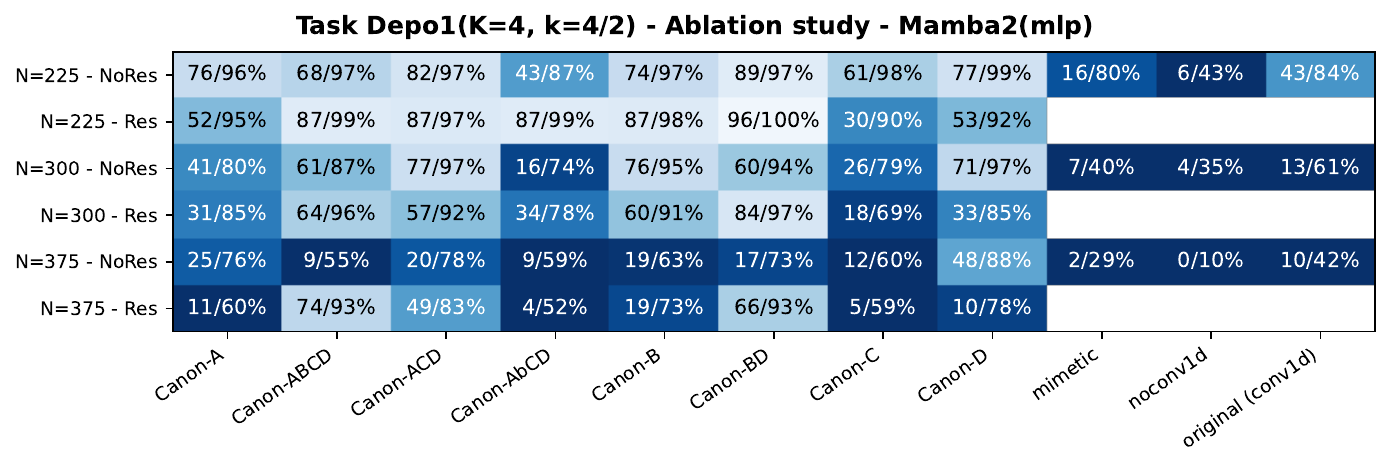}
\includegraphics[page=1,trim={2.5mm 1.5mm 2.5mm 1.5mm},clip,height=\imgwidthBase]{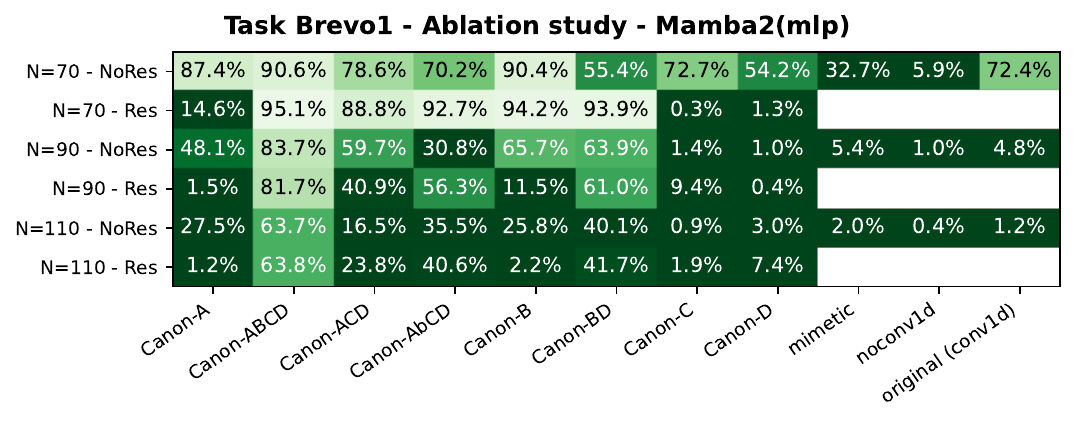}
\\
\includegraphics[page=1,trim={2.5mm 1.5mm 2.5mm 1.5mm},clip,height=\imgwidthBase]{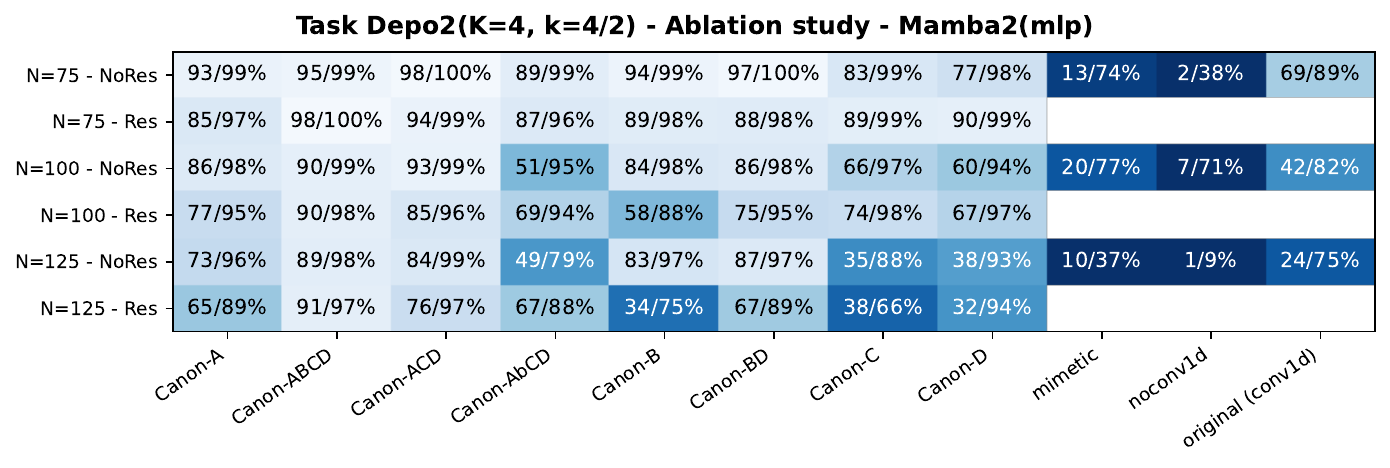}
\includegraphics[page=1,trim={2.5mm 1.5mm 2.5mm 1.5mm},clip,height=\imgwidthBase]{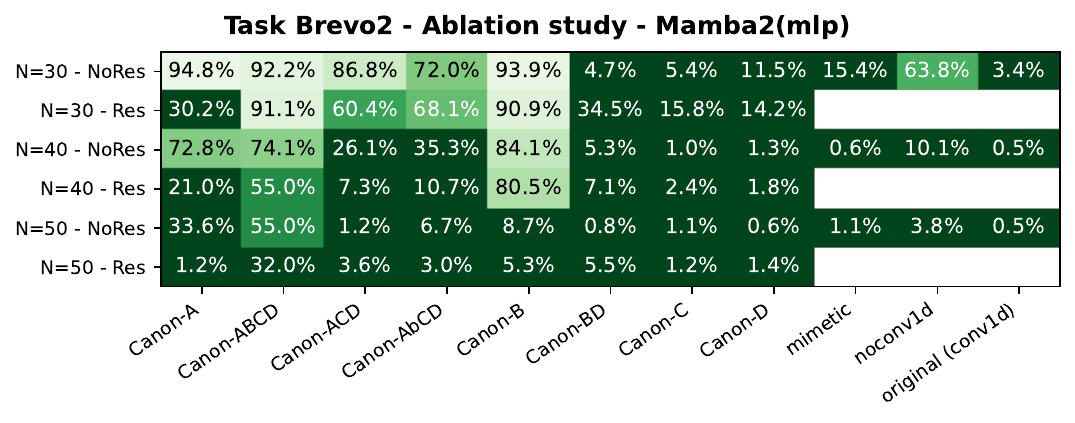}
\\
\includegraphics[page=1,trim={2.5mm 1.5mm 2.5mm 1.5mm},clip,height=\imgwidthBase]{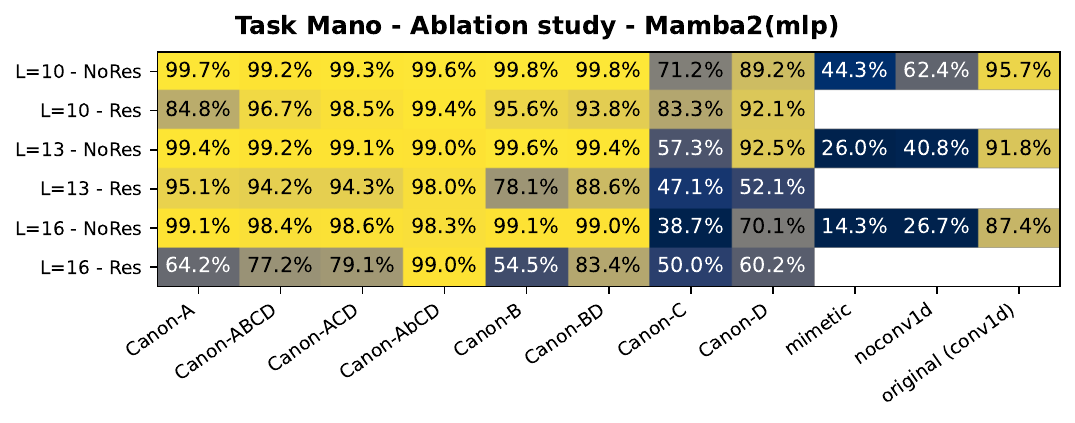}
\includegraphics[page=1,trim={2.5mm 1.5mm 2.5mm 1.5mm},clip,height=\imgwidthBase]{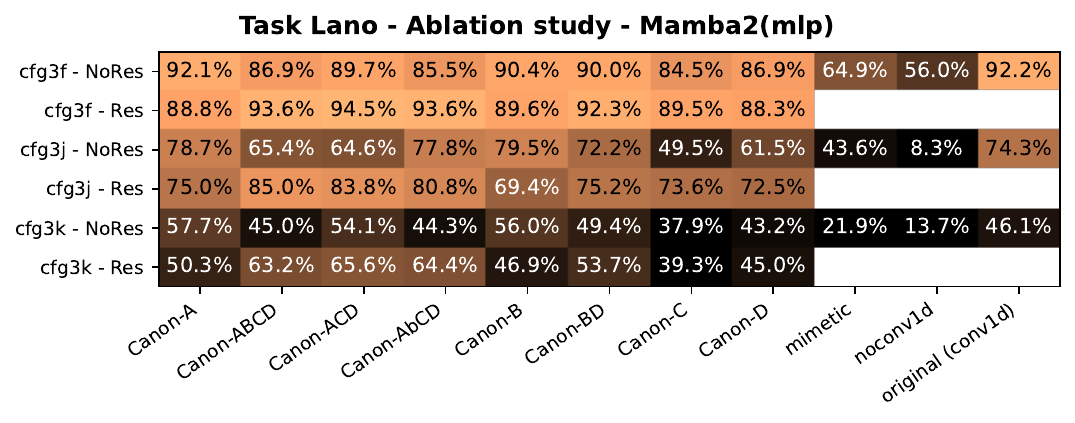}
\caption{\label{fig:mamba-ablation}\textbf{Ablation study of Mamba2 models of 12L768D size} with Canon layers, Canon residuals, original non-linear \texttt{conv1d}, mimetic initialization. Full ablation studies (with additional model sizes, and the effectiveness of Canon-ACD) are in \figureref{fig:full-mamba2}-\ref{fig:full-mamba2-mlp}.}
\end{figure}

\begin{figure}[H]
\centering
\setlength{\imgwidthBase}{0.14\textwidth}
\hspace*{-7mm}
\includegraphics[page=1,trim={2.5mm 1.5mm 2.5mm 1.5mm},clip,width=\imgwidthBase]{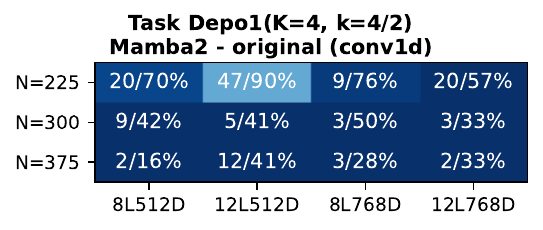}
\includegraphics[page=1,trim={2.5mm 1.5mm 2.5mm 1.5mm},clip,width=\imgwidthBase]{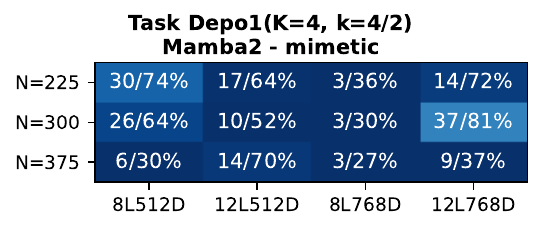}
\includegraphics[page=1,trim={2.5mm 1.5mm 2.5mm 1.5mm},clip,width=\imgwidthBase]{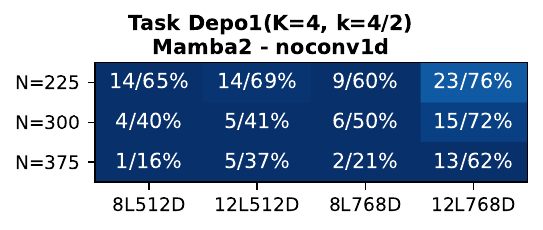}
\includegraphics[page=1,trim={2.5mm 1.5mm 2.5mm 1.5mm},clip,width=\imgwidthBase]{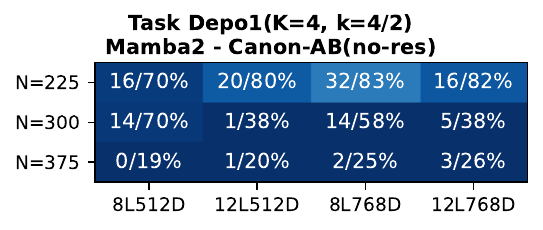}
\includegraphics[page=1,trim={2.5mm 1.5mm 2.5mm 1.5mm},clip,width=\imgwidthBase]{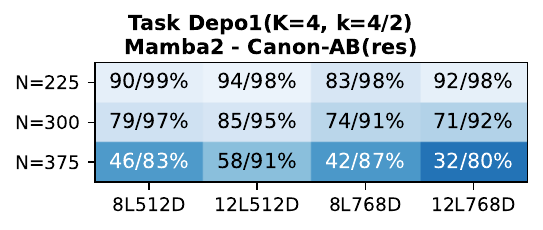}
\includegraphics[page=1,trim={2.5mm 1.5mm 2.5mm 1.5mm},clip,width=\imgwidthBase]{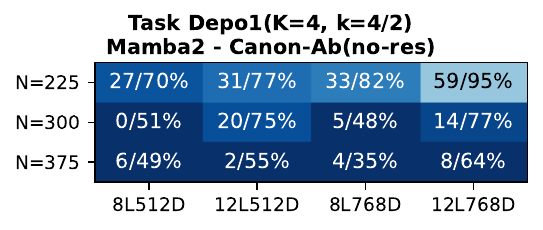}
\includegraphics[page=1,trim={2.5mm 1.5mm 2.5mm 1.5mm},clip,width=\imgwidthBase]{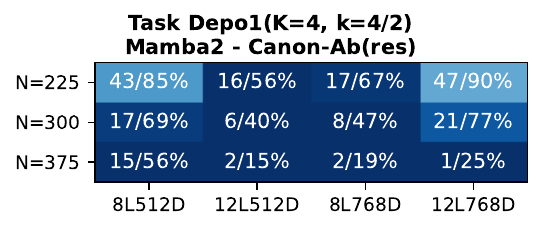}
\hspace*{-7mm}
\\
\hspace*{-7mm}
\includegraphics[page=1,trim={2.5mm 1.5mm 2.5mm 1.5mm},clip,width=\imgwidthBase]{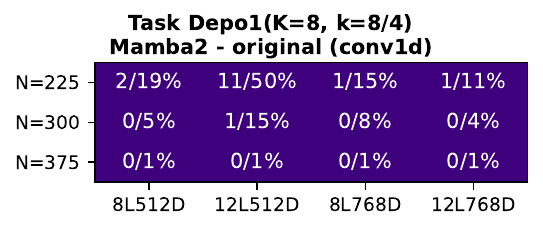}
\includegraphics[page=1,trim={2.5mm 1.5mm 2.5mm 1.5mm},clip,width=\imgwidthBase]{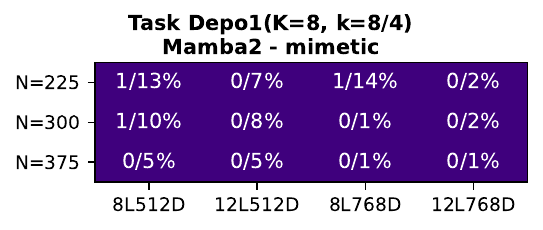}
\includegraphics[page=1,trim={2.5mm 1.5mm 2.5mm 1.5mm},clip,width=\imgwidthBase]{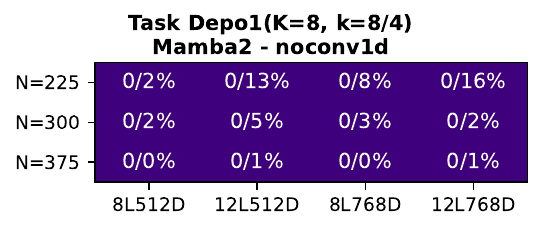}
\includegraphics[page=1,trim={2.5mm 1.5mm 2.5mm 1.5mm},clip,width=\imgwidthBase]{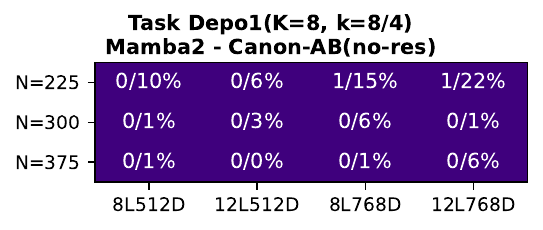}
\includegraphics[page=1,trim={2.5mm 1.5mm 2.5mm 1.5mm},clip,width=\imgwidthBase]{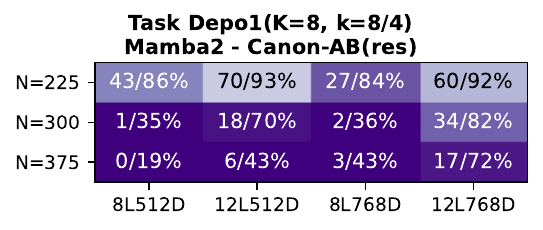}
\includegraphics[page=1,trim={2.5mm 1.5mm 2.5mm 1.5mm},clip,width=\imgwidthBase]{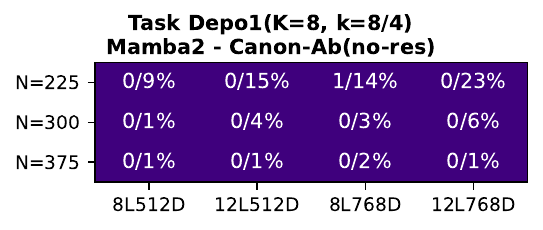}
\includegraphics[page=1,trim={2.5mm 1.5mm 2.5mm 1.5mm},clip,width=\imgwidthBase]{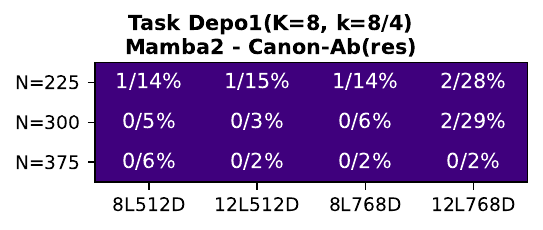}
\hspace*{-7mm}
\\
\hspace*{-7mm}
\includegraphics[page=1,trim={2.5mm 1.5mm 2.5mm 1.5mm},clip,width=\imgwidthBase]{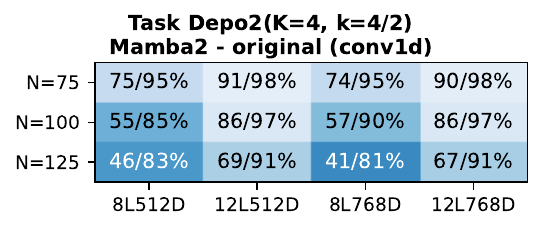}
\includegraphics[page=1,trim={2.5mm 1.5mm 2.5mm 1.5mm},clip,width=\imgwidthBase]{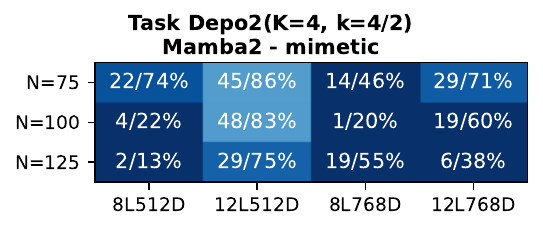}
\includegraphics[page=1,trim={2.5mm 1.5mm 2.5mm 1.5mm},clip,width=\imgwidthBase]{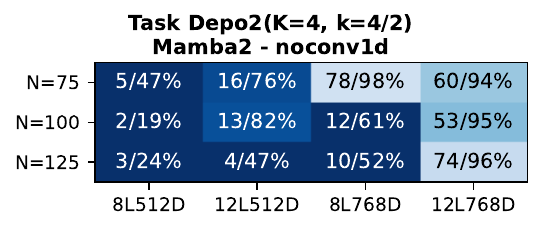}
\includegraphics[page=1,trim={2.5mm 1.5mm 2.5mm 1.5mm},clip,width=\imgwidthBase]{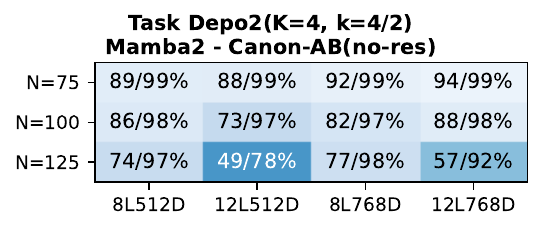}
\includegraphics[page=1,trim={2.5mm 1.5mm 2.5mm 1.5mm},clip,width=\imgwidthBase]{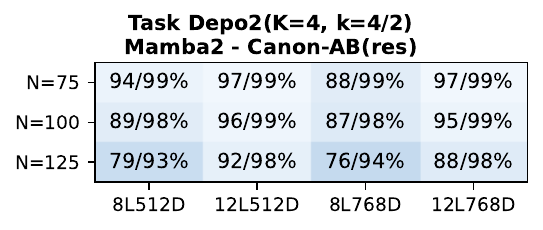}
\includegraphics[page=1,trim={2.5mm 1.5mm 2.5mm 1.5mm},clip,width=\imgwidthBase]{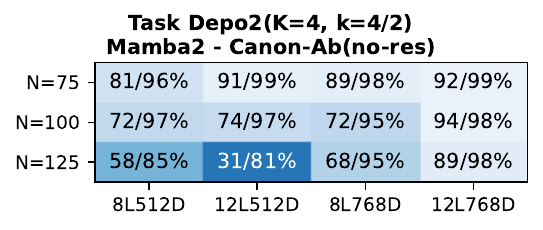}
\includegraphics[page=1,trim={2.5mm 1.5mm 2.5mm 1.5mm},clip,width=\imgwidthBase]{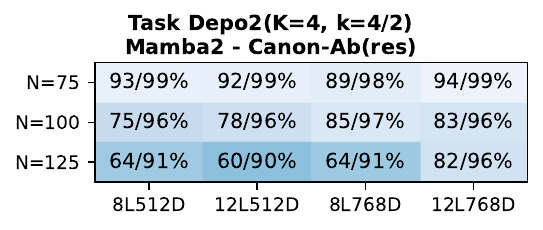}
\hspace*{-7mm}
\\
\hspace*{-7mm}
\includegraphics[page=1,trim={2.5mm 1.5mm 2.5mm 1.5mm},clip,width=\imgwidthBase]{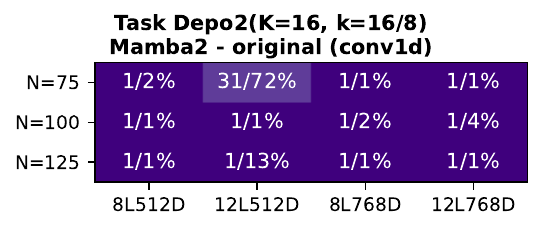}
\includegraphics[page=1,trim={2.5mm 1.5mm 2.5mm 1.5mm},clip,width=\imgwidthBase]{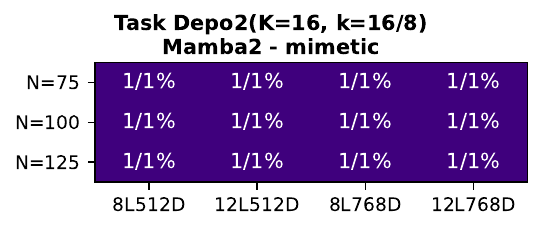}
\includegraphics[page=1,trim={2.5mm 1.5mm 2.5mm 1.5mm},clip,width=\imgwidthBase]{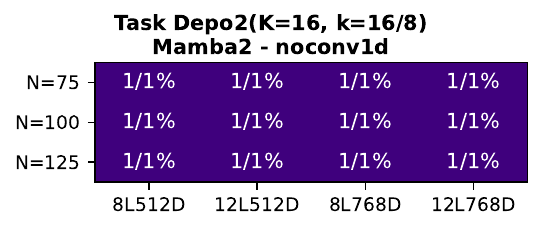}
\includegraphics[page=1,trim={2.5mm 1.5mm 2.5mm 1.5mm},clip,width=\imgwidthBase]{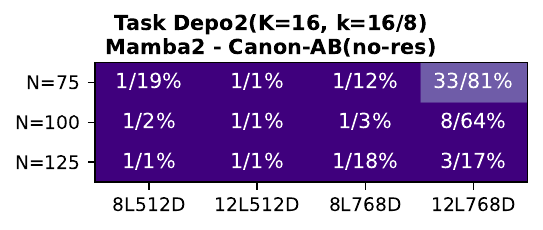}
\includegraphics[page=1,trim={2.5mm 1.5mm 2.5mm 1.5mm},clip,width=\imgwidthBase]{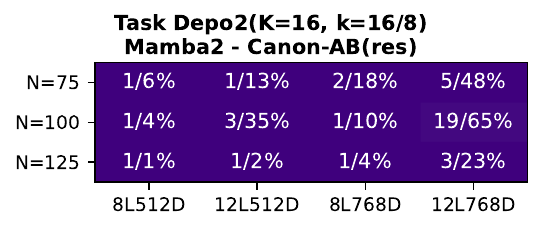}
\includegraphics[page=1,trim={2.5mm 1.5mm 2.5mm 1.5mm},clip,width=\imgwidthBase]{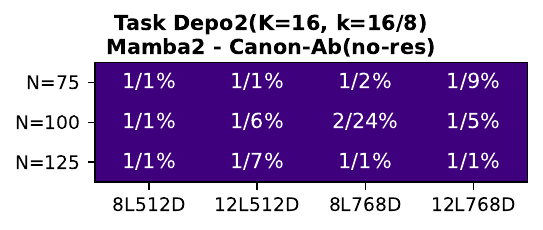}
\includegraphics[page=1,trim={2.5mm 1.5mm 2.5mm 1.5mm},clip,width=\imgwidthBase]{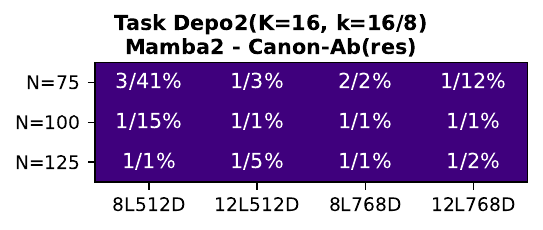}
\hspace*{-7mm}
\\
\hspace*{-7mm}
\includegraphics[page=1,trim={2.5mm 1.5mm 2.5mm 1.5mm},clip,width=\imgwidthBase]{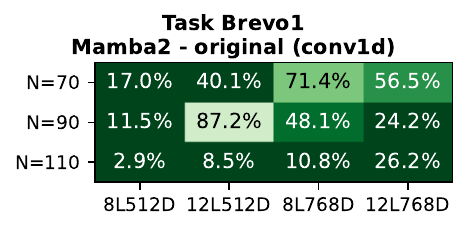}
\includegraphics[page=1,trim={2.5mm 1.5mm 2.5mm 1.5mm},clip,width=\imgwidthBase]{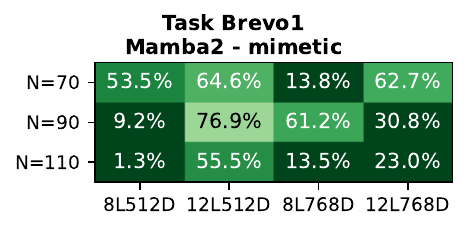}
\includegraphics[page=1,trim={2.5mm 1.5mm 2.5mm 1.5mm},clip,width=\imgwidthBase]{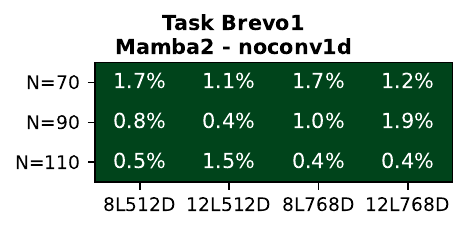}
\includegraphics[page=1,trim={2.5mm 1.5mm 2.5mm 1.5mm},clip,width=\imgwidthBase]{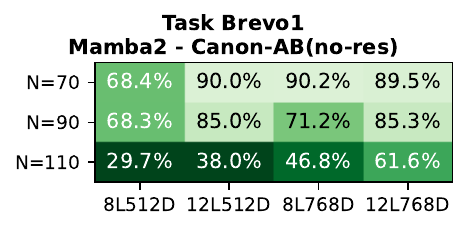}
\includegraphics[page=1,trim={2.5mm 1.5mm 2.5mm 1.5mm},clip,width=\imgwidthBase]{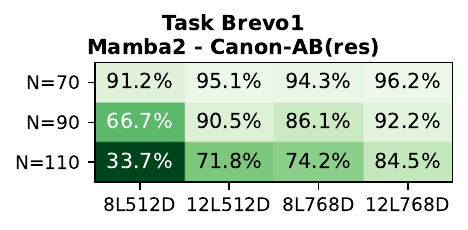}
\includegraphics[page=1,trim={2.5mm 1.5mm 2.5mm 1.5mm},clip,width=\imgwidthBase]{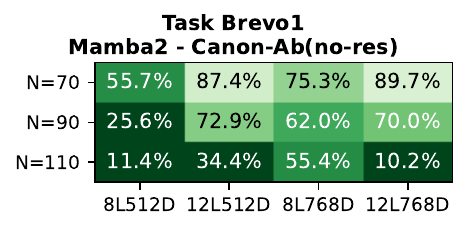}
\includegraphics[page=1,trim={2.5mm 1.5mm 2.5mm 1.5mm},clip,width=\imgwidthBase]{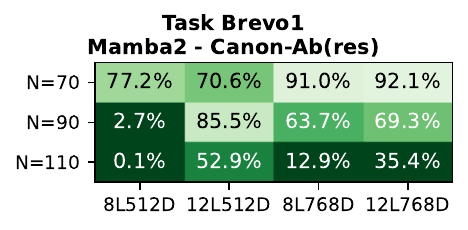}
\hspace*{-7mm}
\\
\hspace*{-7mm}
\includegraphics[page=1,trim={2.5mm 1.5mm 2.5mm 1.5mm},clip,width=\imgwidthBase]{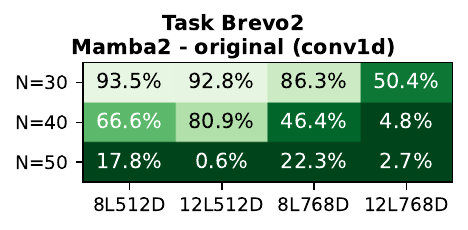}
\includegraphics[page=1,trim={2.5mm 1.5mm 2.5mm 1.5mm},clip,width=\imgwidthBase]{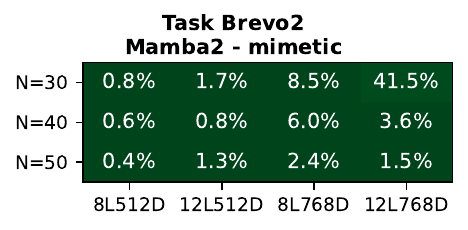}
\includegraphics[page=1,trim={2.5mm 1.5mm 2.5mm 1.5mm},clip,width=\imgwidthBase]{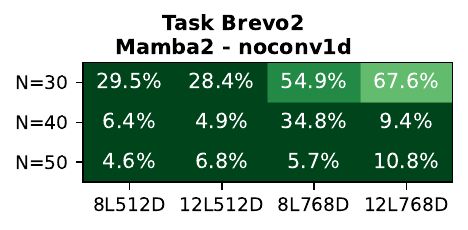}
\includegraphics[page=1,trim={2.5mm 1.5mm 2.5mm 1.5mm},clip,width=\imgwidthBase]{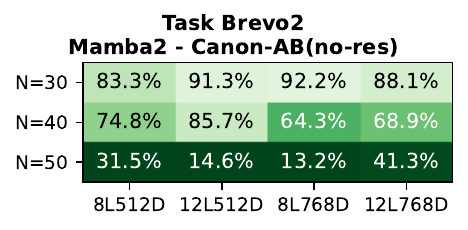}
\includegraphics[page=1,trim={2.5mm 1.5mm 2.5mm 1.5mm},clip,width=\imgwidthBase]{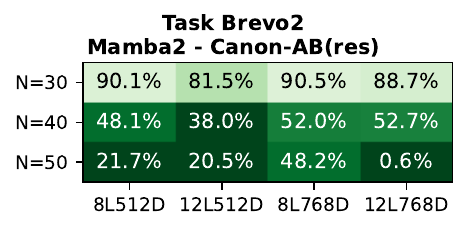}
\includegraphics[page=1,trim={2.5mm 1.5mm 2.5mm 1.5mm},clip,width=\imgwidthBase]{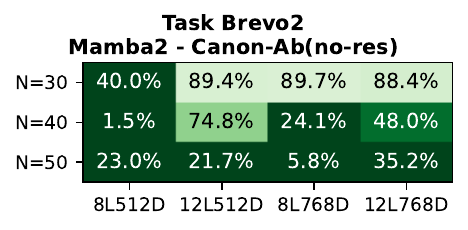}
\includegraphics[page=1,trim={2.5mm 1.5mm 2.5mm 1.5mm},clip,width=\imgwidthBase]{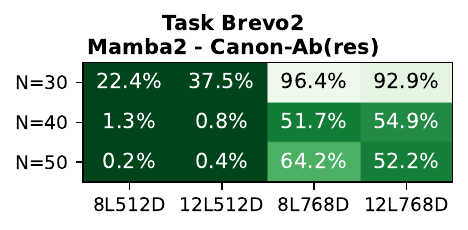}
\hspace*{-7mm}
\\
\hspace*{-7mm}
\includegraphics[page=1,trim={2.5mm 1.5mm 2.5mm 1.5mm},clip,width=\imgwidthBase]{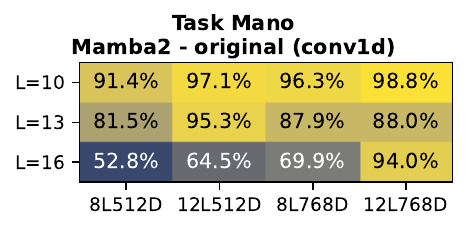}
\includegraphics[page=1,trim={2.5mm 1.5mm 2.5mm 1.5mm},clip,width=\imgwidthBase]{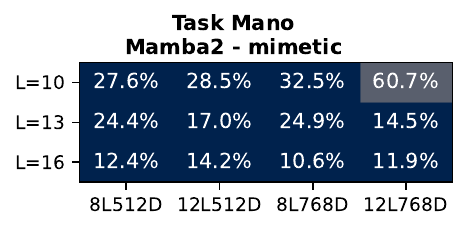}
\includegraphics[page=1,trim={2.5mm 1.5mm 2.5mm 1.5mm},clip,width=\imgwidthBase]{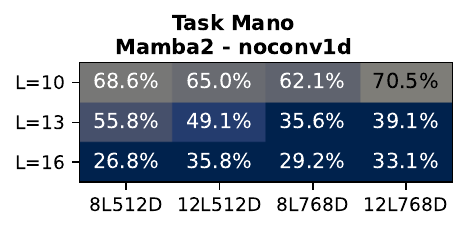}
\includegraphics[page=1,trim={2.5mm 1.5mm 2.5mm 1.5mm},clip,width=\imgwidthBase]{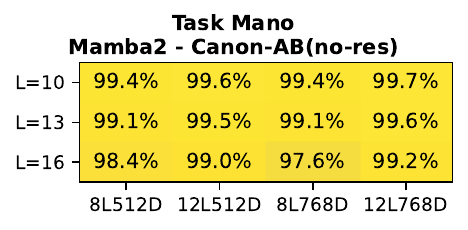}
\includegraphics[page=1,trim={2.5mm 1.5mm 2.5mm 1.5mm},clip,width=\imgwidthBase]{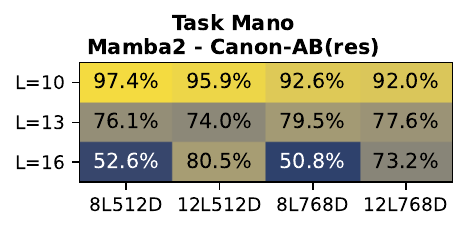}
\includegraphics[page=1,trim={2.5mm 1.5mm 2.5mm 1.5mm},clip,width=\imgwidthBase]{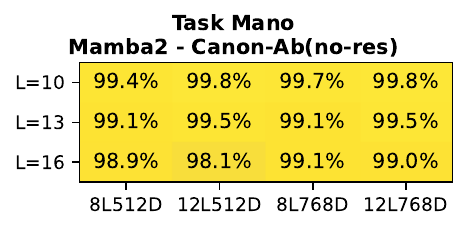}
\includegraphics[page=1,trim={2.5mm 1.5mm 2.5mm 1.5mm},clip,width=\imgwidthBase]{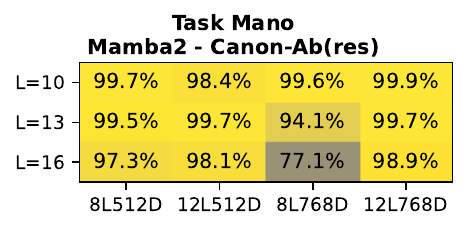}
\hspace*{-7mm}
\\
\hspace*{-7mm}
\includegraphics[page=1,trim={2.5mm 1.5mm 2.5mm 1.5mm},clip,width=\imgwidthBase]{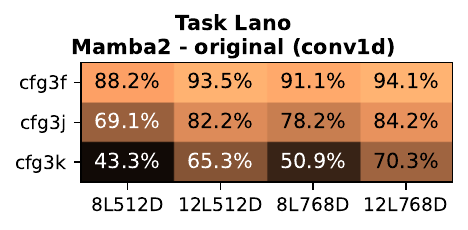}
\includegraphics[page=1,trim={2.5mm 1.5mm 2.5mm 1.5mm},clip,width=\imgwidthBase]{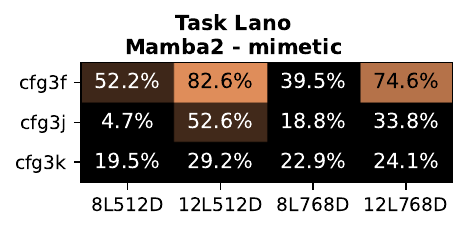}
\includegraphics[page=1,trim={2.5mm 1.5mm 2.5mm 1.5mm},clip,width=\imgwidthBase]{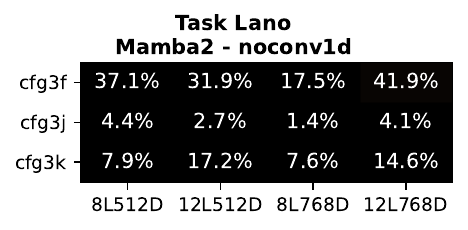}
\includegraphics[page=1,trim={2.5mm 1.5mm 2.5mm 1.5mm},clip,width=\imgwidthBase]{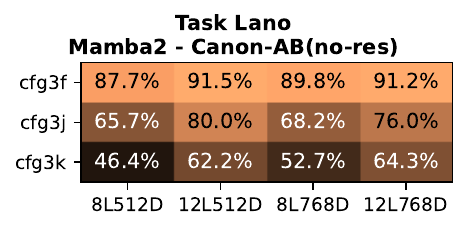}
\includegraphics[page=1,trim={2.5mm 1.5mm 2.5mm 1.5mm},clip,width=\imgwidthBase]{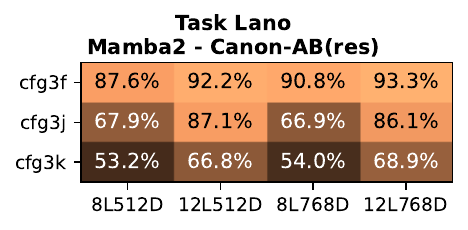}
\includegraphics[page=1,trim={2.5mm 1.5mm 2.5mm 1.5mm},clip,width=\imgwidthBase]{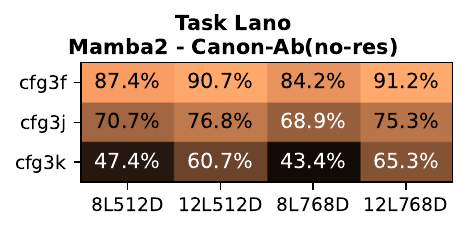}
\includegraphics[page=1,trim={2.5mm 1.5mm 2.5mm 1.5mm},clip,width=\imgwidthBase]{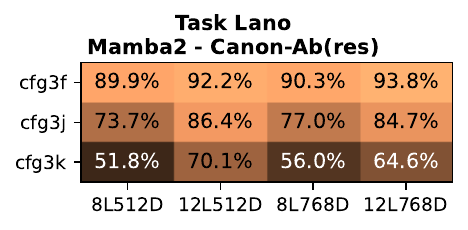}
\hspace*{-7mm}
\\
\hspace*{-7mm}
\includegraphics[page=1,trim={2.5mm 1.5mm 2.5mm 1.5mm},clip,width=\imgwidthBase]{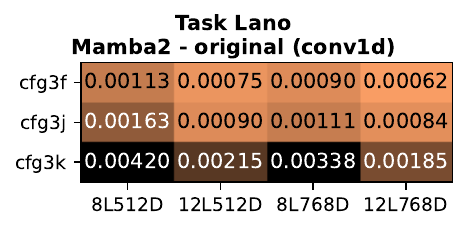}
\includegraphics[page=1,trim={2.5mm 1.5mm 2.5mm 1.5mm},clip,width=\imgwidthBase]{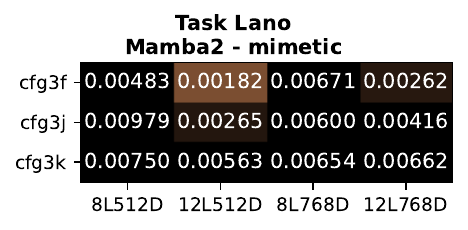}
\includegraphics[page=1,trim={2.5mm 1.5mm 2.5mm 1.5mm},clip,width=\imgwidthBase]{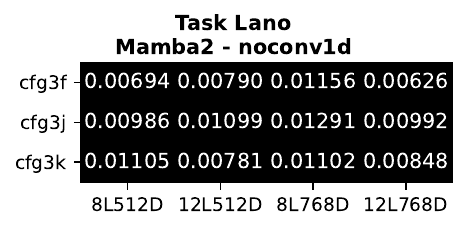}
\includegraphics[page=1,trim={2.5mm 1.5mm 2.5mm 1.5mm},clip,width=\imgwidthBase]{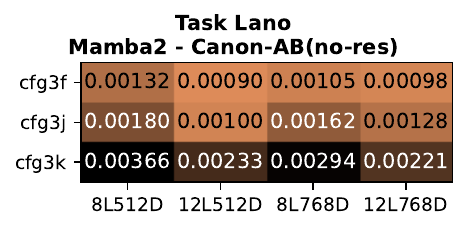}
\includegraphics[page=1,trim={2.5mm 1.5mm 2.5mm 1.5mm},clip,width=\imgwidthBase]{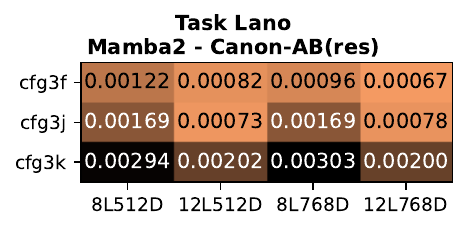}
\includegraphics[page=1,trim={2.5mm 1.5mm 2.5mm 1.5mm},clip,width=\imgwidthBase]{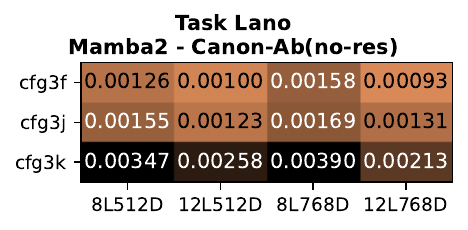}
\includegraphics[page=1,trim={2.5mm 1.5mm 2.5mm 1.5mm},clip,width=\imgwidthBase]{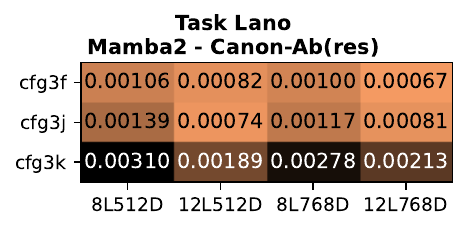}
\hspace*{-7mm}
\caption{\label{fig:full-mamba2}%
\textbf{Mamba2 variants} (left to right): original (conv1d), mimetic (w/ conv1d), no conv1d, Canon-AB(no-res), Canon-AB(res), Canon-Ab(no-res), Canon-Ab(res). }
\end{figure}

\begin{figure}[H]
\centering
\setlength{\imgwidthBase}{0.12\textwidth}
\hspace*{-7mm}
\includegraphics[page=1,trim={2.5mm 1.5mm 2.5mm 1.5mm},clip,width=\imgwidthBase]{perm_4/Mamba2_mlp_-original_conv1d_}
\includegraphics[page=1,trim={2.5mm 1.5mm 2.5mm 1.5mm},clip,width=\imgwidthBase]{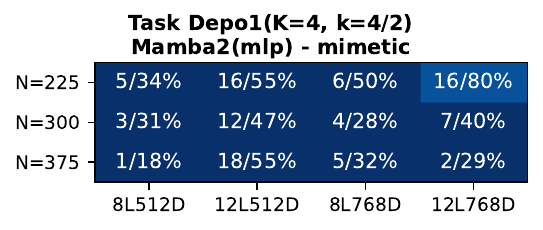}
\includegraphics[page=1,trim={2.5mm 1.5mm 2.5mm 1.5mm},clip,width=\imgwidthBase]{perm_4/Mamba2_mlp_-noconv1d}
\includegraphics[page=1,trim={2.5mm 1.5mm 2.5mm 1.5mm},clip,width=\imgwidthBase]{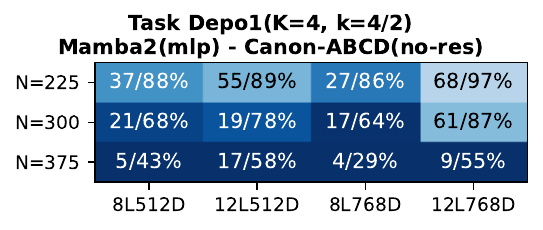}
\includegraphics[page=1,trim={2.5mm 1.5mm 2.5mm 1.5mm},clip,width=\imgwidthBase]{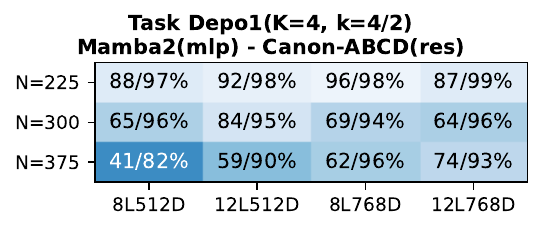}
\includegraphics[page=1,trim={2.5mm 1.5mm 2.5mm 1.5mm},clip,width=\imgwidthBase]{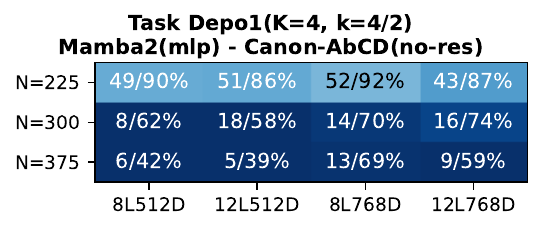}
\includegraphics[page=1,trim={2.5mm 1.5mm 2.5mm 1.5mm},clip,width=\imgwidthBase]{perm_4/Mamba2_mlp_-Res-Canon-AbbCD}
\includegraphics[page=1,trim={2.5mm 1.5mm 2.5mm 1.5mm},clip,width=\imgwidthBase]{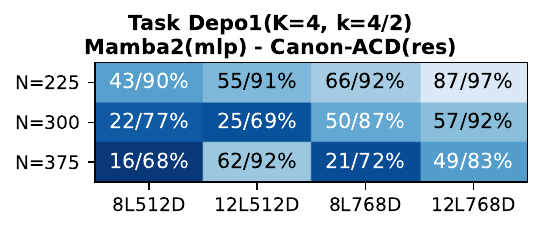}
\hspace*{-7mm}
\\
\hspace*{-7mm}
\includegraphics[page=1,trim={2.5mm 1.5mm 2.5mm 1.5mm},clip,width=\imgwidthBase]{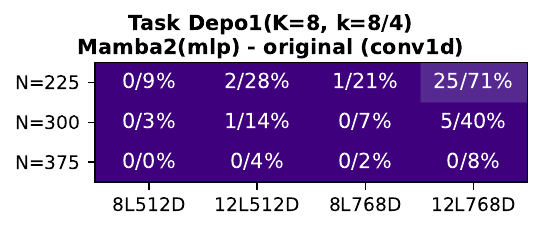}
\includegraphics[page=1,trim={2.5mm 1.5mm 2.5mm 1.5mm},clip,width=\imgwidthBase]{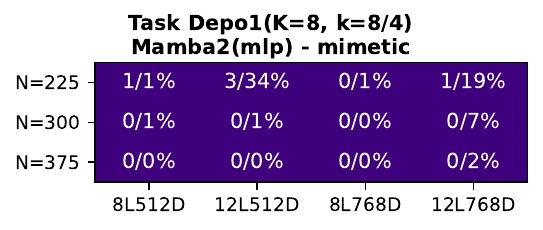}
\includegraphics[page=1,trim={2.5mm 1.5mm 2.5mm 1.5mm},clip,width=\imgwidthBase]{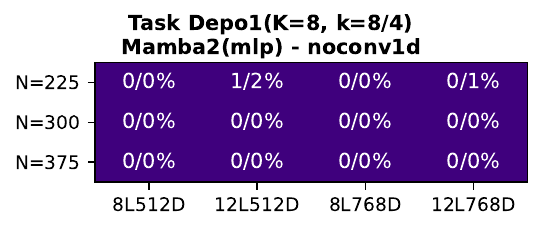}
\includegraphics[page=1,trim={2.5mm 1.5mm 2.5mm 1.5mm},clip,width=\imgwidthBase]{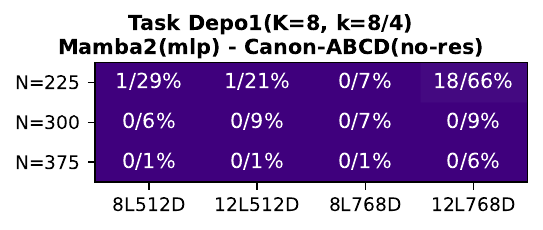}
\includegraphics[page=1,trim={2.5mm 1.5mm 2.5mm 1.5mm},clip,width=\imgwidthBase]{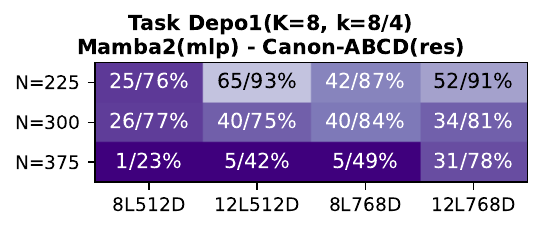}
\includegraphics[page=1,trim={2.5mm 1.5mm 2.5mm 1.5mm},clip,width=\imgwidthBase]{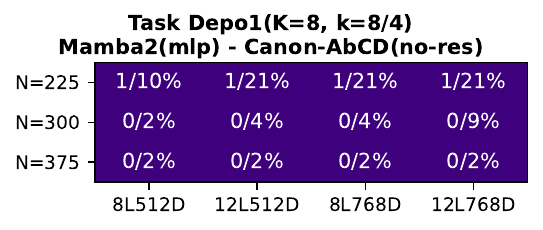}
\includegraphics[page=1,trim={2.5mm 1.5mm 2.5mm 1.5mm},clip,width=\imgwidthBase]{perm/Mamba2_mlp_-Res-Canon-AbbCD}
\includegraphics[page=1,trim={2.5mm 1.5mm 2.5mm 1.5mm},clip,width=\imgwidthBase]{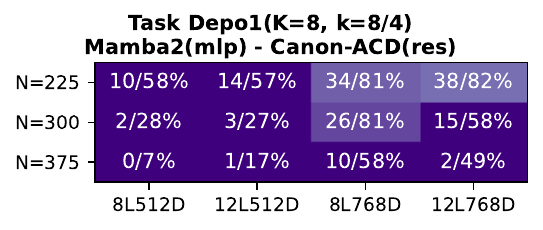}
\hspace*{-7mm}
\\
\hspace*{-7mm}
\includegraphics[page=1,trim={2.5mm 1.5mm 2.5mm 1.5mm},clip,width=\imgwidthBase]{perm_multi_4/Mamba2_mlp_-original_conv1d_}
\includegraphics[page=1,trim={2.5mm 1.5mm 2.5mm 1.5mm},clip,width=\imgwidthBase]{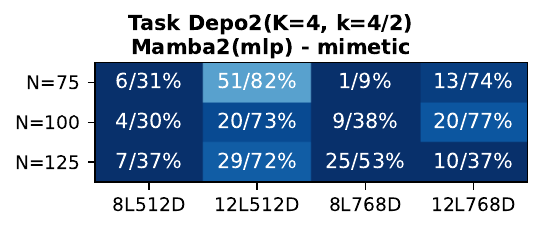}
\includegraphics[page=1,trim={2.5mm 1.5mm 2.5mm 1.5mm},clip,width=\imgwidthBase]{perm_multi_4/Mamba2_mlp_-noconv1d}
\includegraphics[page=1,trim={2.5mm 1.5mm 2.5mm 1.5mm},clip,width=\imgwidthBase]{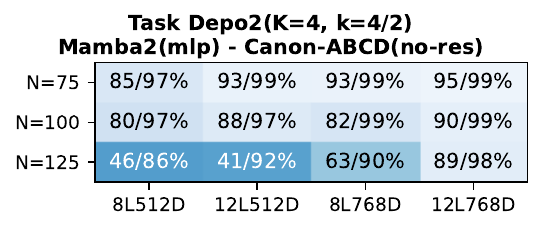}
\includegraphics[page=1,trim={2.5mm 1.5mm 2.5mm 1.5mm},clip,width=\imgwidthBase]{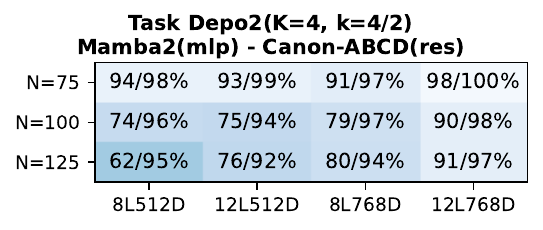}
\includegraphics[page=1,trim={2.5mm 1.5mm 2.5mm 1.5mm},clip,width=\imgwidthBase]{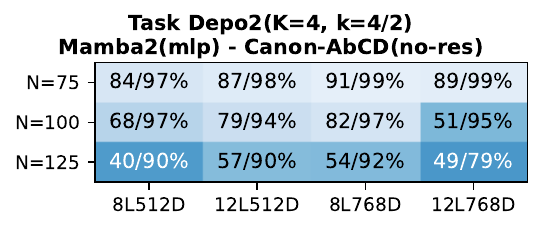}
\includegraphics[page=1,trim={2.5mm 1.5mm 2.5mm 1.5mm},clip,width=\imgwidthBase]{perm_multi_4/Mamba2_mlp_-Res-Canon-AbbCD}
\includegraphics[page=1,trim={2.5mm 1.5mm 2.5mm 1.5mm},clip,width=\imgwidthBase]{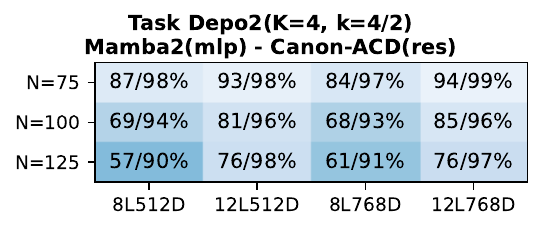}
\hspace*{-7mm}
\\
\hspace*{-7mm}
\includegraphics[page=1,trim={2.5mm 1.5mm 2.5mm 1.5mm},clip,width=\imgwidthBase]{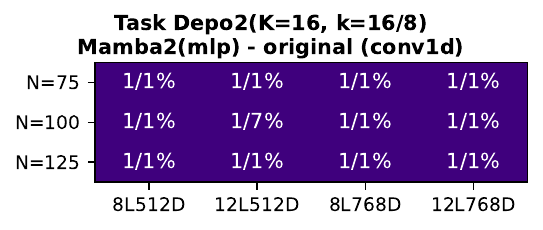}
\includegraphics[page=1,trim={2.5mm 1.5mm 2.5mm 1.5mm},clip,width=\imgwidthBase]{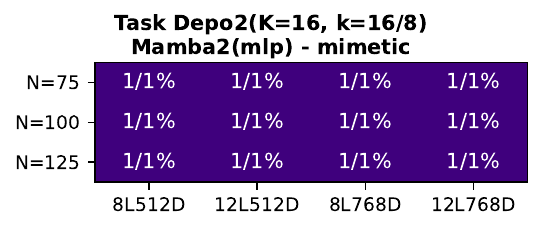}
\includegraphics[page=1,trim={2.5mm 1.5mm 2.5mm 1.5mm},clip,width=\imgwidthBase]{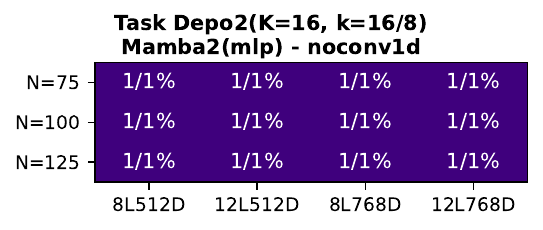}
\includegraphics[page=1,trim={2.5mm 1.5mm 2.5mm 1.5mm},clip,width=\imgwidthBase]{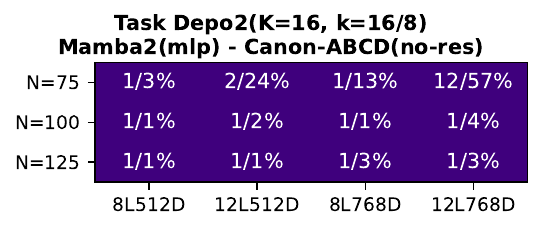}
\includegraphics[page=1,trim={2.5mm 1.5mm 2.5mm 1.5mm},clip,width=\imgwidthBase]{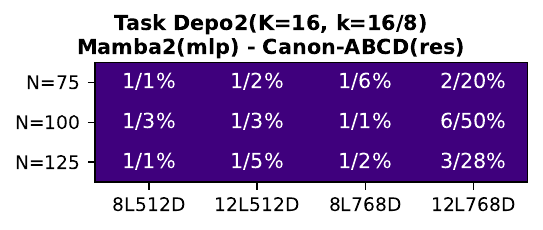}
\includegraphics[page=1,trim={2.5mm 1.5mm 2.5mm 1.5mm},clip,width=\imgwidthBase]{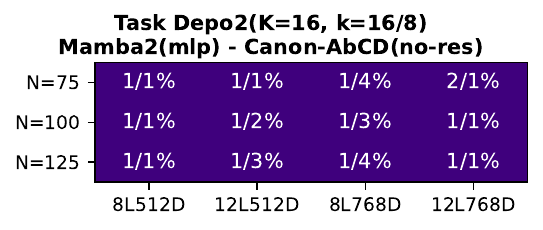}
\includegraphics[page=1,trim={2.5mm 1.5mm 2.5mm 1.5mm},clip,width=\imgwidthBase]{perm_multi/Mamba2_mlp_-Res-Canon-AbbCD}
\includegraphics[page=1,trim={2.5mm 1.5mm 2.5mm 1.5mm},clip,width=\imgwidthBase]{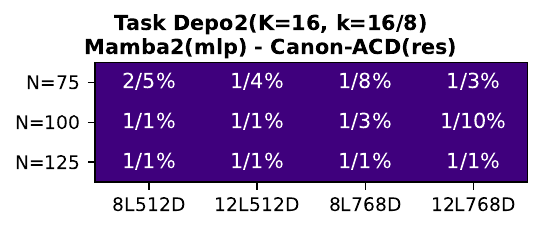}
\hspace*{-7mm}
\\
\hspace*{-7mm}
\includegraphics[page=1,trim={2.5mm 1.5mm 2.5mm 1.5mm},clip,width=\imgwidthBase]{top_sort/Mamba2_mlp_-original_conv1d_}
\includegraphics[page=1,trim={2.5mm 1.5mm 2.5mm 1.5mm},clip,width=\imgwidthBase]{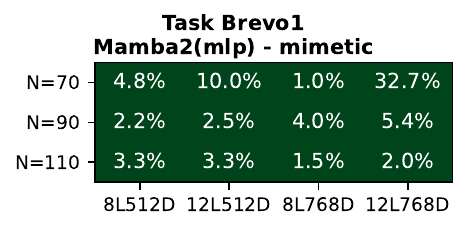}
\includegraphics[page=1,trim={2.5mm 1.5mm 2.5mm 1.5mm},clip,width=\imgwidthBase]{top_sort/Mamba2_mlp_-noconv1d}
\includegraphics[page=1,trim={2.5mm 1.5mm 2.5mm 1.5mm},clip,width=\imgwidthBase]{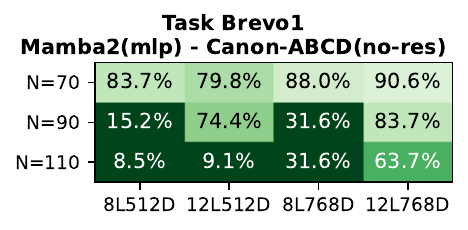}
\includegraphics[page=1,trim={2.5mm 1.5mm 2.5mm 1.5mm},clip,width=\imgwidthBase]{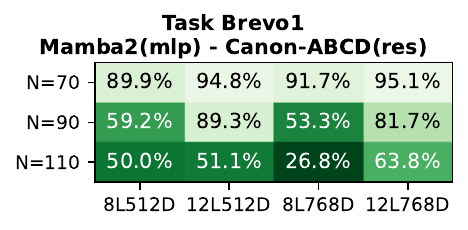}
\includegraphics[page=1,trim={2.5mm 1.5mm 2.5mm 1.5mm},clip,width=\imgwidthBase]{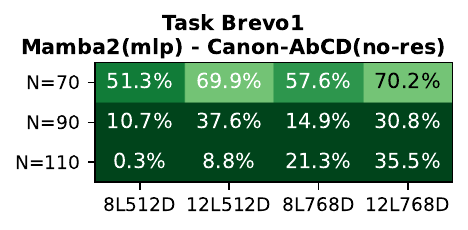}
\includegraphics[page=1,trim={2.5mm 1.5mm 2.5mm 1.5mm},clip,width=\imgwidthBase]{top_sort/Mamba2_mlp_-Res-Canon-AbbCD}
\includegraphics[page=1,trim={2.5mm 1.5mm 2.5mm 1.5mm},clip,width=\imgwidthBase]{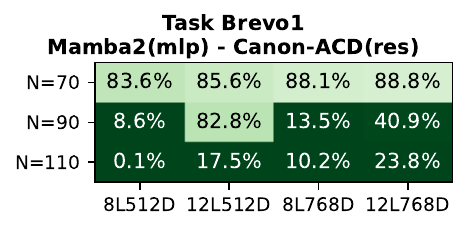}
\hspace*{-7mm}
\\
\hspace*{-7mm}
\includegraphics[page=1,trim={2.5mm 1.5mm 2.5mm 1.5mm},clip,width=\imgwidthBase]{top_sort_multi/Mamba2_mlp_-original_conv1d_}
\includegraphics[page=1,trim={2.5mm 1.5mm 2.5mm 1.5mm},clip,width=\imgwidthBase]{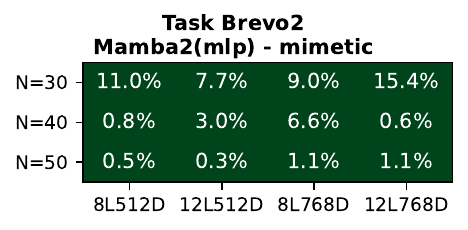}
\includegraphics[page=1,trim={2.5mm 1.5mm 2.5mm 1.5mm},clip,width=\imgwidthBase]{top_sort_multi/Mamba2_mlp_-noconv1d}
\includegraphics[page=1,trim={2.5mm 1.5mm 2.5mm 1.5mm},clip,width=\imgwidthBase]{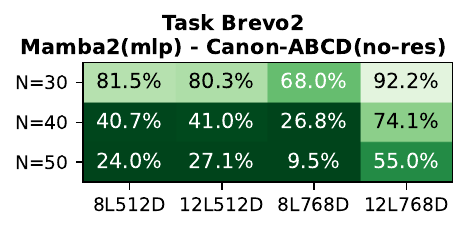}
\includegraphics[page=1,trim={2.5mm 1.5mm 2.5mm 1.5mm},clip,width=\imgwidthBase]{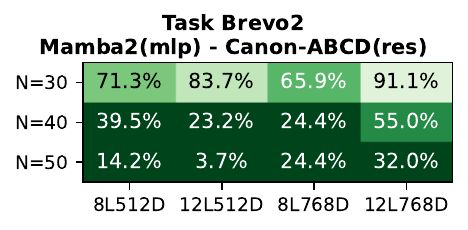}
\includegraphics[page=1,trim={2.5mm 1.5mm 2.5mm 1.5mm},clip,width=\imgwidthBase]{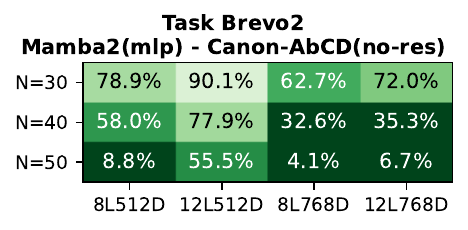}
\includegraphics[page=1,trim={2.5mm 1.5mm 2.5mm 1.5mm},clip,width=\imgwidthBase]{top_sort_multi/Mamba2_mlp_-Res-Canon-AbbCD}
\includegraphics[page=1,trim={2.5mm 1.5mm 2.5mm 1.5mm},clip,width=\imgwidthBase]{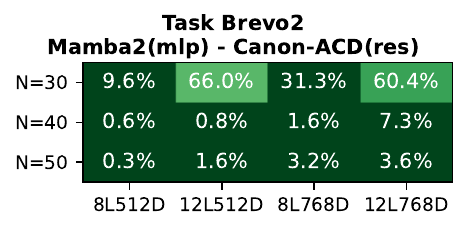}
\hspace*{-7mm}
\\
\hspace*{-7mm}
\includegraphics[page=1,trim={2.5mm 1.5mm 2.5mm 1.5mm},clip,width=\imgwidthBase]{arith/Mamba2_mlp_-original_conv1d_}
\includegraphics[page=1,trim={2.5mm 1.5mm 2.5mm 1.5mm},clip,width=\imgwidthBase]{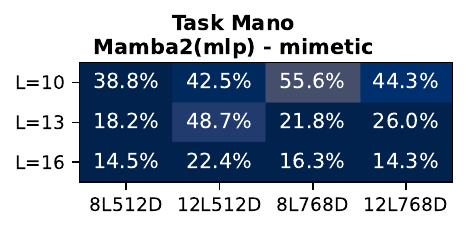}
\includegraphics[page=1,trim={2.5mm 1.5mm 2.5mm 1.5mm},clip,width=\imgwidthBase]{arith/Mamba2_mlp_-noconv1d}
\includegraphics[page=1,trim={2.5mm 1.5mm 2.5mm 1.5mm},clip,width=\imgwidthBase]{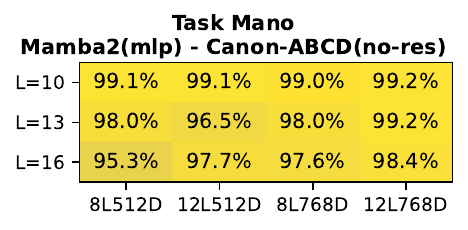}
\includegraphics[page=1,trim={2.5mm 1.5mm 2.5mm 1.5mm},clip,width=\imgwidthBase]{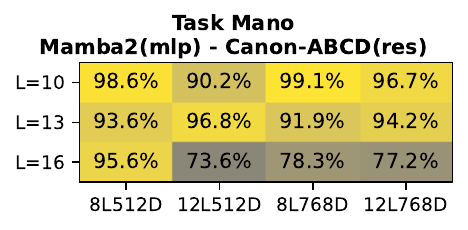}
\includegraphics[page=1,trim={2.5mm 1.5mm 2.5mm 1.5mm},clip,width=\imgwidthBase]{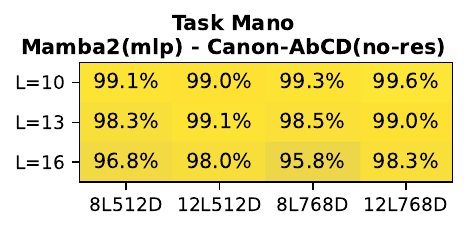}
\includegraphics[page=1,trim={2.5mm 1.5mm 2.5mm 1.5mm},clip,width=\imgwidthBase]{arith/Mamba2_mlp_-Res-Canon-AbbCD}
\includegraphics[page=1,trim={2.5mm 1.5mm 2.5mm 1.5mm},clip,width=\imgwidthBase]{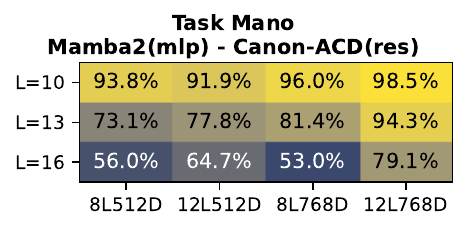}
\hspace*{-7mm}
\\
\hspace*{-7mm}
\includegraphics[page=1,trim={2.5mm 1.5mm 2.5mm 1.5mm},clip,width=\imgwidthBase]{cfg/Mamba2_mlp_-original_conv1d_}
\includegraphics[page=1,trim={2.5mm 1.5mm 2.5mm 1.5mm},clip,width=\imgwidthBase]{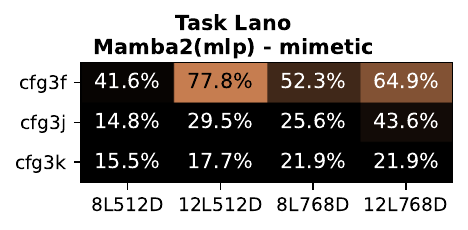}
\includegraphics[page=1,trim={2.5mm 1.5mm 2.5mm 1.5mm},clip,width=\imgwidthBase]{cfg/Mamba2_mlp_-noconv1d}
\includegraphics[page=1,trim={2.5mm 1.5mm 2.5mm 1.5mm},clip,width=\imgwidthBase]{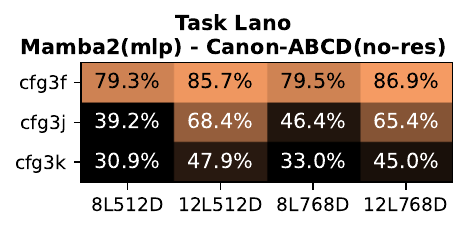}
\includegraphics[page=1,trim={2.5mm 1.5mm 2.5mm 1.5mm},clip,width=\imgwidthBase]{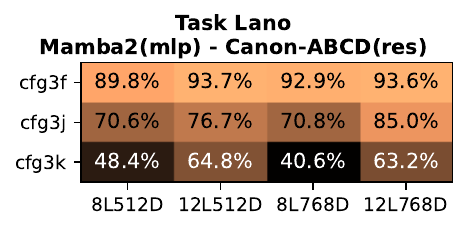}
\includegraphics[page=1,trim={2.5mm 1.5mm 2.5mm 1.5mm},clip,width=\imgwidthBase]{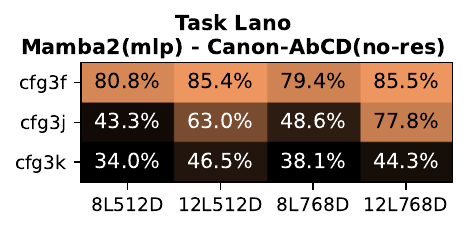}
\includegraphics[page=1,trim={2.5mm 1.5mm 2.5mm 1.5mm},clip,width=\imgwidthBase]{cfg/Mamba2_mlp_-Res-Canon-AbbCD}
\includegraphics[page=1,trim={2.5mm 1.5mm 2.5mm 1.5mm},clip,width=\imgwidthBase]{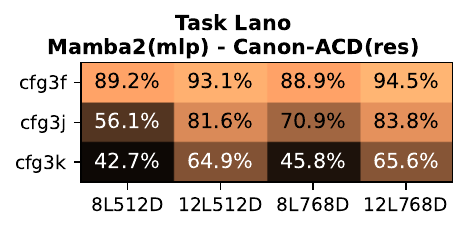}
\hspace*{-7mm}
\\
\hspace*{-7mm}
\includegraphics[page=1,trim={2.5mm 1.5mm 2.5mm 1.5mm},clip,width=\imgwidthBase]{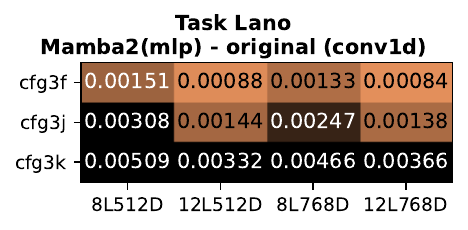}
\includegraphics[page=1,trim={2.5mm 1.5mm 2.5mm 1.5mm},clip,width=\imgwidthBase]{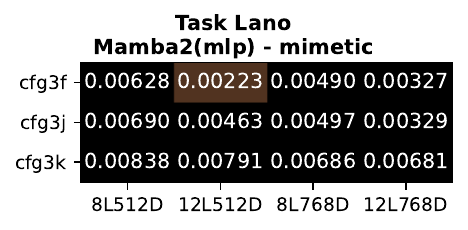}
\includegraphics[page=1,trim={2.5mm 1.5mm 2.5mm 1.5mm},clip,width=\imgwidthBase]{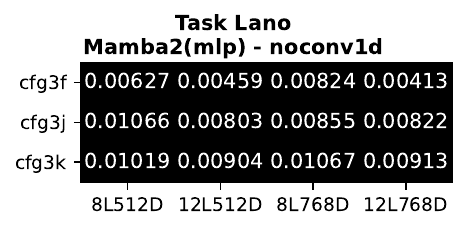}
\includegraphics[page=1,trim={2.5mm 1.5mm 2.5mm 1.5mm},clip,width=\imgwidthBase]{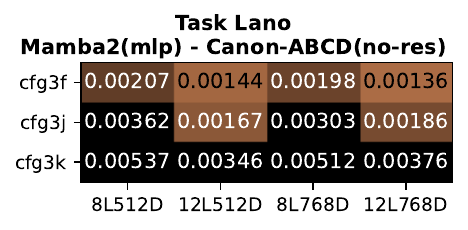}
\includegraphics[page=1,trim={2.5mm 1.5mm 2.5mm 1.5mm},clip,width=\imgwidthBase]{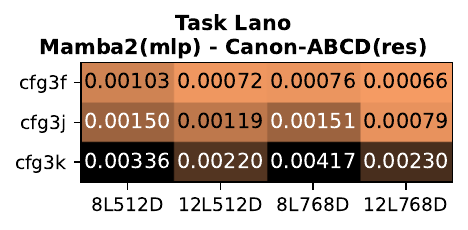}
\includegraphics[page=1,trim={2.5mm 1.5mm 2.5mm 1.5mm},clip,width=\imgwidthBase]{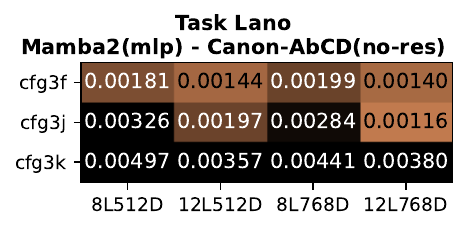}
\includegraphics[page=1,trim={2.5mm 1.5mm 2.5mm 1.5mm},clip,width=\imgwidthBase]{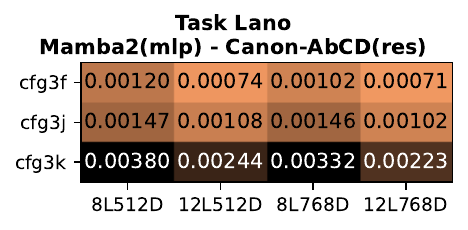}
\includegraphics[page=1,trim={2.5mm 1.5mm 2.5mm 1.5mm},clip,width=\imgwidthBase]{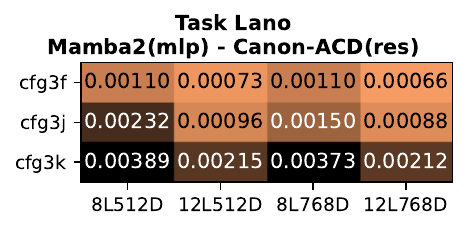}
\hspace*{-7mm}
\caption{\label{fig:full-mamba2-mlp}%
\textbf{Mamba2(mlp) variants} (left to right): original (conv1d), mimetic (w/ conv1d), no conv1d, Canon-ABCD(no-res), Canon-ABCD(res), Canon-AbCD(no-res), Canon-AbCD(res), Canon-ACD(res).}
\end{figure}

\clearpage
\subsection{GLA family}

\begin{figure}[H]
\centering
\setlength{\imgwidthBase}{0.13\textwidth}
\includegraphics[page=1,trim={2.5mm 1.5mm 2.5mm 1.5mm},clip,height=\imgwidthBase]{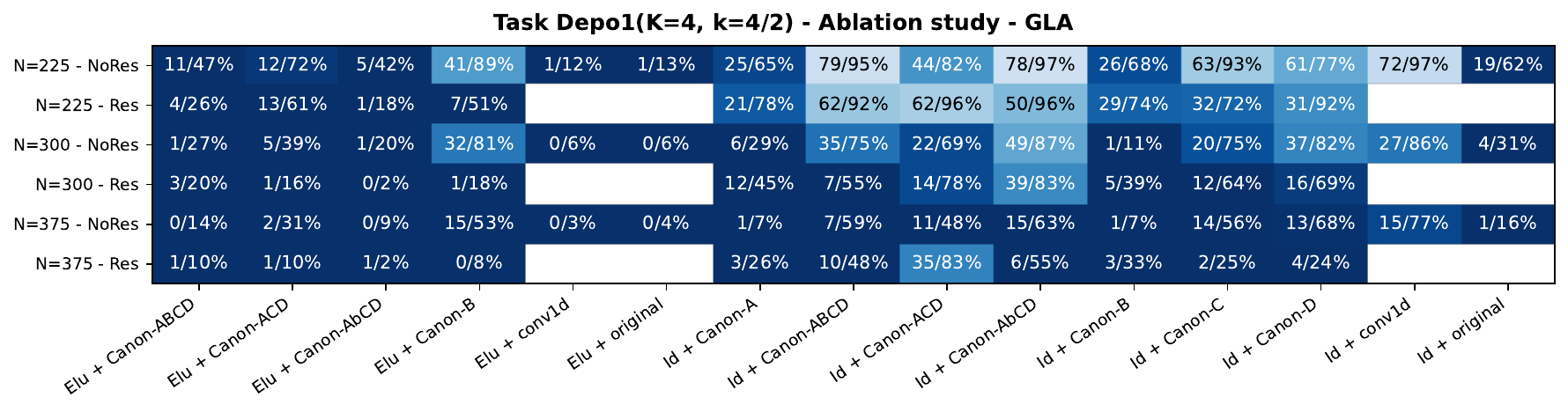}
\includegraphics[page=1,trim={2.5mm 1.5mm 2.5mm 1.5mm},clip,height=\imgwidthBase]{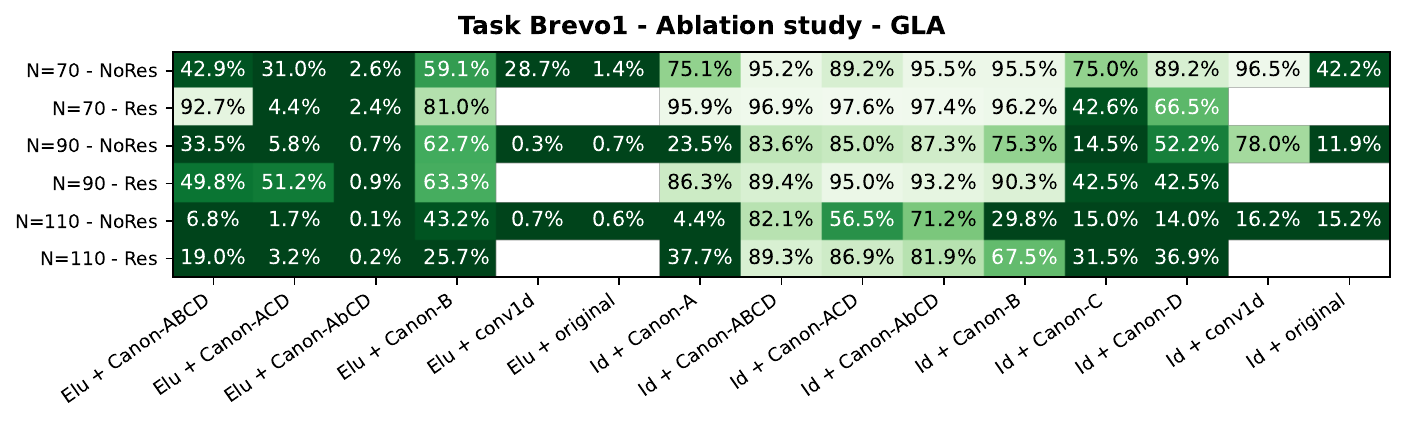}
\includegraphics[page=1,trim={2.5mm 1.5mm 2.5mm 1.5mm},clip,height=\imgwidthBase]{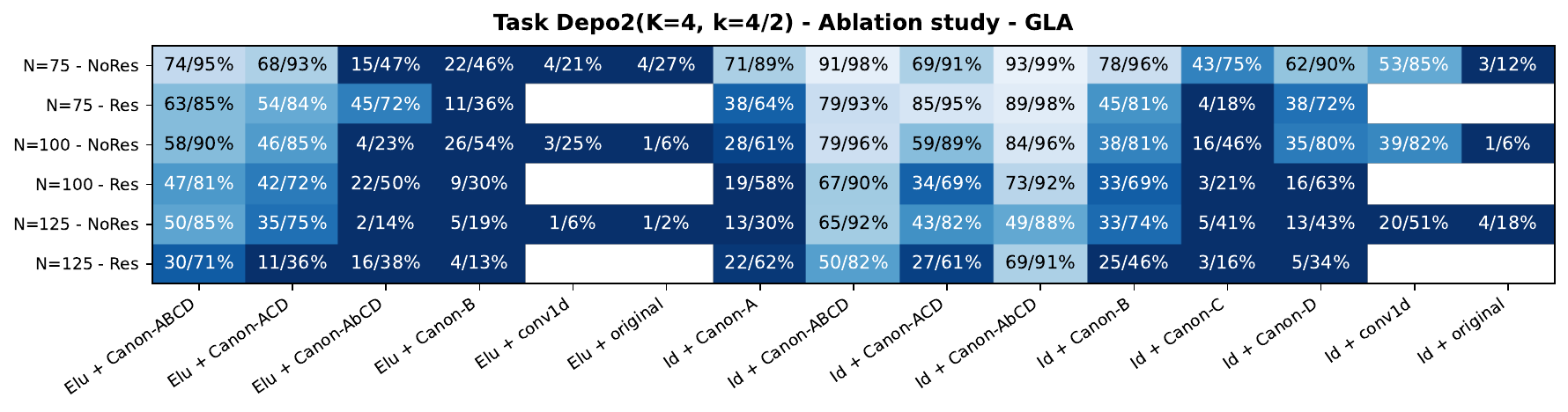}
\includegraphics[page=1,trim={2.5mm 1.5mm 2.5mm 1.5mm},clip,height=\imgwidthBase]{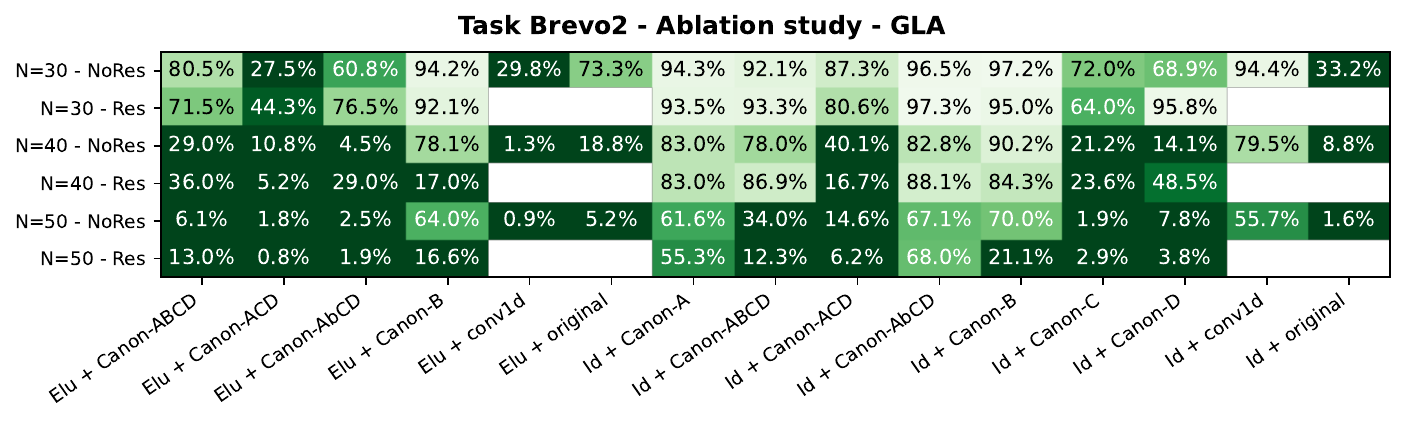}
\includegraphics[page=1,trim={2.5mm 1.5mm 2.5mm 1.5mm},clip,height=\imgwidthBase]{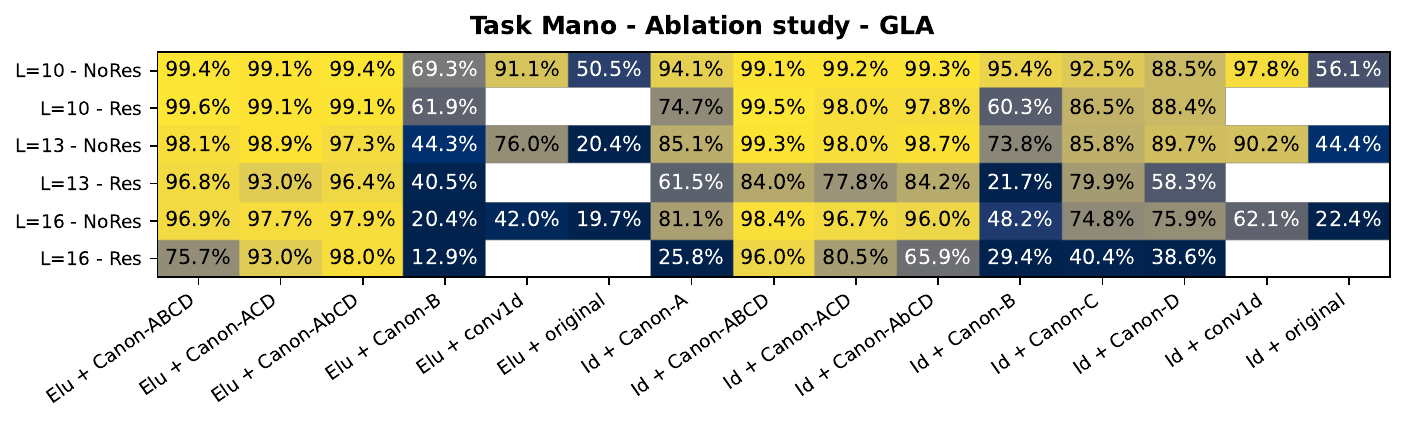}
\includegraphics[page=1,trim={2.5mm 1.5mm 2.5mm 1.5mm},clip,height=\imgwidthBase]{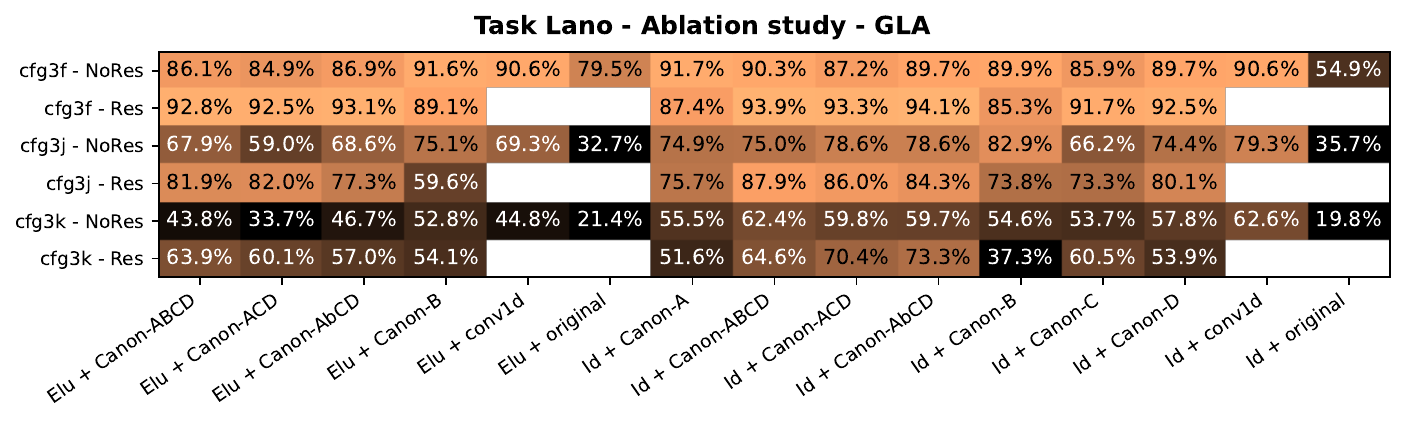}
\caption{\label{fig:gla-ablation}\textbf{Ablation study on 12L768D GLA} with Canon/conv1d layers, residual vs. non-residual, identity feature map vs.\ non-linear ($\phi(x)=\mathrm{elu}(x)+1$) feature map.}
\end{figure}

\begin{figure}[H]
\centering
\setlength{\imgwidthBase}{0.14\textwidth}
\hspace*{-7mm}
\includegraphics[page=1,trim={2.5mm 1.5mm 2.5mm 1.5mm},clip,width=\imgwidthBase]{perm_4/GLA-original}
\includegraphics[page=1,trim={2.5mm 1.5mm 2.5mm 1.5mm},clip,width=\imgwidthBase]{perm_4/GLA-conv1d}
\includegraphics[page=1,trim={2.5mm 1.5mm 2.5mm 1.5mm},clip,width=\imgwidthBase]{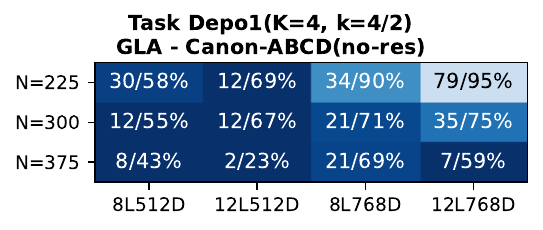}
\includegraphics[page=1,trim={2.5mm 1.5mm 2.5mm 1.5mm},clip,width=\imgwidthBase]{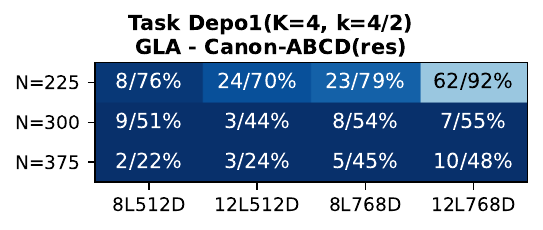}
\includegraphics[page=1,trim={2.5mm 1.5mm 2.5mm 1.5mm},clip,width=\imgwidthBase]{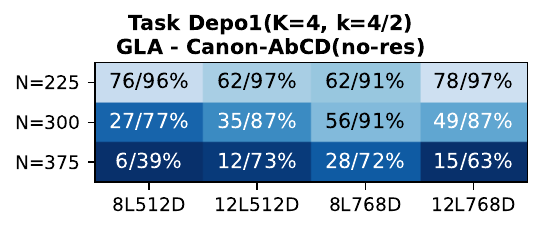}
\includegraphics[page=1,trim={2.5mm 1.5mm 2.5mm 1.5mm},clip,width=\imgwidthBase]{perm_4/GLA-Res-Canon-AbbCD}
\includegraphics[page=1,trim={2.5mm 1.5mm 2.5mm 1.5mm},clip,width=\imgwidthBase]{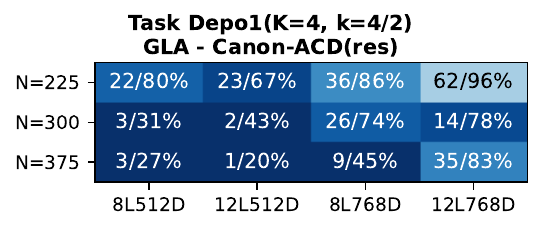}
\hspace*{-7mm}
\\
\hspace*{-7mm}
\includegraphics[page=1,trim={2.5mm 1.5mm 2.5mm 1.5mm},clip,width=\imgwidthBase]{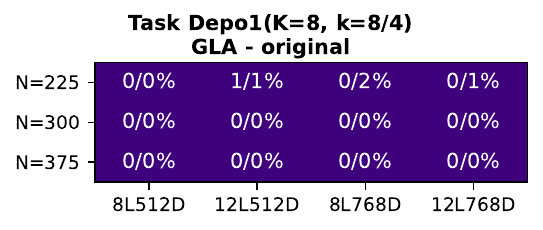}
\includegraphics[page=1,trim={2.5mm 1.5mm 2.5mm 1.5mm},clip,width=\imgwidthBase]{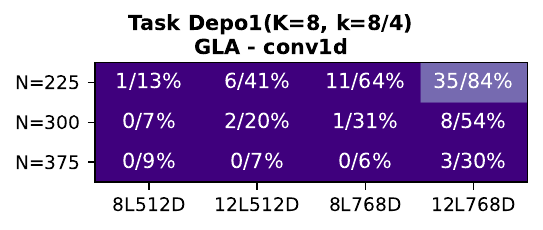}
\includegraphics[page=1,trim={2.5mm 1.5mm 2.5mm 1.5mm},clip,width=\imgwidthBase]{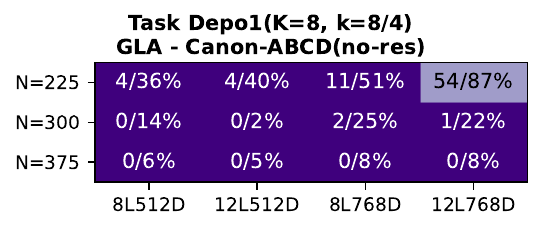}
\includegraphics[page=1,trim={2.5mm 1.5mm 2.5mm 1.5mm},clip,width=\imgwidthBase]{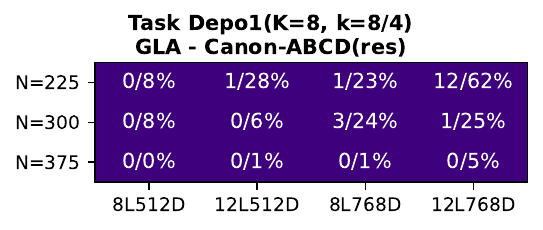}
\includegraphics[page=1,trim={2.5mm 1.5mm 2.5mm 1.5mm},clip,width=\imgwidthBase]{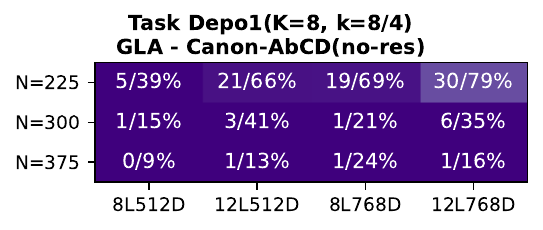}
\includegraphics[page=1,trim={2.5mm 1.5mm 2.5mm 1.5mm},clip,width=\imgwidthBase]{perm/GLA-Res-Canon-AbbCD}
\includegraphics[page=1,trim={2.5mm 1.5mm 2.5mm 1.5mm},clip,width=\imgwidthBase]{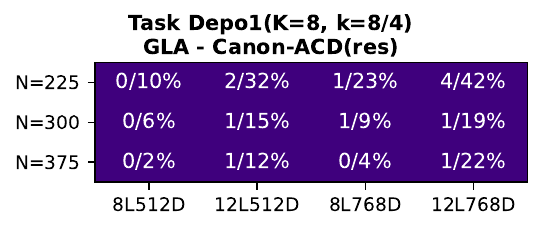}
\hspace*{-7mm}
\\
\hspace*{-7mm}
\includegraphics[page=1,trim={2.5mm 1.5mm 2.5mm 1.5mm},clip,width=\imgwidthBase]{perm_multi_4/GLA-original}
\includegraphics[page=1,trim={2.5mm 1.5mm 2.5mm 1.5mm},clip,width=\imgwidthBase]{perm_multi_4/GLA-conv1d}
\includegraphics[page=1,trim={2.5mm 1.5mm 2.5mm 1.5mm},clip,width=\imgwidthBase]{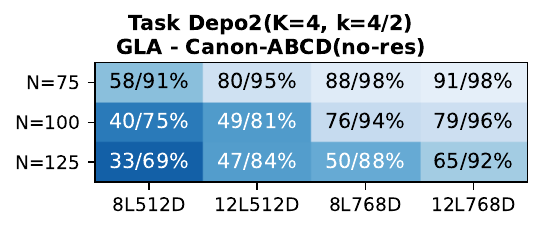}
\includegraphics[page=1,trim={2.5mm 1.5mm 2.5mm 1.5mm},clip,width=\imgwidthBase]{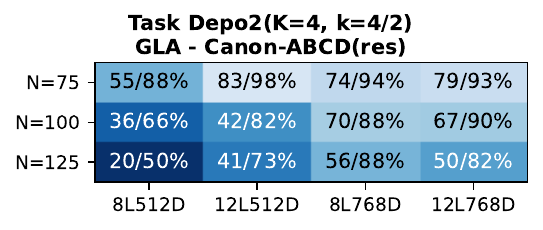}
\includegraphics[page=1,trim={2.5mm 1.5mm 2.5mm 1.5mm},clip,width=\imgwidthBase]{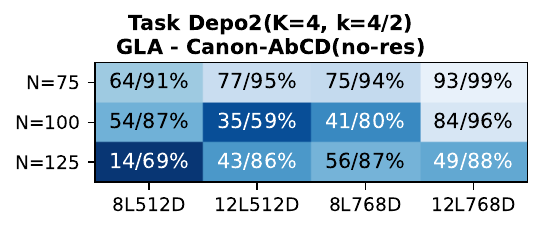}
\includegraphics[page=1,trim={2.5mm 1.5mm 2.5mm 1.5mm},clip,width=\imgwidthBase]{perm_multi_4/GLA-Res-Canon-AbbCD}
\includegraphics[page=1,trim={2.5mm 1.5mm 2.5mm 1.5mm},clip,width=\imgwidthBase]{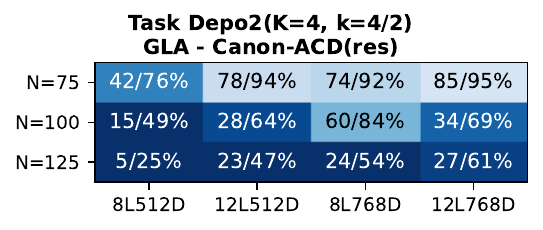}
\hspace*{-7mm}
\\
\hspace*{-7mm}
\includegraphics[page=1,trim={2.5mm 1.5mm 2.5mm 1.5mm},clip,width=\imgwidthBase]{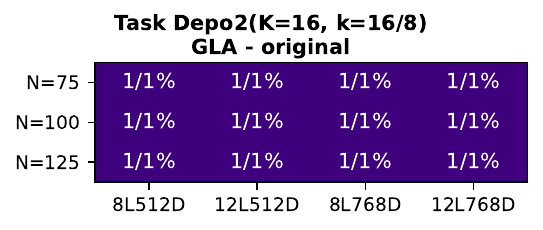}
\includegraphics[page=1,trim={2.5mm 1.5mm 2.5mm 1.5mm},clip,width=\imgwidthBase]{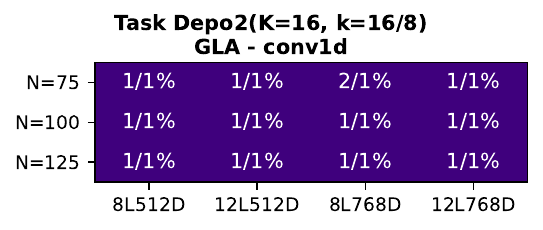}
\includegraphics[page=1,trim={2.5mm 1.5mm 2.5mm 1.5mm},clip,width=\imgwidthBase]{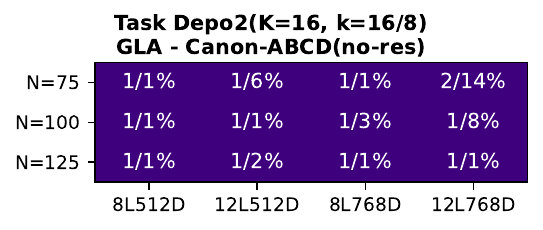}
\includegraphics[page=1,trim={2.5mm 1.5mm 2.5mm 1.5mm},clip,width=\imgwidthBase]{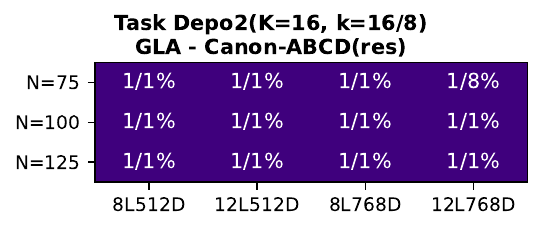}
\includegraphics[page=1,trim={2.5mm 1.5mm 2.5mm 1.5mm},clip,width=\imgwidthBase]{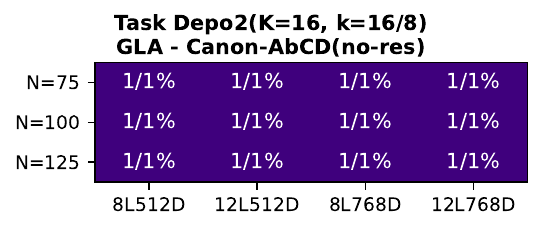}
\includegraphics[page=1,trim={2.5mm 1.5mm 2.5mm 1.5mm},clip,width=\imgwidthBase]{perm_multi/GLA-Res-Canon-AbbCD}
\includegraphics[page=1,trim={2.5mm 1.5mm 2.5mm 1.5mm},clip,width=\imgwidthBase]{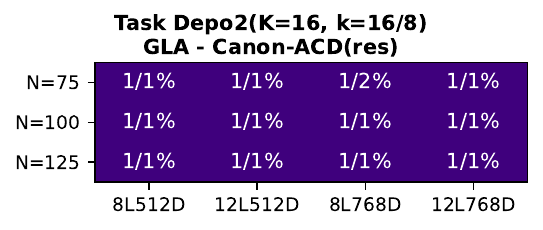}
\hspace*{-7mm}
\\
\hspace*{-7mm}
\includegraphics[page=1,trim={2.5mm 1.5mm 2.5mm 1.5mm},clip,width=\imgwidthBase]{top_sort/GLA-original}
\includegraphics[page=1,trim={2.5mm 1.5mm 2.5mm 1.5mm},clip,width=\imgwidthBase]{top_sort/GLA-conv1d}
\includegraphics[page=1,trim={2.5mm 1.5mm 2.5mm 1.5mm},clip,width=\imgwidthBase]{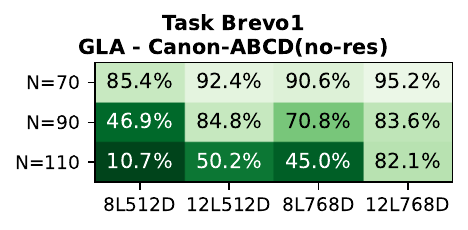}
\includegraphics[page=1,trim={2.5mm 1.5mm 2.5mm 1.5mm},clip,width=\imgwidthBase]{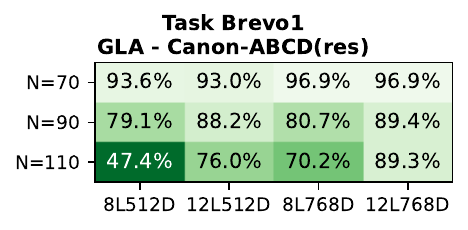}
\includegraphics[page=1,trim={2.5mm 1.5mm 2.5mm 1.5mm},clip,width=\imgwidthBase]{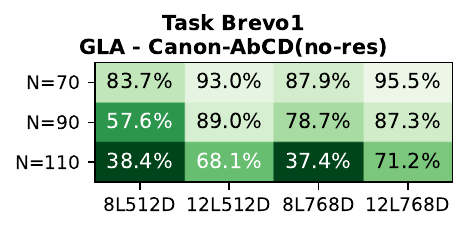}
\includegraphics[page=1,trim={2.5mm 1.5mm 2.5mm 1.5mm},clip,width=\imgwidthBase]{top_sort/GLA-Res-Canon-AbbCD}
\includegraphics[page=1,trim={2.5mm 1.5mm 2.5mm 1.5mm},clip,width=\imgwidthBase]{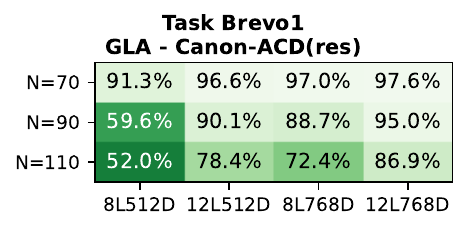}
\hspace*{-7mm}
\\
\hspace*{-7mm}
\includegraphics[page=1,trim={2.5mm 1.5mm 2.5mm 1.5mm},clip,width=\imgwidthBase]{top_sort_multi/GLA-original}
\includegraphics[page=1,trim={2.5mm 1.5mm 2.5mm 1.5mm},clip,width=\imgwidthBase]{top_sort_multi/GLA-conv1d}
\includegraphics[page=1,trim={2.5mm 1.5mm 2.5mm 1.5mm},clip,width=\imgwidthBase]{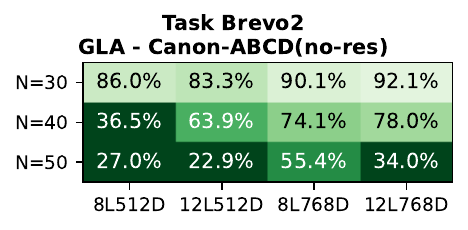}
\includegraphics[page=1,trim={2.5mm 1.5mm 2.5mm 1.5mm},clip,width=\imgwidthBase]{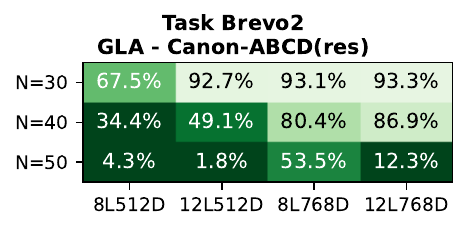}
\includegraphics[page=1,trim={2.5mm 1.5mm 2.5mm 1.5mm},clip,width=\imgwidthBase]{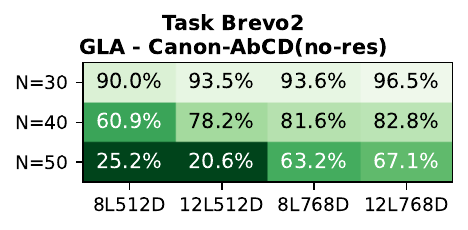}
\includegraphics[page=1,trim={2.5mm 1.5mm 2.5mm 1.5mm},clip,width=\imgwidthBase]{top_sort_multi/GLA-Res-Canon-AbbCD}
\includegraphics[page=1,trim={2.5mm 1.5mm 2.5mm 1.5mm},clip,width=\imgwidthBase]{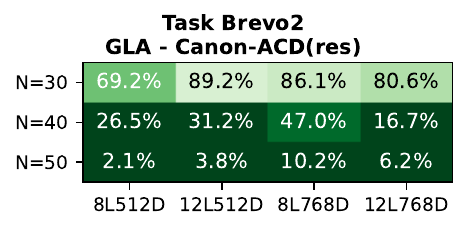}
\hspace*{-7mm}
\\
\hspace*{-7mm}
\includegraphics[page=1,trim={2.5mm 1.5mm 2.5mm 1.5mm},clip,width=\imgwidthBase]{arith/GLA-original}
\includegraphics[page=1,trim={2.5mm 1.5mm 2.5mm 1.5mm},clip,width=\imgwidthBase]{arith/GLA-conv1d}
\includegraphics[page=1,trim={2.5mm 1.5mm 2.5mm 1.5mm},clip,width=\imgwidthBase]{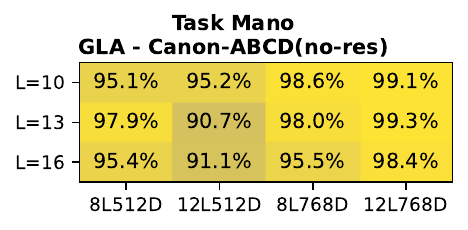}
\includegraphics[page=1,trim={2.5mm 1.5mm 2.5mm 1.5mm},clip,width=\imgwidthBase]{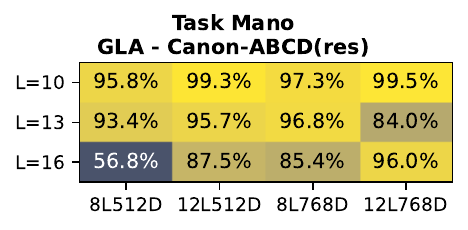}
\includegraphics[page=1,trim={2.5mm 1.5mm 2.5mm 1.5mm},clip,width=\imgwidthBase]{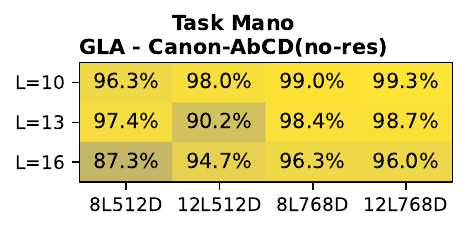}
\includegraphics[page=1,trim={2.5mm 1.5mm 2.5mm 1.5mm},clip,width=\imgwidthBase]{arith/GLA-Res-Canon-AbbCD}
\includegraphics[page=1,trim={2.5mm 1.5mm 2.5mm 1.5mm},clip,width=\imgwidthBase]{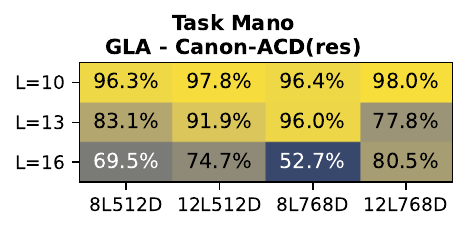}
\hspace*{-7mm}
\\
\hspace*{-7mm}
\includegraphics[page=1,trim={2.5mm 1.5mm 2.5mm 1.5mm},clip,width=\imgwidthBase]{cfg/GLA-original}
\includegraphics[page=1,trim={2.5mm 1.5mm 2.5mm 1.5mm},clip,width=\imgwidthBase]{cfg/GLA-conv1d}
\includegraphics[page=1,trim={2.5mm 1.5mm 2.5mm 1.5mm},clip,width=\imgwidthBase]{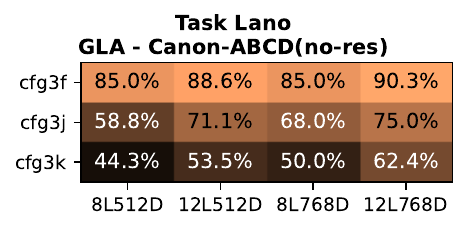}
\includegraphics[page=1,trim={2.5mm 1.5mm 2.5mm 1.5mm},clip,width=\imgwidthBase]{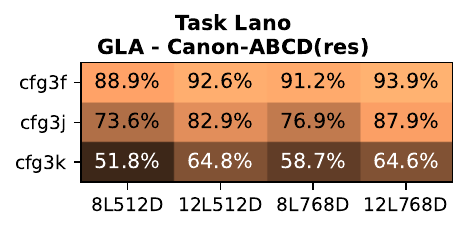}
\includegraphics[page=1,trim={2.5mm 1.5mm 2.5mm 1.5mm},clip,width=\imgwidthBase]{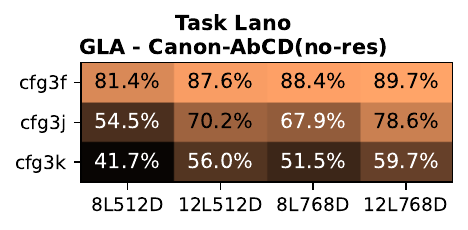}
\includegraphics[page=1,trim={2.5mm 1.5mm 2.5mm 1.5mm},clip,width=\imgwidthBase]{cfg/GLA-Res-Canon-AbbCD}
\includegraphics[page=1,trim={2.5mm 1.5mm 2.5mm 1.5mm},clip,width=\imgwidthBase]{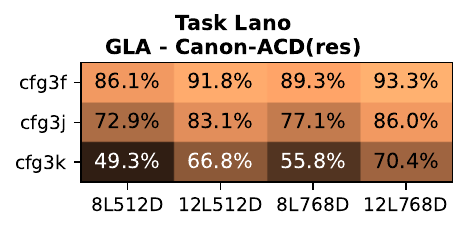}
\hspace*{-7mm}
\\
\hspace*{-7mm}
\includegraphics[page=1,trim={2.5mm 1.5mm 2.5mm 1.5mm},clip,width=\imgwidthBase]{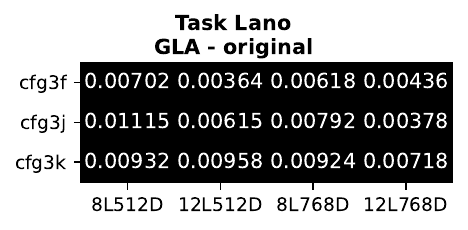}
\includegraphics[page=1,trim={2.5mm 1.5mm 2.5mm 1.5mm},clip,width=\imgwidthBase]{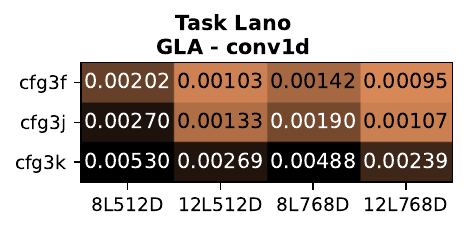}
\includegraphics[page=1,trim={2.5mm 1.5mm 2.5mm 1.5mm},clip,width=\imgwidthBase]{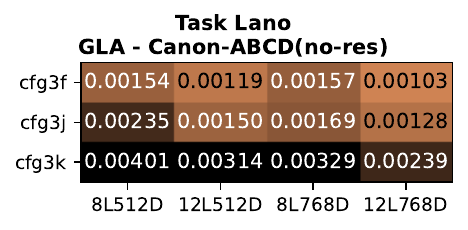}
\includegraphics[page=1,trim={2.5mm 1.5mm 2.5mm 1.5mm},clip,width=\imgwidthBase]{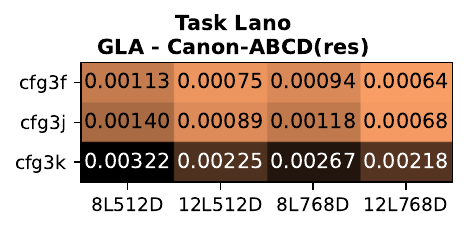}
\includegraphics[page=1,trim={2.5mm 1.5mm 2.5mm 1.5mm},clip,width=\imgwidthBase]{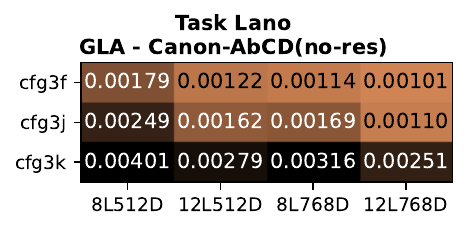}
\includegraphics[page=1,trim={2.5mm 1.5mm 2.5mm 1.5mm},clip,width=\imgwidthBase]{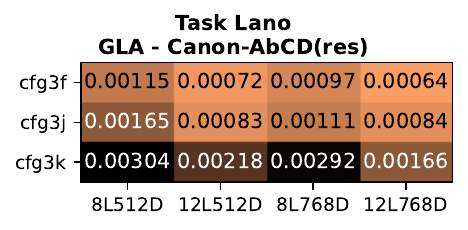}
\includegraphics[page=1,trim={2.5mm 1.5mm 2.5mm 1.5mm},clip,width=\imgwidthBase]{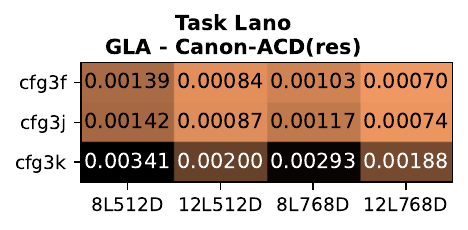}
\hspace*{-7mm}
\caption{\label{fig:full-gla}%
\textbf{GLA variants} (left to right): original, original + conv1d, original + Canon-ABCD(no-res), Canon-ABCD(res), Canon-AbCD(no-res), Canon-AbCD(res), Canon-ACD(res).}
\end{figure}

\subsection{GDN family}

\begin{figure}[H]
\centering
\setlength{\imgwidthBase}{0.17\textwidth}
\includegraphics[page=1,trim={2.5mm 1.5mm 2.5mm 1.5mm},clip,height=\imgwidthBase]{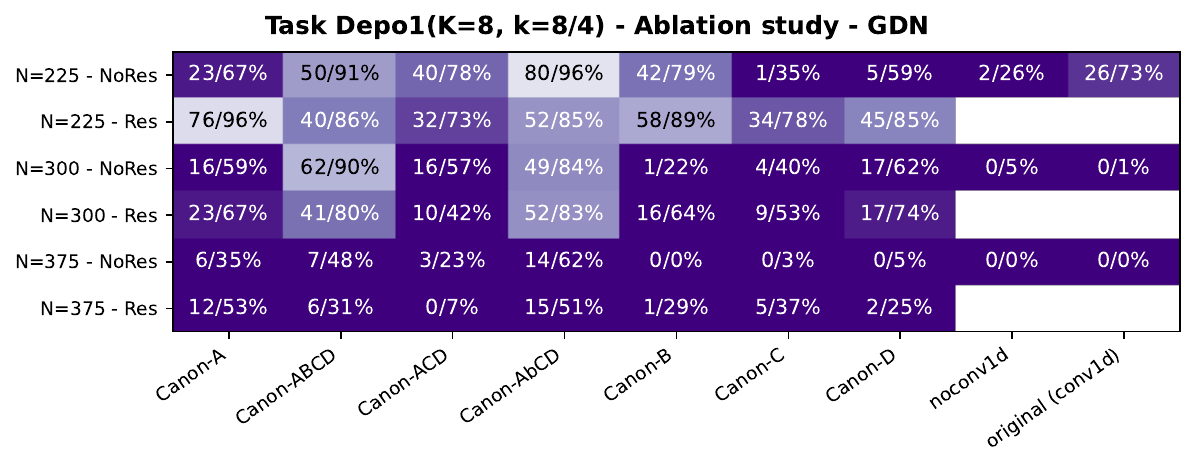}
\includegraphics[page=1,trim={2.5mm 1.5mm 2.5mm 1.5mm},clip,height=\imgwidthBase]{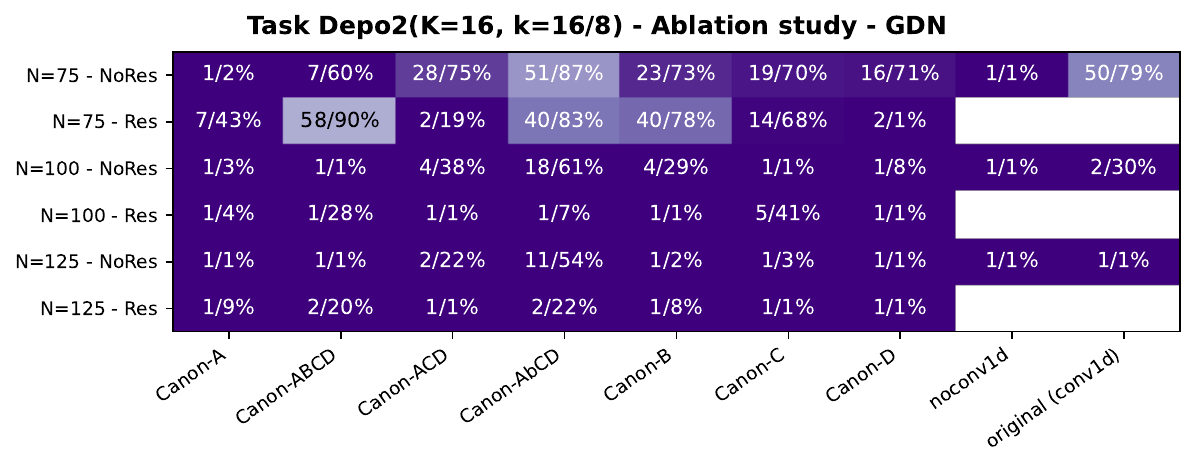}
\includegraphics[page=1,trim={2.5mm 1.5mm 2.5mm 1.5mm},clip,height=\imgwidthBase]{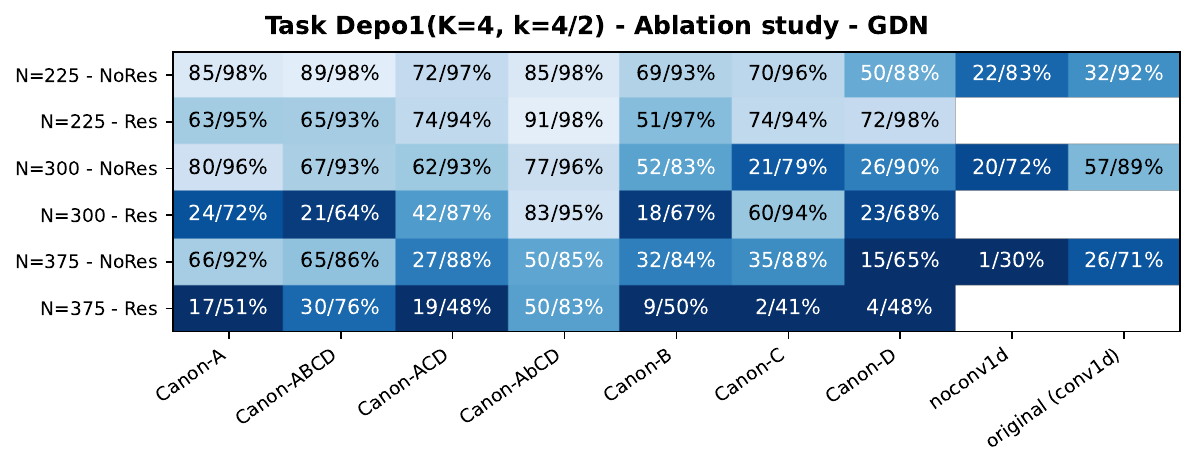}
\includegraphics[page=1,trim={2.5mm 1.5mm 2.5mm 1.5mm},clip,height=\imgwidthBase]{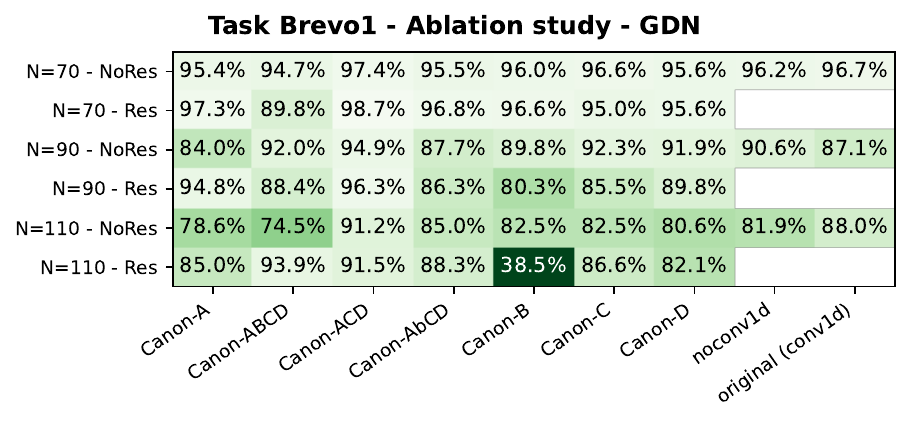}
\includegraphics[page=1,trim={2.5mm 1.5mm 2.5mm 1.5mm},clip,height=\imgwidthBase]{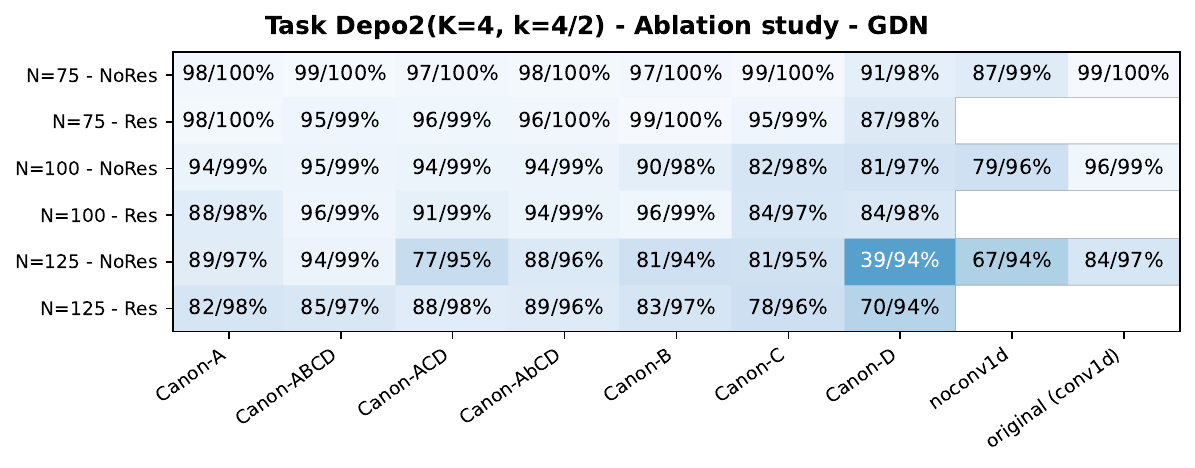}
\includegraphics[page=1,trim={2.5mm 1.5mm 2.5mm 1.5mm},clip,height=\imgwidthBase]{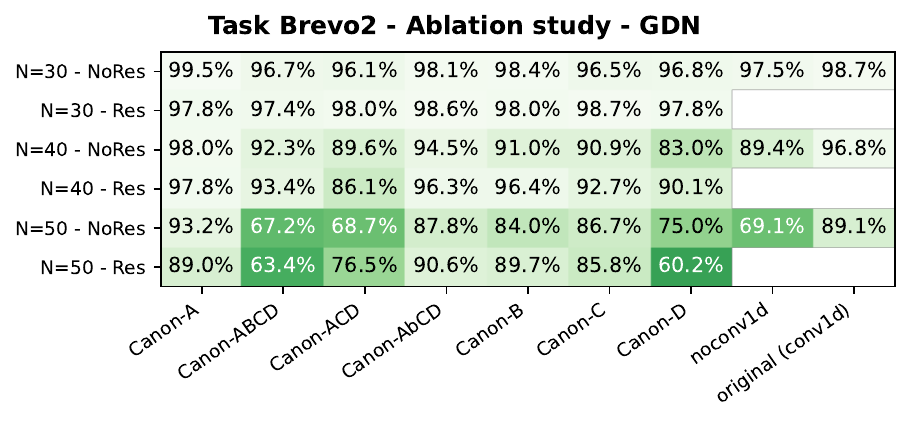}
\includegraphics[page=1,trim={2.5mm 1.5mm 2.5mm 1.5mm},clip,height=\imgwidthBase]{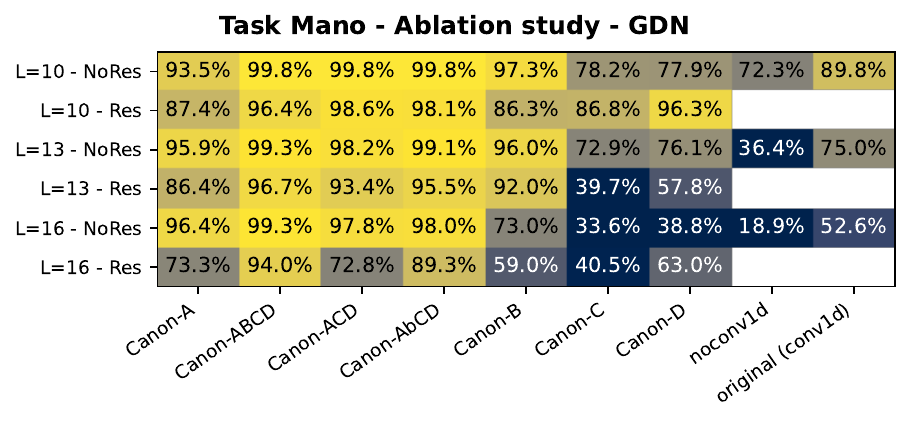}
\includegraphics[page=1,trim={2.5mm 1.5mm 2.5mm 1.5mm},clip,height=\imgwidthBase]{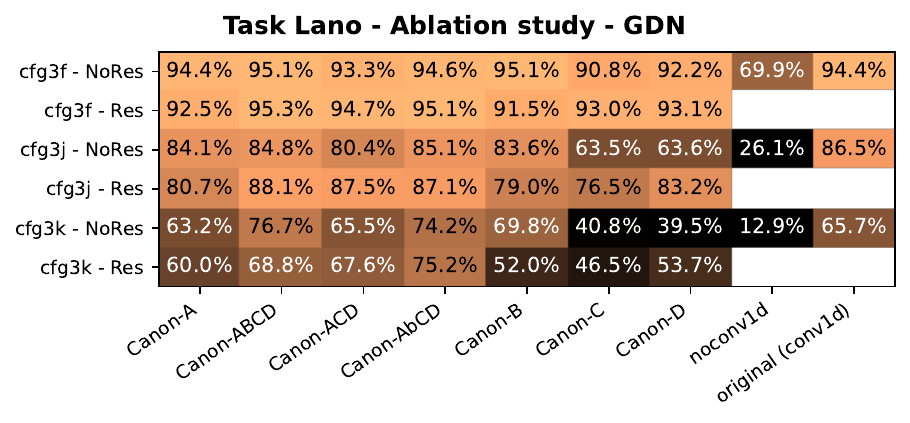}
\caption{\label{fig:gdn-ablation}\textbf{Ablation study on 12L768D GDN} with Canon/conv1d layers, residual vs. non-residual.}
\end{figure}

\begin{figure}[H]
\centering
\setlength{\imgwidthBase}{0.14\textwidth}
\hspace*{-7mm}
\includegraphics[page=1,trim={2.5mm 1.5mm 2.5mm 1.5mm},clip,width=\imgwidthBase]{perm_4/GDN-noconv1d}
\includegraphics[page=1,trim={2.5mm 1.5mm 2.5mm 1.5mm},clip,width=\imgwidthBase]{perm_4/GDN-original_conv1d_}
\includegraphics[page=1,trim={2.5mm 1.5mm 2.5mm 1.5mm},clip,width=\imgwidthBase]{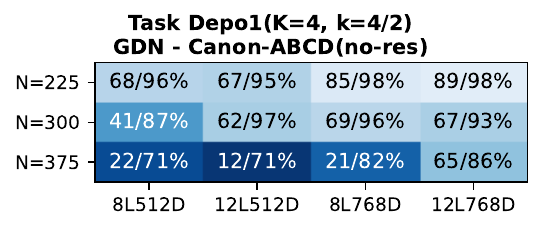}
\includegraphics[page=1,trim={2.5mm 1.5mm 2.5mm 1.5mm},clip,width=\imgwidthBase]{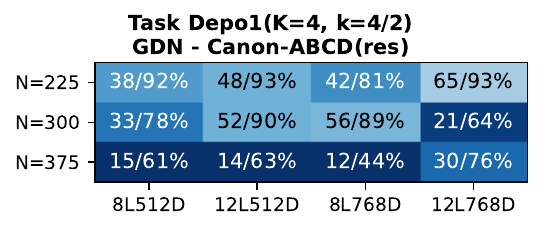}
\includegraphics[page=1,trim={2.5mm 1.5mm 2.5mm 1.5mm},clip,width=\imgwidthBase]{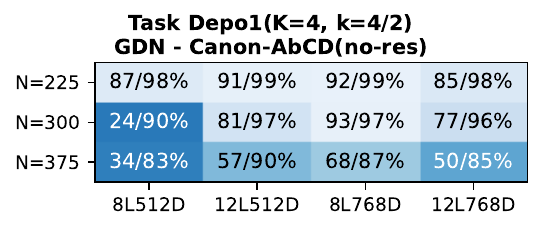}
\includegraphics[page=1,trim={2.5mm 1.5mm 2.5mm 1.5mm},clip,width=\imgwidthBase]{perm_4/GDN-Res-Canon-AbbCD}
\includegraphics[page=1,trim={2.5mm 1.5mm 2.5mm 1.5mm},clip,width=\imgwidthBase]{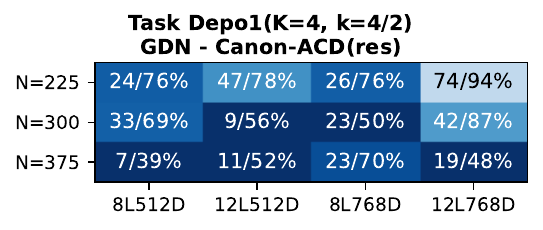}
\hspace*{-7mm}
\\
\hspace*{-7mm}
\includegraphics[page=1,trim={2.5mm 1.5mm 2.5mm 1.5mm},clip,width=\imgwidthBase]{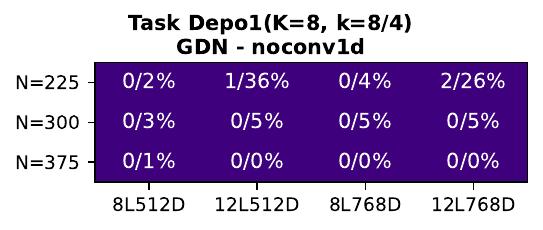}
\includegraphics[page=1,trim={2.5mm 1.5mm 2.5mm 1.5mm},clip,width=\imgwidthBase]{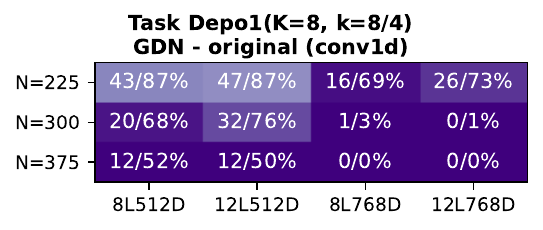}
\includegraphics[page=1,trim={2.5mm 1.5mm 2.5mm 1.5mm},clip,width=\imgwidthBase]{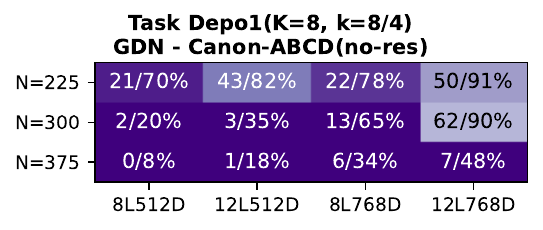}
\includegraphics[page=1,trim={2.5mm 1.5mm 2.5mm 1.5mm},clip,width=\imgwidthBase]{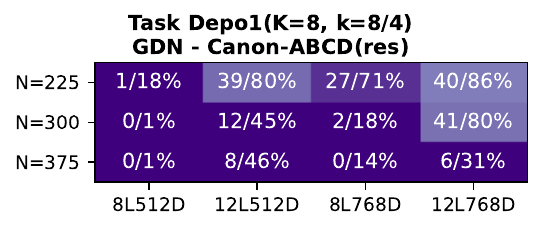}
\includegraphics[page=1,trim={2.5mm 1.5mm 2.5mm 1.5mm},clip,width=\imgwidthBase]{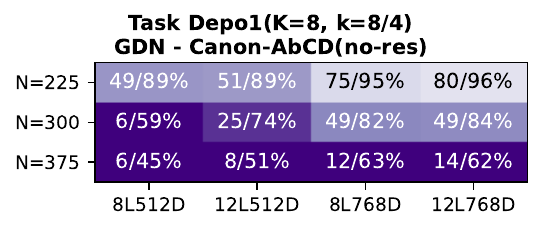}
\includegraphics[page=1,trim={2.5mm 1.5mm 2.5mm 1.5mm},clip,width=\imgwidthBase]{perm/GDN-Res-Canon-AbbCD}
\includegraphics[page=1,trim={2.5mm 1.5mm 2.5mm 1.5mm},clip,width=\imgwidthBase]{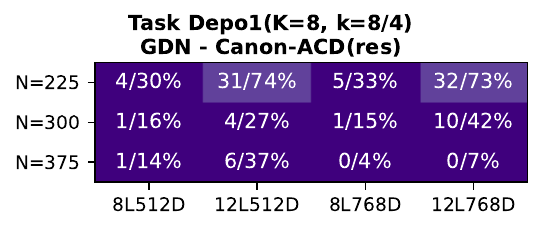}
\hspace*{-7mm}
\\
\hspace*{-7mm}
\includegraphics[page=1,trim={2.5mm 1.5mm 2.5mm 1.5mm},clip,width=\imgwidthBase]{perm_multi_4/GDN-noconv1d}
\includegraphics[page=1,trim={2.5mm 1.5mm 2.5mm 1.5mm},clip,width=\imgwidthBase]{perm_multi_4/GDN-original_conv1d_}
\includegraphics[page=1,trim={2.5mm 1.5mm 2.5mm 1.5mm},clip,width=\imgwidthBase]{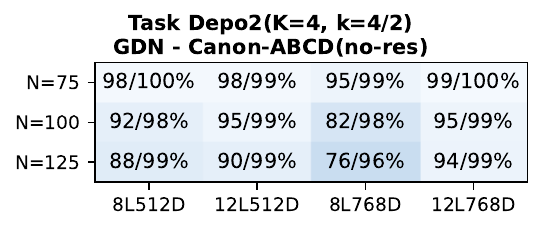}
\includegraphics[page=1,trim={2.5mm 1.5mm 2.5mm 1.5mm},clip,width=\imgwidthBase]{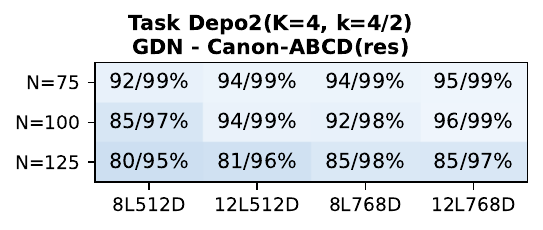}
\includegraphics[page=1,trim={2.5mm 1.5mm 2.5mm 1.5mm},clip,width=\imgwidthBase]{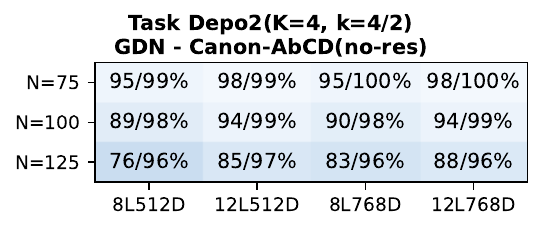}
\includegraphics[page=1,trim={2.5mm 1.5mm 2.5mm 1.5mm},clip,width=\imgwidthBase]{perm_multi_4/GDN-Res-Canon-AbbCD}
\includegraphics[page=1,trim={2.5mm 1.5mm 2.5mm 1.5mm},clip,width=\imgwidthBase]{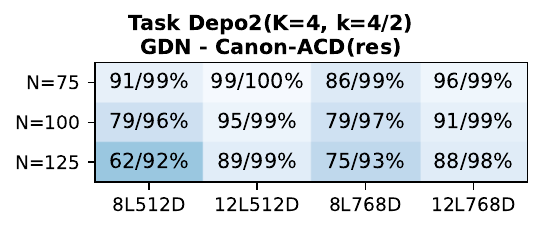}
\hspace*{-7mm}
\\
\hspace*{-7mm}
\includegraphics[page=1,trim={2.5mm 1.5mm 2.5mm 1.5mm},clip,width=\imgwidthBase]{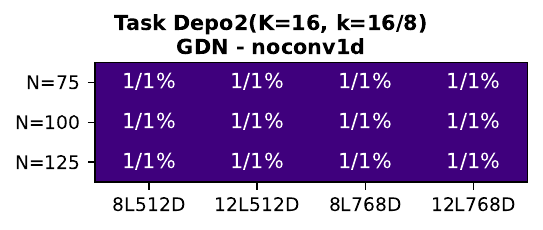}
\includegraphics[page=1,trim={2.5mm 1.5mm 2.5mm 1.5mm},clip,width=\imgwidthBase]{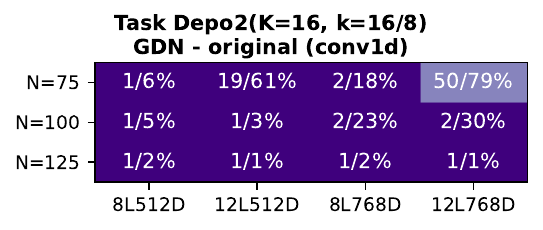}
\includegraphics[page=1,trim={2.5mm 1.5mm 2.5mm 1.5mm},clip,width=\imgwidthBase]{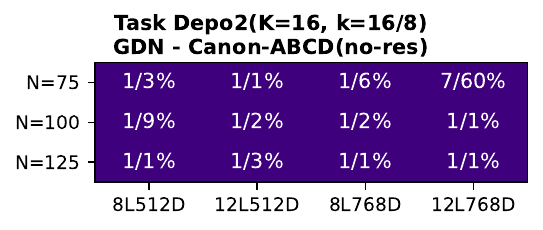}
\includegraphics[page=1,trim={2.5mm 1.5mm 2.5mm 1.5mm},clip,width=\imgwidthBase]{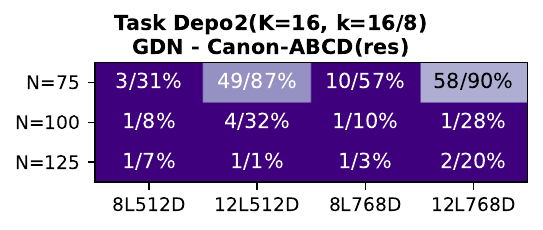}
\includegraphics[page=1,trim={2.5mm 1.5mm 2.5mm 1.5mm},clip,width=\imgwidthBase]{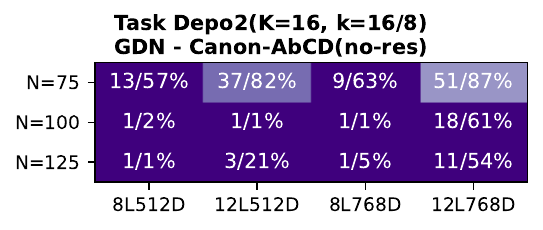}
\includegraphics[page=1,trim={2.5mm 1.5mm 2.5mm 1.5mm},clip,width=\imgwidthBase]{perm_multi/GDN-Res-Canon-AbbCD}
\includegraphics[page=1,trim={2.5mm 1.5mm 2.5mm 1.5mm},clip,width=\imgwidthBase]{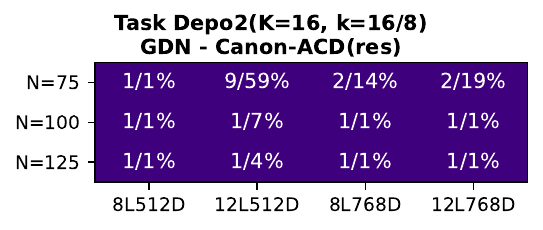}
\hspace*{-7mm}
\\
\hspace*{-7mm}
\includegraphics[page=1,trim={2.5mm 1.5mm 2.5mm 1.5mm},clip,width=\imgwidthBase]{top_sort/GDN-noconv1d}
\includegraphics[page=1,trim={2.5mm 1.5mm 2.5mm 1.5mm},clip,width=\imgwidthBase]{top_sort/GDN-original_conv1d_}
\includegraphics[page=1,trim={2.5mm 1.5mm 2.5mm 1.5mm},clip,width=\imgwidthBase]{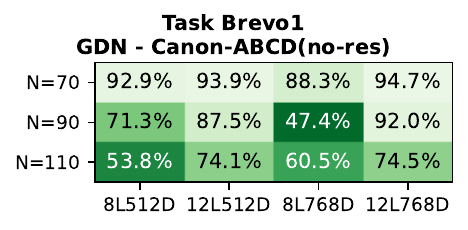}
\includegraphics[page=1,trim={2.5mm 1.5mm 2.5mm 1.5mm},clip,width=\imgwidthBase]{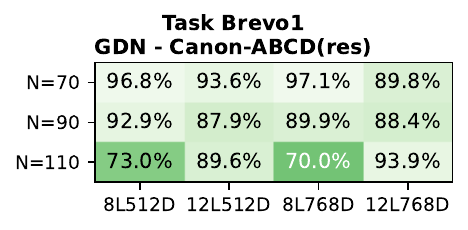}
\includegraphics[page=1,trim={2.5mm 1.5mm 2.5mm 1.5mm},clip,width=\imgwidthBase]{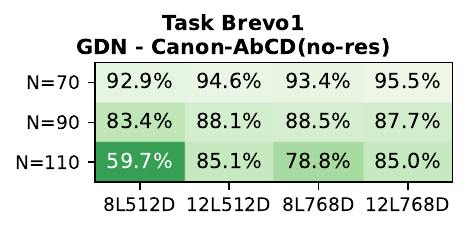}
\includegraphics[page=1,trim={2.5mm 1.5mm 2.5mm 1.5mm},clip,width=\imgwidthBase]{top_sort/GDN-Res-Canon-AbbCD}
\includegraphics[page=1,trim={2.5mm 1.5mm 2.5mm 1.5mm},clip,width=\imgwidthBase]{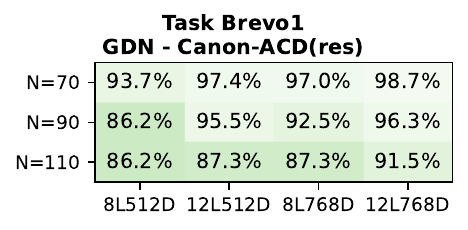}
\hspace*{-7mm}
\\
\hspace*{-7mm}
\includegraphics[page=1,trim={2.5mm 1.5mm 2.5mm 1.5mm},clip,width=\imgwidthBase]{top_sort_multi/GDN-noconv1d}
\includegraphics[page=1,trim={2.5mm 1.5mm 2.5mm 1.5mm},clip,width=\imgwidthBase]{top_sort_multi/GDN-original_conv1d_}
\includegraphics[page=1,trim={2.5mm 1.5mm 2.5mm 1.5mm},clip,width=\imgwidthBase]{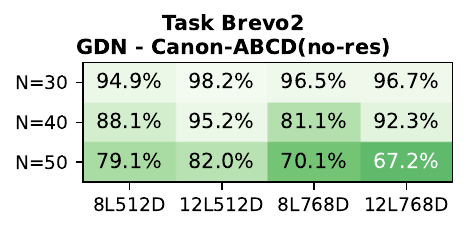}
\includegraphics[page=1,trim={2.5mm 1.5mm 2.5mm 1.5mm},clip,width=\imgwidthBase]{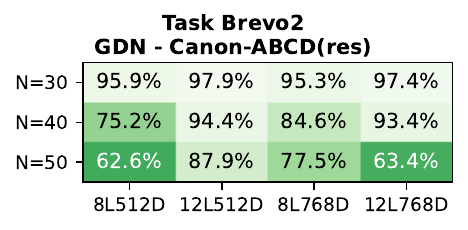}
\includegraphics[page=1,trim={2.5mm 1.5mm 2.5mm 1.5mm},clip,width=\imgwidthBase]{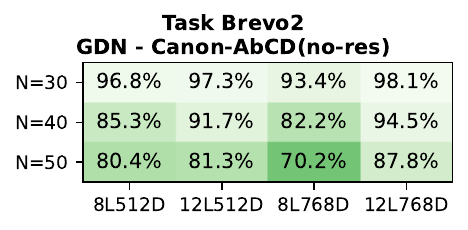}
\includegraphics[page=1,trim={2.5mm 1.5mm 2.5mm 1.5mm},clip,width=\imgwidthBase]{top_sort_multi/GDN-Res-Canon-AbbCD}
\includegraphics[page=1,trim={2.5mm 1.5mm 2.5mm 1.5mm},clip,width=\imgwidthBase]{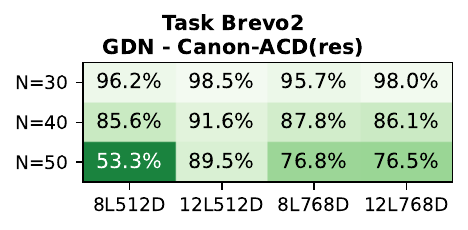}
\hspace*{-7mm}
\\
\hspace*{-7mm}
\includegraphics[page=1,trim={2.5mm 1.5mm 2.5mm 1.5mm},clip,width=\imgwidthBase]{arith/GDN-noconv1d}
\includegraphics[page=1,trim={2.5mm 1.5mm 2.5mm 1.5mm},clip,width=\imgwidthBase]{arith/GDN-original_conv1d_}
\includegraphics[page=1,trim={2.5mm 1.5mm 2.5mm 1.5mm},clip,width=\imgwidthBase]{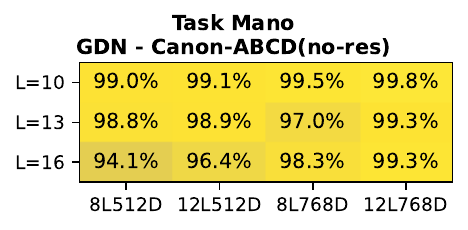}
\includegraphics[page=1,trim={2.5mm 1.5mm 2.5mm 1.5mm},clip,width=\imgwidthBase]{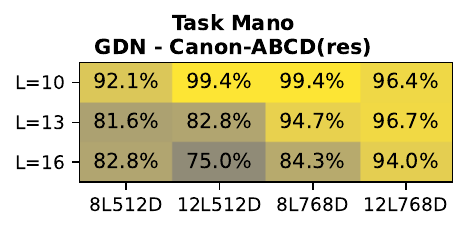}
\includegraphics[page=1,trim={2.5mm 1.5mm 2.5mm 1.5mm},clip,width=\imgwidthBase]{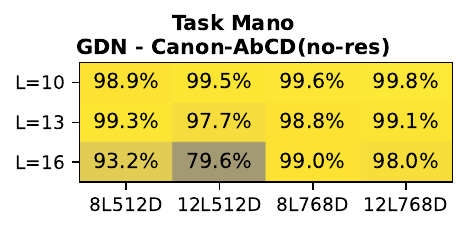}
\includegraphics[page=1,trim={2.5mm 1.5mm 2.5mm 1.5mm},clip,width=\imgwidthBase]{arith/GDN-Res-Canon-AbbCD}
\includegraphics[page=1,trim={2.5mm 1.5mm 2.5mm 1.5mm},clip,width=\imgwidthBase]{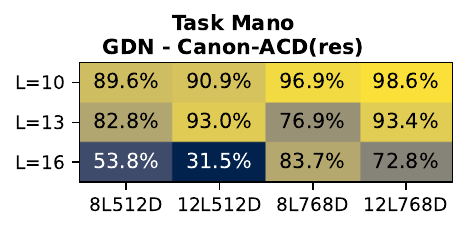}
\hspace*{-7mm}
\\
\hspace*{-7mm}
\includegraphics[page=1,trim={2.5mm 1.5mm 2.5mm 1.5mm},clip,width=\imgwidthBase]{cfg/GDN-noconv1d}
\includegraphics[page=1,trim={2.5mm 1.5mm 2.5mm 1.5mm},clip,width=\imgwidthBase]{cfg/GDN-original_conv1d_}
\includegraphics[page=1,trim={2.5mm 1.5mm 2.5mm 1.5mm},clip,width=\imgwidthBase]{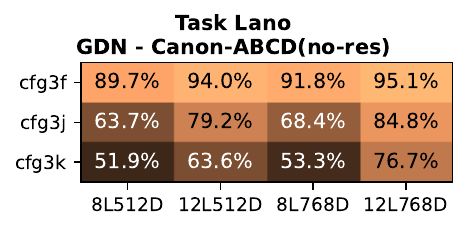}
\includegraphics[page=1,trim={2.5mm 1.5mm 2.5mm 1.5mm},clip,width=\imgwidthBase]{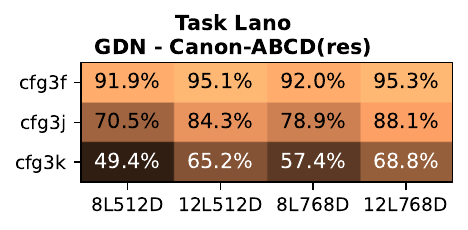}
\includegraphics[page=1,trim={2.5mm 1.5mm 2.5mm 1.5mm},clip,width=\imgwidthBase]{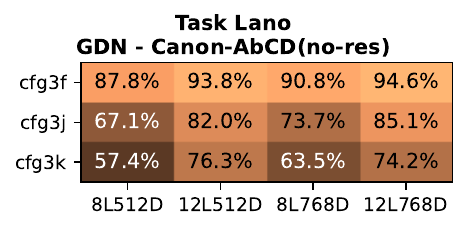}
\includegraphics[page=1,trim={2.5mm 1.5mm 2.5mm 1.5mm},clip,width=\imgwidthBase]{cfg/GDN-Res-Canon-AbbCD}
\includegraphics[page=1,trim={2.5mm 1.5mm 2.5mm 1.5mm},clip,width=\imgwidthBase]{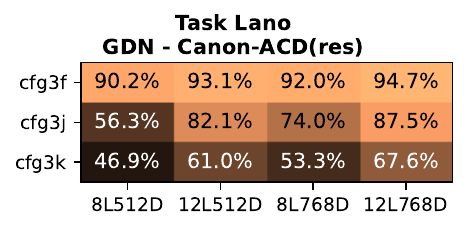}
\hspace*{-7mm}
\\
\hspace*{-7mm}
\includegraphics[page=1,trim={2.5mm 1.5mm 2.5mm 1.5mm},clip,width=\imgwidthBase]{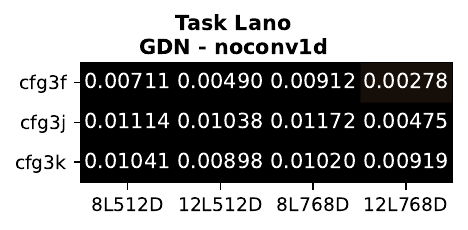}
\includegraphics[page=1,trim={2.5mm 1.5mm 2.5mm 1.5mm},clip,width=\imgwidthBase]{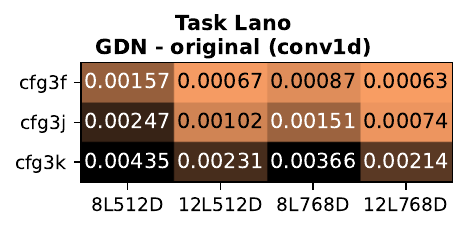}
\includegraphics[page=1,trim={2.5mm 1.5mm 2.5mm 1.5mm},clip,width=\imgwidthBase]{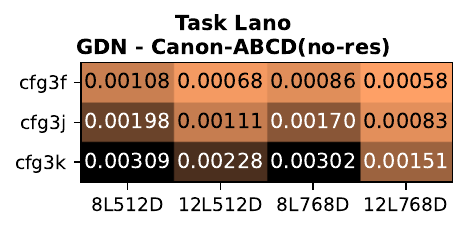}
\includegraphics[page=1,trim={2.5mm 1.5mm 2.5mm 1.5mm},clip,width=\imgwidthBase]{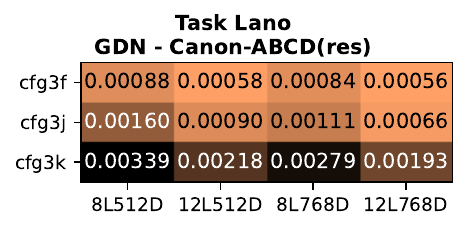}
\includegraphics[page=1,trim={2.5mm 1.5mm 2.5mm 1.5mm},clip,width=\imgwidthBase]{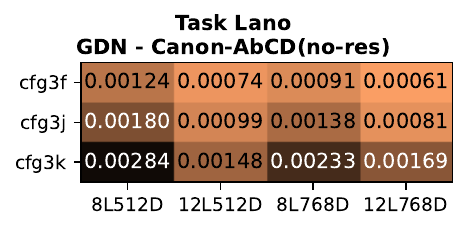}
\includegraphics[page=1,trim={2.5mm 1.5mm 2.5mm 1.5mm},clip,width=\imgwidthBase]{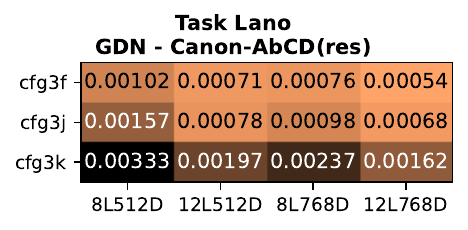}
\includegraphics[page=1,trim={2.5mm 1.5mm 2.5mm 1.5mm},clip,width=\imgwidthBase]{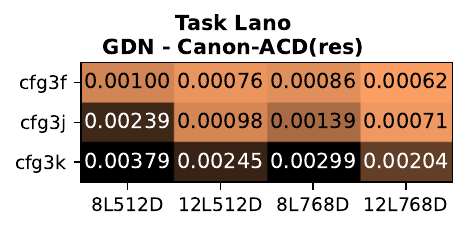}
\hspace*{-7mm}
\caption{\label{fig:full-gdn}%
\textbf{GDN variants} (left to right): no conv1d, original (w/ conv1d), Canon-ABCD(no-res), Canon-ABCD(res), Canon-AbCD(no-res), Canon-AbCD(res), Canon-ACD(res).}
\end{figure}

\small
\setlength{\bibsep}{3pt}

\bibliographystyle{plainnat}

\bibliography{../canon-paper/canon}

\end{document}